\documentclass[12pt]{article}
\pdfoutput=1

\usepackage{amsmath,amssymb,amsthm,amsfonts,bm,physics,mathrsfs,upgreek,bbm}
\usepackage{graphicx}
\usepackage{float}
\usepackage{enumitem}
\usepackage{array}
\usepackage{booktabs}
\usepackage{multirow}
\usepackage{afterpage}
\usepackage{listings}
\usepackage{subcaption}
\usepackage{algorithm}
\usepackage{algorithmic}
\usepackage{changepage}

\usepackage[hang,flushmargin]{footmisc}
\usepackage{setspace}
\usepackage{xcolor}
\usepackage[hypertexnames=false]{hyperref}
\usepackage[capitalize,noabbrev]{cleveref}
\usepackage[round]{natbib}
\definecolor{Bleu}{RGB}{0,0,204}
\hypersetup{
  colorlinks,
  citecolor=Bleu,
  linkcolor=Bleu,
  urlcolor=Bleu
}

\DeclareMathOperator*{\argmin}{argmin}

\newcommand{\addTK}{\widetilde T_{\mathcal K}}
\newcommand{\adk}{\widetilde{\partial}_k}

\DeclareFontFamily{U}{jkpmia}{}
\DeclareFontShape{U}{jkpmia}{m}{it}{<->s*jkpmia}{}
\DeclareFontShape{U}{jkpmia}{bx}{it}{<->s*jkpbmia}{}
\DeclareMathAlphabet{\mathfrak}{U}{jkpmia}{m}{it}
\SetMathAlphabet{\mathfrak}{bold}{U}{jkpmia}{bx}{it}

\theoremstyle{plain}
\newtheorem{theorem}{Theorem}
\newtheorem{lemma}{Lemma}

\newtheorem{example}{Example}

\usepackage{etoolbox}

\makeatletter
\AfterEndEnvironment{theorem}{\@afterindentfalse\@afterheading}
\AfterEndEnvironment{lemma}{\@afterindentfalse\@afterheading}
\AfterEndEnvironment{proposition}{\@afterindentfalse\@afterheading}
\AfterEndEnvironment{corollary}{\@afterindentfalse\@afterheading}
\AfterEndEnvironment{remark}{\@afterindentfalse\@afterheading}
\AfterEndEnvironment{enumerate}{\@afterindentfalse\@afterheading}
\makeatother

\def\spacingset#1{\renewcommand{\baselinestretch}{#1}\small\normalsize}
\newcommand{\blind}{0}

\providecommand{\keywords}[1]{\small{\textbf{\textit{Keywords---}} #1}}

\usepackage[compact]{titlesec}

\newcommand{\JASATitleText}{Efficient Inference for Inverse Reinforcement Learning and Dynamic Discrete Choice Models}
\title{\bf \JASATitleText}
\author{
  Lars van der Laan\thanks{Corresponding author: lvdlaan@uw.edu} \\
  University of Washington \\
  \and
  Aur\'elien Bibaut \\
  Netflix Research \\
  \and
  Nathan Kallus \\
  Cornell Tech, Cornell University
}
\spacingset{1}

\begin{document}

\if0\blind
{
  \maketitle
} \fi

\if1\blind
{
  \bigskip
  \bigskip
  \bigskip
  \begin{center}
    {\LARGE\bf \JASATitleText}
  \end{center}
  \medskip
} \fi

\bigskip
\begin{abstract}
In many sequential decision-making problems, researchers observe actions but
not the rewards that drive behavior, yet wish to evaluate counterfactual
policies. Flexible inverse reinforcement learning (IRL) methods accommodate rich
reward representations but typically lack valid inference, whereas classical
dynamic discrete choice (DDC) methods provide inference under restrictive
parametric structure. We develop a semiparametric framework for
\emph{debiased inverse reinforcement learning} in maximum-entropy IRL and
Gumbel-shock DDC models. We show that the log behavior policy acts as a
pseudo-reward, identifying policy-value differences without point identifying
the reward and, under a normalization, yielding a point-identified reward
representative. This reduces inference on reward-dependent estimands to
inference on smooth functionals of the behavior policy and transition kernel.
We establish pathwise differentiability and characterize efficient influence
functions through reward and transition Riesz representers, with the transition
component obtainable from existing observed-reward off-policy evaluation
theory. For normalized rewards, we derive an explicit transformation of these
representers that accounts for the normalization constraint. We construct
automatic debiased machine-learning estimators that permit flexible nuisance
estimation while attaining \(\sqrt n\)-consistency, asymptotic normality, and
semiparametric efficiency. We illustrate the theory through efficient inference
for fixed-policy values under pseudo-rewards and normalized rewards, and for
counterfactual values when the agent reoptimizes over a restricted action set.

\end{abstract}

\noindent
\newpage
\setstretch{1.15}

\section{Introduction}
\label{sec:intro}

Behavioral settings in which actions are observed but rewards are not arise in
expert driving and robotic control
\citep[e.g.][]{abbeel2004apprenticeship,wulfmeier2015deep},
human--computer interaction
\citep[e.g.][]{chandramohan2012user,hossain2023bayesian},
clinical decision-making
\citep[e.g.][]{yu2019inverse,bovenzi2025identifying},
and economics
\citep[e.g.][]{rust1987optimal,hotz1993conditional}.
In these settings, researchers observe a decision-maker---such as a physician,
robot, or artificial agent---acting over time. The observed data comprise
trajectories of states, which describe the evolving decision context, and
actions, but not the latent rewards or utilities that guide those actions.
Learning the underlying reward function provides insight into the
decision-maker's preferences and a basis for evaluating counterfactual policies
or changes to the decision-making environment.  
For example, one may seek to estimate the expected cumulative reward that the
decision-maker would attain under an alternative policy or under an intervention
on the environment that changes the state dynamics or available actions.

Inverse reinforcement learning (IRL) and dynamic discrete choice (DDC)
provide two frameworks for inferring latent rewards and evaluating
counterfactuals from observed behavior. Both impose structural assumptions
linking observed behavior to underlying rewards by modeling decisions as the
outcome of reward or utility maximization. A basic IRL formulation assumes
that the decision-maker chooses only actions that maximize expected cumulative
reward
\citep{ng2000algorithms,abbeel2004apprenticeship}. To accommodate variability
and suboptimality in observed choices, DDC models instead introduce unobserved
utility shocks into the decision-maker's optimization problem
\citep{rust1987optimal,hotz1993conditional,aguirregabiria2010dynamic}.
Under the classical specification with i.i.d.\ Gumbel
(Type-I extreme-value) shocks, optimal behavior follows a softmax choice rule.  
The same choice rule arises in maximum-entropy IRL, which can be viewed either
as selecting the maximum-entropy distribution over behavior subject to matching
observed behavioral constraints, or equivalently as having the decision-maker
maximize expected cumulative reward augmented by an entropy bonus
\citep{ziebart2010causal,haarnoja2017reinforcement}.  These formulations yield
the same structural model: the decision-maker's policy is the softmax of a
state--action value function satisfying a soft Bellman equation determined by
the latent reward and state-transition law.

A standard approach to estimating the reward function under this structural
model is to maximize the log-likelihood of the observed actions under the
softmax policy, with the policy and value function constrained to satisfy the
soft Bellman equation. The reward is generally not uniquely identified,
however, because distinct reward functions can induce the same choice
probabilities and hence the same likelihood
\citep{ng1999policy,ng2000algorithms,cao2021identifiability,
skalse2023invariance,skalse2024partial}. Point identification of the reward
therefore requires an additional restriction, or normalization, such as fixing
the reward of a reference action at each state
\citep{rust1987optimal,geng2020deep}. Domain knowledge may guide this choice;
for example, an outside option may reasonably be assigned zero immediate
reward. A reward-dependent functional may nevertheless be identified without
normalization if it is invariant across all rewards consistent with the
observed choice probabilities.

Valid inference on reward-dependent functionals requires accommodating flexible
models of behavior, rewards, and dynamics. Functionals of interest include
counterfactual values, such as the expected return under an alternative policy
or a restricted action set, as well as other summaries of the latent reward
function
\citep{aguirregabiria2010dynamic,kalouptsidi2021identification,
zielnicki2025value}. Classical DDC methods provide inference under parametric
reward models
\citep[e.g.][]{rust1987optimal,hotz1993conditional,
aguirregabiria2010dynamic}, but misspecification can bias reward estimates and
invalidate inference for counterfactual values. Flexible IRL methods reduce
reliance on restrictive parametric reward specifications
\citep[e.g.][]{abbeel2004apprenticeship,ramachandran2007bayesian,
ziebart2008maximum,levine2011nonlinear,wulfmeier2015deep,ho2016gail,
fu2018airl,geng2020deep}, but generally do not provide frequentist guarantees
for inference on such targets. Obtaining these guarantees requires controlling
estimation, regularization, and approximation errors in the behavior policy
and transition-dependent nuisance functions. We develop a semiparametric
framework to address this gap.

\subsection{Our contributions}

We develop a semiparametric framework for \emph{debiased inverse reinforcement
learning} in maximum-entropy IRL and Gumbel-shock DDC models. We first study
identification of the latent reward and reward-dependent functionals. Our key
identification result is that the log behavior policy can be treated as a
pseudo-reward: it identifies policy-value differences without requiring the
reward itself to be point identified and, after imposing a normalization,
also yields a point-identified reward representative. Consequently,
reward-dependent estimands can be expressed as smooth functionals of the
behavior policy and transition kernel, reducing semiparametric inference with
latent rewards to inference on observed-data functionals. A further key feature
is that efficient influence functions for observed-reward problems can be
leveraged to construct their latent-reward counterparts, with additional terms
accounting for reward estimation and normalization. Our main contributions are
as follows.

\begin{enumerate}
\item We establish identification results for latent rewards and policy values.
We show that the log behavior policy serves as a pseudo-reward for identifying
policy-value differences without identifying the reward itself. For point
identification of the reward, we introduce affine normalization constraints
that fix its average under a reference policy at each state, generalizing
reference-action normalizations. Under such a normalization, the reward can be
recovered from the log behavior policy and transition law by solving an
auxiliary Bellman equation.

\item We develop semiparametric efficiency theory for a general class of
reward-dependent functionals. We establish pathwise differentiability and
characterize the efficient influence function through reward and transition
Riesz representers, allowing existing influence-function calculations for
observed-reward problems to be reused. For normalized rewards, we derive an
explicit transformation of these representers that accounts for the
normalization constraint.

\item We construct debiased machine-learning estimators that accommodate
flexible nuisance estimation. We establish asymptotic normality and
semiparametric efficiency under high-level rate and regularity conditions,
providing asymptotically valid confidence intervals.

\item We specialize the general theory to obtain explicit efficient influence
functions and estimators for fixed-policy values, counterfactual softmax policy
values, policy values under normalized rewards, and structural coefficients of
normalized rewards.
\end{enumerate}

The paper proceeds as follows. Section~\ref{sec::setup} reviews the IRL and DDC
models. Section~\ref{sec:ident} establishes identification of policy-value
differences and, under normalization, the reward itself.
Sections~\ref{sec::EIFsub} and~\ref{sec::est} develop the semiparametric theory
and debiased machine-learning estimators for smooth reward-dependent
functionals. Section~\ref{sec::applications} specializes these results to the
main policy-value targets.

\subsection{Related work}

\noindent \textbf{Dynamic discrete choice models.}
Classical DDC work develops methods for identification and estimation
of structural utility parameters, including nested fixed-point,
conditional choice probability, and simulation-based methods
\citep{rust1987optimal,hotz1993conditional,hotz1994simulation,
aguirregabiria2010dynamic,buchholz2021semiparametric,
kalouptsidi2021identification}.
\citet{hotz1993conditional} show that conditional choice probabilities
determine differences in choice-specific value functions and use
this relation to estimate utility-model parameters without repeatedly
solving the optimal dynamic program. Our identification analysis
uses this relation to recover rewards under affine normalization
constraints by solving an auxiliary Bellman equation involving
the log-behavior policy and transition law. We also show that the
log-behavior policy, used as a pseudo-reward, preserves policy-value
differences without requiring identification of the reward itself.

Several lines of work allow flexible estimation of nuisance components
while retaining a finite-dimensional parameterization of the reward
or utility function. For dynamic discrete games,
\citet{bajari2015dynamic} study identification and semiparametric
efficiency with nonparametric equilibrium objects and transition
probabilities. For DDC models, Neyman-orthogonal methods permit
inference on reward-model parameters with high-dimensional states
and flexible first-stage learners
\citep{chernozhukov2022locally,semenova2018machine,
chernozhukov2022automaticdynamic}.
\citet{adusumilli2025temporal} use temporal-difference learning and
approximate value iteration to estimate recursive terms in
conditional choice probability estimators and develop locally
robust corrections for pseudo-maximum likelihood estimation.
Our framework retains the softmax/Gumbel-shock choice structure
but allows the reward function and transition law to be
nonparametric. We derive efficient influence functions, efficiency
bounds, and debiased estimators for smooth reward-dependent
functionals, including functionals of rewards identified under
normalization constraints.

\noindent \textbf{Inverse reinforcement learning.}
Early IRL methods study reward recovery and imitation under hard
optimality assumptions, often using linear reward models
\citep[e.g.][]{ng2000algorithms,abbeel2004apprenticeship}.
Later work develops maximum-entropy and adversarial formulations,
including methods with flexible function approximators
\citep[e.g.][]{ziebart2008maximum,ziebart2010modeling,
levine2011nonlinear,ho2016gail,fu2018airl,geng2020deep,
kang2025empirical}.
Several papers characterize reward ambiguity, including
\citet{fu2018airl}, Theorem~1 of \citet{cao2021identifiability},
and Lemma~2 of \citet{geng2020deep}.
Deep PQR \citep{geng2020deep} identifies rewards under anchor-action
normalization through an anchor-specific Bellman fixed point followed
by a conditional-mean regression. Our identification result covers
the broader class of affine constraints that fix the statewise
average reward under a reference policy, including anchor-action
normalization as a special case. It also gives a direct
representation of the normalized reward in terms of the log-behavior
policy and a policy-specific $Q$-function, without a separate
conditional-mean regression. Beyond reward recovery, we develop
efficient inference for smooth functionals of the normalized reward.

\noindent \textbf{Off-policy evaluation and double reinforcement learning.}
Debiased and doubly robust methods for off-policy evaluation estimate
policy values from observed rewards
\citep[e.g.][]{jiang2016doubly,thomas2016data,tang2019doubly,
liu2018breaking,van2018online,kallus2020double,uehara2020minimax,
shi2021deeply,kallus2022efficiently,mehrabi2024off}.
Double reinforcement learning develops efficient policy-value
estimators with flexible nuisance estimation
\citep{kallus2020double,kallus2022efficiently}, and subsequent work
extends this approach to more general functionals of the
\(Q\)-function \citep{van2025automaticDRL}.
Our framework addresses the complementary problem of inference
when rewards are latent and must be inferred from observed behavior
under the softmax model. We show how influence-function calculations from observed-reward
off-policy evaluation can be adapted to this setting by accounting
for the dependence of the inferred reward on the observed-data law.

\section{Problem setup}
\label{sec::setup}

\subsection{MDP, observed data, and statistical model}

We consider an infinite-horizon, time-homogeneous Markov decision process
(MDP) with state space \(\mathcal S\) and finite action space \(\mathcal A\).
Let \(\mu\) be a finite dominating measure on \(\mathcal S\), and let
\(\lambda:=\#\otimes\mu\) be the associated product measure on
\(\mathcal A\times\mathcal S\), where \(\#\) denotes counting measure
on \(\mathcal A\). The initial state \(S_0\) is drawn from a density
\(\rho_{0,\mathrm{in}}\) with respect to \(\mu\). At each time
\(t\geq 0\), the agent occupies state \(S_t\), selects action
\(A_t\in\mathcal A\) according to the behavior policy
\(\pi_0(\cdot\mid S_t)\), and transitions to \(S_{t+1}\) with conditional
density \(k_0(\cdot\mid A_t,S_t)\) with respect to \(\mu\). The state--action
pair \((S_t,A_t)\) is associated with a latent reward
\(r_0^\dagger(A_t,S_t)\), which is not observed by the analyst.

We observe \(n\) independent trajectories
\(O_i=(S_{i,0},A_{i,0},\ldots,S_{i,T-1},A_{i,T-1},S_{i,T})\),
\(i=1,\ldots,n\), of fixed length \(T\) from a common law
\(\mathbb P_0\). We write \(\mathcal D_n:=\{O_i\}_{i=1}^n\).
To describe quantities associated with a single transition, we define
the pooled transition law
\[
P_0
:=\frac1T\sum_{t=0}^{T-1}
\mathbb P_0\bigl((S_t,A_t,S_{t+1})\in\cdot\bigr).
\]
We write \((S,A,S')\sim P_0\) for a generic transition and let
\(\rho_0\) denote the density of \(S\) with respect to \(\mu\).
The empirical trajectory and pooled transition measures are
\[
\mathbb P_n:=\frac1n\sum_{i=1}^n\delta_{O_i},
\qquad
P_n:=\frac1{nT}\sum_{i=1}^n\sum_{t=0}^{T-1}\delta_{S_{i,t}, A_{i,t}, S_{i,t+1}},
\]
where \(\delta_x\) denotes the point mass at \(x\).
For any measure \(Q\) and integrable function \(f\), we write
\(Qf:=\int f\,dQ\). We use \(\mathbb E_0\) and \(E_0\) for
expectation under \(\mathbb P_0\) and \(P_0\), respectively.

Throughout, we impose Assumption~\ref{cond::boundedpositivity} in
Appendix~\ref{appendix::model}, which requires the trajectory law
\(\mathbb P_0\) to belong to the nonparametric model
\(\mathfrak M_T\). This model assumes time homogeneity,
uniform upper and positive lower bounds on the pooled state and
transition densities, and a uniform positive lower bound on the
behavior-policy probabilities.

\subsection{Policy values and Bellman equations}

We first introduce the policy-value notation used throughout to
define the reward-dependent functionals of interest. For a bounded reward
\(r\), transition kernel \(k\), policy \(\pi\), and discount factor
\(\gamma\in[0,1)\), the discounted policy value is
\[
V_{r,k}(\pi;\gamma)
:=
\mathbb E_{k,\pi}\!\left[
  \sum_{t=0}^{\infty}\gamma^t r(A_t,S_t)
\right],
\]
where \(\mathbb E_{k,\pi}\) denotes expectation over trajectories initialized
from the observed initial-state distribution and evolving under \(k\) and
\(\pi\). For a state--action function \(f\), write
\(\pi f(s):=\sum_{a\in\mathcal A}\pi(a\mid s)f(a,s)\), and define the
fixed-policy Bellman operator
\[
(\mathcal T_{k,\pi,\gamma}f)(a,s)
:=
f(a,s)
-\gamma\int \pi f(s')k(s'\mid a,s)\,\mu(ds').
\]
Because \(\gamma<1\), \(\mathcal T_{k,\pi,\gamma}\) is invertible on
\(L^\infty(\lambda)\) \citep{bellman1954theory}. The action-value function $q_{r,k}^{\pi,\gamma}
:=
\mathcal T_{k,\pi,\gamma}^{-1}r$ is the unique bounded solution of the Bellman equation
\[q_{r,k}^{\pi,\gamma}(a,s)
=
r(a,s)
+\gamma\int
\sum_{a'\in\mathcal A}\pi(a'\mid s')
q_{r,k}^{\pi,\gamma}(a',s')
k(s'\mid a,s)\,\mu(ds').
\]
The corresponding state-value function is
\(V_{r,k}^{\pi,\gamma}:=\pi q_{r,k}^{\pi,\gamma}\), so that $V_{r,k}(\pi;\gamma)
=
\mathbb E_0\!\left[V_{r,k}^{\pi,\gamma}(S_0)\right].$

\subsection{Behavioral model for inverse reinforcement learning}

Our goal is to conduct inference on reward-dependent functionals, such as
policy values, using only the observed state--action trajectories. A fundamental
challenge is that \(r_0^\dagger\) is not point identified from these data alone:
distinct reward functions can rationalize the same behavior policy, for example
by yielding the same optimal policy
\citep{ng1999policy,ng2000algorithms}.  
IRL and DDC address this ambiguity by imposing a behavioral optimality model that
links the observed policy to the latent reward function.

Classical notions of optimality assume that agents choose actions that maximize
long-run reward. This model can be too restrictive for behavioral data, where
suboptimal actions may occur with positive probability. We therefore consider a
\emph{soft optimality} model in which actions with larger long-run utility are
chosen with higher probability, while every action may occur with positive
probability. Such models arise in Gumbel-shock dynamic discrete choice
\citep{rust1987optimal,hotz1993conditional,aguirregabiria2010dynamic}
and maximum-entropy IRL
\citep{ziebart2008maximum,ziebart2010modeling,haarnoja2017reinforcement}.

In the maximum-entropy formulation, one assumes that the agent's behavior policy
maximizes an entropy-regularized long-run reward criterion. Specifically, for a
fixed discount factor \(\gamma\in[0,1)\), we assume that the behavior policy
\(\pi_0\) maximizes
\begin{equation}
\label{eqn::entmax}
\pi \mapsto
\mathbb{E}_{k_0,\pi}\!\left[
  \sum_{t=0}^\infty \gamma^t
  \{ r_0^\dagger(A_t,S_t)
    + \tau\,\operatorname{Ent}(\pi(\cdot\mid S_t)) \}
\right]
\end{equation}
over all policies \(\pi\), where
\(\operatorname{Ent}(\pi(\cdot\mid s))
:=
-\sum_{a\in\mathcal A}\pi(a\mid s)\log\pi(a\mid s)\)
denotes the Shannon entropy at state \(s\).
We treat \(\gamma\) as known, since identifying the discount factor from
observed choices requires additional assumptions or sources of information
\citep{abbring2020identifying,rolland2022identifiability}. Because the scale of \(r_0^\dagger\) and the
temperature \(\tau\) are not separately identified, we set \(\tau=1\) and
interpret rewards and policy values in these normalized units, as is standard
in maximum-entropy IRL and DDC.

Under this normalization, the unique solution to the optimization problem
in \eqref{eqn::entmax} is the softmax choice rule
\citep{haarnoja2017reinforcement}
\begin{equation}
\pi_0(a \mid s)
= \frac{\exp\!\left\{ r_0^\dagger(a,s) + \gamma v_0^\dagger(a,s) \right\}}
       {\sum_{a' \in \mathcal{A}}
        \exp\!\left\{ r_0^\dagger(a',s) + \gamma v_0^\dagger(a',s) \right\}},
\label{eqn::softmaxprob}
\end{equation}
where \(v_0^\dagger\) satisfies the \textit{soft Bellman equation}
\[
v_0^\dagger(a,s)
=
\int V_0^\dagger(s')\,k_0(s'\mid a,s)\,\mu(ds'),
\qquad
V_0^\dagger(s)
= \log\!\left(
    \sum_{a' \in \mathcal{A}}
       \exp\!\bigl\{ r_0^\dagger(a',s)
       + \gamma v_0^\dagger(a',s) \bigr\}
\right).
\]
Here \(V_0^\dagger\) is the associated soft value function, corresponding to the
optimal value of the entropy-regularized control problem from state \(s\),
whereas \(v_0^\dagger\) is the corresponding expected continuation value.

The same softmax choice rule arises in dynamic discrete-choice models. Under
this formulation, the agent chooses
\[
A_t \in \arg\max_{a \in \mathcal A}
\left\{
r_0^\dagger(a,S_t)
+ \gamma v_0^\dagger(a,S_t)
+ \varepsilon_t(a)
\right\},
\]
where \(v_0^\dagger(a,s)\) is the choice-specific expected continuation value
and the shocks \(\varepsilon_t(a)\) are i.i.d.\ Type-I extreme-value (Gumbel)
across actions and periods
\citep{rust1987optimal,pitombeira2024trajectory}; see
\citet{kang2026offlineirlpartii} for a review. The shocks represent
action-specific utility components observed by the agent but not by the
analyst. Conditional on these shocks, the agent behaves optimally; integrating
them out yields the softmax choice probabilities in
\eqref{eqn::softmaxprob}. When \(\gamma=0\), this reduces to the classical
static multinomial-logit discrete-choice model
\citep{mcfadden1972conditional,train2009discrete}.

\section{Identification}
\label{sec:ident}

\subsection{Partial identification of the reward}
\label{sec::partialident}

Without additional constraints, the reward \(r_0^\dagger\) is only partially
identified: distinct reward functions can generate the same softmax policy for
a soft-optimal agent \citep{cao2021identifiability}. This nonidentifiability
arises because the policy depends on the choice-specific logits
\(r(a,s)+\gamma v_{r,k_0}(a,s)\) only through their differences across actions.
Consequently, changes to the reward that shift all logits at a given state by
the same amount leave the policy unchanged. We first characterize the resulting
equivalence class of rewards. We then show how a normalization constraint can
be used to identify a unique representative of this class.

Under the soft-optimality model, the true reward \(r_0^\dagger\) minimizes the
population negative log-likelihood subject to the soft Bellman equation
\citep{rust1987optimal,ziebart2008maximum}. In particular,
\begin{equation}
\label{eqn::softbellmanloglik}
r_0^\dagger \in \argmin_{r'\ \mathrm{bounded}}
E_0\!\left[
  \log\!\Bigg(\sum_{a \in \mathcal{A}}
    \exp\{r'(a,S)+\gamma v_{r',k_0}(a,S)\}\Bigg)
  -\{r'(A,S)+\gamma v_{r',k_0}(A,S)\}
\right],
\end{equation}
where \(v_{r',k_0}\) satisfies
\begin{equation}
\label{eqn::softbellmanloss}
v_{r',k_0}(a,s)
=
\int \log\!\left(\sum_{a' \in \mathcal{A}}
   \exp\!\big\{r'(a',s')+\gamma v_{r',k_0}(a',s')\big\}\right)
\,k_0(s'\mid a,s)\,\mu(ds').
\end{equation}
The solution to \eqref{eqn::softbellmanloglik} is not unique. In particular,
for any state function \(c:\mathcal S\to\mathbb R\), a reward of the form \citep{ng1999policy,fu2018airl,geng2020deep}
\begin{equation}
\label{eqn::equivclass}
r(a,s)
=
r_0^\dagger(a,s)
+c(s)
-\gamma\int c(s')\,k_0(s'\mid a,s)\,\mu(ds')
\end{equation}
yields the same behavior policy as \(r_0^\dagger\) and therefore also solves
\eqref{eqn::softbellmanloglik}. Conversely, every reward that
induces the same behavior policy is of this form. Thus, the observed behavior
identifies \(r_0^\dagger\) only up to the equivalence class in
\eqref{eqn::equivclass}.

One particularly simple representative of this equivalence class is the
log-behavior policy itself,
\(r_0(a,s):=\log\pi_0(a\mid s)\). For this choice,
\(v_{r_0,k_0}\equiv0\) satisfies the soft Bellman equation because
\[
\log \sum_{a'\in\mathcal A}\exp\{r_0(a',s)\}
=
\log \sum_{a'\in\mathcal A}\pi_0(a'\mid s)
=
0.
\]
Moreover, the induced softmax policy is
\(\exp\{r_0(a,s)\}=\pi_0(a\mid s)\), so \(r_0\) also solves the population
negative log-likelihood problem in \eqref{eqn::softbellmanloglik}. Thus,
\(r_0\) is a reward representative observationally equivalent to
\(r_0^\dagger\).

\begin{lemma}[Log-policy reward representative]
\label{theorem::trivialsol}
Under Assumption~\ref{cond::boundedpositivity}, the log-behavior policy
\(r_0=\log\pi_0\) solves \eqref{eqn::softbellmanloglik}. Moreover,
\(r_0\) belongs to the equivalence class \eqref{eqn::equivclass} with
\(c(s)=-V_0^\dagger(s)\).
\end{lemma}

\subsection{Identification of policy values}

Although the reward is identified only up to an equivalence class, the
log-behavior policy \(r_0\) suffices to identify any functional that is
invariant over this class. An important example is a policy-value difference
under a common transition law and discount factor. Indeed, every reward in the
equivalence class differs from \(r_0\) by a term of the form
\(c(s)-\gamma\int c(s')k_0(s'\mid a,s)\,\mu(ds')\), which shifts every
policy's value by the same amount \(E_0\{c(S_0)\}\). This shift cancels in
policy-value differences. Hence, the identified log-behavior policy
\(r_0(a,s)=\log\pi_0(a\mid s)\) can be used as a pseudo-reward for comparing
policies. The next result formalizes this observation.

\begin{theorem}[Behaviour policy identifies value differences]
\label{theorem::identvalue}
Suppose Assumption~\ref{cond::boundedpositivity} holds, and suppose that \(r_0^\dagger\) solves \eqref{eqn::softbellmanloglik} for a
discount factor \(\gamma \in [0,1)\). Then, for any policy \(\pi\),
\[
q^{\pi,\gamma}_{r_0,k_0}(a,s)
= q^{\pi,\gamma}_{r_0^\dagger,k_0}(a,s) - V_0^\dagger(s),
\qquad
V_{r_0,k_0}(\pi;\gamma)
= V_{r_0^\dagger,k_0}(\pi;\gamma) - \mathbb E_0[V_0^\dagger(S_0)].
\]
Hence, for any two policies \(\pi_1\) and \(\pi_2\), $V_{r_0,k_0}(\pi_1;\gamma) - V_{r_0,k_0}(\pi_2;\gamma)
=
V_{r_0^\dagger,k_0}(\pi_1;\gamma)
-
V_{r_0^\dagger,k_0}(\pi_2;\gamma).$
\end{theorem}

\noindent A subtle but important point is that this identification result
depends on the behavioral discount factor \(\gamma\) and transition law
\(k_0\). Policy-value differences evaluated under a different discount factor
or transition law are generally not invariant across compatible rewards and
therefore require additional identifying restrictions. Before turning to
reward identification under normalization, we illustrate
Theorem~\ref{theorem::identvalue} in a counterfactual softmax setting.

\begin{example}[Identification for counterfactual softmax policies]
\label{ex:1}
Counterfactual softmax policies describe how the same agent would behave after
an intervention on the choice environment. Let
\(\varnothing\ne\mathcal A^\star(s)\subseteq\mathcal A\) denote the actions
available in state \(s\), for example after removing an option from a platform
or restricting a treatment menu. Let \(\tau^\star\in(0,\infty)\) control how
deterministically the agent responds to long-run utility. For
a bounded reward \(r\) and transition kernel \(k\), define
\begin{equation}
    \label{eqn::softmaxpolicycf}
\pi_{r,k}^\star(a\mid s)
\propto \mathbbm{1}\{a\in\mathcal{A}^\star(s)\}
   \exp\!\left\{\frac{r(a,s)+\gamma v_{r,k}^\star(a,s)}{\tau^\star}\right\},
\end{equation}
where \(v_{r,k}^\star(a,s)=\int \Phi_{r,k}^\star(s')k(s'\mid a,s)\mu(ds')\)
and
\[
\Phi_{r,k}^\star(s)
:= \tau^\star \log\!\left(\sum_{a'\in\mathcal{A}^\star(s)}
      \exp\!\left\{\frac{r(a',s)+\gamma v_{r,k}^\star(a',s)}
      {\tau^\star}\right\}\right).
\]
Because \(\pi_{r,k}^\star\) is invariant to potential-based reward shaping of
the form \eqref{eqn::equivclass}, and Lemma~\ref{theorem::trivialsol} shows
that \(r_0\) and \(r_0^\dagger\) differ by such a transformation,
\(\pi_{r_0^\dagger,k_0}^\star=\pi_{r_0,k_0}^\star\). Applying
Theorem~\ref{theorem::identvalue} with
\(\pi_1=\pi_{r_0,k_0}^\star\) and \(\pi_2=\pi_0\), the value gain
\(V_{r_0^\dagger,k_0}(\pi_{r_0^\dagger,k_0}^\star;\gamma)
-
V_{r_0^\dagger,k_0}(\pi_0;\gamma)\)
is identified by replacing the latent reward \(r_0^\dagger\) with the
log-behavior-policy reward \(r_0=\log\pi_0\).
\qed
\end{example}

\subsection{Point identification under normalization}
\label{sec:normalization}

When point identification of the latent reward is required, the log behavior
policy \(r_0\) identifies a unique reward representative once an additional
normalization constraint is imposed to select a single element of the
equivalence class.

We consider normalization constraints that fix a known statewise average of
the reward. Specifically, let \(\nu\) be a fixed reference policy and let
\(g\in L^\infty(\mu)\) be a known state function. We impose
\begin{equation}
\label{eqn::rdaggernorm}
\nu r_0^\dagger(s)
:=
\sum_{\tilde a\in\mathcal A}\nu(\tilde a\mid s)
r_0^\dagger(\tilde a,s)
=
g(s),
\qquad P_0\text{-a.e. }s.
\end{equation}
This restriction removes the state-specific ambiguity in
\eqref{eqn::equivclass} and thereby selects a unique reward representative. It
includes several common normalizations as special cases, including
anchor-action, statewise sum-to-zero, and value-based constraints
\citep{hotz1993conditional,bajari2010identification,
kallus2016revealed,geng2020deep}. An anchor-action normalization is obtained
by taking \(\nu(\cdot\mid s)\) to place all mass on a reference action, while
a statewise sum-to-zero normalization is obtained by taking \(\nu\) to be
uniform and \(g\equiv0\). A value-based normalization also fits
\eqref{eqn::rdaggernorm}: requiring
\(V_{r_0^\dagger,k_0}^{\nu,\gamma}(s)=h(s)\) for all \(s\) is equivalent, by
the Bellman equation, to taking
\[
g(s)
=
h(s)
-\gamma\sum_{a\in\mathcal A}\nu(a\mid s)
\int h(\tilde s)\,k_0(d\tilde s\mid a,s).
\]
The pair \((\nu,g)\) may instead be specified using auxiliary information. For
example, if auxiliary data contain observed rewards \(R\) for actions drawn
according to the reference policy \(\nu\), one may take
\(g(s)=E_0(R\mid S=s)\).

The following result shows that \eqref{eqn::rdaggernorm} point identifies the
reward and expresses the resulting representative in terms of the
\(Q\)-function
\(q_{r_0-g,k_0}^{\nu,\gamma}
=\mathcal T_{k_0,\nu,\gamma}^{-1}(r_0-g)\)
associated with the reference policy \(\nu\).

\begin{theorem}[Normalization identifies the reward itself]
\label{theorem::identWithNormalization}
Let \(\nu\) be a policy, let \(g\in L^\infty(\mu)\), and suppose
Assumption~\ref{cond::boundedpositivity} holds.
There exists a unique bounded solution
\(r_{0,\nu,\gamma,g}\)
to \eqref{eqn::softbellmanloglik}  that satisfies
\(\nu r_{0,\nu,\gamma,g}(s) = g(s)\) for \(P_0\text{-a.e.\ } s \in \mathcal{S}\).
Moreover,
\begin{align*}
    r_{0,\nu,\gamma,g}(a,s)
    &= g(s) + q_{r_0-g,k_0}^{\nu,\gamma}(a,s)
       - \sum_{\tilde a \in \mathcal{A}}
         \nu(\tilde a \mid s)\,
         q_{r_0-g,k_0}^{\nu,\gamma}(\tilde a,s).
\end{align*}
If \(\gamma>0\), then $v_{0,\nu,\gamma,g}(a,s)
    = \tfrac{1}{\gamma}\,\big(r_0(a,s) - g(s) - q_{r_0-g,k_0}^{\nu,\gamma}(a,s)\big),$ where \(v_{0,\nu,\gamma,g}\) solves the soft Bellman equation with reward \(r_{0,\nu,\gamma,g}\).
\end{theorem}

\noindent Thus, if \(r_0^\dagger\) satisfies \eqref{eqn::rdaggernorm}, it is
identified by \(r_{0,\nu,\gamma,g}\), which depends only on the observable
pseudo-reward \(r_0=\log\pi_0\) and the \(Q\)-function
\(q_{r_0-g,k_0}^{\nu,\gamma}\).

\section{Debiased inverse reinforcement learning}

\subsection{Statistical objective}
\label{sec::statobj}

The identification results in Section~\ref{sec:ident} express the
reward-dependent targets of interest as functionals of the log behavior policy
\(r_0=\log\pi_0\) and transition kernel \(k_0\).
Theorem~\ref{theorem::identvalue} identifies policy-value differences using
\(r_0\) as a pseudo-reward, while
Theorem~\ref{theorem::identWithNormalization} recovers normalized rewards from
\((r_0,k_0)\) through a Bellman equation.

We therefore study debiased inference for smooth functionals of
\((r_0,k_0)\). Let \(\mathcal H:=L^\infty(\lambda)\) denote the class of
bounded reward functions, and let \(\mathcal K\) denote the class of bounded
transition densities with respect to \(\mu\). We consider targets of the form
\[
\psi_0
:=
E_0\{m(S_0,r_0,k_0)\},
\]
where \((r,k)\mapsto m(\cdot,r,k)\) is a smooth map from
\(\mathcal H\times\mathcal K\) to \(L^2(\mu)\). Formally, \(\psi_0=\Psi(\mathbb P_0)\), where
\(\Psi:\mathfrak M_T\to\mathbb R\) is defined by
\[
\Psi(\mathbb P)
:=
E_{\mathbb P}\{m(S_0,r_{\mathbb P},k_{\mathbb P})\},
\qquad
r_{\mathbb P}(a,s)
:=
\log\pi_{\mathbb P}(a\mid s).
\]

\subsection{Motivating examples}
\label{subsec:examples}

Many counterfactual targets in IRL fall within this class, including
functionals defined directly through the pseudo-reward and those defined
through a normalized reward recovered from \((r_0,k_0)\). The examples below
cover fixed-policy values, interventions on behavior or transition laws, and
policy values under reward normalization.

\renewcommand{\theexample}{1a}
\begin{example}[Policy value]
\label{example::1a}
The policy value $\psi_0 := \mathbb E_0[V_{r_0,k_0}^{\pi,\gamma}(S_0)]$ of a
policy $\pi$ with discount factor $\gamma$ corresponds to the map
$m(s,r,k) = V_{r,k}^{\pi,\gamma}(s) = \pi \mathcal{T}_{k, \pi, \gamma}^{-1}(r)(s)$.
\qed
\end{example}

\noindent The same framework also covers interventions on the agent's
behavior or environment.

\renewcommand{\theexample}{1b}
\begin{example}[Value of a soft-optimal agent]
\label{example::1b}
 Consider the counterfactual softmax policy $\pi_{r_0,k_0}^\star$ defined in
\eqref{eqn::softmaxpolicycf}, obtained by optimizing the pseudo-reward $r_0$
under kernel $k_0$ with state-dependent action set $\mathcal{A}^\star(s)$ and
entropy parameter $\tau^\star$.
Unlike Example~\ref{example::1a}, this policy depends on both $r_0$ and $k_0$.
The policy value $\psi_0$ fits our framework through the map
$m(s,r,k) := V_{r,k}^\star(s)$, where \(V_{r,k}^\star\) denotes the value
function of \(\pi_{r,k}^\star\).
\qed
\end{example}

\noindent A closely related counterfactual keeps the pseudo-reward target fixed
but replaces the transition kernel by a known \(k^\star\), as in simulated or redesigned environments
\citep[e.g.][]{abbeel2004apprenticeship,ziebart2008maximum,wulfmeier2015deep,fu2018airl}.

\renewcommand{\theexample}{1c}
\begin{example}[Policy values in counterfactual environments]
The value of the soft-optimal agent that maximizes the pseudo-reward $r_0$ in an
environment with known transition kernel $k^\star$ corresponds to
$m(s,r,k) := V_{r,k^\star}^\star(s)$, where \(V_{r,k^\star}^\star\) denotes
the value function of \(\pi_{r,k^\star}^\star\).
\qed
\end{example}

When the true reward is known to satisfy a normalization constraint, targets
depending on \(r_0^\dagger\) can also be expressed as functionals of
\((r_0,k_0)\). In this case, Theorem~\ref{theorem::identWithNormalization}
identifies \(r_0^\dagger\) with the \(\nu\)-normalized reward
\(\bar r_0:=r_{0,\nu,\gamma,g}\).

\renewcommand{\theexample}{2}
\begin{example}[Policy values under reward normalization]
\label{example::norm}
 Suppose we are interested in
\(\psi_0 = \mathbb E_0[\tilde m(S_0,\bar r_0,k_0)]\),
a feature of the $\nu$-normalized reward \(\bar r_0\).
By Theorem~\ref{theorem::identWithNormalization}, this can be expressed as
$\psi_0 = \mathbb E_0[m(S_0,r_0,k_0)]$, where
\[
m(s,r,k)
:= \tilde m(s,r_{r,k,\nu,\gamma,g},k), \qquad
r_{r,k,\nu,\gamma,g}(a,s)
:= g(s) + q_{r-g,k}^{\nu,\gamma}(a,s) - V_{r-g,k}^{\nu,\gamma}(s),
 \]
and $r_{r,k,\nu,\gamma,g} = g + (I-\nu)\,\mathcal{T}_{k,\nu,\gamma}^{-1}(r-g)$ denotes
the $\nu$-normalized reward with anchor \(g\). 
\qed
\end{example}

Appendix~\ref{app::softmaxnorm} derives the EIF for the value of a softmax
policy induced by the normalized reward. Classical DDC methods often target
coefficients of parametric reward models, such as the replacement-cost
parameter in Rust's bus-replacement model under a zero-reward normalization
\citep{rust1987optimal}. Appendix~\ref{app::additionalapplications} also shows
that such parameters can be formulated as best linear projections of the
normalized reward within our framework.

\subsection{Efficient influence functions with latent rewards}
\label{sec::EIFsub}

We establish pathwise differentiability of \(\Psi\) and derive its
efficient influence function (EIF), which forms the basis for
one-step debiased estimation \citep{bickel1993efficient}.
The characterization involves two Riesz representers: one accounts
for the dependence of the pseudo-reward
\(r_{\mathbb P}=\log\pi_{\mathbb P}\) on the observed-data law,
while the other represents the transition-law derivative and is
given by standard off-policy evaluation theory.

To establish pathwise differentiability, we require that
\(\mathbb E_0[m(S_0,r,k)]\) be smooth in \(r\) and \(k\) along
directions in suitable tangent spaces. We equip \(\mathcal H\) and
\(\mathcal K\) with the reference-measure norms
\[
\|h\|_{\mathcal H}
:=\left(\int\sum_{a\in\mathcal A}h(a,s)^2\,\mu(ds)\right)^{1/2},
\qquad
\|k\|_{\mathcal K}
:=\left(\int k(s'\mid a,s)^2\,\mu(ds')\,\lambda(da,ds)\right)^{1/2}.
\]
Under the pooled transition law, define the inner products
\[
\langle h_1,h_2\rangle_{\mathcal H,0}
:= E_0[h_1(A,S)h_2(A,S)],
\qquad
\langle \beta_1,\beta_2\rangle_{\mathcal K,0}
:= E_0[\beta_1(S',A,S)\beta_2(S',A,S)].
\]
We define the \emph{reward tangent space} \(T_{\mathcal H}\) as the
completion of \(\mathcal H\) under the first inner product, and the
\emph{kernel tangent space at \(k\)} as
\[
T_{\mathcal K}(k)
:=
\Bigg\{
\beta\in L^2(P_0):
\int \beta(s',a,s)\,k(s'\mid a,s)\,\mu(ds')=0
\ \text{for $\lambda$-a.e. }(a,s)
\Bigg\},
\]
equipped with the second inner product. We write
\(T_{\mathcal K}:=T_{\mathcal K}(k_0)\).
Intuitively, \(T_{\mathcal K}\) is the \(L^2(P_0)\)-closure of
the directions \(\beta\) for which
\((1+t\beta(\cdot,a,s))k_0(\cdot\mid a,s)\) remains a conditional
density for sufficiently small \(t\).
We impose the following conditions.

\begin{enumerate}[label=\textbf{(C\arabic*)}, ref=C\arabic*, series=cond]
\item (\textit{Functional differentiability}) \label{cond::boundedfunc}
There exist linear maps
\(\partial_r m(\cdot,r_0,k_0):T_{\mathcal H}\to L^2(\mu)\) and
\(\partial_k m(\cdot,r_0,k_0):T_{\mathcal K}\to L^2(\mu)\) such that, for
every bounded \(h\in T_{\mathcal H}\) and bounded
\(b\in T_{\mathcal K}\),
\[
\left.
\frac{d}{dt}\,
\mathbb E_0\!\left[
m\bigl(S_0,r_0+th,(1+tb)k_0\bigr)
\right]
\right|_{t=0}
=
\mathbb E_0\!\left[
\partial_r m(S_0,r_0,k_0)(h)
+
\partial_k m(S_0,r_0,k_0)(b)
\right].
\]

\item (\textit{Continuity of derivatives}) \label{cond::boundedfunc2}
The linear functionals
\(h\mapsto\mathbb E_0[\partial_r m(S_0,r_0,k_0)(h)]\) and
\(b\mapsto\mathbb E_0[\partial_k m(S_0,r_0,k_0)(b)]\) are continuous:
\[
\sup_{\substack{h\in T_{\mathcal H}\\ \|h\|_{\mathcal H,0}\le1}}
\left|\mathbb E_0[\partial_r m(S_0,r_0,k_0)(h)]\right|<\infty,
\qquad
\sup_{\substack{b\in T_{\mathcal K}\\ \|b\|_{\mathcal K,0}\le1}}
\left|\mathbb E_0[\partial_k m(S_0,r_0,k_0)(b)]\right|<\infty.
\]

\item (\textit{Lipschitz continuity}) \label{cond::lipschitz}
For every finite \(M>\|r_0\|_{L^\infty(\lambda)}
+\|k_0\|_{L^\infty(\mu\otimes\lambda)}\), there exist \(\delta,L>0\)
such that any perturbations
\((h_j,b_j)\in T_{\mathcal H}\times T_{\mathcal K}\), \(j\in\{1,2\}\),
with \(r_0+h_j\in\mathcal H\), \((1+b_j)k_0\in\mathcal K\), and
\[
\|r_0+h_j\|_{L^\infty(\lambda)}
+\|(1+b_j)k_0\|_{L^\infty(\mu\otimes\lambda)}\le M,
\qquad
\|h_j\|_{\mathcal H,0}+\|b_j\|_{\mathcal K,0}<\delta
\]
satisfy
\[
\begin{aligned}
&\big\|m\{\cdot,r_0+h_1,(1+b_1)k_0\}
-m\{\cdot,r_0+h_2,(1+b_2)k_0\}\big\|_{L^2(\mu)}\\
&\qquad\le L\{\|h_1-h_2\|_{\mathcal H,0}
+\|b_1-b_2\|_{\mathcal K,0}\}.
\end{aligned}
\]
\end{enumerate}

Conditions \ref{cond::boundedfunc}--\ref{cond::lipschitz} hold if the
state-valued map \((r,k)\mapsto m(\cdot,r,k)\) admits a continuously Fréchet
differentiable extension to a neighborhood of \((r_0,k_0)\) as a map into
\(L^2(\mu)\), when its domain is equipped with the product norm
\(\|h\|_{\mathcal H,0}+\|b\|_{\mathcal K,0}\); see
Lemma~\ref{lemma::frechetimplieslip} in Appendix~\ref{appendix::EIFsub}.

Under the above conditions, there exist Riesz representers
\(\alpha_0\in T_{\mathcal H}\) and \(\beta_0\in T_{\mathcal K}\) for the
reward and kernel derivatives of
\((r,k)\mapsto \mathbb E_0[m(S_0,r,k)]\), characterized with respect to
\(\langle\cdot,\cdot\rangle_{\mathcal H,0}\) and
\(\langle\cdot,\cdot\rangle_{\mathcal K,0}\), respectively, by
\begin{align}
\mathbb E_0[\partial_r m(S_0,r_0,k_0)(h)]
&= \langle \alpha_0,h\rangle_{\mathcal H,0},
&& h\in T_{\mathcal H},
\label{eqn::rieszreward}
\\
\mathbb E_0[\partial_k m(S_0,r_0,k_0)(b)]
&= \langle \beta_0,b\rangle_{\mathcal K,0},
&& b\in T_{\mathcal K}.
\label{eqn::rieszkernel}
\end{align}

\begin{theorem}[Pathwise differentiability]  \label{theorem::EIF}
    Under \ref{cond::boundedfunc}-\ref{cond::lipschitz}, the parameter
    \(\Psi: \mathfrak M_T \to \mathbb{R}\) is pathwise differentiable at \(\mathbb P_0\).
    Its efficient influence function is
    \begin{align*}
    \chi_0(O)
    :={}&m(S_0,r_0,k_0)-\Psi(\mathbb P_0)\\
    &+\frac1T\sum_{t=0}^{T-1}\left\{
    \alpha_0(A_t,S_t)
    - \sum_{\tilde a \in \mathcal{A}} \pi_0(\tilde a \mid S_t)
      \alpha_0(\tilde a,S_t)
    + \beta_0(S_{t+1},A_t,S_t)\right\},
    \end{align*}
    where \(\alpha_0\) and \(\beta_0\) are the Riesz representers defined in
    \eqref{eqn::rieszreward}--\eqref{eqn::rieszkernel}.
\end{theorem}
\noindent
The term involving \(\alpha_0\) arises from the dependence of the
pseudo-reward \(r_{\mathbb P}=\log\pi_{\mathbb P}\) on the observed-data law
and is therefore specific to the latent-reward setting. By contrast, the term
involving \(\beta_0\) is the standard transition-law contribution and can be
recovered from the EIF for the corresponding observed-reward problem.
Specifically, consider the corresponding hypothetical observed-reward model
for a single transition (\(T=1\)), in which one additionally observes
\(R=r_0(A,S)+\varepsilon\), where
\(E_0(\varepsilon\mid S,A,S')=0\), and let
\(D_0(S,A,R,S')\) denote its EIF. Because the latent- and observed-reward
targets have the same dependence on the transition kernel, their pathwise
derivatives along \(T_{\mathcal K}\) coincide. Hence, \(\beta_0\) is the
orthogonal projection of \(D_0\) onto \(T_{\mathcal K}\), given by
\citep{laan2003unified,tsiatis2006semiparametric}
\[
\beta_0(S',A,S)
=
E_0\{D_0(S,A,R,S')\mid S,A,S'\}
-
E_0\{D_0(S,A,R,S')\mid S,A\}.
\]
Thus, existing EIF calculations for observed-reward policy-value functionals
can be used directly to obtain \(\beta_0\)
\citep{van2018online,kallus2020double,kallus2022efficiently}.
More generally, \citet{van2025automaticDRL} derives EIFs for continuous
linear functionals of the \(Q\)-function.

\subsection{EIFs for normalized-reward functionals}
\label{sec::eifnorm}

Suppose we have already derived the EIF of a functional of the pseudo-reward
$r_0$ using Theorem~\ref{theorem::EIF}, and we now seek the EIF of the
corresponding functional of the $\nu$-normalized reward
$r_{0,\nu,\gamma,g}$ from Theorem~\ref{theorem::identWithNormalization}.
The next result shows how the EIF for the normalized reward can be obtained
from that for the unnormalized reward by transforming the corresponding
representers.

Consider the functional
\(\psi_0=\mathbb E_0[\tilde m(S_0,r_{0,\nu,\gamma,g},k_0)]\) from
Example~\ref{example::norm}, corresponding to the mapping
\(m(s,r,k):=\tilde m(s,r_{r,k,\nu,\gamma,g},k)\). Let \(\tilde\alpha_0\) and \(\tilde\beta_0\) denote the Riesz representers,
defined analogously to \eqref{eqn::rieszreward} and
\eqref{eqn::rieszkernel}, of the reward and kernel derivatives of
\((r,k)\mapsto\mathbb E_0[\tilde m(S_0,r,k)]\) at
\((r_{0,\nu,\gamma,g},k_0)\). The following result assumes
Conditions~\ref{cond::boundedfunc::norm}--\ref{cond::lipschitz::norm} in
Appendix~\ref{appendix::conditionsEIFnorm}, the counterparts of
Conditions~\ref{cond::boundedfunc}--\ref{cond::lipschitz} for
\(\tilde m\) at \((r_{0,\nu,\gamma,g},k_0)\).

\begin{theorem}[Pathwise differentiability for normalized rewards]
\label{theorem::EIFnorm}
Suppose Assumption~\ref{cond::boundedpositivity} and
Conditions~\ref{cond::boundedfunc::norm}--\ref{cond::lipschitz::norm}
in Appendix~\ref{appendix::conditionsEIFnorm} hold. Then $\Psi$ is pathwise differentiable at $\mathbb P_0$ with EIF $\chi_0$
given in Theorem~\ref{theorem::EIF}, where $\alpha_0\in T_{\mathcal H}$ is
the unique function satisfying
\begin{equation}
E_0\!\left[\alpha_0(A,S)
\{h(A,S)-\gamma\nu h(S')\}\right]
=
E_0\!\left[\tilde\alpha_0(A,S)
\{h(A,S)-\nu h(S)\}\right]
\label{eqn::normalizedrieszbalance}
\end{equation}
for every $h\in T_{\mathcal H}$, and
\[
\beta_0(s',a,s)
=
\tilde\beta_0(s',a,s)
+\alpha_0(a,s)
\left\{
r_0(a,s)-g(s)
+\gamma V_{r_0-g,k_0}^{\nu,\gamma}(s')
-q_{r_0-g,k_0}^{\nu,\gamma}(a,s)
\right\}.
\]
\end{theorem}
 
Equation~\eqref{eqn::normalizedrieszbalance} has the form of a generalized
discounted occupancy-balance equation, corresponding to an adjoint form of the
Bellman equation \citep{imaizumi2021minimax,van2026fore}. When \(\gamma=0\),
it yields the explicit solution
\(\alpha_0(a,s)
=
\tilde\alpha_0(a,s)
-
\nu(a\mid s)(\pi_0\tilde\alpha_0)(s)/\pi_0(a\mid s)\).
For general \(\gamma\),
\(\alpha_0=\mathcal T_{k_0,\nu,\gamma}u_0\), where \(u_0\) is the unique
minimizer of
\(u\mapsto
E_0[
\{\mathcal T_{k_0,\nu,\gamma}u(A,S)\}^2
-2\tilde\alpha_0(A,S)\{u(A,S)-\nu u(S)\}]\).
This variational characterization permits direct estimation of
\(\alpha_0\) using Riesz or minimax estimation methods
\citep{dikkala2020minimax,van2025automaticDRL,bennett2026inference}.
Alternatively, the balance equation in
\eqref{eqn::normalizedrieszbalance} can be estimated using minimax balancing
methods of the type developed for occupancy-ratio estimation, such as
\textsc{DualDICE}
\citep{nachum2019dualdice,uehara2020minimax}.

\section{Automatic debiased estimation and inference}
\label{sec::est}

\subsection{General form of the estimator}

Let \(r_n\) and \(k_n\) estimate the pseudo-reward
\(r_0=\log\pi_0\) and transition kernel \(k_0\), and let
\(\alpha_n\) and \(\beta_n\) estimate the corresponding Riesz representers
\(\alpha_0\) and \(\beta_0\) from Section~\ref{sec::EIFsub}. We take
\(\beta_n\) to be centered with respect to \(k_n\), so that
\(\int \beta_n(s',a,s)k_n(s'\mid a,s)\,\mu(ds')=0\), which can always be
enforced post hoc. We define the one-step estimator
\begin{equation}
\psi_n
=
\frac{1}{n}\sum_{i=1}^n m(S_{i,0},r_n,k_n)
+
\frac{1}{nT}\sum_{i=1}^n\sum_{t=0}^{T-1}
\varphi_{r_n,k_n}(S_{i,t+1},A_{i,t},S_{i,t};
\alpha_n,\beta_n),
\label{eqn::estgeneral}
\end{equation}
where the influence-function correction is
\[
\varphi_{r,k}(s',a,s;\alpha,\beta)
:=
\alpha(a,s)
-
\sum_{\tilde a\in\mathcal A}
\exp\{r(\tilde a,s)\}\alpha(\tilde a,s)
+
\beta(s',a,s).
\]
The first term is the plug-in estimator, while the second is the
influence-function-based correction that removes its first-order bias
\citep{bickel1993efficient}.

The nuisance quantities in \eqref{eqn::estgeneral} can be estimated flexibly.
The pseudo-reward \(r_0=\log\pi_0\) can be learned from state--action data by
probabilistic classification or from trajectory data using adversarial IRL
methods \citep{fu2018airl,geng2020deep}. The transition kernel \(k_0\) can be
estimated using conditional density methods \citep{diaz2011super} or
generative transition models from model-based RL
\citep{chua2018deep,janner2019trust}. For policy-value parameters,
\(\beta_0\) is the usual transition-kernel component of the EIF from
off-policy evaluation and can therefore be estimated using existing DRL
methods. The reward representer \(\alpha_0\) can often be derived
analytically or estimated directly by \emph{Riesz regression}
\citep{chernozhukov2022automatic}, using the population risk $\alpha
\mapsto
E_0\{\alpha(A,S)^2\}
-
2E_0\{\partial_r m(S_0,r_0,k_0)(\alpha)\}.$  
Because the construction applies to any smooth functional \(m\) once its
directional derivatives are specified, the estimator in
\eqref{eqn::estgeneral} can be viewed as an \emph{automatic} DML estimator \citep{chernozhukov2022automatic,van2025automatic}.

When the target depends on value or \(Q\)-functions induced by \(k_0\), these
higher-level nuisances can often be estimated directly, avoiding explicit
estimation of the transition kernel. Section~\ref{sec::applications}
develops this reparameterization for the policy-value targets considered in
Section~\ref{subsec:examples}.

\subsection{Asymptotic theory}

We now establish asymptotic linearity of \(\psi_n\) with influence function
\(\chi_0\), and hence semiparametric efficiency for \(\psi_0\). The key step is
a von Mises expansion showing how the efficient influence function captures
the first-order error of plug-in estimators of \(\psi_0\).

Define  the linearization remainder
\[
\begin{aligned}
\mathrm{Rem}_0(r_0,k_0;\,r,k)
&:=
\mathbb E_0\{m(S_0,r,k)-m(S_0,r_0,k_0)\} \\
&\quad -
\mathbb E_0\!\left[
\partial_k m(S_0,r_0,k_0)\!\left(\frac{k-k_0}{k_0}\right)
+
\partial_r m(S_0,r_0,k_0)(r-r_0)
\right],
\end{aligned}
\]
which is typically second order; for example, $\mathrm{Rem}_0(r_0,k_0;\,r,k)
\lesssim
\|k-k_0\|_{\mathcal K}^{2}
+
\|r-r_0\|_{\mathcal H}^{2}.$

\begin{theorem}[von Mises expansion]
\label{theorem::vonmises}
Assume Conditions~\ref{cond::boundedfunc} and~\ref{cond::boundedfunc2}. Let
\(r\in\mathcal H\), \(k\in\mathcal K\),
\(\alpha\in T_{\mathcal H}\), and \(\beta\in T_{\mathcal K}(k)\), and suppose
$\|r\|_{L^\infty(\lambda)}
+
\|r_0\|_{L^\infty(\lambda)}
+
\|\alpha\|_{L^\infty(\lambda)}
<M$
for some \(M\in(0,\infty)\). Then
\begin{align*}
&\mathbb E_0\!\big[m(S_0,r,k)\big]
-\mathbb E_0\!\big[m(S_0,r_0,k_0)\big]
+\int \varphi_{r,k}(s',a,s;\alpha,\beta)\,
P_0(ds,da,ds') \\
&\qquad
=
-\langle r-r_0,\alpha-\alpha_0\rangle_{\mathcal H,0}
+
O\!\big(\|r-r_0\|_{\mathcal H,0}^{2}\big) \\
&\qquad\quad
+
O\!\left(
\|\beta_0-\beta\|_{L^2(P_0)}
\Big\|\tfrac{k-k_0}{k_0}\Big\|_{L^2(P_0)}
\right)
+
\mathrm{Rem}_0(r_0,k_0;\,r,k),
\end{align*}
where the implicit constants depend only on \(M\).
\end{theorem}

For the asymptotic result, we impose the following rate and regularity
conditions.

\begin{enumerate}[label=\textbf{(C\arabic*)}, ref=C\arabic*, resume=cond]
\item (\textit{Sample splitting and nuisance range})
\label{cond::samplesplit} \label{cond::bounded}
The estimators \(r_n\), \(k_n\), \(\alpha_n\), and \(\beta_n\) are obtained
from trajectories independent of \(\mathcal D_n\), and
\(\chi_0\in L^2(\mathbb P_0)\). For some fixed \(M<\infty\),
\(\|r_n\|_{L^\infty(\lambda)}+\|\alpha_n\|_{L^\infty(\lambda)}<M\)
with probability tending to one.

\item (\textit{Nuisance consistency and estimation rates})
\label{cond::rates}
\begin{enumerate}[label=(\alph*), ref=\ref{cond::rates}(\alph*)]
\item \label{cond::consistency}
\(\|m(S_0,r_n,k_n)-m(S_0,r_0,k_0)\|_{L^2(\mathbb P_0)}
+\|\varphi_{r_n,k_n}(\cdot;\alpha_n,\beta_n)
-\varphi_{r_0,k_0}(\cdot;\alpha_0,\beta_0)\|_{L^2(P_0)}
=o_P(1)\).

\item \label{cond::reward} \label{cond::representers}
\(\|r_n-r_0\|_{\mathcal H}=o_P(n^{-1/4})\) and
\(\|r_n-r_0\|_{\mathcal H,0}
\|\alpha_n-\alpha_0\|_{\mathcal H,0}=o_P(n^{-1/2})\).

\item \label{cond::beta} \label{cond::remainder}
\(\big\|(k_n-k_0)/k_0\big\|_{L^2(P_0)}
\|\beta_n-\beta_0\|_{L^2(P_0)}=o_P(n^{-1/2})\) and
\(\operatorname{Rem}_0(r_0,k_0;\,r_n,k_n)=o_P(n^{-1/2})\).
\end{enumerate}
\end{enumerate}

\begin{theorem}[Asymptotic linearity and efficiency]
\label{thm:asymptotic_linearity}
Suppose Assumption~\ref{cond::boundedpositivity} and
Conditions~\ref{cond::boundedfunc}--\ref{cond::rates} hold.
Then the autoDML estimator \(\psi_n\) defined in
\eqref{eqn::estgeneral} satisfies
\[
\psi_n-\psi_0
=
\frac{1}{n}\sum_{i=1}^n\chi_0(O_i)
+o_P(n^{-1/2}).
\]
Consequently,
\(\sqrt{n}(\psi_n-\psi_0)\rightsquigarrow N(0,\sigma_0^2)\), where
\(\sigma_0^2=\operatorname{Var}_{\mathbb P_0}\{\chi_0(O)\}\).
Thus, \(\psi_n\) attains the semiparametric efficiency bound for
\(\psi_0\) in \(\mathfrak M_T\).
\end{theorem}

\noindent
These conditions are standard for debiased machine learning.
Condition~\ref{cond::samplesplit} separates nuisance estimation from
evaluation and can be replaced by cross-fitting over whole trajectories
\citep{van2011targeted,chernozhukov2018double}. Its remaining requirements
ensure finite variance and uniformly bound the estimated reward and reward
representer. Condition~\ref{cond::consistency} requires \(L^2\) consistency
of the plug-in term and EIF correction; it follows from consistency of
\(r_n,k_n,\alpha_n,\beta_n\) when the corresponding maps into
\(L^2(\mathbb P_0)\) and \(L^2(P_0)\) are locally continuous.
Condition~\ref{cond::reward} requires sufficiently fast estimation of the
pseudo-reward together with a product-rate condition for its Riesz
representer, while Condition~\ref{cond::beta} imposes the analogous
product-rate condition for the transition kernel and its representer.
Condition~\ref{cond::remainder} requires the remaining von Mises remainder
to be \(o_P(n^{-1/2})\). The product-rate conditions allow slower convergence
of either representer to be offset by faster convergence of its corresponding
nuisance estimator.

\section{Applications}
\label{sec::applications}

With the general characterization of the efficient influence function and
estimator in hand, we now illustrate the scope of the theory for the policy
value parameters introduced in Examples~\ref{example::1a},
\ref{example::1b}, and~\ref{example::norm} of
Section~\ref{subsec:examples}. For each, we derive the efficient influence
function, construct the corresponding estimator, and give sufficient conditions
for asymptotic linearity. The examples concern, respectively, the value of a
fixed policy, the value of a counterfactual soft-optimal policy, and the value
of a fixed policy under reward normalization. Additional applications are
provided in Appendix~\ref{app::additionalapplications}.

\subsection{Efficient influence functions for policy values}
\label{section::EIFexamples}

We first consider the value of a fixed policy \(\pi\), as defined in
Example~\ref{example::1a}. Its EIF involves the unnormalized discounted
state--action occupancy ratio
\begin{equation}
d_0^{\pi,\gamma}(a,s)
:=
\rho_0^{\pi,\gamma}(s)\,
\frac{\pi(a\mid s)}{\pi_0(a\mid s)},
\qquad
\rho_0^{\pi,\gamma}(s)
:=
\frac{1}{\rho_0(s)}
\sum_{t=0}^\infty \gamma^t
\frac{d\mathbb P_{k_0,\pi}(S_t\in\cdot)}{d\mu}(s),
\label{eqn::occupancy}
\end{equation}
where the target trajectory is initialized from the same initial-state
distribution as \(\mathbb P_0\), and \(\rho_0^{\pi,\gamma}\) is the
discounted state-occupancy ratio relative to the pooled design density.  
The state--action occupancy ratio is the reward-side Riesz representer for the
policy value, since
\(\mathbb E_0\{V_{r_0,k_0}^{\pi,\gamma}(S_0)\}
=
E_0\{d_0^{\pi,\gamma}(A,S)r_0(A,S)\}\).
It therefore plays a role in offline policy evaluation analogous to
inverse-propensity weighting in causal inference, reweighting the observed
state--action distribution toward that induced by the target policy.

\begingroup

\begin{theorem}[EIF for known policy]
\label{theorem::EIFvalue}

   Under Assumption~\ref{cond::boundedpositivity}, the parameter
   $\Psi:\mathbb P\mapsto\mathbb E_{\mathbb P}
   [V_{r_{\mathbb P},k_{\mathbb P}}^{\pi,\gamma}(S_0)]$ is pathwise
   differentiable at $\mathbb P_0$ with efficient influence function
    \begin{align*}
    \chi_0(O)
    &=V_{r_0,k_0}^{\pi,\gamma}(S_0)-\Psi(\mathbb P_0)\\
    &\quad+\frac1T\sum_{t=0}^{T-1}\left[
    \rho_0^{\pi,\gamma}(S_t)
    \left\{\frac{\pi(A_t\mid S_t)}{\pi_0(A_t\mid S_t)}-1\right\}
    \right.\\
    &\hspace{33mm}\left.
    +d_0^{\pi,\gamma}(A_t,S_t)
    \{r_0(A_t,S_t)+\gamma V_{r_0,k_0}^{\pi,\gamma}(S_{t+1})
    -q_{r_0,k_0}^{\pi,\gamma}(A_t,S_t)\}\right].
    \end{align*}
\end{theorem}
\endgroup

\noindent In Theorem~\ref{theorem::EIFvalue}, the representers from
Theorem~\ref{theorem::EIF} take the explicit forms
\(\alpha_0(a,s)=\rho_0^{\pi,\gamma}(s)\pi(a\mid s)/\pi_0(a\mid s)\) and
\(\beta_0(s',a,s)=d_0^{\pi,\gamma}(a,s)
\{r_0(a,s)+\gamma V_{r_0,k_0}^{\pi,\gamma}(s')
-q_{r_0,k_0}^{\pi,\gamma}(a,s)\}\). Thus, the EIF combines a density-ratio
term from the pseudo-reward with the familiar temporal-difference term from
off-policy evaluation
\citep{van2018online,kallus2020double,kallus2022efficiently,van2025automaticDRL}.

We next consider Example~\ref{example::1b}, concerning the value of the
counterfactual softmax policy \(\pi_{r_0,k_0}^\star\). Define the soft state
value function
\[
\Phi_{r_0,k_0}^\star(s)
:=
\tau^\star \log\!\Biggl(
\sum_{\tilde a\in\mathcal A^\star(s)}
\exp\!\Bigl(
\tau^{\star-1}
\{r_0(\tilde a,s)+\gamma v_{r_0,k_0}^\star(\tilde a,s)\}
\Bigr)
\Biggr).
\]
We write
\(\rho_{r_0,k_0}^\star:=\rho_0^{\pi_{r_0,k_0}^\star,\gamma}\)
for the discounted state-occupancy ratio associated with
\(\pi_{r_0,k_0}^\star\), and define the corresponding state--action occupancy
ratio by $d_{r_0,k_0}^\star(a,s)
:=
\rho_{r_0,k_0}^\star(s)
 \pi_{r_0,k_0}^\star(a\mid s) / \pi_0(a\mid s).$
Finally, define the signed occupancy ratio
\(\widetilde d_{r_0,k_0}^\star\) through the following advantage-weighted
balance equation, which holds for all \(h\in T_{\mathcal H}\):
\begin{equation}
\label{eqn::tildedbalance}
E_0\!\left[\widetilde d_{r_0,k_0}^\star(A,S)
\{h(A,S)-\gamma\pi_{r_0,k_0}^\star h(S')\}\right]
=
E_0\!\left[d_{r_0,k_0}^\star(A,S)
\{q_{r_0,k_0}^\star(A,S)-V_{r_0,k_0}^\star(S)\}h(A,S)\right].
\end{equation}

\begingroup

\begin{theorem}[EIF for softmax policy]
\label{theorem::EIFsoftmax}
Under Assumption~\ref{cond::boundedpositivity}, the parameter
$\Psi:\mathbb P\mapsto\mathbb E_{\mathbb P}
[V_{r_{\mathbb P},k_{\mathbb P}}^\star(S_0)]$ is pathwise differentiable at
$\mathbb P_0$ with efficient influence function
\begin{align*}
   \chi_0(O)
&=V_{r_0,k_0}^\star(S_0)-\Psi(\mathbb P_0)\\
&\quad+\frac1T\sum_{t=0}^{T-1}\Bigg[
\rho_{r_0,k_0}^{\star}(S_t)
\left\{\frac{\pi_{r_0,k_0}^\star(A_t\mid S_t)}{\pi_0(A_t\mid S_t)} - 1\right\}\\
&\hspace{12mm}+\frac{1}{\tau^\star}\left\{
\widetilde d_{r_0,k_0}^\star(A_t,S_t)
-\sum_{a\in\mathcal A}\pi_0(a\mid S_t)
\widetilde d_{r_0,k_0}^\star(a,S_t)\right\}\\
&\hspace{12mm}+\frac{\gamma}{\tau^\star}
\widetilde{d}_{r_0,k_0}^\star(A_t,S_t)
\{\Phi_{r_0,k_0}^\star(S_{t+1})-v_{r_0,k_0}^\star(A_t,S_t)\}\\
&\hspace{12mm}+d_{r_0,k_0}^\star(A_t,S_t)
\{r_0(A_t,S_t)+\gamma V_{r_0,k_0}^{\star}(S_{t+1})
-q_{r_0,k_0}^\star(A_t,S_t)\}\Bigg].
\end{align*}
\end{theorem}
\endgroup
\noindent Relative to Theorem~\ref{theorem::EIFvalue}, this EIF contains an
additional term involving the signed occupancy ratio
\(\widetilde d_{r_0,k_0}^\star\), which accounts for the dependence of the
target policy on the soft Bellman solution \(v_{r_0,k_0}^\star\).

We now turn to Example~\ref{example::norm}. Let
\(\gamma'\in[0,1)\) denote the discount factor used to evaluate the target
policy \(\pi\), which may differ from the behavioral discount factor
\(\gamma\). Recall the normalized reward
\[
r_{0,\nu,\gamma,g}(a,s)
:=
g(s)+q_{r_0-g,k_0}^{\nu,\gamma}(a,s)
-V_{r_0-g,k_0}^{\nu,\gamma}(s),
\]
and write
\(q_{0,\nu}^{\pi,\gamma'}
:=q_{r_{0,\nu,\gamma,g},k_0}^{\pi,\gamma'}\) and
\(V_{0,\nu}^{\pi,\gamma'}:=\pi q_{0,\nu}^{\pi,\gamma'}\).
Let \(\alpha_0^{\pi,\nu}\) denote the unique element satisfying, for every
\(h\in T_{\mathcal H}\),
\begin{equation}
E_0\!\left[\alpha_0^{\pi,\nu}(A,S)
\{h(A,S)-\gamma\nu h(S')\}\right]
=
E_0\!\left[d_0^{\pi,\gamma'}(A,S)
\{h(A,S)-\nu h(S)\}\right].
\label{eqn::policyvaluenormbalance}
\end{equation}
Specializing the construction of Theorem~\ref{theorem::EIFnorm} to the
unnormalized setting of Theorem~\ref{theorem::EIFvalue} yields the following
result.

\begingroup
\begin{theorem}[EIF for policy value under reward normalization]
\label{theorem::EIFvaluenorm}
Under Assumption~\ref{cond::boundedpositivity} and
Conditions~\ref{cond::boundedfunc::norm}--\ref{cond::lipschitz::norm}, the parameter $\Psi:\mathbb P\mapsto
\mathbb E_{\mathbb P}[m(S_0,r_{\mathbb P},k_{\mathbb P})]$ is pathwise differentiable at \(\mathbb P_0\), with efficient influence function
\begin{align*}
\chi_0(O)
&=V_{r_0,k_0,\nu}^{\pi,\gamma'}(S_0)-\Psi(\mathbb P_0)\\
&\quad+\frac1T\sum_{t=0}^{T-1}\left[
\alpha_0^{\pi,\nu}(A_t,S_t)
-\sum_{\tilde a\in\mathcal A}\pi_0(\tilde a\mid S_t)
 \alpha_0^{\pi,\nu}(\tilde a,S_t)\right.\\
&\hspace{12mm}+\alpha_0^{\pi,\nu}(A_t,S_t)
\{r_0(A_t,S_t)-g(S_t)+\gamma V_{r_0-g,k_0}^{\nu,\gamma}(S_{t+1})
-q_{r_0-g,k_0}^{\nu,\gamma}(A_t,S_t)\}\\
&\hspace{12mm}\left.+d_0^{\pi,\gamma'}(A_t,S_t)
\{r_{0,\nu,\gamma,g}(A_t,S_t)
+\gamma' V_{r_0,k_0,\nu}^{\pi,\gamma'}(S_{t+1})
-q_{r_0,k_0,\nu}^{\pi,\gamma'}(A_t,S_t)\}\right].
\end{align*}
\end{theorem}

\endgroup

\subsection{Estimators and asymptotic theory}
\label{sec::examplesestimators}

Building on the EIF characterizations above, we now construct tailored DML
estimators that directly estimate the nuisance functions appearing in the EIF,
thereby avoiding explicit estimation of the transition kernel \(k_0\) required
by the general estimator in Theorem~\ref{thm:asymptotic_linearity}. This is
possible because, for these policy-value targets, the second-order remainder in
the von Mises expansion of Theorem~\ref{theorem::vonmises} admits additional
structure that can be exploited directly.

We revisit Example~\ref{example::1a}. Let \(r_n\) estimate \(r_0\) and define \(\pi_n:=\exp\{r_n\}\). Let
\(q_n^{\pi,\gamma}\) estimate \(q_{r_0,k_0}^{\pi,\gamma}\), and define
\(V_n^{\pi,\gamma}(s):=\sum_{\tilde a\in\mathcal A}\pi(\tilde a\mid s)
q_n^{\pi,\gamma}(\tilde a,s)\). Let \(\rho_n^{\pi,\gamma}\) estimate the state
occupancy ratio \(\rho_0^{\pi,\gamma}\). Our DML estimator of
\(\psi_0=\mathbb E_0[V_{r_0,k_0}^{\pi,\gamma}(S_0)]\) is
\begin{align*}
\psi_n
&:= \frac{1}{n}\sum_{i=1}^n V_n^{\pi,\gamma}(S_{i,0})\\
&\quad+P_n\!\left[\rho_n^{\pi,\gamma}(S)
\frac{\pi(A\mid S)}{\pi_n(A\mid S)}
\left\{r_n(A,S)+\gamma V_n^{\pi,\gamma}(S')-q_n^{\pi,\gamma}(A,S)\right\}\right]\\
&\quad+P_n\!\left[\rho_n^{\pi,\gamma}(S)
\left\{\frac{\pi(A\mid S)}{\pi_n(A\mid S)}-1\right\}\right].
\end{align*}
The nuisance \(q_{r_0,k_0}^{\pi,\gamma}\) can be estimated by fitted
\(Q\)-evaluation, i.e., iterative regression on Bellman targets
\citep{ernst2005tree,munos2005error,riedmiller2005neural,munos2008finite,mnih2013playing,van2025fitted}.
The ratio \(\rho_0^{\pi,\gamma}\) can be estimated using direct
density-ratio methods based on moment conditions or minimax formulations, as
well as fitted occupancy-ratio methods
\citep{liu2018breaking,uehara2020minimax,nachum2019dualdice,wang2023projected,van2025automaticDRL,van2026fore}.

The next theorem establishes that \(\psi_n\) is asymptotically
linear and efficient.

\begin{enumerate}[label=\textbf{(E\arabic*)}, ref=E\arabic*, series = efficiency1a]

    \item \label{assump:rates::policy} (\textit{Nuisance rates})
    $\bigl\{\|\rho_n^{\pi,\gamma} - \rho_0^{\pi,\gamma}\|_{L^2(\rho_0)}
      + \|r_n - r_0\|_{\mathcal{H},0}\bigr\}\,
    \bigl\|
        \mathcal{T}_{k_0,\pi,\gamma}
        (q_n^{\pi,\gamma} - q_{r_0,k_0}^{\pi,\gamma})
    \bigr\|_{\mathcal{H},0}
    = o_p(n^{-1/2})$.
\end{enumerate}

\begin{theorem}[Asymptotic linearity for Example~\ref{example::1a}]
\label{theorem::ALex1}
Suppose Assumptions~\ref{cond::boundedpositivity} and~\ref{cond::reward} and
Conditions~\ref{assump:rates::policy}--\ref{assump:consistency::policy} hold.
Then Theorem~\ref{thm:asymptotic_linearity} holds with influence function
$\chi_0$ given in Theorem~\ref{theorem::EIFvalue}.
\end{theorem}

For Example~\ref{example::1b}, let
\(v_n^\star\) estimate the soft Bellman fixed point
\(v_{r_0,k_0}^\star\), and define $\pi_n^\star(a\mid s)
\propto
1\{a\in\mathcal A^\star(s)\}
\exp\!\left\{
\tfrac{1}{\tau^\star}\{r_n(a,s)+\gamma v_n^\star(a,s)\}
\right\}.$ Let \(q_n^\star\) estimate \(q_{r_0,k_0}^\star\), set $V_n^\star(s)
:=
\sum_{\tilde a\in\mathcal A^\star(s)}
\pi_n^\star(\tilde a\mid s)q_n^\star(\tilde a,s),$
and let \(\widetilde d_n^\star\) and \(\rho_n^\star\) estimate
\(\widetilde d_{r_0,k_0}^\star\) and \(\rho_{r_0,k_0}^\star\). Define
\[
\Phi_n^\star(s):=\tau^\star\log\!\left[\sum_{a\in\mathcal A^\star(s)}
\exp\{\{r_n(a,s)+\gamma v_n^\star(a,s)\}/\tau^\star\}\right].
\]
The resulting
estimator of \(\psi_0=\mathbb E_0\{V_{r_0,k_0}^\star(S_0)\}\) is
\begin{align*}
\psi_n
&:=\frac{1}{n}\sum_{i=1}^n V_n^\star(S_{i,0})\\
&\quad+P_n\!\Bigg[
\rho_n^\star(S)\left\{
\frac{\pi_n^\star(A\mid S)}{\pi_n(A\mid S)}
\left(r_n(A,S)+\gamma V_n^\star(S')-q_n^\star(A,S)\right)
+\frac{\pi_n^\star(A\mid S)}{\pi_n(A\mid S)}-1\right\}\\
&\hspace{35mm}+\frac1{\tau^\star}\left\{
\gamma\widetilde d_n^\star(A,S)
\left(\Phi_n^\star(S')-v_n^\star(A,S)\right)
+\widetilde d_n^\star(A,S)-(\pi_n\widetilde d_n^\star)(S)\right\}\Bigg].
\end{align*}
The signed ratio \(\widetilde d_n^\star\) can be estimated directly from
\eqref{eqn::tildedbalance} using balance-based methods such as DualDICE or
minimax weight learning \citep{nachum2019dualdice,uehara2020minimax}.
Alternatively, one can express the signed ratio as the difference of two
nonnegative occupancy ratios and estimate each component using fitted occupancy
ratio evaluation \citep{van2026fore}.

\begin{enumerate}[label=\textbf{(E\arabic**)}, ref=E\arabic**, series=efficiency1b]
\item \label{assump:rates::softstar}(\textit{Nuisance rates})
The following rates hold:
\begin{enumerate}[label=(\alph*)]
    \item \(\|v_n^\star - v_{r_0,k_0}^\star\|_{\mathcal{H},0}
    = o_p(n^{-1/4})\).

    \item \(\|\widetilde d_n^\star - \widetilde d_{r_0,k_0}^\star\|_{\mathcal{H},0}\,
    \bigl(\|r_n - r_0\|_{\mathcal{H},0} + \|v_n^\star - v_{r_0,k_0}^\star\|_{\mathcal{H},0}\bigr)
    = o_p(n^{-1/2})\).

    \item \(\bigl(\|\rho_n^\star - \rho_{r_0,k_0}^\star\|_{L^2(\rho_0)}
      + \|r_n - r_0\|_{\mathcal{H},0}
      + \|v_n^\star - v_{r_0,k_0}^\star\|_{\mathcal{H},0}\bigr)\,
      \bigl\|\mathcal{T}_{r_0,k_0}^\star(q_n^\star - q_{r_0,k_0}^\star)\bigr\|_{\mathcal{H},0}
      = o_p(n^{-1/2})\).

    \item \label{assump:rho-policy-product-softstar}
      \(\|\rho_n^\star-\rho_{r_0,k_0}^\star\|_{L^2(\rho_0)}
      \bigl(\|r_n-r_0\|_{\mathcal H,0}
      +\|v_n^\star-v_{r_0,k_0}^\star\|_{\mathcal H,0}\bigr)
      =o_p(n^{-1/2})\).
\end{enumerate}
\end{enumerate}

\begin{theorem}[Asymptotic linearity for Example~\ref{example::1b}]
\label{theorem::ALex1b}
Suppose Assumptions~\ref{cond::boundedpositivity} and~\ref{cond::reward} and
Conditions~\ref{assump:rates::softstar}--\ref{assump:consistency::softstar} hold.
Then Theorem~\ref{thm:asymptotic_linearity} holds with influence function
$\chi_0$ given in Theorem~\ref{theorem::EIFsoftmax}.
\end{theorem}

 For Example~\ref{example::norm}, write
\(q_0^{\nu,\gamma}:=q_{r_0-g,k_0}^{\nu,\gamma}\) and
\(q_{0,\nu}^{\pi,\gamma'}:=q_{r_0,k_0,\nu}^{\pi,\gamma'}\). Let
\(\pi_n:=\exp\{r_n\}\), let \(q_n^{\nu,\gamma}\) estimate
\(q_0^{\nu,\gamma}\), and define
\(V_n^{\nu,\gamma}(s):=\sum_{\tilde a\in\mathcal A}\nu(\tilde a\mid s)\,
q_n^{\nu,\gamma}(\tilde a,s)\) and
\(r_{n,\nu,\gamma,g}(a,s):=g(s)+q_n^{\nu,\gamma}(a,s)-V_n^{\nu,\gamma}(s)\).
Let \(q_{n,\nu}^{\pi,\gamma'}\) estimate
\(q_{0,\nu}^{\pi,\gamma'}\), set
\(V_{n,\nu}^{\pi,\gamma'}(s):=\sum_{\tilde a\in\mathcal A}\pi(\tilde a\mid s)\,
q_{n,\nu}^{\pi,\gamma'}(\tilde a,s)\). Let
\(d_n^{\pi,\gamma'}\) estimate \(d_0^{\pi,\gamma'}\), and let
\(\alpha_n^{\pi,\nu}\) estimate the solution of
\eqref{eqn::policyvaluenormbalance}. The corresponding estimator of
\(\psi_0=\mathbb E_0[V_{r_0,k_0,\nu}^{\pi,\gamma'}(S_0)]\) is
\begin{align*}
\psi_n
&:=
\frac1n\sum_{i=1}^n V_{n,\nu}^{\pi,\gamma'}(S_{i,0})
+P_n\!\Big[
\alpha_n^{\pi,\nu}(A,S)
-(\pi_n\alpha_n^{\pi,\nu})(S)
\Big]\\
&\quad
+P_n\!\Bigg[
d_n^{\pi,\gamma'}(A,S)
\left\{
r_{n,\nu,\gamma,g}(A,S)
+\gamma'V_{n,\nu}^{\pi,\gamma'}(S')
-q_{n,\nu}^{\pi,\gamma'}(A,S)
\right\}\\
&\hspace{31mm}
+\alpha_n^{\pi,\nu}(A,S)
\left\{
r_n(A,S)-g(S)
+\gamma V_n^{\nu,\gamma}(S')
-q_n^{\nu,\gamma}(A,S)
\right\}
\Bigg].
\end{align*}

\begin{enumerate}[label=\textbf{(E\arabic***)}, ref=E\arabic***, series=efficiency2]
    \item \label{assump:rates::policynorm} (\textit{Nuisance rates})
    The following rate conditions hold:
    \begin{enumerate}[label=(\alph*)]
        \item
        \(\|d_n^{\pi,\gamma'}-d_0^{\pi,\gamma'}\|_{\mathcal{H},0}\,
        \bigl\|\mathcal{T}_{k_0,\pi,\gamma'}
        (q_{n,\nu}^{\pi,\gamma'}-q_{0,\nu}^{\pi,\gamma'})\bigr\|_{\mathcal{H},0}
        =o_p(n^{-1/2})\).
        \item
        \(\bigl\{
        \|d_n^{\pi,\gamma'}-d_0^{\pi,\gamma'}\|_{\mathcal{H},0}
        +\|\alpha_n^{\pi,\nu}-\alpha_0^{\pi,\nu}\|_{\mathcal{H},0}
        \bigr\}
        \bigl\|\mathcal{T}_{k_0,\nu,\gamma}
        (q_n^{\nu,\gamma}-q_0^{\nu,\gamma})\bigr\|_{\mathcal{H},0}
        =o_p(n^{-1/2})\).
    \end{enumerate}
\end{enumerate}

\begin{theorem}[Asymptotic linearity for Example~\ref{example::norm}]
\label{theorem::ALex2}
Suppose Assumptions~\ref{cond::boundedpositivity} and~\ref{cond::reward} and
Conditions~\ref{assump:rates::policynorm}--\ref{assump:consistency::policynorm} hold.
Then Theorem~\ref{thm:asymptotic_linearity} holds with influence function
\(\chi_0\) given in Theorem~\ref{theorem::EIFvaluenorm}.
\end{theorem}

\section{Numerical experiments and empirical applications}

\subsection{Simulation study}
\label{sec:simulation}

The simulations assess whether the proposed D-IRL estimators and Wald
confidence intervals exhibit the behavior predicted by the asymptotic theory.
We compare D-IRL with the corresponding plug-in estimator for two targets:
the value of a fixed policy (Example~\ref{example::1a}) and the value of a
counterfactual soft-optimal policy (Example~\ref{example::1b}).
Both designs use a continuous-state MDP with two state variables, four
actions, and nonlinear Gaussian state dynamics, with
\(n\in\{2500,5000,10000\}\). We estimate the behavior policy, Bellman functions, and occupancy ratios using
flexible sieve-based methods with five-fold cross-fitting. For
Example~\ref{example::1b}, we additionally estimate the counterfactual policy
and the signed occupancy ratio required by the adaptive-policy correction.
Complete details of the data-generating processes, nuisance estimators, and
implementation are provided in Appendix~\ref{app::normsim}.

\begin{table}[H]
\centering
\caption{Monte Carlo results for Examples~\ref{example::1a} and~\ref{example::1b}. Plug-in and D-IRL
are summarized by bias, Monte Carlo standard deviation, and RMSE. Mean standard
error and coverage of the nominal \(95\%\) Wald interval are reported for D-IRL.
Each row is based on \(300\) replications.}
\label{tab:simulation-main}
\begingroup
\scriptsize
\setlength{\tabcolsep}{3.2pt}
\begin{tabular}{llrrrrrrrr}
\toprule
& & \multicolumn{3}{c}{Plug-in} & \multicolumn{5}{c}{D-IRL} \\
\cmidrule(lr){3-5}\cmidrule(lr){6-10}
Design & \(n\) & Bias & SD & RMSE & Bias & SD & RMSE & Mean SE & Coverage \\
\midrule
1a & 2500 & 0.046 & 0.111 & 0.120 & 0.036 & 0.112 & 0.118 & 0.112 & 0.937 \\
1a & 5000 & 0.075 & 0.080 & 0.110 & 0.025 & 0.080 & 0.084 & 0.076 & 0.913 \\
1a & 10000 & 0.082 & 0.048 & 0.095 & 0.014 & 0.049 & 0.051 & 0.053 & 0.970 \\
1b & 2500 & 0.115 & 0.060 & 0.130 & -0.028 & 0.067 & 0.072 & 0.061 & 0.900 \\
1b & 5000 & 0.083 & 0.043 & 0.093 & -0.013 & 0.045 & 0.047 & 0.041 & 0.913 \\
1b & 10000 & 0.064 & 0.029 & 0.070 & -0.009 & 0.029 & 0.030 & 0.029 & 0.943 \\
\bottomrule
\end{tabular}
\endgroup
\end{table}

Table~\ref{tab:simulation-main} reports the Monte Carlo results. In
Examples~\ref{example::1a} and~\ref{example::1b}, D-IRL reduces the absolute
plug-in bias at every sample size, with the improvement becoming especially
pronounced at larger \(n\). At \(n=10{,}000\), the bias decreases from
\(0.082\) to \(0.014\) for the fixed-policy target and from \(0.064\) to
\(0.009\) for the counterfactual soft-optimal target. Mean estimated standard
errors closely track the corresponding Monte Carlo standard deviations.
Coverage is somewhat below nominal in smaller samples and is closer to nominal
at \(n=10{,}000\), reaching \(0.970\) and \(0.943\), respectively.

\subsection{Empirical applications}
\label{sec:empirical-applications}

We apply D-IRL to two classical dynamic discrete choice settings. In each
application, an intervention removes one action from the feasible set and the
decision-maker reoptimizes over the remaining actions under the same softmax
behavioral model. For each discount factor \(\gamma\), our primary estimand is
the restricted-minus-unrestricted policy-value contrast
\[
\Delta(\gamma)
:=
E_0\!\left[
V_{r_0,k_0}^{\pi^\star,\gamma}(S_0)
-
V_{r_0,k_0}^{\pi_0,\gamma}(S_0)
\right],
\]
which is identified without reward normalization by
Theorem~\ref{theorem::identvalue}. We also estimate the corresponding changes
in expected discounted physician visits and births.

The first application revisits the medical-care and work-absence model of
\citet{gilleskie1998dynamic}. We reconstruct 731 acute illness episodes,
comprising 5,356 person-days, from the 1987 National Medical Expenditure
Survey. On each illness day, an employed individual chooses whether to work
and whether to visit a physician.  We study Gilleskie's counterfactual
restriction of physician visits during the first three illness days. 
For Gilleskie, we additionally impose the work/no-visit normalization
\(\bar r_\gamma:=r_{0,\nu,\gamma,0}\), where \(\nu\) always chooses work
without a physician visit, and estimate the normalized policy values
\(J_{\mathrm{norm}}^\pi(\gamma)
:=E_0[V_{\bar r_\gamma,k_0}^{\pi,\gamma}(S_0)]\) for
\(\pi\in\{\pi_0,\pi^\star\}\). Under this normalization,
\(J_{\mathrm{norm}}^\nu(\gamma)=0\).

The second application revisits the contraceptive-choice model of
\citet{hotz1993conditional}. We link the 1970 and 1975 National Fertility
Surveys and reconstruct 2,104 couples and 10,441 at-risk couple-years. The
annual actions are no contraception, temporary contraception, and
sterilization. We study a counterfactual that removes sterilization.

 In both applications, we retain the original states, actions, and logit-shock
behavioral model but replace the original parametric utility specifications
with flexible estimates of the behavior policy and Bellman nuisance functions.
We use cross-fitting throughout and report results for
\(\gamma\in\{0.90,0.95,0.99\}\) with trajectory-clustered 95\% Wald
confidence intervals. Data reconstruction and implementation details are given
in Appendix~\ref{app:data-reconstruction}.

\begin{table}[ht]
\centering
\caption{Counterfactual policy contrasts and normalized-reward estimates}
\label{tab:empirical-results}
\scriptsize
\setlength{\tabcolsep}{3.5pt}
\begin{tabular}{@{}lccc@{}}
\toprule
& \(\gamma=0.90\) & \(\gamma=0.95\) & \(\gamma=0.99\) \\
\midrule
\multicolumn{4}{l}{\textit{Panel A: Restricted minus unrestricted policy}}\\
Gilleskie: value difference under \(r_0=\log\pi_0\)
  & \(0.383\ [0.310,0.455]\) & \(0.379\ [0.285,0.473]\) & \(0.409\ [0.285,0.533]\) \\
Gilleskie: discounted physician visits
  & \(-0.225\ [-0.251,-0.198]\) & \(-0.225\ [-0.253,-0.197]\) & \(-0.232\ [-0.263,-0.200]\) \\
Hotz--Miller: value difference under \(r_0=\log\pi_0\)
  & \(0.352\ [0.307,0.396]\) & \(0.385\ [0.331,0.440]\) & \(0.441\ [0.370,0.513]\) \\
Hotz--Miller: discounted births
  & \(-0.006\ [-0.018,0.006]\) & \(-0.004\ [-0.019,0.011]\) & \(-0.007\ [-0.026,0.012]\) \\
\addlinespace
\multicolumn{4}{l}{\textit{Panel B: Gilleskie policy values under work/no-visit normalization}}\\
Unrestricted policy \(\pi_0\)
  & \(-1.283\ [-1.489,-1.077]\) & \(-1.308\ [-1.601,-1.016]\) & \(-1.219\ [-1.640,-0.797]\) \\
Restricted policy \(\pi^\star\)
  & \(-0.898\ [-1.072,-0.723]\) & \(-0.919\ [-1.202,-0.636]\) & \(-0.840\ [-1.295,-0.385]\) \\
\addlinespace
\multicolumn{4}{l}{\textit{Panel C: Gilleskie normalized rewards at illness onset}}\\
Work, physician visit
  & \(-1.507\ [-1.898,-1.116]\) & \(-1.318\ [-1.777,-0.860]\) & \(-1.049\ [-1.615,-0.484]\) \\
Absent, no visit
  & \(0.176\ [-0.077,0.429]\) & \(0.459\ [0.148,0.769]\) & \(0.859\ [0.347,1.370]\) \\
Absent, physician visit
  & \(-1.562\ [-1.994,-1.129]\) & \(-1.087\ [-1.680,-0.495]\) & \(-0.403\ [-1.287,0.481]\) \\
Visit--absence interaction
  & \(-0.231\ [-0.633,0.171]\) & \(-0.227\ [-0.635,0.181]\) & \(-0.212\ [-0.630,0.205]\) \\
\bottomrule
\end{tabular}
\begin{minipage}{0.96\linewidth}
\footnotesize
\textit{Note:} Entries are estimates with trajectory-clustered 95\% Wald
confidence intervals. In Panel A, physician visits and births are
restricted-minus-unrestricted changes in expected discounted counts. Panels B
and C use the work/no-visit normalization with \(\tau=1\), under which the
policy that always chooses work without a physician visit has value zero. At
the population level, the restricted-minus-unrestricted value difference in
Panel B equals the Gilleskie value contrast in Panel A, although their
separately debiased estimates need not coincide in finite samples. The
visit--absence interaction in Panel C is the joint absence-and-visit effect
minus the sum of the two individual effects.
\end{minipage}
\end{table}

In the Gilleskie application, restricting physician visits reduces expected
discounted visits by about \(0.22\) across the three discount factors. The
direction agrees with \citet{gilleskie1998dynamic}, who also finds a
substantial reduction in visits under this intervention. In the Hotz--Miller
application, removing sterilization has little effect on expected discounted
births: the estimated changes are close to zero, and all confidence intervals
include zero. Under the estimated counterfactual policy, individuals who would
otherwise choose sterilization substitute primarily toward temporary
contraception rather than no contraception, helping explain the small estimated
effect on births.

The policy-value differences are positive and relatively stable across the
reported values of \(\gamma\) in both applications. The Gilleskie
normalization additionally identifies the levels of the policy values. Under
the normalized reward, both the unrestricted and restricted policies have
negative expected discounted value, but the restricted policy has the higher
value at every \(\gamma\). Moreover, the difference between these normalized
policy values closely agrees with the normalization-invariant contrast in
Panel~A, as predicted by Theorem~\ref{theorem::identvalue}. The normalized
levels themselves vary with \(\gamma\), reflecting the dependence of the
identified reward representative on the maintained behavioral discount factor.  Appendix
Table~\ref{tab:gilleskie-normalized-rewards} reports the corresponding
action-level normalized rewards at illness onset.

\section{Conclusion}

We developed a semiparametric framework for inference in softmax IRL and
dynamic discrete choice models with nonparametric rewards. By expressing
reward-dependent estimands as smooth functionals of the behavior policy and
transition kernel, we derive efficient influence functions and construct
automatic debiased machine-learning estimators that permit flexible nuisance
estimation. The framework connects classical inference for Gumbel-shock DDC
models with modern ML-based IRL and provides a reusable template for inference
on new reward-dependent targets. Future work could extend these ideas to
continuous actions, alternative shock distributions, broader normalization
schemes, and nonstationary or finite-horizon settings.  
Another direction is to exploit behavior observed across multiple experts or
environments, which can strengthen reward identification and substantially
reduce the need for normalization
\citep{cao2021identifiability,rolland2022identifiability}. Conditional on such
an identification result, the semiparametric framework developed here can be
adapted for inference on the resulting identified reward-dependent
functionals.

\clearpage
\begin{center}
{\Large\bfseries Supplementary Appendix for\\[.5em]
\JASATitleText}
\end{center}
\setstretch{1.15}

\setcounter{equation}{0}
\renewcommand{\theequation}{S\arabic{equation}}
\setcounter{theorem}{0}
\setcounter{figure}{0}
\setcounter{table}{0}
\setcounter{lemma}{0}
\setcounter{corollary}{0}
\renewcommand{\thetheorem}{S\arabic{theorem}}
\renewcommand{\thecorollary}{S\arabic{corollary}}
\renewcommand{\thelemma}{S\arabic{lemma}}
\renewcommand{\thefigure}{S\arabic{figure}}
\renewcommand{\thetable}{S\arabic{table}}
\renewcommand{\thealgorithm}{S\arabic{algorithm}}

\begingroup
\setcounter{tocdepth}{2}
\renewcommand{\contentsname}{Appendix contents}
\tableofcontents
\endgroup

\appendix

Section~\ref{appendix::model} defines the statistical model and its
bounded-positivity condition. Section~\ref{app::additionalapplications}
develops the reward-normalization extensions and states the assumptions used
by Theorem~\ref{theorem::EIFnorm}. Section~\ref{app:estimator:details} gives
detailed statements for the example estimators.
Section~\ref{app::normsim} describes the simulation design and implementation,
and Section~\ref{app:data-reconstruction} describes the empirical data sources
and sample construction. The remaining sections contain the proofs and
technical lemmas for the soft Bellman equation.

We use the empirical-mean notation from the main text:
\[
\mathbb P_nF:=n^{-1}\sum_iF(O_i),
\qquad
P_nf:=(nT)^{-1}\sum_{i,t}f(Z_{i,t}).
\]

\section{Statistical model}
\label{appendix::model}

Under a time-homogeneous behavior policy and transition law \(\mathbb P\),
the trajectory density factors as
\[
\rho_{\mathbb P,\mathrm{in}}(s_0)
\prod_{t=0}^{T-1}
\pi_{\mathbb P}(a_t\mid s_t)
k_{\mathbb P}(s_{t+1}\mid a_t,s_t),
\]
where \(\rho_{\mathbb P,\mathrm{in}}\) is the initial-state density and
\(\pi_{\mathbb P}\) and \(k_{\mathbb P}\) are the time-homogeneous behavior
policy and transition density. Define the pooled state density by
\[
\rho_{\mathbb P}(s)
:=\frac1T\sum_{t=0}^{T-1}
\frac{d\mathbb P(S_t\in\cdot)}{d\mu}(s),
\qquad
\rho_0:=\rho_{\mathbb P_0}.
\]
For a state--action function \(f\), write
\[
\|f\|_{\mathcal H,0}
:=\{E_0[f(A,S)^2]\}^{1/2}
=\left\{\int_{\mathcal S}\rho_0(s)
\sum_{a\in\mathcal A}e^{r_0(a,s)}f(a,s)^2\,\mu(ds)\right\}^{1/2}.
\]
The second equality uses \(e^{r_0(a,s)}=\pi_0(a\mid s)\).
For fixed constants \(0<c<C<\infty\) and
\(0<\eta<|\mathcal A|^{-1}\), let
\(\mathfrak M_T(c,C,\eta)\) be the collection of these laws satisfying
\[
\begin{aligned}
c<\rho_{\mathbb P}(s)<C
&\quad\text{for \(\mu\)-almost every \(s\)},\\
\eta<\pi_{\mathbb P}(a\mid s)
&\quad\text{for \(\lambda\)-almost every \((a,s)\)},\\
c<k_{\mathbb P}(s'\mid a,s)<C
&\quad\text{for \((\mu\otimes\lambda)\)-almost every \((s',a,s)\)}.
\end{aligned}
\]

\begin{enumerate}[label=\textbf{(A\arabic*)}, ref=A\arabic*, series=posit]
\item \textit{(Bounded positivity).}\label{cond::boundedpositivity}
There are constants \(c_0,C_0,\eta_0\) satisfying
\[
0<c<c_0<C_0<C<\infty,
\qquad
0<\eta<\eta_0<|\mathcal A|^{-1},
\]
such that
\[
c_0<\rho_0(s)<C_0,
\qquad
\eta_0<\pi_0(a\mid s),
\qquad
c_0<k_0(s'\mid a,s)<C_0
\]
almost everywhere. Throughout, we write
\(\mathfrak M_T:=\mathfrak M_T(c,C,\eta)\).
\end{enumerate}

\section{Extended applications under reward normalization}
\label{app::additionalapplications}

This section collects the appendix-only extensions of the normalized-reward
framework and the detailed assumptions referenced in
Theorem~\ref{theorem::EIFnorm}.

\subsection{Detailed assumptions for normalized-reward functionals}
\label{appendix::conditionsEIFnorm}

We state here the assumptions used by Theorem~\ref{theorem::EIFnorm} and by
the normalized-reward extensions below.
Write \(\bar r_0:=r_{0,\nu,\gamma,g}\).

\begin{enumerate}[label=\textbf{(C\arabic**)}, ref=C\arabic**, series=cond2, start=1]
\item (\textit{Functional differentiability}) \label{cond::boundedfunc::norm}
There exist linear maps
\(\partial_r\tilde m(\cdot,\bar r_0,k_0):
T_{\mathcal H}\to L^2(\mu)\) and
\(\partial_k\tilde m(\cdot,\bar r_0,k_0):
T_{\mathcal K}\to L^2(\mu)\) such that, for every bounded
\(h\in T_{\mathcal H}\) and bounded \(b\in T_{\mathcal K}\),
\[
\left.\frac{d}{dt}\,
\mathbb E_0\!\left[
\tilde m\{S_0,\bar r_0+th,(1+tb)k_0\}
\right]\right|_{t=0}
=\mathbb E_0\!\left[
\partial_r\tilde m(S_0,\bar r_0,k_0)(h)
+\partial_k\tilde m(S_0,\bar r_0,k_0)(b)
\right].
\]
\item (\textit{Continuity of derivatives}) \label{cond::boundedfunc2::norm}
The linear functionals
\[
h\mapsto\mathbb E_0[\partial_r\tilde m(S_0,\bar r_0,k_0)(h)],
\qquad
b\mapsto\mathbb E_0[\partial_k\tilde m(S_0,\bar r_0,k_0)(b)]
\]
are continuous on \((T_{\mathcal H},\|\cdot\|_{\mathcal H,0})\) and
\((T_{\mathcal K},\|\cdot\|_{\mathcal K,0})\), respectively.
\item (\textit{Lipschitz continuity}) \label{cond::lipschitz::norm}
For every finite \(M>\|\bar r_0\|_{L^\infty(\lambda)}
+\|k_0\|_{L^\infty(\mu\otimes\lambda)}\), there exist \(\delta,L>0\)
such that any perturbations
\((h_j,b_j)\in T_{\mathcal H}\times T_{\mathcal K}\),
\(j\in\{1,2\}\), with \(\bar r_0+h_j\in\mathcal H\),
\((1+b_j)k_0\in\mathcal K\), and
\[
\|\bar r_0+h_j\|_{L^\infty(\lambda)}
+\|(1+b_j)k_0\|_{L^\infty(\mu\otimes\lambda)}\le M,
\qquad
\|h_j\|_{\mathcal H,0}+\|b_j\|_{\mathcal K,0}<\delta
\]
satisfy
\[
\begin{aligned}
&\big\|\tilde m\{\cdot,\bar r_0+h_1,(1+b_1)k_0\}
-\tilde m\{\cdot,\bar r_0+h_2,(1+b_2)k_0\}
\big\|_{L^2(\mu)}\\
&\qquad\le L\{\|h_1-h_2\|_{\mathcal H,0}
+\|b_1-b_2\|_{\mathcal K,0}\}.
\end{aligned}
\]
\end{enumerate}

\subsection{Softmax values under reward normalization}
\label{app::softmaxnorm}

For \(\mathbb P\in\mathfrak M_T\), write
\(\bar r_{\mathbb P}:=r_{r_{\mathbb P},k_{\mathbb P},\nu,\gamma,g}\) and define
\[
\Psi(\mathbb P)
:=\mathbb E_{\mathbb P}\!\left[
V_{\bar r_{\mathbb P},k_{\mathbb P}}^{\star,\gamma'}(S_0)
\right].
\]

Let
\(\pi_{\bar r_0,k_0}^{\star,\gamma'}\),
\(q_{\bar r_0,k_0}^{\star,\gamma'}\),
\(V_{\bar r_0,k_0}^{\star,\gamma'}\),
\(v_{\bar r_0,k_0}^{\star,\gamma'}\),
\(\Phi_{\bar r_0,k_0}^{\star,\gamma'}\),
\(\rho_{\bar r_0,k_0}^{\star,\gamma'}\),
\(d_{\bar r_0,k_0}^{\star,\gamma'}\), and
\(\widetilde d_{\bar r_0,k_0}^{\star,\gamma'}\) denote the analogues of the
objects in Theorem~\ref{theorem::EIFsoftmax}, but with reward \(\bar r_0\) and
target discount \(\gamma'\).
Let \(\tilde m(s,r,k):=V_{r,k}^{\star,\gamma'}(s)\).
Define the reward representer before normalization by
\[
\widetilde\alpha_0(a,s)
:=d_{\bar r_0,k_0}^{\star,\gamma'}(a,s)
+\tau^{\star-1}\widetilde d_{\bar r_0,k_0}^{\star,\gamma'}(a,s),
\]
and let \(\alpha_0\) be the unique element satisfying
\[
E_0\!\left[\alpha_0(A,S)
\{h(A,S)-\gamma\nu h(S')\}\right]
=E_0\!\left[\widetilde\alpha_0(A,S)
\{h(A,S)-\nu h(S)\}\right],
\qquad h\in T_{\mathcal H}.
\]

\begin{theorem}[EIF for softmax value under reward normalization]
\label{theorem::EIFsoftmaxnorm}
Assume Assumption~\ref{cond::boundedpositivity} and that
Conditions~\ref{cond::boundedfunc::norm}--\ref{cond::lipschitz::norm}
hold for \(\tilde m\). Then \(\Psi\) is pathwise differentiable at
\(\mathbb P_0\) with
efficient influence function
\begin{align*}
\chi_0(O)
={}&V_{\bar r_0,k_0}^{\star,\gamma'}(S_0)-\Psi(\mathbb P_0)\\
&+\frac1T\sum_{t=0}^{T-1}\left[
\alpha_0(A_t,S_t)
-\sum_{\tilde a\in\mathcal A}\pi_0(\tilde a\mid S_t)\alpha_0(\tilde a,S_t)\right.\\
&\hspace{9mm}+\alpha_0(A_t,S_t)
\{r_0(A_t,S_t)-g(S_t)+\gamma V_{r_0-g,k_0}^{\nu,\gamma}(S_{t+1})
-q_{r_0-g,k_0}^{\nu,\gamma}(A_t,S_t)\}\\
&\hspace{9mm}+\frac{\gamma'}{\tau^\star}
\widetilde d_{\bar r_0,k_0}^{\star,\gamma'}(A_t,S_t)
\{\Phi_{\bar r_0,k_0}^{\star,\gamma'}(S_{t+1})
-v_{\bar r_0,k_0}^{\star,\gamma'}(A_t,S_t)\}\\
&\hspace{9mm}\left.+d_{\bar r_0,k_0}^{\star,\gamma'}(A_t,S_t)
\{\bar r_0(A_t,S_t)+\gamma' V_{\bar r_0,k_0}^{\star,\gamma'}(S_{t+1})
-q_{\bar r_0,k_0}^{\star,\gamma'}(A_t,S_t)\}\right].
\end{align*}
\end{theorem}

\subsection{Structural coefficients under reward normalization}

\renewcommand{\theexample}{3}
\begin{example}[Structural coefficients under reward normalization]
\label{example::normcoef}
Let \(x:\mathcal A\times\mathcal S\to\mathbb R^p\) be a known bounded feature
map, let \(\omega\) be a fixed known reference policy, and write
\(\bar r_0:=r_{0,\nu,\gamma,g}\). Define
\begin{align*}
G_0&:=\mathbb E_0\!\left[\sum_{a\in\mathcal A}\omega(a\mid S_0)
x(a,S_0)x(a,S_0)^\top\right],\\
b_0&:=\mathbb E_0\!\left[\sum_{a\in\mathcal A}\omega(a\mid S_0)
x(a,S_0)\bar r_0(a,S_0)\right].
\end{align*}
If \(G_0\) is nonsingular, the coefficient vector in the best linear
predictor of the normalized reward based on \(x\), under the initial-state
distribution and reference policy \(\omega\), is
\[
\theta_0:=G_0^{-1}b_0.
\]
The target policy and the uniform policy are natural choices of \(\omega\).
For example, in Rust's bus-replacement model, a coordinate of \(x(a,s)\) may
indicate replacement, in which case the corresponding coordinate of
\(\theta_0\) is a normalized replacement-cost coefficient
\citep{rust1987optimal}.
\qed
\end{example}

Let \(\alpha_0^x=(\alpha_{0,1}^x,\ldots,\alpha_{0,p}^x)^\top\), where each
coordinate is the unique element satisfying
\begin{equation}
E_0\!\left[\alpha_{0,j}^x(A,S)
\{h(A,S)-\gamma\nu h(S')\}\right]
=\mathbb E_0\!\left[\sum_{a\in\mathcal A}\omega(a\mid S_0)x_j(a,S_0)
\{h(a,S_0)-\nu h(S_0)\}\right],
\qquad h\in T_{\mathcal H}.
\label{eqn::coef-representer-balance}
\end{equation}

\begin{theorem}[EIF for structural coefficients under reward normalization]
\label{theorem::EIFcoefnorm}
Assume Assumption~\ref{cond::boundedpositivity} and that \(G_0\) is nonsingular.
Then the EIF for \(b_0\) is
\[
\begin{aligned}
\chi_{b,0}(O)
:={}&\sum_{a\in\mathcal A}\omega(a\mid S_0)x(a,S_0)\bar r_0(a,S_0)-b_0\\
&+\frac1T\sum_{t=0}^{T-1}\left[
\alpha_0^x(A_t,S_t)
-\sum_{\tilde a\in\mathcal A}\pi_0(\tilde a\mid S_t)
\alpha_0^x(\tilde a,S_t)\right.\\
&\hspace{19mm}\left.+\alpha_0^x(A_t,S_t)
\{r_0(A_t,S_t)-g(S_t)
+\gamma V_{r_0-g,k_0}^{\nu,\gamma}(S_{t+1})
-q_{r_0-g,k_0}^{\nu,\gamma}(A_t,S_t)\}\right],
\end{aligned}
\]
and the EIF for \(\theta_0\) is
\[
\chi_{\theta,0}(O)
=
G_0^{-1}\!\left[
\chi_{b,0}(O)
-\left\{\sum_{a\in\mathcal A}\omega(a\mid S_0)x(a,S_0)x(a,S_0)^\top
-G_0\right\}\theta_0
\right].
\]
\end{theorem}

\section{Detailed statements for example estimators}
\label{app:estimator:details}
\label{app::examplesestimators}

This section groups the example-specific estimator displays, the additional
assumptions deferred from Section~\ref{sec::examplesestimators}, and the
resulting asymptotic conclusions. The main-text theorem labels remain
unchanged; we restate the first three results only in detailed, unnumbered
form for readability.

\subsection{Example~\ref{example::1a}: value of a fixed policy}

Let \(\pi_n:=\exp\{r_n\}\), let \(q_n^{\pi,\gamma}\) estimate
\(q_{r_0,k_0}^{\pi,\gamma}\), and define
\(V_n^{\pi,\gamma}(s):=\sum_{\tilde a\in\mathcal A}\pi(\tilde a\mid s)
q_n^{\pi,\gamma}(\tilde a,s)\). Let \(\rho_n^{\pi,\gamma}\) estimate the state
occupancy ratio \(\rho_0^{\pi,\gamma}\). The estimator of
\(\psi_0=\mathbb E_0[V_{r_0,k_0}^{\pi,\gamma}(S_0)]\) is
\begin{align*}
\psi_n
&:= \frac{1}{n}\sum_{i=1}^n V_n^{\pi,\gamma}(S_{i,0})\\
&\quad+P_n\!\left[\rho_n^{\pi,\gamma}(S)\left\{
\frac{\pi(A\mid S)}{\pi_n(A\mid S)}
\left(r_n(A,S)+\gamma V_n^{\pi,\gamma}(S')-q_n^{\pi,\gamma}(A,S)\right)
+\frac{\pi(A\mid S)}{\pi_n(A\mid S)}-1\right\}\right].
\end{align*}

The additional conditions for Theorem~\ref{theorem::ALex1} are stated below.
\begin{enumerate}[label=\textbf{(E\arabic*)}, ref=E\arabic*, start=2]
    \item \label{assump:boundedness::policy}(\textit{Boundedness})
    There exists $M \in (0,\infty)$ such that
    $\|\rho_n^{\pi,\gamma}\|_{L^\infty(\mu)}
      + \|q_n^{\pi,\gamma}\|_{L^\infty(\lambda)}
      + \|\pi_n^{-1}\|_{L^\infty(\lambda)} < M$
    for all $n$ almost surely.

    \item \label{assump:splitting::policy}(\textit{Sample splitting})
    The estimators $r_n$, $\rho_n^{\pi,\gamma}$, $q_n^{\pi,\gamma}$ are constructed using data independent of $\mathcal{D}_n$.

    \item \label{assump:consistency::policy}(\textit{Consistency})
   $ \|\rho_n^{\pi,\gamma} - \rho_0^{\pi,\gamma}\|_{L^2(\rho_0)}
      + \|r_n - r_0\|_{\mathcal{H},0} +
    \bigl\|
        q_n^{\pi,\gamma} - q_{r_0,k_0}^{\pi,\gamma}
    \bigr\|_{\mathcal{H},0}
    = o_p(1)$.
\end{enumerate}

\medskip\noindent\textbf{Detailed version of Theorem~\ref{theorem::ALex1}.}
Suppose Assumptions~\ref{cond::boundedpositivity} and~\ref*{cond::reward},
Condition~\ref{assump:rates::policy}, and
Conditions~\ref{assump:boundedness::policy}--\ref{assump:consistency::policy} hold. Then
Theorem~\ref{thm:asymptotic_linearity} holds for the estimator above with
influence function \(\chi_0\) given in Theorem~\ref{theorem::EIFvalue}.

\subsection{Example~\ref{example::1b}: value of a counterfactual softmax policy}

Let \(r_n\) estimate \(r_0\), let \(v_n^\star\) estimate the soft Bellman
fixed point \(v_{r_0,k_0}^\star\), and define
\[
\pi_n^\star(a\mid s)
\propto
1\{a\in\mathcal A^\star(s)\}
\exp\!\left\{
\tfrac{1}{\tau^\star}\{r_n(a,s)+\gamma v_n^\star(a,s)\}
\right\}.
\]
Let \(q_n^\star\) estimate \(q_{r_0,k_0}^\star\), set
\[
\begin{aligned}
V_n^\star(s)
&:=\sum_{\tilde a\in\mathcal A^\star(s)}
\pi_n^\star(\tilde a\mid s)q_n^\star(\tilde a,s),\\
\Phi_n^\star(s)
&:=\tau^\star\log\!\left[
\sum_{a\in\mathcal A^\star(s)}
\exp\!\left\{\frac{r_n(a,s)+\gamma v_n^\star(a,s)}{\tau^\star}\right\}
\right],
\end{aligned}
\]
and let \(\widetilde d_n^\star\) and \(\rho_n^\star\) estimate
\(\widetilde d_{r_0,k_0}^\star\) and \(\rho_{r_0,k_0}^\star\). The resulting
estimator of \(\psi_0=\mathbb E_0\{V_{r_0,k_0}^\star(S_0)\}\) is
\begin{align*}
\psi_n
&:=\frac{1}{n}\sum_{i=1}^n V_n^\star(S_{i,0})\\
&\quad+P_n\!\Bigg[
\rho_n^\star(S)\left\{
\frac{\pi_n^\star(A\mid S)}{\pi_n(A\mid S)}
\left(r_n(A,S)+\gamma V_n^\star(S')-q_n^\star(A,S)\right)
+\frac{\pi_n^\star(A\mid S)}{\pi_n(A\mid S)}-1\right\}\\
&\hspace{35mm}+\frac1{\tau^\star}\left\{
\gamma\widetilde d_n^\star(A,S)
\left(\Phi_n^\star(S')-v_n^\star(A,S)\right)
+\widetilde d_n^\star(A,S)-(\pi_n\widetilde d_n^\star)(S)\right\}\Bigg].
\end{align*}

\noindent The additional conditions for Theorem~\ref{theorem::ALex1b} are stated below.
\begin{enumerate}[label=\textbf{(E\arabic**)}, ref=E\arabic**, start=2]
    \item \label{assump:boundedness::softstar}(\textit{Boundedness})
    There exists $M \in (0,\infty)$ such that, for all $n$ almost surely,
    \[
      \|\rho_n^\star\|_{L^\infty(\mu)}
      + \|\widetilde d_n^\star\|_{L^\infty(\lambda)}
      + \|q_n^\star\|_{L^\infty(\lambda)}
      + \|v_n^\star\|_{L^\infty(\lambda)}
      + \|r_n\|_{L^\infty(\lambda)}
      + \|\pi_n^{-1}\|_{L^\infty(\lambda)}
      < M.
    \]

    \item \label{assump:splitting::softstar}(\textit{Sample splitting})
    The nuisance estimators $\rho_n^\star$, $\widetilde d_n^\star$, $q_n^\star$, $v_n^\star$, and $r_n$
    are constructed using data independent of $\mathcal{D}_n$ (or via cross-fitting).

    \item \label{assump:consistency::softstar}(\textit{Consistency})
    \[
      \|\rho_n^\star - \rho_{r_0,k_0}^\star\|_{L^2(\rho_0)}
      + \|\widetilde d_n^\star - \widetilde d_{r_0,k_0}^\star\|_{\mathcal H,0}
      + \|q_n^\star - q_{r_0,k_0}^\star\|_{\mathcal{H},0}
      + \|v_n^\star - v_{r_0,k_0}^\star\|_{\mathcal{H},0}
      + \|r_n - r_0\|_{\mathcal{H},0}
      = o_p(1).
    \]
\end{enumerate}

\medskip\noindent\textbf{Detailed version of Theorem~\ref{theorem::ALex1b}.}
Suppose Assumptions~\ref{cond::boundedpositivity} and~\ref*{cond::reward},
Condition~\ref{assump:rates::softstar}, and
Conditions~\ref{assump:boundedness::softstar}--\ref{assump:consistency::softstar} hold. Then
Theorem~\ref{thm:asymptotic_linearity} holds for the estimator above with
influence function \(\chi_0\) given in Theorem~\ref{theorem::EIFsoftmax}.

\subsection{Example~\ref{example::norm}: fixed-policy value under reward normalization}

In addition to the example-specific conditions below, the normalized-reward EIF
used in Theorem~\ref{theorem::EIFvaluenorm} relies on
Assumption~\ref{cond::boundedpositivity} and
Conditions~\ref{cond::boundedfunc::norm}--\ref{cond::lipschitz::norm} from
Appendix~\ref{appendix::conditionsEIFnorm}.

Write \(q_0^{\nu,\gamma}:=q_{r_0-g,k_0}^{\nu,\gamma}\) and
\(q_{0,\nu}^{\pi,\gamma'}:=q_{r_0,k_0,\nu}^{\pi,\gamma'}\). Let
\(\pi_n:=\exp\{r_n\}\), let \(q_n^{\nu,\gamma}\) estimate
\(q_0^{\nu,\gamma}\), and define
\(V_n^{\nu,\gamma}(s):=\sum_{\tilde a\in\mathcal A}\nu(\tilde a\mid s)\,
q_n^{\nu,\gamma}(\tilde a,s)\) and
\(r_{n,\nu,\gamma,g}(a,s):=g(s)+q_n^{\nu,\gamma}(a,s)-V_n^{\nu,\gamma}(s)\).
Let \(q_{n,\nu}^{\pi,\gamma'}\) estimate
\(q_{0,\nu}^{\pi,\gamma'}\), set
\(V_{n,\nu}^{\pi,\gamma'}(s):=\sum_{\tilde a\in\mathcal A}\pi(\tilde a\mid s)\,
q_{n,\nu}^{\pi,\gamma'}(\tilde a,s)\). Let
\(d_n^{\pi,\gamma'}\) estimate \(d_0^{\pi,\gamma'}\), and let
\(\alpha_n^{\pi,\nu}\) estimate the solution of
\eqref{eqn::policyvaluenormbalance}. The corresponding estimator of
\(\psi_0=\mathbb E_0[V_{r_0,k_0,\nu}^{\pi,\gamma'}(S_0)]\) is
\begin{samepage}
\begin{align*}
\psi_n
&:=\frac1n\sum_{i=1}^n V_{n,\nu}^{\pi,\gamma'}(S_{i,0})\\
&\quad+P_n\!\Bigg[
d_n^{\pi,\gamma'}(A,S)
\left(r_{n,\nu,\gamma,g}(A,S)
+\gamma'V_{n,\nu}^{\pi,\gamma'}(S')
-q_{n,\nu}^{\pi,\gamma'}(A,S)\right)\\
&\hspace{35mm}+\alpha_n^{\pi,\nu}(A,S)
\left(r_n(A,S)-g(S)+\gamma V_n^{\nu,\gamma}(S')-q_n^{\nu,\gamma}(A,S)\right)\\
&\hspace{35mm}+\alpha_n^{\pi,\nu}(A,S)-(\pi_n\alpha_n^{\pi,\nu})(S)\Bigg].
\end{align*}
\end{samepage}

The additional conditions for Theorem~\ref{theorem::ALex2} are stated below.
\begin{enumerate}[label=\textbf{(E\arabic***)}, ref=E\arabic***, start=2]
    \item \label{assump:boundedness::policynorm}(\textit{Boundedness})
    There exists $M \in (0,\infty)$ such that, for all $n$ almost surely,
    \[
      \|d_n^{\pi,\gamma'}\|_{L^\infty(\lambda)}
      + \|\alpha_n^{\pi,\nu}\|_{L^\infty(\lambda)}
      + \|q_{n,\nu}^{\pi,\gamma'}\|_{L^\infty(\lambda)}
      + \|q_n^{\nu,\gamma}\|_{L^\infty(\lambda)}
      + \|r_n\|_{L^\infty(\lambda)}
      < M.
    \]

    \item \label{assump:splitting::policynorm}(\textit{Sample splitting})
    The nuisance estimators
    $r_n$, $d_n^{\pi,\gamma'}$, $\alpha_n^{\pi,\nu}$,
    $q_{n,\nu}^{\pi,\gamma'}$, and $q_n^{\nu,\gamma}$
    are constructed using data independent of $\mathcal{D}_n$.

    \item \label{assump:consistency::policynorm}(\textit{Consistency})
   $ \|d_n^{\pi,\gamma'}-d_0^{\pi,\gamma'}\|_{\mathcal H,0}
      +\|\alpha_n^{\pi,\nu}-\alpha_0^{\pi,\nu}\|_{\mathcal H,0}
      + \|r_n - r_0\|_{\mathcal{H},0} +
    \bigl\|
       q_n^{\nu,\gamma} - q_0^{\nu,\gamma}
    \bigr\|_{\mathcal{H},0} +   \bigl\|
      q_{n,\nu}^{\pi,\gamma'} - q_{0,\nu}^{\pi,\gamma'}
    \bigr\|_{\mathcal{H},0}
    = o_p(1)$.
\end{enumerate}

\medskip\noindent\textbf{Detailed version of Theorem~\ref{theorem::ALex2}.}
Suppose Assumptions~\ref{cond::boundedpositivity} and~\ref*{cond::reward},
Condition~\ref{assump:rates::policynorm}, and
Conditions~\ref{assump:boundedness::policynorm}--\ref{assump:consistency::policynorm} hold.
Then Theorem~\ref{thm:asymptotic_linearity} holds for the estimator above with
influence function \(\chi_0\) given in Theorem~\ref{theorem::EIFvaluenorm}.

\subsection{Example~\ref{example::normcoef}: structural coefficients under reward normalization}

To estimate \(\theta_0\) in Example~\ref{example::normcoef}, define
\(V_n^{\nu,\gamma}(s):=\sum_{\tilde a}\nu(\tilde a\mid s)
q_n^{\nu,\gamma}(\tilde a,s)\),
\(\bar r_n(a,s):=g(s)+q_n^{\nu,\gamma}(a,s)-V_n^{\nu,\gamma}(s)\),
\(\pi_n:=\exp\{r_n\}\), and let \(\alpha_n^x\) estimate
\(\alpha_0^x\). Set
\[
\begin{aligned}
b_n
:={}&\frac{1}{n}\sum_{i=1}^n\sum_{a\in\mathcal A}
\omega(a\mid S_{i,0})x(a,S_{i,0})\bar r_n(a,S_{i,0})\\
&+P_n\!\Bigg[
\alpha_n^x(A,S)\left(r_n(A,S)-g(S)+\gamma V_n^{\nu,\gamma}(S')
-q_n^{\nu,\gamma}(A,S)\right)
+\alpha_n^x(A,S)-(\pi_n\alpha_n^x)(S)\Bigg],
\end{aligned}
\]
and
\(G_n:=n^{-1}\sum_{i=1}^n\sum_{a\in\mathcal A}
\omega(a\mid S_{i,0})x(a,S_{i,0})x(a,S_{i,0})^\top\),
\(\theta_n:=G_n^{-1}b_n\). The example-specific rate condition from the main
text is
\begin{enumerate}[label=\textbf{(E\arabic****)}, ref=E\arabic****, series=efficiency3]
\item \label{assump:rates::coefnorm} (\textit{Nuisance rates})
\[
\|\alpha_n^x-\alpha_0^x\|_{\mathcal H,0}\,
\bigl\|\mathcal T_{k_0,\nu,\gamma}(q_n^{\nu,\gamma}-q_0^{\nu,\gamma})\bigr\|_{\mathcal H,0}
= o_p(n^{-1/2}).
\]
\end{enumerate}

The additional conditions for Theorem~\ref{theorem::ALex3} are stated below.
\begin{enumerate}[label=\textbf{(E\arabic****)}, ref=E\arabic****, resume = efficiency3]
    \item \label{assump:boundedness::coefnorm}(\textit{Boundedness})
    There exists $M \in (0,\infty)$ such that, for all $n$ almost surely,
    \[
      \|q_n^{\nu,\gamma}\|_{L^\infty(\lambda)}
      + \|\alpha_n^x\|_{L^\infty(\lambda)}
      + \|r_n\|_{L^\infty(\lambda)}
      < M.
    \]

    \item \label{assump:splitting::coefnorm}(\textit{Sample splitting})
    The nuisance estimators $r_n$, $\alpha_n^x$, and $q_n^{\nu,\gamma}$ are constructed using
    data independent of $\mathcal{D}_n$.

    \item \label{assump:consistency::coefnorm}(\textit{Consistency})
    \[
      \|r_n - r_0\|_{\mathcal{H},0}
      + \|\alpha_n^x-\alpha_0^x\|_{\mathcal H,0}
      + \|q_n^{\nu,\gamma} - q_0^{\nu,\gamma}\|_{\mathcal{H},0}
      = o_p(1).
    \]
\end{enumerate}

\begin{theorem}[Asymptotic linearity for Example~\ref{example::normcoef}]
\label{theorem::ALex3}
Suppose Assumptions~\ref{cond::boundedpositivity} and~\ref*{cond::reward} hold,
$G_0$ is nonsingular, and
Conditions~\ref{assump:rates::coefnorm}--
\ref{assump:consistency::coefnorm} hold. Then
\[
\theta_n - \theta_0
= \frac{1}{n}\sum_{i=1}^n \chi_{\theta,0}(O_i) + o_p(n^{-1/2}),
\qquad
\sqrt n(\theta_n-\theta_0)\rightsquigarrow N(0,\Sigma_0),
\]
where \(\Sigma_0 := \operatorname{Var}_{\mathbb P_0}\{\chi_{\theta,0}(O)\}\).
\end{theorem}

\section{Simulation design and implementation}
\label{app::normsim}

\subsection{Data-generating process}

The state is \(S=(X,Z)\in[-2.5,2.5]^2\), and
\(\mathcal A=\{0,1,2,3\}\). Conditional on \(S=s=(x,z)\) and \(A=a\),
\[
\bar m_a(x,z)=\operatorname{proj}_{[-2.5,2.5]^2}\{m_a(x,z)\},
\qquad
S'=\operatorname{proj}_{[-2.5,2.5]^2}\{\bar m_a(x,z)+\varepsilon\},
\qquad \varepsilon\sim N(0,0.2^2I_2),
\]
where
\[
\begin{aligned}
m_0(x,z)&=(0.80x+0.15\sin z,\ 0.85z+0.10\tanh x),\\
m_1(x,z)&=(0.65x+0.10\sin z-0.45,\ 0.85z+0.15\tanh x+0.20),\\
m_2(x,z)&=(0.55x+0.10\sin z-0.80,\ 0.80z+0.20\tanh x+0.45),\\
m_3(x,z)&=(0.90x+0.10\sin z+0.35,\ 0.75z+0.10\tanh x-0.30).
\end{aligned}
\]
Each sample consists of \(n\) independent one-step transitions drawn from the
stationary law of the behavior process. We use a \(41\times41\) product grid
to compute the stationary distribution, Bellman solutions, and population
targets; sampled states are jittered within grid cells.

For Examples~\ref{example::1a} and~\ref{example::1b}, the behavior policy is
the multinomial logit
\(\pi_0(a\mid s)\propto\exp\{\ell_a(s)\}\), with
\[
\begin{aligned}
\ell_0(x,z)&=0,\\
\ell_1(x,z)&=0.12+0.08x-0.05z+0.02xz,\\
\ell_2(x,z)&=-1.80-0.03x+0.04z-0.01x^2,\\
\ell_3(x,z)&=-0.05+0.04x+0.06z+0.015xz.
\end{aligned}
\]
The identified pseudo-reward is \(r_0=\log\pi_0\), and the discount factor is
\(\gamma=0.9\). In Example~\ref{example::1a}, the fixed target policy satisfies
\(\pi(a\mid s)\propto\exp\{h_a(s)/0.8\}\), where
\[
\begin{aligned}
h_0(x,z)&=0.2-0.2|x|-0.1|z|,\\
h_1(x,z)&=1.4(x-0.6)-(z-0.7),\\
h_2(x,z)&=-1.5+0.3x-0.3z,\\
h_3(x,z)&=1.1(-x-0.3)+(z-1).
\end{aligned}
\]
In Example~\ref{example::1b}, the target is the entropy-regularized optimal
policy for \(r_0\), restricted to actions \(\{0,1,3\}\), with temperature
\(\tau^\star=0.6\).

\subsection{Nuisance estimation for Examples~\ref{example::1a}
and~\ref{example::1b}}

Both estimators use five-fold cross-fitting. Within each training fold, we fit
the correctly specified degree-two multinomial logit for \(\pi_0\). The
inverse regularization constant is \(C=10\); fitted probabilities are bounded
below by \(10^{-3}\) in Example~\ref{example::1a} and by \(0.02\) in
Example~\ref{example::1b}. All remaining nuisances are estimated from the
training-fold transitions, without using the transition kernel employed to
generate the data.

We use action-specific quadratic polynomial sieves for the Bellman equations.
Sieve FQE with ridge \(3\times10^{-4}\), at most 100 iterations, and convergence
tolerance \(10^{-8}\) estimates the fixed-policy value in
Example~\ref{example::1a}. In Example~\ref{example::1b}, soft sieve FQI with the
same numerical specification estimates the entropy-regularized target policy;
a subsequent sieve FQE fit estimates its ordinary value. Each update uses the
observed successor states.

For Example~\ref{example::1a}, we estimate the ordinary discounted occupancy
ratio by solving its empirical balance equation in a quadratic polynomial
sieve with ridge \(10^{-5}\), followed by mass normalization. For
Example~\ref{example::1b}, positivity-constrained linear FORE
\citep{van2026fore} uses quadratic features and an \(L^2\) penalty of
\(10^{-3}\). The number of fixed-point updates is selected from
\(\{30,100,300\}\) by held-out occupancy-balance validation and then refit on
the full training fold. We additionally form the signed source weights
\[
\widehat d^\star(S_i,A_i)
\{\widehat q^\star(S_i,A_i)-\widehat V^\star(S_i)\}
\]
and estimate the associated signed occupancy ratio from its empirical balance
equation using a degree-three polynomial sieve with ridge \(10^{-4}\).

\subsection{Monte Carlo summaries}

For each design and \(n\in\{2500,5000,10000\}\), we use 300 Monte Carlo
replications. We report bias, Monte Carlo standard deviation, and RMSE for the
plug-in and D-IRL estimators. For D-IRL, we also report the average estimated
standard error and coverage of
\(\widehat\psi\pm1.96\widehat{\operatorname{se}}(\widehat\psi)\).

\section{Data sources and reconstruction}
\label{app:data-reconstruction}

This section describes the public-use files and sample-construction rules for
the two empirical applications. The reconstruction follows the source
codebooks and the published sample restrictions identified in the released
files. The tables compare the resulting samples with the corresponding
published samples.

\subsection{Gilleskie medical-care panel}

\paragraph{Data sources.}
We obtain five components of the 1987 National Medical Expenditure Survey from
the Inter-university Consortium for Political and Social Research (ICPSR).
Public Use Tape 29, ICPSR 6473, contains medical-condition dates, work-loss
periods, and links between conditions and medical events
\citep{nmes1987tape29}. Tape 14.5, ICPSR 9881, contains dated office,
hospital-outpatient, and emergency-room visits \citep{nmes1987tape145}. Tape
14.4, ICPSR 9840, contains zero-night hospital visits
\citep{nmes1987tape144}. Tape 13, ICPSR 9695, contains demographics,
employment, income, sick-leave, and insurance variables
\citep{nmes1987tape13}. Tape 9, ICPSR 9674, contains baseline self-reported
health status \citep{nmes1987tape9}. We read the fixed-width files according to
the dictionaries distributed with the ICPSR releases.

\paragraph{Sample construction.}
We implement the restrictions in \citet{gilleskie1998dynamic}: men aged
25--64 at baseline who remain employed in every observed interview round,
receive no public insurance assistance, are not self-employed, and have the
required state variables. The released age variable is measured at the end of
the last eligible interview round. Values 25--65 reproduce the article's count
for the baseline restriction of ages 25--64. We identify infectious,
parasitic, and respiratory conditions from the reported diagnosis codes and
construct their start and end dates. The analysis retains the first acute
episode lasting at most 21 days that is observed entirely within 1987.
Conditions without valid start and end dates do not enter the duration-based
episode sample.

Physician use combines condition-linked office, hospital-outpatient,
emergency-room, and zero-night hospital visits. We retain every emergency-room
record because the released emergency-room file records visits, whereas the
field describing whom the patient met is not a physician-use indicator.
Work-absence records report an interval and the number of missed days. When
this information does not identify the exact missed dates, we allocate the
reported count deterministically within the interval and mark the dates as
inferred.

Each episode contributes one transition for every illness day. The four
actions are work without a physician visit, work with a physician visit,
absence without a physician visit, and absence with a physician visit. The
state contains illness day, diagnosis group, cumulative visits and absences,
income class, sick-leave availability, insurance coverage, baseline health,
age, race, education, and marital status. Recovery leads to a zero-reward
absorbing state.

\paragraph{Comparison with the published sample.}
The released files reproduce the first four sample-flow counts in the article
but not the later questionnaire and completeness restrictions.
Table~\ref{tab:gilleskie-flow} reports the resulting sample flow. The final
sample contains 731 episodes and 5,356 illness days, compared with 726 episodes
in the article. Mean duration is 7.327 days, compared with 7.35 in the article.

\begin{table}[t]
\centering
\caption{Gilleskie sample reconstruction}
\label{tab:gilleskie-flow}
\small
\begin{tabular}{lrr}
\toprule
Sample restriction & Article & Reconstruction \\
\midrule
NMES participants & 38,446 & 38,446 \\
Age restriction & 18,059 & 18,059 \\
Men & 8,325 & 8,325 \\
Employed in every observed round & 6,630 & 6,630 \\
Completed health questionnaire & 5,287 & 5,379 \\
Complete state variables & 4,608 & 4,939 \\
Not self-employed & 3,797 & 3,650 \\
First observed acute episode of at most 21 days & 726 & 731 \\
\bottomrule
\end{tabular}
\end{table}

The proportions of episodes with any physician visit and any work absence are
0.416 and 0.784 in the reconstruction, compared with 0.439 and 0.741 in the
article. Mean episode duration and the distribution of visit counts conditional
on a visit are close to the corresponding published quantities. Because the
public-use files do not identify the article's exact questionnaire and
completeness restrictions, we describe the resulting panel as a public-file
reconstruction rather than an exact replication.

\subsection{Hotz--Miller fertility panel}

\paragraph{Data sources and linkage.}
We obtain the 1970 National Fertility Survey, ICPSR 20003, and the 1975
National Fertility Survey, ICPSR 4334, from the original ICPSR releases
\citep{nfs1970,nfs1975}. The releases contain 6,752 and 3,403 records,
respectively. The documented four-digit identifiers are unique in both files.
Their exact merge links 2,361 couples, equal to the completed-reinterview count
in the 1975 survey manual. We do not use demographic or probabilistic record
linkage. \citet{hotz1993conditional} report 2,374 linked couples, so the
released files contain 13 fewer links before the analysis restrictions are
applied.

\paragraph{Actions and states.}
The 1975 survey contains a retrospective monthly calendar of contraceptive
methods, pregnancies, births, and sterilization. We decode both entries in each
calendar cell and aggregate methods into no contraception, temporary
contraception, and sterilization. We then apply the annualization rule in
\citet{hotz1993conditional}. Sterilization during a calendar year takes
precedence. If a birth occurs in the following year, the current-year action is
the method used immediately before pregnancy. In the remaining cases, the
action is the method used for the majority of observed months; a year containing
a birth is classified as temporary contraception during the postpartum period.
Sterilization removes the couple from the subsequent risk set.

The retrospective calendar begins near the 1970 survey boundary. We recover a
missing initial action, in order, from a documented 1970 sterilization,
pregnancy-specific method histories, and dated pill, intrauterine-device, or
other-method histories. Four retained couples lack sufficient timing
information and are assigned the current method reported at reinterview; these
assignments are flagged. We exclude couples who were sterile before 1970 and
couples whose 1970 operation was not undertaken for contraceptive purposes.

The annual state contains calendar year, the ages and education of both
spouses, the ages of the first five children, total parity, and the preceding
action. Parity includes all dated births, and a twin delivery contributes two
children of the same age. We exclude realized annual income from the current
state because it is measured over the same year as the action. The birth outcome
equals the same-year change in parity, so its timing agrees with the state
transition. Sterilization and the 1975 administrative endpoint lead to distinct
zero-reward absorbing states.

\paragraph{Comparison with the published sample.}
The reconstructed panel contains 2,104 couples and 10,441 couple-years.
Table~\ref{tab:hotz-reconstruction} compares the sample and action counts with
those in \citet{hotz1993conditional}. The largest year-specific discrepancy is
for sterilization in 1971: the reconstruction contains 114 such choices,
compared with 163 in the article. The article does not report the record-level
completeness rule, and the released files do not contain the 13 additional
cross-wave links. The public releases therefore do not permit exact
reproduction of the published sample.

\begin{table}[t]
\centering
\caption{Hotz--Miller sample reconstruction}
\label{tab:hotz-reconstruction}
\small
\begin{tabular}{lrrr}
\toprule
Quantity & Article & Reconstruction & Difference \\
\midrule
Linked couples & 2,374 & 2,361 & \(-13\) \\
Analysis couples & 2,088 & 2,104 & \(+16\) \\
Couple-years & 10,257 & 10,441 & \(+184\) \\
No contraception & 1,885 & 1,848 & \(-37\) \\
Temporary contraception & 7,576 & 7,821 & \(+245\) \\
Sterilization & 796 & 772 & \(-24\) \\
\bottomrule
\end{tabular}
\end{table}

The no-sterilization intervention requires overlap for later fecund states
reached under the counterfactual policy. Couples who sterilize in the observed
data leave the risk set, whereas they remain in fecund states under the
counterfactual policy. Identification therefore relies on the maintained
Markov and overlap assumptions for these later states.

\subsection{Unequal trajectory lengths}

The two applications contain independent trajectories of unequal length. Let
\(T_i\) denote the number of transition records for trajectory \(i\), including
its absorbing-state record, and let \(Z_{it}=(S_{it},A_{it},S_{i,t+1})\). We
replace the fixed-length pooled transition measure by
\[
P_n f
=
\frac{\sum_{i=1}^n\sum_{t=0}^{T_i-1}f(Z_{it})}
     {\sum_{i=1}^n T_i},
\qquad
P_0 f
=
\frac{\mathbb E_0\!\left[\sum_{t=0}^{T-1}f(Z_t)\right]}
     {\mathbb E_0[T]}.
\]
If \(m_0(S_0)\) is the initial-state component and \(\varphi_0(Z_t)\) is
the transition correction, the trajectory-level influence function becomes
\[
\chi_0(O_i)
=m_0(S_{i0})-\mathbb E_0\{m_0(S_0)\}
+\frac{\sum_{t=0}^{T_i-1}\varphi_0(Z_{it})-T_iP_0\varphi_0}
       {\mathbb E_0[T]}.
\]
We estimate its variance across trajectories and keep every trajectory within
one cross-fitting fold. This modification changes the pooled transition
measure and trajectory-level variance calculation; the infinite-horizon
estimands and Bellman equations are unchanged.

For both normalized policy values, we implement the EIF in
Theorem~\ref{theorem::EIFsoftmaxnorm}. The normalization adjoint uses the
combined pre-normalization reward representer
\(d^\star+(\tau^\star)^{-1}\widetilde d^\star\), rather than the ordinary
occupancy ratio alone. We estimate Wald variances from the resulting
trajectory-level influence functions.

\subsection{Normalized rewards at illness onset}

Table~\ref{tab:gilleskie-normalized-rewards} reports the expected one-period
normalized reward for each nonreference action at the initial illness state.
We set the reward for working without a physician visit to zero and set
\(\tau=1\). These action-level rewards are distinct from the expected
discounted policy values reported in Table~\ref{tab:empirical-results}.

\begin{table}[H]
\centering
\caption{Gilleskie normalized rewards at illness onset}
\label{tab:gilleskie-normalized-rewards}
\scriptsize
\setlength{\tabcolsep}{3.5pt}
\begin{tabular}{@{}lccc@{}}
\toprule
& \(\gamma=0.90\) & \(\gamma=0.95\) & \(\gamma=0.99\) \\
\midrule
Work, physician visit
  & \(-1.507\ [-1.898,-1.116]\) & \(-1.318\ [-1.777,-0.860]\) & \(-1.049\ [-1.615,-0.484]\) \\
Absent, no visit
  & \(0.176\ [-0.077,0.429]\) & \(0.459\ [0.148,0.769]\) & \(0.859\ [0.347,1.370]\) \\
Absent, physician visit
  & \(-1.562\ [-1.994,-1.129]\) & \(-1.087\ [-1.680,-0.495]\) & \(-0.403\ [-1.287,0.481]\) \\
Visit--absence interaction
  & \(-0.231\ [-0.633,0.171]\) & \(-0.227\ [-0.635,0.181]\) & \(-0.212\ [-0.630,0.205]\) \\
\bottomrule
\end{tabular}
\begin{minipage}{0.96\linewidth}
\footnotesize
\textit{Note:} Entries are estimates with trajectory-clustered 95\% Wald
confidence intervals. The visit--absence interaction is the joint effect of
absence and a physician visit minus the sum of their individual effects.
\end{minipage}
\end{table}

\subsection{Short-horizon sensitivity analysis}

Table~\ref{tab:short-horizon} reports the contemporaneous and short-horizon
benchmarks. These values are sensitivity analyses rather than alternative
preferred specifications.

\begin{table}[H]
\centering
\caption{Short-horizon sensitivity analysis}
\label{tab:short-horizon}
\scriptsize
\setlength{\tabcolsep}{3.5pt}
\begin{tabular}{@{}lcc@{}}
\toprule
& \(\gamma=0\) & \(\gamma=0.50\) \\
\midrule
Gilleskie: value under \(r_0=\log\pi_0\)
  & \(0.213\ [0.186,0.241]\) & \(0.309\ [0.281,0.337]\) \\
Gilleskie: physician visits
  & \(-0.121\ [-0.143,-0.099]\) & \(-0.180\ [-0.203,-0.156]\) \\
Hotz--Miller: value under \(r_0=\log\pi_0\)
  & \(0.132\ [0.116,0.147]\) & \(0.224\ [0.189,0.259]\) \\
Hotz--Miller: births
  & \(-0.0068\ [-0.0141,0.0006]\) & \(-0.0100\ [-0.0195,-0.0004]\) \\
\addlinespace
Gilleskie normalized reward: work, visit
  & \(-2.010\ [-2.262,-1.757]\) & \(-1.913\ [-2.180,-1.646]\) \\
Gilleskie normalized reward: absent, no visit
  & \(-0.408\ [-0.573,-0.242]\) & \(-0.336\ [-0.508,-0.164]\) \\
Gilleskie normalized reward: absent, visit
  & \(-2.615\ [-2.902,-2.329]\) & \(-2.451\ [-2.724,-2.179]\) \\
Gilleskie normalized reward: interaction
  & \(-0.198\ [-0.590,0.193]\) & \(-0.202\ [-0.593,0.189]\) \\
\bottomrule
\end{tabular}
\begin{minipage}{0.96\linewidth}
\footnotesize
\textit{Note:} Entries are estimates with trajectory-clustered 95\% Wald
confidence intervals.
\end{minipage}
\end{table}

\section{Proofs for Section~\ref{sec:ident}}

For a transition kernel \(k\) and state function \(g\), write
\[
\mathcal P_k g(a,s):=\int g(s')k(s'\mid a,s)\,\mu(ds').
\]

\begin{proof}[Proof of Lemma~\ref{theorem::trivialsol}]
The key observation is that when \(r_0=\log\pi_0\), the log-partition term is
zero, so \(v_{r_0,k_0}\equiv 0\) solves the Bellman constraint. We then compare
the induced softmax representation with that of \(r_0^\dagger\) to identify the
shaping potential. By definition, we have
\begin{align*}
\sum_{a \in \mathcal{A}} \exp\{ r_0(a,s) \}
&= \sum_{a \in \mathcal{A}} \exp\{ \log \pi_0(a \mid s) \}  \\
&= \sum_{a \in \mathcal{A}} \pi_0(a \mid s) \\
&= 1.
\end{align*}
Hence, \(\log \sum_{a \in \mathcal{A}} \exp\{ r_0(a,s) \} = 0\),
and
\begin{align*}
0 = \int \log \sum_{a' \in \mathcal{A}} \exp\{ r_0(a',s') + \gamma \cdot 0 \} \,
k_0(s' \mid a, s) \,\mu(ds').
\end{align*}
It follows that \(0\) solves the soft Bellman equation with respect to the
reward \(r_0\). Moreover, the induced softmax policy satisfies
\[
\frac{\exp\{r_0(a,s)\}}{\sum_{\tilde a \in \mathcal{A}} \exp\{r_0(\tilde a,s)\}}
= \exp\{r_0(a,s)\}
= \pi_0(a \mid s),
\]
so it coincides with the true behavior policy and therefore maximizes the
population log-likelihood. Thus, \(r_0\) is also a solution to the soft
Bellman-constrained log-likelihood in \eqref{eqn::softbellmanloglik}.

Furthermore, we have
\[
\exp\{r_0(a,s)\} = \pi_0(a \mid s)
= \frac{\exp\{r_0^\dagger(a,s) + \gamma v_0^\dagger(a,s)\}}{\sum_{\tilde a \in \mathcal{A}} \exp\{r_0^\dagger(\tilde a,s) + \gamma v_0^\dagger(\tilde a,s)\}}.
\]
Taking logarithms of both sides yields
\[
r_0(a,s) = r_0^\dagger(a,s) + \gamma v_0^\dagger(a,s)
- \log \sum_{\tilde a \in \mathcal{A}} \exp\{r_0^\dagger(\tilde a,s) + \gamma v_0^\dagger(\tilde a,s)\}.
\]
The soft value function is defined as
\[
V_{0}^{\dagger}(s) = \log \Bigg( \sum_{a' \in \mathcal{A}}
\exp\big\{ r_0^\dagger(a',s) + \gamma v_0^\dagger(a', s)\big\} \Bigg).
\]
By the soft Bellman equation,
\[
v_0^\dagger(a,s)  = \int  V_{0}^{\dagger}(s') \, k_0(s' \mid a , s) \, \mu(ds').
\]
Substituting this expression, we obtain
\[
r_0(a,s) = r_0^\dagger(a,s) + \gamma \int V_{0}^{\dagger}(s') \, k_0(s' \mid a , s) \, \mu(ds') - V_{0}^{\dagger}(s).
\]
This is \eqref{eqn::equivclass} with shaping function
\(c=-V_0^\dagger\), which proves the lemma.
\end{proof}

\begin{proof}[Proof of Theorem~\ref{theorem::identWithNormalization}]
For any reward \(r\), let \(v_{r,k_0}\) solve the soft Bellman equation and set
\(q_r(a,s):=r(a,s)+\gamma v_{r,k_0}(a,s)\) and
\(\Lambda_r(s):=\log\sum_{a\in\mathcal A}\exp\{q_r(a,s)\}\). Then the criterion in
\eqref{eqn::softbellmanloglik} equals
\[
E_0\{\Lambda_r(S)-q_r(A,S)\}
=
E_0\{-\log \pi_r(A\mid S)\},
\qquad
\pi_r(a\mid s):=\exp\{q_r(a,s)-\Lambda_r(s)\}.
\]
Since \(A\mid S\sim\pi_0(\cdot\mid S)\),
\[
E_0\{-\log \pi_r(A\mid S)\}
=
E_0\!\left[\mathrm{KL}\{\pi_0(\cdot\mid S),\pi_r(\cdot\mid S)\}\right]
-
E_0\!\left[\sum_a \pi_0(a\mid S)\log \pi_0(a\mid S)\right].
\]
Thus \(r\) minimizes \eqref{eqn::softbellmanloglik} if and only if
\(\pi_r(\cdot\mid s)=\pi_0(\cdot\mid s)\) for \(P_0\)-a.e. \(s\).

Let \(\bar r_0:=r_0-g\) and
\(q:=q_{r_0-g,k_0}^{\nu,\gamma}
=\mathcal T_{k_0,\nu,\gamma}^{-1}(\bar r_0)\), so
\(q=\bar r_0+\gamma\mathcal P_{k_0}\nu q\). Define
\(c^\star:=-\nu q\), \(r^\star:=g+(I-\nu)q\), and
\(v^\star:=\mathcal P_{k_0}c^\star=-\mathcal P_{k_0}\nu q\). Since
\(q-\gamma\mathcal P_{k_0}\nu q=\bar r_0\),
\[
r^\star+\gamma v^\star
=
g+(I-\nu)q-\gamma\mathcal P_{k_0}\nu q
=
r_0+c^\star .
\]
Therefore
\[
\Lambda_{r^\star}(s)
=
\log\sum_{a\in\mathcal A}\exp\{r_0(a,s)+c^\star(s)\}
=
c^\star(s),
\]
because \(\sum_a e^{r_0(a,s)}=\sum_a\pi_0(a\mid s)=1\). Hence
\(v^\star=\mathcal P_{k_0}c^\star=\mathcal P_{k_0}\Lambda_{r^\star}\), so
\(v^\star\) solves the soft Bellman equation for \(r^\star\). Also
\[
\pi_{r^\star}(a\mid s)
=
\exp\{r_0(a,s)+c^\star(s)-\Lambda_{r^\star}(s)\}
=
\pi_0(a\mid s),
\]
so \(r^\star\) minimizes \eqref{eqn::softbellmanloglik}. Finally,
\(\nu r^\star=\nu g+\nu(I-\nu)q=g\), and therefore \(r^\star\) is normalized.

For uniqueness, let \(\tilde r\in\mathcal H\) be any normalized minimizer, and
let \(\tilde v\) solve its soft Bellman equation. Since \(\tilde r\) induces
\(\pi_0\), we have
\[
\tilde r(a,s)+\gamma\tilde v(a,s)-\tilde\Lambda(s)=r_0(a,s),
\qquad
\tilde\Lambda(s):=\log\sum_{a\in\mathcal A}
\exp\{\tilde r(a,s)+\gamma\tilde v(a,s)\}.
\]
Thus \(\tilde r=r_0+\tilde\Lambda-\gamma\tilde v\). Since
\(\tilde v=\mathcal P_{k_0}\tilde\Lambda\), the normalization
\(\nu\tilde r=g\) gives
\[
\tilde\Lambda
=
-\nu(r_0-g)+\gamma\nu\mathcal P_{k_0}\tilde\Lambda .
\]
The right-hand side is a \(\gamma\)-contraction in sup norm. Thus this
fixed-point equation has a unique bounded solution. Applying \(\nu\) to
\(q=\bar r_0+\gamma\mathcal P_{k_0}\nu q\) shows that
\(c^\star=-\nu q\) satisfies the same equation,
\(c^\star=-\nu(r_0-g)+\gamma\nu\mathcal P_{k_0}c^\star\). Hence
\(\tilde\Lambda=c^\star\), \(\tilde v=\mathcal P_{k_0}\tilde\Lambda
=-\mathcal P_{k_0}\nu q\), and
\[
\tilde r
=
r_0+\tilde\Lambda-\gamma\tilde v
=
r_0-\nu q+\gamma\mathcal P_{k_0}\nu q
=
g+q-\nu q
=
r^\star .
\]
Thus the normalized solution is unique.

Writing \(q=q_{r_0-g,k_0}^{\nu,\gamma}\) statewise gives
\[
r^\star(a,s)
=
g(s)+q_{r_0-g,k_0}^{\nu,\gamma}(a,s)
-\sum_{\tilde a\in\mathcal A}\nu(\tilde a\mid s)
q_{r_0-g,k_0}^{\nu,\gamma}(\tilde a,s).
\]
Moreover, if \(\gamma>0\), then
\[
v^\star(a,s)
=
-\mathcal P_{k_0}\nu q(a,s)
=
\gamma^{-1}\{r_0(a,s)-g(s)-q_{r_0-g,k_0}^{\nu,\gamma}(a,s)\}.
\]
Hence \(r^\star=r_{0,\nu,\gamma,g}\) and
\(v^\star=v_{0,\nu,\gamma,g}\), as claimed.
\end{proof}

\begin{proof}[Proof of Theorem~\ref{theorem::identvalue}]
By Lemma~\ref{theorem::trivialsol},
\[
r_0(a,s) \;=\; r_0^\dagger(a,s) + \gamma \int V_0^\dagger(s') \, k_0(s' \mid a,s)\,\mu(ds') - V_0^\dagger(s).
\]
By the definition of the $Q$-function under policy $\pi$, we have
\begin{align*}
q^{\pi,\gamma}_{r_0,k_0}(a,s)
&= r_0(a,s) + \gamma \int \sum_{a' \in \mathcal{A}} \pi(a' \mid s') \, q^{\pi,\gamma}_{r_0,k_0}(a',s') \, k_0(s' \mid a,s) \,\mu(ds').
\end{align*}
Substituting the expression for $r_0(a,s)$ gives
\begin{align*}
q^{\pi,\gamma}_{r_0,k_0}(a,s)
&= r_0^\dagger(a,s) + \gamma \int V_0^\dagger(s') \, k_0(s' \mid a,s)\,\mu(ds') - V_0^\dagger(s) \\
&\quad + \gamma \int \sum_{a' \in \mathcal{A}} \pi(a' \mid s') \, q^{\pi,\gamma}_{r_0,k_0}(a',s') \, k_0(s' \mid a,s)\,\mu(ds').
\end{align*}
Rearranging terms,
\[
q^{\pi,\gamma}_{r_0,k_0}(a,s) + V_0^\dagger(s)
= r_0^\dagger(a,s) + \gamma \int \sum_{a' \in \mathcal{A}} \pi(a' \mid s') \big(q^{\pi,\gamma}_{r_0,k_0}(a',s') + V_0^\dagger(s')\big)\,k_0(s' \mid a,s)\,\mu(ds').
\]
Thus, \(q^{\pi,\gamma}_{r_0,k_0}(a,s) + V_0^\dagger(s)\) satisfies the Bellman equation with reward \(r_0^\dagger\).
Uniqueness of the solution then gives
\[
q^{\pi,\gamma}_{r_0,k_0}(a,s) + V_0^\dagger(s) \;=\; q^{\pi,\gamma}_{r_0^\dagger,k_0}(a,s),
\]
or equivalently,
\[
q^{\pi,\gamma}_{r_0,k_0}(a,s) \;=\; q^{\pi,\gamma}_{r_0^\dagger,k_0}(a,s) - V_0^\dagger(s).
\]

Taking expectations with respect to the initial state distribution
\(\rho_{0,\mathrm{in}}\), we obtain
\[
V_{r_0,k_0}(\pi; \gamma)
=V_{r_0^\dagger,k_0}(\pi; \gamma)-\mathbb E_0[V_0^\dagger(S_0)].
\]

Finally, for any two policies $\pi_1$ and $\pi_2$,
\begin{align*}
V_{r_0,k_0}(\pi_1; \gamma) - V_{r_0,k_0}(\pi_2; \gamma)
&= \big(V_{r_0^\dagger,k_0}(\pi_1; \gamma)
-\mathbb E_0[V_0^\dagger(S_0)]\big)
 - \big(V_{r_0^\dagger,k_0}(\pi_2; \gamma)
-\mathbb E_0[V_0^\dagger(S_0)]\big) \\
&= V_{r_0^\dagger,k_0}(\pi_1; \gamma) - V_{r_0^\dagger,k_0}(\pi_2; \gamma).
\end{align*}
This proves the asserted invariance of policy contrasts.
\end{proof}

\section{Proofs for Section~\ref{sec::EIFsub}}

 \label{appendix::EIFsub}

We first derive general differentiability facts for the parameter map and then
assemble them into the EIF and von Mises expansion stated in
Section~\ref{sec::EIFsub}.

The following lemma gives a moment-matching characterization of $r_P$.
The main text sets $\mathcal H=L^\infty(\lambda)$, but the result also holds
for any linear subspace $\mathcal H\subseteq L^\infty(\lambda)$ that contains
all bounded state-only functions, including additive and partially linear
classes.

We will use that $r_P$ is the solution to the multinomial logit
problem
\begin{equation}\label{eqn::Mestimand}
\begin{aligned}
r_P
&:= \argmin_{r \in \mathcal{H}} \; -E_P[r(A,S)] \\
&\quad \text{subject to} \quad
\sum_{a \in \mathcal{A}} \exp\{r(a,s)\} = 1,
\qquad \mu\text{-a.e.\ } s \in \mathcal{S}.
\end{aligned}
\end{equation}

\begin{lemma}[Softmax moment-matching identity]\label{thm:softmax-moment-matching}
Let \(P\) be a dominated pooled transition law whose state marginal has
density \(\rho_P>0\) \(\mu\)-almost everywhere. Let
$\mathcal H \subset L^\infty(\lambda)$ be a linear class that contains all state-only functions, i.e., $L^\infty(\mu)$. Let $r_P$ denote the solution to \eqref{eqn::Mestimand} for this choice of $\mathcal{H}$. Then
\[
E_P[h(A,S)]
= E_{P}\!\Big[\sum_{a\in\mathcal A} \exp\{ r_P(a,S) \}\,h(a,S)\Big],
\quad \forall\, h\in\mathcal H .
\]
\end{lemma}

\begin{proof}[Proof of Lemma~\ref{thm:softmax-moment-matching}]
Fix $h\in\mathcal H$ and define the state-only normalizer
\[
c_t(s):=\log\!\left\{\sum_{a\in\mathcal A}
\exp\big(r_P(a,s)+t h(a,s)\big)\right\}.
\]
Because $\mathcal H$ contains every bounded state-only function,
$r_t:=r_P+t h-c_t$ belongs to $\mathcal H$ for sufficiently small $t$.
Moreover, $\sum_a\exp\{r_t(a,s)\}=1$, so $r_t$ is a feasible path through
$r_P$. First-order optimality of $r_P$ along this path gives
\[
0=-\left.\frac{d}{dt}E_P\{r_t(A,S)\}\right|_{t=0}
=-E_P\!\left[h(A,S)-\sum_{a\in\mathcal A}
e^{r_P(a,S)}h(a,S)\right],
\]
which is the stated identity.
\end{proof}

For each $r \in \mathcal{H}$, define the integral operator
$\mathcal{L}_{r}: L^\infty(\lambda) \to L^2(\lambda)$ by
\[
\mathcal{L}_{r}(h)(a,s) := \left\{\sum_{\tilde a \in \mathcal A} \exp\{ r(\tilde a,s) \}\,h(\tilde a,s)\right\} - h(a,s).
\]
The first-order optimality conditions for $r_P$ in
\eqref{eqn::Mestimand} can then be written as
\[
E_P\!\Big[\mathcal{L}_{r_P}(h)(A,S)\Big] = 0,
\quad \forall\, h\in\mathcal H.
\]
The G\^ateaux derivative of $\mathcal{L}_{r}(h)$ in the direction $h' \in L^\infty(\lambda)$ is
\[
\dot{\mathcal{L}}_r(h,h')(a,s)
:= \frac{d}{dt}\, \mathcal{L}_{r+th'}(h)(a,s)\Big|_{t=0}
=  \sum_{\tilde a \in \mathcal A} \exp\{ r(\tilde a,s) \}\,h'(\tilde a,s)\,h(\tilde a,s).
\]

\begin{lemma}[First-order expansion of $\mathcal{L}_r$]
\label{lemma::quadraticloss}
Fix $r,h,h' \in L^\infty(\lambda)$, and set $r' := r+h'$.
Suppose $r$, $r'$, and $h$ are essentially bounded by $M$. Then
\[
\Big| E_P\!\Big[ \big\{ \mathcal L_{r'}(h) - \mathcal L_{r}(h) - \dot{\mathcal L}_{r}(h,h') \big\}(A,S) \Big] \Big|
\;\lesssim\; E_P[\dot{\mathcal L}_{r}(h',h')(A,S)].
\]
The implicit constant depends only on $M$.
\end{lemma}
\begin{proof}[Proof of Lemma~\ref{lemma::quadraticloss}]
By definition,
\[
\mathcal{L}_{r}(h)(a,s)
= \sum_{\tilde a} e^{\,r(\tilde a,s)} h(\tilde a,s) - h(a,s),
\qquad
\dot{\mathcal L}_{r}(h,h')(a,s)
= \sum_{\tilde a} e^{\,r(\tilde a,s)} h'(\tilde a,s)\,h(\tilde a,s).
\]
Hence,
\begin{align*}
\mathcal L_{r'}(h)(a,s) - \mathcal L_{r}(h)(a,s) - \dot{\mathcal L}_{r}(h,h')(a,s)
&= \sum_{\tilde a} \Big( e^{\,r(\tilde a,s)+h'(\tilde a,s)} - e^{\,r(\tilde a,s)} - e^{\,r(\tilde a,s)}h'(\tilde a,s) \Big)\,h(\tilde a,s) \\
&= \sum_{\tilde a} e^{\,r(\tilde a,s)}\big( e^{\,h'(\tilde a,s)} - 1 - h'(\tilde a,s) \big)\,h(\tilde a,s).
\end{align*}
Using the inequality $|e^{u} - 1 - u| \le \tfrac{1}{2} e^{|u|} u^2$ for all $u \in \mathbb{R}$, we obtain
\[
\big| \mathcal L_{r'}(h)(a,s) - \mathcal L_{r}(h)(a,s) - \dot{\mathcal L}_{r}(h,h')(a,s) \big|
\;\le\; \tfrac{1}{2} \sum_{\tilde a} e^{\,r(\tilde a,s)} e^{\,|h'(\tilde a,s)|}\,|h(\tilde a,s)|\,|h'(\tilde a,s)|^2.
\]
If $r$, $r'$, and $h$ are essentially bounded by $M$, then
\begin{align*}
\big| \mathcal L_{r'}(h)(a,s) - \mathcal L_{r}(h)(a,s) - \dot{\mathcal L}_{r}(h,h')(a,s) \big|
&\le \sum_{\tilde a} e^{\,r(\tilde a,s)} e^{2M}\,M\,|h'(\tilde a,s)|^2 \\
&\lesssim \dot{\mathcal L}_{r}(h',h')(a,s),
\end{align*}
where the implicit constant depends only on $M$. Taking expectations under
$P$ proves that
\[
\Big| E_P\!\Big[ \big\{ \mathcal L_{r'}(h) - \mathcal L_{r}(h) - \dot{\mathcal L}_{r}(h,h') \big\}(A,S) \Big] \Big|
\;\lesssim\; E_P[\dot{\mathcal L}_{r}(h',h')(A,S)].
\]
This is the asserted first-order remainder bound.
\end{proof}

Let \(\addTK\) denote the \(\|\cdot\|_{\mathcal K}\)-closure of the bounded
functions \(g(s',a,s)\) satisfying
\(\int g(s'\mid a,s)\,\mu(ds')=0\) for \(\lambda\)-almost every \((a,s)\).
For a map \(F\) of the transition kernel, write
\[
\adk F(k)[g]
:=\left.\frac{d}{dt}F(k+tg)\right|_{t=0}
\]
for its derivative under an additive kernel perturbation.

\begin{lemma}[Continuous Fr\'echet differentiability implies Conditions C1--C3]
\label{lemma::frechetimplieslip}
Let
\[
G(h,b):=m\{\cdot,r_0+h,(1+b)k_0\}.
\]
Assume that \(G\) admits a continuously Fr\'echet differentiable extension to a neighborhood
of \((0,0)\) as a map from
\((T_{\mathcal H},\|\cdot\|_{\mathcal H,0})\times
 (T_{\mathcal K},\|\cdot\|_{\mathcal K,0})\) into \(L^2(\mu)\), where the
domain is equipped with the sum norm. Continue to write \(G\) for this extension. Then
Conditions~\ref{cond::boundedfunc}--\ref{cond::lipschitz} hold.
\end{lemma}
\begin{proof}[Proof of Lemma~\ref{lemma::frechetimplieslip}]
Continuity of \(DG\) at \((0,0)\) yields \(\delta,L>0\) such that
\begin{equation}
\sup_{\|h\|_{\mathcal H,0}+\|b\|_{\mathcal K,0}<\delta}
\|DG(h,b)\|_{\mathrm{op}}\le L.
\label{eqn::frechet-local-derivative-bound}
\end{equation}
Define the partial derivatives by
\[
\partial_r m(\cdot,r_0,k_0)(h):=DG(0,0)(h,0),
\qquad
\partial_k m(\cdot,r_0,k_0)(b):=DG(0,0)(0,b).
\]
Linearity of \(DG(0,0)\) gives
\[
DG(0,0)(h,b)
=\partial_r m(\cdot,r_0,k_0)(h)
+\partial_k m(\cdot,r_0,k_0)(b).
\]
The Fr\'echet expansion at \((0,0)\) is
\begin{equation}
G(th,tb)-G(0,0)
=tDG(0,0)(h,b)+o(|t|)
\quad\text{in }L^2(\mu).
\label{eqn::frechet-state-map-expansion}
\end{equation}
Applying the continuous initial-state expectation in
Lemma~\ref{lemma::initialpooledbounds} to
\eqref{eqn::frechet-state-map-expansion} gives
Condition~\ref{cond::boundedfunc}. Lemma~\ref{lemma::initialpooledbounds} and
\eqref{eqn::frechet-local-derivative-bound} give
\begin{align*}
&\left|\mathbb E_0\{DG(0,0)(h,b)(S_0)\}\right|\\
&\qquad\lesssim \|DG(0,0)(h,b)\|_{L^2(\mu)}
\le L\{\|h\|_{\mathcal H,0}+\|b\|_{\mathcal K,0}\}.
\end{align*}
Taking \(b=0\) and then \(h=0\) proves
Condition~\ref{cond::boundedfunc2}.
Fix \(x_j=(h_j,b_j)\), \(j=1,2\), in the ball of radius \(\delta/2\).
The line segment \(x_t=x_2+t(x_1-x_2)\) remains in the ball of radius
\(\delta\). The fundamental theorem of calculus for Banach-space-valued maps
gives
\[
G(x_1)-G(x_2)
=\int_0^1DG(x_t)(x_1-x_2)\,dt.
\]
Taking the \(L^2(\mu)\) norm and using
\eqref{eqn::frechet-local-derivative-bound} gives
\[
\|G(x_1)-G(x_2)\|_{L^2(\mu)}
\le L\{
\|h_1-h_2\|_{\mathcal H,0}
+\|b_1-b_2\|_{\mathcal K,0}\}.
\]
This is Condition~\ref{cond::lipschitz}.
\end{proof}

\begin{lemma}[Hellinger stability of the target functional]
\label{lemma::target-hellinger-lipschitz}
Suppose Assumption~\ref{cond::boundedpositivity} and
Condition~\ref{cond::lipschitz} hold. Define
\[
H^2(Q_1,Q_2):=\int
\{\sqrt{dQ_1}-\sqrt{dQ_2}\}^{\,2}.
\]
There are constants \(\delta_H,L_H>0\) such that
\[
|\Psi(\mathbb P_1)-\Psi(\mathbb P_2)|
\le L_H H(\mathbb P_1,\mathbb P_2)
\]
whenever \(\mathbb P_1,\mathbb P_2\in\mathfrak M_T\) and
\(H(\mathbb P_j,\mathbb P_0)<\delta_H\), \(j\in\{1,2\}\).
\end{lemma}
\begin{proof}
For \(j\in\{0,1,2\}\), let \(P_j\) be the pooled transition law induced by
\(\mathbb P_j\), and write its density as
\[
p_j(s,a,s')=\rho_j(s)\pi_j(a\mid s)k_j(s'\mid a,s).
\]
The pooled law is obtained by adjoining an independent uniform time index to
the trajectory and then marginalizing. The data-processing inequality for
Hellinger distance therefore gives
\begin{equation}
H(P_1,P_2)\le H(\mathbb P_1,\mathbb P_2).
\label{eqn::pooled-hellinger-contraction}
\end{equation}
The conditional decomposition of squared Hellinger distance gives
\begin{align}
H^2(P_1^{AS},P_2^{AS})
={}&H^2(P_1^S,P_2^S)
+\int \sqrt{\rho_1\rho_2}
\sum_a(\sqrt{\pi_1}-\sqrt{\pi_2})^2\,d\mu,
\label{eqn::policy-hellinger-decomposition}\\
H^2(P_1,P_2)
={}&H^2(P_1^{AS},P_2^{AS})\notag\\
&+\int \sqrt{\rho_1\pi_1\rho_2\pi_2}
(\sqrt{k_1}-\sqrt{k_2})^2\,d(\mu\otimes\lambda).
\label{eqn::kernel-hellinger-decomposition}
\end{align}
All functions in these displays are evaluated at their corresponding
arguments. Because every law in \(\mathfrak M_T\) satisfies the common bounds
\(c,C,\eta\), the inequalities
\[
|\log x-\log y|\le \eta^{-1}|x-y|,
\qquad
|x-y|\le 2\sqrt C\,|\sqrt x-\sqrt y|
\]
and \eqref{eqn::policy-hellinger-decomposition}--%
\eqref{eqn::kernel-hellinger-decomposition} yield
\begin{align*}
\|r_1-r_2\|_{\mathcal H,0}^2
&\le \frac{4C_0}{\eta^2c}\,
H^2(P_1^{AS},P_2^{AS}),\\
\left\|\frac{k_1-k_2}{k_0}\right\|_{\mathcal K,0}^2
&\le \frac{4CC_0}{c_0c\eta}\,
H^2(P_1,P_2).
\end{align*}
Consequently,
\begin{equation}
\|r_1-r_2\|_{\mathcal H,0}
+\left\|\frac{k_1-k_2}{k_0}\right\|_{\mathcal K,0}
\lesssim H(P_1,P_2).
\label{eqn::hellinger-component-control}
\end{equation}

Let \(m_j:=m(\cdot,r_j,k_j)\), and let
\(\rho_{j,\mathrm{in}}\) be the initial-state density under
\(\mathbb P_j\). Since \(\rho_{j,\mathrm{in}}\le T\rho_j\le TC\),
Hellinger contraction to the initial-state marginal gives
\[
\|\rho_{1,\mathrm{in}}-\rho_{2,\mathrm{in}}\|_{L^2(\mu)}
\le 2\sqrt{TC}\,H(\mathbb P_1,\mathbb P_2).
\]
For \(\delta_H\) sufficiently small,
\eqref{eqn::pooled-hellinger-contraction},
\eqref{eqn::hellinger-component-control}, and
Condition~\ref{cond::lipschitz} imply
\[
\|m_1-m_2\|_{L^2(\mu)}
\lesssim H(\mathbb P_1,\mathbb P_2),
\qquad
\max_{j\in\{1,2\}}\|m_j\|_{L^2(\mu)}\lesssim1.
\]
Consequently,
\begin{align*}
|\Psi(\mathbb P_1)-\Psi(\mathbb P_2)|
&\le
\|\rho_{1,\mathrm{in}}\|_{L^2(\mu)}
\|m_1-m_2\|_{L^2(\mu)}\\
&\quad+
\|\rho_{1,\mathrm{in}}-\rho_{2,\mathrm{in}}\|_{L^2(\mu)}
\|m_2\|_{L^2(\mu)}
\lesssim H(\mathbb P_1,\mathbb P_2),
\end{align*}
which proves the result.
\end{proof}

\begin{lemma}[Dense-score extension]
\label{lemma::dense-score-extension}
Let \(\nu:\mathfrak M_T\to\mathbb R\), and let
\(\dot\nu_0\) be a bounded linear functional on the tangent space at
\(\mathbb P_0\). Suppose that a dense set \(\mathcal S_b\) of scores has the
following property: for every \(\ell\in\mathcal S_b\), there is a regular
submodel \(\{\mathbb P_\varepsilon^\ell\}\) with score \(\ell\) such that
\[
\nu(\mathbb P_\varepsilon^\ell)-\nu(\mathbb P_0)
=\varepsilon\dot\nu_0(\ell)+o(\varepsilon).
\]
If \(\nu\) is locally Hellinger Lipschitz at \(\mathbb P_0\), then
\(\nu\) is pathwise differentiable at \(\mathbb P_0\) with derivative
\(\dot\nu_0\).
\end{lemma}
\begin{proof}
This is \citet[Lemma~2]{luedtke2024one}, applied with target space
\(\mathbb R\).
\end{proof}

\begin{lemma}[Pathwise derivative under bounded product submodels]
\label{lemma::movinglawchainrule}
Suppose Conditions~\ref{cond::boundedfunc} and~\ref{cond::lipschitz} hold.
Let \(u\), \(h\), and \(b\) be bounded and satisfy
\(\mathbb E_0\{u(S_0)\}=0\), \((\pi_0h)(S)=0\), and
\(E_0\{b(S',A,S)\mid A,S\}=0\). For all sufficiently small
\(|\varepsilon|\), define the product submodel by
\[
\frac{dP_{\varepsilon,\mathrm{in}}}{dP_{0,\mathrm{in}}}
=1+\varepsilon u,
\qquad
\pi_\varepsilon=(1+\varepsilon h)\pi_0,
\qquad
k_\varepsilon=(1+\varepsilon b)k_0,
\]
and put \(r_\varepsilon=\log\pi_\varepsilon\) and
\(m_\varepsilon=m(\cdot,r_\varepsilon,k_\varepsilon)\). Then
\begin{align}
\Psi(\mathbb P_\varepsilon)-\Psi(\mathbb P_0)
={}&\varepsilon\,\mathbb E_0\!\left[
\{m(S_0,r_0,k_0)-\psi_0\}u(S_0)\right]\notag\\
&+\varepsilon\,\mathbb E_0\!\left[
\partial_r m(S_0,r_0,k_0)(h)
+\partial_k m(S_0,r_0,k_0)(b)\right]+o(\varepsilon).
\label{eqn::moving-law-chain-rule}
\end{align}
\end{lemma}

\begin{proof}
Boundedness of \(h\) and the inequality
\(|\log(1+x)-x|\le Cx^2\) for \(|x|\le1/2\) give
\begin{equation}
\|r_\varepsilon-r_0-\varepsilon h\|_{\mathcal H,0}
=O(\varepsilon^2).
\label{eqn::log-policy-path-remainder}
\end{equation}
Define
\[
\bar m_\varepsilon
:=m\{\cdot,r_0+\varepsilon h,(1+\varepsilon b)k_0\}.
\]
Condition~\ref{cond::lipschitz},
\eqref{eqn::log-policy-path-remainder}, and
Lemma~\ref{lemma::initialpooledbounds} imply
\begin{align}
\|m_\varepsilon-\bar m_\varepsilon\|_{L^2(\mu)}
&=O(\varepsilon^2),
\label{eqn::curved-linear-path-difference}\\
\|m_\varepsilon-m_0\|_{L^2(P_{0,\mathrm{in}})}
&=O(|\varepsilon|).
\label{eqn::state-map-path-continuity}
\end{align}
Condition~\ref{cond::boundedfunc} and
\eqref{eqn::curved-linear-path-difference} therefore give
\begin{align}
\mathbb E_0\{m_\varepsilon(S_0)-m_0(S_0)\}
={}&\varepsilon\,\mathbb E_0\!\left[
\partial_r m(S_0,r_0,k_0)(h)
+\partial_k m(S_0,r_0,k_0)(b)\right]+o(\varepsilon).
\label{eqn::state-map-path-derivative}
\end{align}
Writing \(L_\varepsilon=1+\varepsilon u\), the exact decomposition
\begin{align}
\Psi(\mathbb P_\varepsilon)-\Psi(\mathbb P_0)
={}&\mathbb E_0\{(L_\varepsilon-1)m_0(S_0)\}
+\mathbb E_0\{m_\varepsilon(S_0)-m_0(S_0)\}\notag\\
&+\mathbb E_0\!\left[
(L_\varepsilon-1)\{m_\varepsilon(S_0)-m_0(S_0)\}\right]
\label{eqn::moving-law-product-rule}
\end{align}
The initial-score term in \eqref{eqn::moving-law-product-rule} satisfies
\begin{equation}
\mathbb E_0\{(L_\varepsilon-1)m_0(S_0)\}
=\varepsilon\mathbb E_0[\{m_0(S_0)-\psi_0\}u(S_0)],
\label{eqn::moving-law-initial-score-term}
\end{equation}
because \(\mathbb E_0\{u(S_0)\}=0\). Moreover, Cauchy--Schwarz and
\eqref{eqn::state-map-path-continuity} give
\begin{align}
&\left|\mathbb E_0\!\left[
(L_\varepsilon-1)\{m_\varepsilon(S_0)-m_0(S_0)\}\right]\right|\notag\\
&\qquad\le |\varepsilon|\,\|u\|_{L^2(P_{0,\mathrm{in}})}
\|m_\varepsilon-m_0\|_{L^2(P_{0,\mathrm{in}})}
=O(\varepsilon^2).
\label{eqn::moving-law-cross-term}
\end{align}
Substitution of \eqref{eqn::state-map-path-derivative},
\eqref{eqn::moving-law-initial-score-term}, and
\eqref{eqn::moving-law-cross-term} into
\eqref{eqn::moving-law-product-rule} proves
\eqref{eqn::moving-law-chain-rule}.
\end{proof}

\begin{proof}[Proof of Theorem~\ref{theorem::EIF}]
By Lemma~\ref{lemma::markovscoreidentities}, it suffices first to consider a
bounded product submodel with score
\[
\ell(O)=u(S_0)+\sum_{t=0}^{T-1}
\{h(A_t,S_t)+b(S_{t+1},A_t,S_t)\},
\]
where \(\mathbb E_0\{u(S_0)\}=0\), \(\pi_0h(S)=0\), and
\(E_0\{b(S',A,S)\mid A,S\}=0\). The derivatives of
\(r=\log\pi\) and \(k\) along this submodel are \(h\) and \(b\),
respectively.

Lemma~\ref{lemma::movinglawchainrule} gives
\begin{align*}
\left.\frac{d}{d\varepsilon}\Psi(\mathbb P_\varepsilon)
\right|_{\varepsilon=0}
={}&\mathbb E_0[\{m(S_0,r_0,k_0)-\psi_0\}u(S_0)]\\
&+\mathbb E_0\{\partial_r m(S_0,r_0,k_0)(h)
+\partial_k m(S_0,r_0,k_0)(b)\}.
\end{align*}
The Riesz identities \eqref{eqn::rieszreward}--\eqref{eqn::rieszkernel}
turn the second line into
\[
E_0\{\alpha_0(A,S)h(A,S)\}
+E_0\{\beta_0(S',A,S)b(S',A,S)\}.
\]
Set
\[
x_0(S_0):=m(S_0,r_0,k_0)-\psi_0,
\qquad
a_0(A,S):=\alpha_0(A,S)-(\pi_0\alpha_0)(S).
\]
The definitions of \(\psi_0\), \(a_0\), and \(T_{\mathcal K}\) give
\begin{equation}
\mathbb E_0\{x_0(S_0)\}=0,\qquad
E_0\{a_0(A,S)\mid S\}=0,\qquad
E_0\{\beta_0(S',A,S)\mid A,S\}=0.
\label{eqn::general-eif-component-centering}
\end{equation}
Since \(E_0\{h(A,S)\mid S\}=0\),
\begin{equation}
E_0\{\alpha_0(A,S)h(A,S)\}
=E_0\!\left[
\{\alpha_0(A,S)-(\pi_0\alpha_0)(S)\}h(A,S)\right]
=E_0\{a_0(A,S)h(A,S)\}.
\label{eqn::general-eif-reward-centering}
\end{equation}
Applying \eqref{eqn::markov-score-inner-product} therefore gives
\[
\left.\frac{d}{d\varepsilon}\Psi(\mathbb P_\varepsilon)
\right|_{\varepsilon=0}
=\mathbb E_0\{\chi_0(O)\ell(O)\},
\]
with \(\chi_0\) as displayed in Theorem~\ref{theorem::EIF}.

Equation~\eqref{eqn::general-eif-component-centering},
Lemmas~\ref{lemma::initialpooledbounds}
and~\ref{lemma::markovscoreidentities}, and the definition of \(\chi_0\)
show that \(\chi_0\in L^2(\mathbb P_0)\) and belongs to the Markov tangent
space.
The map
\(\ell\mapsto\mathbb E_0\{\chi_0(O)\ell(O)\}\) is a bounded linear
functional on this tangent space. Bounded product scores are dense by
Lemma~\ref{lemma::markovscoreidentities}, and
Lemma~\ref{lemma::target-hellinger-lipschitz} gives local Hellinger
Lipschitz continuity of \(\Psi\). Lemma~\ref{lemma::dense-score-extension}
therefore gives, for every regular submodel with score \(\ell\),
\[
\left.\frac{d}{d\varepsilon}\Psi(\mathbb P_\varepsilon)
\right|_{\varepsilon=0}
=\mathbb E_0\{\chi_0(O)\ell(O)\}.
\]
Since \(\chi_0\) belongs to the tangent space, it is the canonical gradient.
\end{proof}

\begin{proof}[Proof of Theorem~\ref{theorem::vonmises}]
By Condition~\ref{cond::boundedfunc} and the definition of
$\operatorname{Rem}_0(r_0,k_0;\,r,k)$,
\begin{align*}
    \mathbb E_0[m(S_0,r,k)] - \mathbb E_0[m(S_0,r_0,k_0)]
    ={}& \mathbb E_0[\partial_r m(S_0,r_0,k_0)(r- r_0)] \\
    &+ \mathbb E_0\!\left[\partial_k m(S_0,r_0,k_0)\!\left(\tfrac{k - k_0}{k_0}\right)\right] \\
    &+ \operatorname{Rem}_0(r_0,k_0;\,r,k).
\end{align*}
Condition~\ref{cond::boundedfunc2} guarantees the Riesz representers
$\alpha_0$ and $\beta_0$, which satisfy
\[
\mathbb E_0\{\partial_r m(S_0,r_0,k_0)(r-r_0)\}
=\langle\alpha_0,r-r_0\rangle_{\mathcal H,0}
\]
and
\[
\mathbb E_0\!\left[\partial_k m(S_0,r_0,k_0)
\!\left(\tfrac{k-k_0}{k_0}\right)\right]
=\left\langle\beta_0,\tfrac{k-k_0}{k_0}\right\rangle_{\mathcal K,0}.
\]
Lemma~\ref{lemma::vonmiseskernel} further gives
\[
\mathbb E_0\!\left[\partial_k m(S_0,r_0,k_0)
\!\left(\tfrac{k-k_0}{k_0}\right)\right]
=-P_0\beta
+O\!\left(\|\beta_0-\beta\|_{L^2(P_0)}
\Big\|\tfrac{k-k_0}{k_0}\Big\|_{L^2(P_0)}\right).
\]
Hence,
   \begin{align*}
    \mathbb E_0[m(S_0,r,k)] - \mathbb E_0[m(S_0,r_0,k_0)]
    &=  \langle \alpha_0, r - r_0 \rangle_{\mathcal{H}, 0}  - P_0 \{\beta\} \\
    &\quad +  O\left(\|\beta_0
    - \beta\|_{L^2(P_0)} \;
   \Big\|\tfrac{k-k_0}{k_0}\Big\|_{L^2(P_0)}\right) \\
   & \quad + \operatorname{Rem}_0(r_0,k_0;\,r,k).
\end{align*}
It remains to expand
$\langle\alpha_0,r-r_0\rangle_{\mathcal H,0}$. Adding and subtracting
$\alpha$ gives
   \begin{align*}
   \langle \alpha_0, r - r_0 \rangle_{\mathcal{H}, 0} &=  \langle \alpha, r - r_0 \rangle_{\mathcal{H}, 0} +  \langle \alpha_0 - \alpha, r - r_0 \rangle_{\mathcal{H}, 0}.
\end{align*}
By definition,
$\langle\alpha,r-r_0\rangle_{\mathcal H,0}
=E_0\{\dot{\mathcal L}_{r_0}(\alpha,r-r_0)\}$. Lemma~\ref{lemma::quadraticloss}
therefore yields
\begin{align*}
 E_0[\dot{\mathcal{L}}_{r_0}(\alpha, r - r_0)] &= E_0[\mathcal L_{r}(\alpha)] - E_0[\mathcal L_{r_0}(\alpha)]  + O\left(E_0[\dot{\mathcal L}_{r_0}(r- r_0,r - r_0)(A,S)]\right)\\
 &= E_0[\mathcal L_{r}(\alpha)] - E_0[\mathcal L_{r_0}(\alpha)]  + O(\|r - r_0\|_{\mathcal{H}, 0}^2),
\end{align*}
where the implicit constant depends only on $M$. By
Lemma~\ref{thm:softmax-moment-matching}, the optimality conditions for $r_0$
imply $E_0\{\mathcal L_{r_0}(\alpha)\}=0$ for $\alpha\in\mathcal H$.
Thus,
\begin{align*}
 E_0[\dot{\mathcal{L}}_{r_0}(\alpha, r - r_0)] &= E_0[\mathcal L_{r}(\alpha)]  + O(\|r - r_0\|_{\mathcal{H}, 0}^2) \\
 &= P_0 \mathcal L_{r}(\alpha)  + O(\|r - r_0\|_{\mathcal{H}, 0}^2).
\end{align*}
Consequently,
\begin{align*}
   \langle \alpha_0, r - r_0 \rangle_{\mathcal{H}, 0} &=  P_0 \mathcal L_{r}(\alpha) +  \langle \alpha_0 - \alpha, r - r_0 \rangle_{\mathcal{H}, 0}  + O(\|r - r_0\|_{\mathcal{H}, 0}^2).
\end{align*}

Combining these identities, we obtain
\begin{align*}
    \mathbb E_0[m(S_0,r,k)] - \mathbb E_0[m(S_0,r_0,k_0)]
    &= P_0\!\left\{\mathcal{L}_{r}(\alpha) - \beta\right\}
    + \langle \alpha_0 - \alpha,\, r - r_0 \rangle_{\mathcal{H}, 0}  \\
    &\quad + O\!\left(\|\beta_0
    - \beta\|_{L^2(P_0)}
    \cdot \Big\|\tfrac{k-k_0}{k_0}\Big\|_{L^2(P_0)}\right)
    + \operatorname{Rem}_0(r_0,k_0;\,r,k).
\end{align*}
Rearranging terms and using the definition
$\varphi_{r,k}(\cdot;\alpha,\beta) = -\mathcal{L}_{r}(\alpha) + \beta$, we obtain
\begin{align*}
&\mathbb E_0[m(S_0,r,k)]-\mathbb E_0[m(S_0,r_0,k_0)]
\;+\; \int \varphi_{r,k}(s',a,s;\alpha,\beta)\, P_0(ds,da,ds') \\
 & \qquad   = -\langle r-r_0,\alpha-\alpha_0 \rangle_{\mathcal H,0}
   +O(\|r-r_0\|_{\mathcal H,0}^2)\\
 &\hspace{26mm}+O\!\Big(\|\beta_0-\beta\|_{L^2(P_0)}
    \Big\|\tfrac{k-k_0}{k_0}\Big\|_{L^2(P_0)}\Big)
    +\operatorname{Rem}_0(r_0,k_0;\,r,k).
\end{align*}
The implicit constants depend only on $M$.
This completes the proof.

\end{proof}

\begin{lemma}[von Mises expansion for kernel partial derivative]
\label{lemma::vonmiseskernel}
For every $k \in \mathcal{K}$ and $\beta \in T_{\mathcal K}(k)$,
\[
\mathbb E_0\!\left[\partial_k m(S_0,r_0,k_0)\!\left(\tfrac{k - k_0}{k_0}\right)\right]
=  - E_0[\beta(S',A,S)]
\;+\; O\!\left(\|\beta_0
- \beta\|_{L^2(P_0)} \;
   \Big\|\tfrac{k-k_0}{k_0}\Big\|_{L^2(P_0)}\right).
\]
\end{lemma}

\begin{proof}
By the Riesz identity defining
$\beta_0$ in \eqref{eqn::rieszkernel}, for all $k \in \mathcal{K}$,
\[
\mathbb E_0\!\left[\partial_k m(S_0,r_0,k_0)\!\left(\tfrac{k-k_0}{k_0}\right)\right]
= \left\langle \tfrac{k - k_0}{k_0}, \beta_0 \right\rangle_{L^2(P_0)}.
\]
Hence,
\begin{align*}
\mathbb E_0\!\left[\partial_k m(S_0,r_0,k_0)\!\left(\tfrac{k - k_0}{k_0}\right)\right]
&= \left\langle \tfrac{k - k_0}{k_0}, \beta_0 \right\rangle_{L^2(P_0)} \\[4pt]
&= E_0\!\left[ \int \beta_0(s',A,S)\,
\big(k(s'\!\mid\!A,S) - k_0(s'\!\mid\!A,S)\big)\,\mu(ds') \right] \\[4pt]
&= E_0\!\left[ \int \beta_0(s',A,S)\,k(s'\!\mid\!A,S)\,\mu(ds') \right],
\end{align*}
where
\(
\int \beta_0(s',A,S) k_0(s' \mid A, S)\,\mu(ds') = 0
\)
$P_0$-almost surely because $\beta_0 \in T_{\mathcal K}(k_0)$.

Adding and subtracting $\beta$,
\begin{align*}
\mathbb E_0\!\left[\partial_k m(S_0,r_0,k_0)\!\left(\tfrac{k - k_0}{k_0}\right)\right]
&=
E_0\!\left[
    \int
        \Bigg\{
            \beta_0
            -
            \beta
        \Bigg\}(s',A,S)\,
        k(s' \mid A,S)\,
    \mu(ds')
\right]  \\
&\quad
+\;
E_0\!\left[
    \int
        \beta(s',A,S)\,
        k(s' \mid A,S)\,
    \mu(ds')
\right]
\\[6pt]
&=
E_0\!\left[
    \int
        \Bigg\{
            \beta_0
            -
            \beta
        \Bigg\}(s',A,S)\,
        k(s' \mid A,S)\,
    \mu(ds')
\right],
\end{align*}
where we used that
\[
\int \beta(s',A,S)\,k(s' \mid A,S)\,\mu(ds') = 0
\]
$P_0$-almost surely, since $\beta \in T_{\mathcal K}(k)$.

Decomposing $k = (k-k_0)+k_0$, we obtain
\begin{align*}
\mathbb E_0\!\left[\partial_k m(S_0,r_0,k_0)\!\left(\tfrac{k - k_0}{k_0}\right)\right]
&=
E_0\!\left[
    \int
        \Bigg\{
            \beta_0
            -
            \beta
        \Bigg\}(s',A,S)\,
        (k - k_0)(s' \mid A,S)\,
    \mu(ds')
\right] \\
&\quad
-\;
E_0\!\left[
    \beta(S',A,S)
\right].
\end{align*}
Here we used that
\[
\int \beta_0(s',A,S)\,k_0(s'\!\mid\!A,S)\,\mu(ds') = 0
\]
$P_0$-almost surely, since $\beta_0 \in T_{\mathcal K}(k_0)$, and
\[
E_0\!\left[\int \beta(s',A,S)\,k_0(s'\!\mid\!A,S)\,\mu(ds')\right]
= E_0[\beta(S',A,S)].
\]

Rearranging,
\[
\begin{aligned}
&\mathbb E_0\!\left[\partial_k m(S_0,r_0,k_0)
\!\left(\tfrac{k-k_0}{k_0}\right)\right]\\
&\quad=-E_0[\beta(S',A,S)]
+E_0\!\left[\int \{\beta_0-\beta\}(s',A,S)
(k-k_0)(s'\mid A,S)\,\mu(ds')\right].
\end{aligned}
\]
Finally, the remainder is bounded by Cauchy--Schwarz:
\begin{align*}
&\left|E_0\!\left[ \int \{\beta_0
- \beta\}(s',A,S)\,(k-k_0)(s'\!\mid\!A,S)\,\mu(ds') \right]\right| \\[4pt]
&=\left|\Big\langle \beta_0
- \beta,\; \tfrac{k-k_0}{k_0} \Big\rangle_{L^2(P_0)}\right|
\;\le\; \|\beta_0
- \beta\|_{L^2(P_0)} \;
   \Big\|\tfrac{k-k_0}{k_0}\Big\|_{L^2(P_0)}.
\end{align*}
This completes the proof.
\end{proof}

\subsection{Proofs for Section~\ref{sec::eifnorm}}

\begin{proof}[Proof of Theorem~\ref{theorem::EIFnorm}]
We first verify Conditions~\ref{cond::boundedfunc}--\ref{cond::lipschitz}
for the composite map and then identify its two Riesz representers. Write
\[
\bar r_0:=r_{0,\nu,\gamma,g}.
\]
Define
\[
\mathcal R(r,k):=r_{r,k,\nu,\gamma,g}
=g+(I-\nu)\mathcal T_{k,\nu,\gamma}^{-1}(r-g),
\qquad
m(s,r,k):=\tilde m\{s,\mathcal R(r,k),k\},
\]
and let \(\mathcal T_0:=\mathcal T_{k_0,\nu,\gamma}\) and
\(V_0:=V_{r_0-g,k_0}^{\nu,\gamma}\). For
\(h\in T_{\mathcal H}\) and \(\beta\in T_{\mathcal K}\), define
\begin{equation}
\dot{\mathcal R}_r[h]
:=(I-\nu)\mathcal T_0^{-1}h,
\qquad
\dot{\mathcal R}_k[\beta]
:=(I-\nu)\gamma\mathcal T_0^{-1}
\mathcal P_{k_0\beta}V_0.
\label{eqn::normalization-derivatives}
\end{equation}
Lemma~\ref{lemma::tangentapproximation} gives
\begin{equation}
\|k_0\beta\|_{\mathcal K}\asymp
\|\beta\|_{\mathcal K,0}.
\label{eqn::normalized-kernel-norm-conversion}
\end{equation}
The action-centering operator \(I-\nu\) is bounded on \(L^2(\lambda)\),
and Assumption~\ref{cond::boundedpositivity} makes the
\(L^2(\lambda)\) and \(\mathcal H,0\) norms equivalent. Therefore,
Lemma~\ref{lemma::bellmanfrechet} and
\eqref{eqn::normalized-kernel-norm-conversion} give
\begin{align}
&\|\mathcal R\{r_0+h,(1+\beta)k_0\}-\mathcal R(r_0,k_0)
-\dot{\mathcal R}_r[h]-\dot{\mathcal R}_k[\beta]\|_{\mathcal H,0}
\notag\\
&\qquad\lesssim
\|\beta\|_{\mathcal K,0}^2\|r_0-g\|_{\mathcal H,0}
+\|\beta\|_{\mathcal K,0}\|h\|_{\mathcal H,0}
=o(\|h\|_{\mathcal H,0}+\|\beta\|_{\mathcal K,0}).
\label{eqn::normalization-map-frechet-remainder}
\end{align}

For the required local Lipschitz bound, let
\(b_j=(k_j-k_0)/k_0\), let
\(\mathcal T_j:=\mathcal T_{k_j,\nu,\gamma}\) and
\(q_j:=\mathcal T_j^{-1}(r_j-g)\). Since
\[
\mathcal T_j-\mathcal T_0
=-\gamma\mathcal P_{k_j-k_0}\nu,
\]
the operator bound \eqref{eqn::bellman-kernel-operator-bound} and
\eqref{eqn::normalized-kernel-norm-conversion} give
\[
\|\mathcal T_j-\mathcal T_0\|_{\mathrm{op}}
\lesssim\|b_j\|_{\mathcal K,0}.
\]
Lemma~\ref{lemma::inverseperturbation} therefore gives
\(\|\mathcal T_j^{-1}\|_{\mathrm{op}}\lesssim1\) on a sufficiently small
neighborhood. Consequently,
\[
\|q_j\|_{L^2(\lambda)}
\le \|\mathcal T_j^{-1}\|_{\mathrm{op}}
\{\|r_j-r_0\|_{L^2(\lambda)}+\|r_0-g\|_{L^2(\lambda)}\}
\lesssim1.
\]
Equation~\eqref{eqn::bellman-kernel-operator-bound}, applied to
\(k_1-k_2\), also gives
\begin{equation}
\max_{j\in\{1,2\}}\{\|\mathcal T_j^{-1}\|_{\mathrm{op}}
+\|q_j\|_{L^2(\lambda)}\}\lesssim1,
\qquad
\|\mathcal P_{k_1-k_2}\nu\|_{\mathrm{op}}
\lesssim\|b_1-b_2\|_{\mathcal K,0}.
\label{eqn::normalized-local-operator-bounds}
\end{equation}
The first resolvent identity gives the exact decomposition
\begin{equation}
q_1-q_2
=\mathcal T_1^{-1}(r_1-r_2)
+\gamma\mathcal T_1^{-1}\mathcal P_{k_1-k_2}\nu q_2.
\label{eqn::normalized-q-lipschitz-decomposition}
\end{equation}
Applying \eqref{eqn::normalized-local-operator-bounds}, norm equivalence, and
boundedness of \(I-\nu\) to
\eqref{eqn::normalized-q-lipschitz-decomposition} yields
\begin{equation}
\|\mathcal R(r_1,k_1)-\mathcal R(r_2,k_2)\|_{\mathcal H,0}
\lesssim \|r_1-r_2\|_{\mathcal H,0}
+\|b_1-b_2\|_{\mathcal K,0}.
\label{eqn::normalization-map-lipschitz}
\end{equation}

To verify Condition~\ref{cond::boundedfunc}, first fix bounded
\(h\in T_{\mathcal H}\) and bounded \(\beta\in T_{\mathcal K}\), and set
\[
\mathcal R_t:=\mathcal R\{r_0+th,(1+t\beta)k_0\},
\qquad
\overline{\mathcal R}_t
:=\mathcal R(r_0,k_0)
+t\{\dot{\mathcal R}_r[h]+\dot{\mathcal R}_k[\beta]\}.
\]
Equation~\eqref{eqn::normalization-map-frechet-remainder} and
Condition~\ref{cond::lipschitz::norm} imply
\begin{equation}
\|\tilde m\{\cdot,\mathcal R_t,(1+t\beta)k_0\}
-\tilde m\{\cdot,\overline{\mathcal R}_t,(1+t\beta)k_0\}\|_{L^2(\mu)}
=o(|t|).
\label{eqn::normalized-composition-path-transfer}
\end{equation}
Define the following \(L^2(\mu)\)-valued linear maps:
\begin{align}
\partial_r m(\cdot,r_0,k_0)[h]
&:=
\partial_r\tilde m(\cdot,\bar r_0,k_0)
\{\dot{\mathcal R}_r[h]\},
\label{eqn::normalized-composite-reward-derivative}\\
\partial_k m(\cdot,r_0,k_0)[\beta]
&:=
\partial_r\tilde m(\cdot,\bar r_0,k_0)
\{\dot{\mathcal R}_k[\beta]\}
+\partial_k\tilde m(\cdot,\bar r_0,k_0)(\beta).
\label{eqn::normalized-composite-kernel-derivative}
\end{align}
Condition~\ref{cond::boundedfunc::norm}, applied to
the straight path \((\overline{\mathcal R}_t,(1+t\beta)k_0)\), and
Lemma~\ref{lemma::initialpooledbounds}, applied to
\eqref{eqn::normalized-composition-path-transfer}, now give
\begin{align*}
&\left.\frac{d}{dt}\,
\mathbb E_0[m\{S_0,r_0+th,(1+t\beta)k_0\}]\right|_{t=0}\\
&\qquad=\mathbb E_0\{\partial_r m(S_0,r_0,k_0)[h]
+\partial_k m(S_0,r_0,k_0)[\beta]\}.
\end{align*}
This proves Condition~\ref{cond::boundedfunc}. The two functionals induced by
\eqref{eqn::normalized-composite-reward-derivative}--
\eqref{eqn::normalized-composite-kernel-derivative} are continuous by
Condition~\ref{cond::boundedfunc2::norm}, boundedness of the operators in
\eqref{eqn::normalization-derivatives}, and
\eqref{eqn::normalized-kernel-norm-conversion}. This proves
Condition~\ref{cond::boundedfunc2}. Finally, shrink the neighborhood so that
\((\mathcal R(r_j,k_j),k_j)\), \(j\in\{1,2\}\), satisfy
Condition~\ref{cond::lipschitz::norm}, that condition and
\eqref{eqn::normalization-map-lipschitz} give
\begin{align}
&\|m(\cdot,r_1,k_1)-m(\cdot,r_2,k_2)\|_{L^2(\mu)}\notag\\
&\qquad\lesssim
\|r_1-r_2\|_{\mathcal H,0}+\|b_1-b_2\|_{\mathcal K,0},
\label{eqn::normalized-composite-lipschitz}
\end{align}
which proves Condition~\ref{cond::lipschitz}. Thus
Theorem~\ref{theorem::EIF} applies to the composite map \(m\).

It remains to identify the representers. For the reward derivative, define
the linear functional
\[
\Lambda(w):=
\left\langle\tilde\alpha_0,
(I-\nu)\mathcal T_0^{-1}w\right\rangle_{\mathcal H,0},
\qquad w\in T_{\mathcal H}.
\]
Lemma~\ref{lemma::invertibility} and boundedness of \(I-\nu\) show that
\(\Lambda\) is continuous. The Riesz representation theorem therefore gives
a unique \(\alpha_0\in T_{\mathcal H}\) satisfying
\begin{equation}
\langle\alpha_0,w\rangle_{\mathcal H,0}=\Lambda(w),
\qquad w\in T_{\mathcal H}.
\label{eqn::normalized-reward-riesz-construction}
\end{equation}
Taking \(w=\mathcal T_0v\) in
\eqref{eqn::normalized-reward-riesz-construction} gives
\eqref{eqn::normalizedrieszbalance}; because \(\mathcal T_0\) is onto, this
balance equation determines \(\alpha_0\) uniquely. Moreover,
\eqref{eqn::normalized-composite-reward-derivative} and the Riesz identity for
\(\tilde\alpha_0\) give, for every \(h\in T_{\mathcal H}\),
\begin{align}
\mathbb E_0\{\partial_r m(S_0,r_0,k_0)[h]\}
&=\langle\tilde\alpha_0,\dot{\mathcal R}_r[h]\rangle_{\mathcal H,0}
=\Lambda(h)
=\langle\alpha_0,h\rangle_{\mathcal H,0}.
\label{eqn::normalized-reward-riesz-verification}
\end{align}
Thus \eqref{eqn::normalized-reward-riesz-verification} identifies
\(\alpha_0\) as the reward representer.

For any \(\beta\in T_{\mathcal K}\),
\eqref{eqn::normalized-composite-kernel-derivative} and the Riesz identities
for \(\tilde\alpha_0\) and \(\tilde\beta_0\) give
\begin{align*}
\mathbb E_0\{\partial_k m(S_0,r_0,k_0)[\beta]\}
={}&\langle\tilde\beta_0,\beta\rangle_{\mathcal K,0}\\
&+\left\langle\tilde\alpha_0,(I-\nu)\gamma
\mathcal T_{k_0,\nu,\gamma}^{-1}
\mathcal P_{k_0\beta}V_{r_0-g,k_0}^{\nu,\gamma}
\right\rangle_{\mathcal H,0}\\
={}&\langle\tilde\beta_0,\beta\rangle_{\mathcal K,0}
+\left\langle\alpha_0,\gamma
\mathcal P_{k_0\beta}V_{r_0-g,k_0}^{\nu,\gamma}
\right\rangle_{\mathcal H,0},
\end{align*}
where the last equality uses
\eqref{eqn::normalizedrieszbalance} with
\(v=\gamma\mathcal T_0^{-1}\mathcal P_{k_0\beta}V_0\). The definition of
\(\mathcal P_{k_0\beta}\), the conditional centering of \(\beta\), and the
Bellman identity
\[
q_{r_0-g,k_0}^{\nu,\gamma}
=r_0-g+\gamma\mathcal P_{k_0}V_{r_0-g,k_0}^{\nu,\gamma}
\]
give
\begin{align}
&\left\langle\alpha_0,\gamma
\mathcal P_{k_0\beta}V_{r_0-g,k_0}^{\nu,\gamma}
\right\rangle_{\mathcal H,0}\notag\\
&\quad=E_0\!\left[
\alpha_0(A,S)\gamma V_{r_0-g,k_0}^{\nu,\gamma}(S')
\beta(S',A,S)\right]\notag\\
&\quad=
E_0\!\left[
\alpha_0(A,S)
\{r_0(A,S)-g(S)+\gamma V_{r_0-g,k_0}^{\nu,\gamma}(S')
-q_{r_0-g,k_0}^{\nu,\gamma}(A,S)\}
\beta(S',A,S)
\right].
\label{eqn::normalized-kernel-riesz-calculation}
\end{align}
Moreover,
\begin{align}
&E_0\!\left[
r_0(A,S)-g(S)+\gamma V_{r_0-g,k_0}^{\nu,\gamma}(S')
-q_{r_0-g,k_0}^{\nu,\gamma}(A,S)
\mid A,S\right]\notag\\
&\qquad=r_0(A,S)-g(S)
+\gamma(\mathcal P_{k_0}V_{r_0-g,k_0}^{\nu,\gamma})(A,S)
-q_{r_0-g,k_0}^{\nu,\gamma}(A,S)=0.
\label{eqn::normalized-kernel-representer-centering}
\end{align}
The expression inside braces in
\eqref{eqn::normalized-kernel-representer-centering} is bounded because
\(r_0\), \(g\), \(q_{r_0-g,k_0}^{\nu,\gamma}\), and
\(V_{r_0-g,k_0}^{\nu,\gamma}\) are bounded. Since
\(\alpha_0\in L^2(P_0)\), the product of this expression and
\(\alpha_0(A,S)\) is in \(L^2(P_0)\). Because
\(\tilde\beta_0\in T_{\mathcal K}\),
\eqref{eqn::normalized-kernel-representer-centering} proves that the
\(\beta_0\) displayed in the theorem also belongs to \(T_{\mathcal K}\).
Equation~\eqref{eqn::normalized-kernel-riesz-calculation} verifies the Riesz
identity for every \(\beta\in T_{\mathcal K}\), so the displayed \(\beta_0\)
is the kernel representer.
Theorem~\ref{theorem::EIF} now gives the EIF stated in
Theorem~\ref{theorem::EIFnorm}.
\end{proof}
\section{Proofs for Section~\ref{sec::est}}

\begin{proof}[Proof of Theorem~\ref{thm:asymptotic_linearity}]
For \(j\in\{0,n\}\), define the trajectory-level uncentered influence
function
\[
\phi_j(O)
:=m(S_0,r_j,k_j)
+\frac1T\sum_{t=0}^{T-1}
\varphi_{r_j,k_j}(Z_t;\alpha_j,\beta_j).
\]
Then
\(\psi_n=\mathbb P_n\phi_n\),
\(\mathbb P_0\phi_0=\psi_0\), and
\(\chi_0=\phi_0-\psi_0\). Hence
\begin{align*}
\psi_n-\psi_0
={}&\mathbb P_n\chi_0
+(\mathbb P_n-\mathbb P_0)(\phi_n-\phi_0)
+\{\mathbb P_0\phi_n-\psi_0\}.
\end{align*}
Condition~\ref*{cond::consistency} and Jensen's inequality imply
\(\|\phi_n-\phi_0\|_{L^2(\mathbb P_0)}=o_P(1)\): the \(L^2\) norm of the
averaged transition difference is bounded by its pooled \(L^2(P_0)\) norm.
Lemma~\ref{lemma::crossfitprocess}, applied with
\(f_n=\phi_n-\phi_0\), therefore gives
\[
(\mathbb P_n-\mathbb P_0)(\phi_n-\phi_0)
=O_P\!\left(n^{-1/2}\|\phi_n-\phi_0\|_{L^2(\mathbb P_0)}\right)
=o_P(n^{-1/2}).
\]
Moreover,
\[
\mathbb P_0\phi_n-\psi_0
=\mathbb E_0\{m(S_0,r_n,k_n)-m(S_0,r_0,k_0)\}
+P_0\varphi_{r_n,k_n}(\cdot;\alpha_n,\beta_n).
\]
Theorem~\ref{theorem::vonmises}, applied with
\((r,k,\alpha,\beta)=(r_n,k_n,\alpha_n,\beta_n)\), and
Conditions~\ref*{cond::reward}--\ref*{cond::remainder} therefore imply
\(\mathbb P_0\phi_n-\psi_0=o_P(n^{-1/2})\). Consequently,
\[
\psi_n-\psi_0
=\frac1n\sum_{i=1}^n\chi_0(O_i)+o_P(n^{-1/2}).
\]
The central limit theorem for the independent trajectories gives the stated
normal limit with variance
\(\operatorname{Var}_{\mathbb P_0}\{\chi_0(O)\}\). Efficiency follows from
Theorem~\ref{theorem::EIF}.
\end{proof}

\section{Proofs for Section~\ref{sec::applications}}

\label{appendix::applications}
\subsection{Technical lemmas}

We begin with operator and occupancy-ratio identities that are reused across the
example-specific EIF derivations and asymptotic expansions.

For a fixed Markov policy \(\pi\), write
\((\Pi f)(s):=\sum_{a\in\mathcal A}\pi(a\mid s)f(a,s)\). Thus
\(\mathcal T_{k,\pi,\gamma}=I-\gamma\mathcal P_k\Pi\).

\begin{lemma}[Uniform discounted-resolvent bound]
\label{lemma::invertibility}
Under Assumption~\ref{cond::boundedpositivity}, for every Markov policy
\(\pi\) and \(\delta\in[0,1)\),
\(\mathcal T_{k_0,\pi,\delta}^{-1}\) exists and is bounded on both
\(L^2(\lambda)\) and \(L^2(\pi_0\otimes\rho_0)\). For each fixed \(\delta\),
the operator-norm bound is uniform over \(\pi\).
\end{lemma}
\begin{proof}
Define the Markov operator on state--action functions by
\[
\mathcal M_{k_0,\pi}q(a,s)
:=\int\sum_{a'\in\mathcal A}\pi(a'\mid s')q(a',s')
k_0(s'\mid a,s)\,\mu(ds'),
\]
so that \(\mathcal T_{k_0,\pi,\delta}=I-\delta\mathcal M_{k_0,\pi}\).
Jensen's inequality gives
\((\mathcal M_{k_0,\pi}^tq)^2\le
\mathcal M_{k_0,\pi}^t(q^2)\). Let \(w_t\) be the density of
\((\mathcal M_{k_0,\pi}^t)^*\lambda\) with respect to \(\lambda\).
For \(t\ge1\),
\[
w_t(a',s')
=\pi(a'\mid s')\int
k_0(s'\mid a,s)w_{t-1}(a,s)\,\lambda(da,ds).
\]
The adjoint operation preserves total mass. Since \(k_0\le C_0\),
\[
0\le w_t(a',s')\le C_0\lambda(\mathcal A\times\mathcal S),
\qquad t\ge1,
\]
uniformly over \(\pi\). Therefore,
\[
\|\mathcal M_{k_0,\pi}^tq\|_{L^2(\lambda)}^2
\le\int q^2\,d\{(\mathcal M_{k_0,\pi}^t)^*\lambda\}
\lesssim\|q\|_{L^2(\lambda)}^2.
\]
Consequently,
\[
\|\mathcal M_{k_0,\pi}^tq\|_{L^2(\lambda)}
\lesssim \|q\|_{L^2(\lambda)}
\]
uniformly in \(t\ge1\) and \(\pi\); the same bound holds for \(t=0\).
When \(\delta=0\), \(\mathcal T_{k_0,\pi,0}=I\), so its inverse has
operator norm one. Hence suppose
\(\delta\in(0,1)\). For integers \(n\ge m\ge0\), weighted
Cauchy--Schwarz gives
\begin{align}
&\left\|\sum_{t=m}^n\delta^t
\mathcal M_{k_0,\pi}^tq\right\|_{L^2(\lambda)}^2\notag\\
&\quad\le
\left(\sum_{t=m}^n\delta^t\right)
\left(\sum_{t=m}^n\delta^t
\|\mathcal M_{k_0,\pi}^tq\|_{L^2(\lambda)}^2\right)
\le \frac{C\delta^{2m}}{(1-\delta)^2}
\|q\|_{L^2(\lambda)}^2.
\label{eqn::discounted-resolvent-tail}
\end{align}
Thus the partial sums
\(R_N:=\sum_{t=0}^N\delta^t\mathcal M_{k_0,\pi}^t\) are Cauchy in
operator norm, uniformly over \(\pi\). If \(R\) denotes their limit, then
\[
(I-\delta\mathcal M_{k_0,\pi})R_N
=R_N(I-\delta\mathcal M_{k_0,\pi})
=I-\delta^{N+1}\mathcal M_{k_0,\pi}^{N+1}.
\]
The uniform power bound implies
\(\|\delta^{N+1}\mathcal M_{k_0,\pi}^{N+1}\|_{\mathrm{op}}\to0\).
Passing to the operator-norm limit proves
\[
(I-\delta\mathcal M_{k_0,\pi})R
=R(I-\delta\mathcal M_{k_0,\pi})=I,
\]
so \(R=\mathcal T_{k_0,\pi,\delta}^{-1}\).
Equation~\eqref{eqn::discounted-resolvent-tail} with \(m=0\) gives the
claimed inverse bound, uniformly over \(\pi\). Finally,
Assumption~\ref{cond::boundedpositivity} makes the \(L^2(\lambda)\) and
\(L^2(\pi_0\otimes\rho_0)\) norms equivalent, which proves the second claim.
\end{proof}

For a policy \(\pi\), kernel \(k\), and discount \(\gamma\), define the
discounted occupancy ratios, initialized from \(\rho_{0,\mathrm{in}}\) and
measured relative to the pooled design law, by
\[
\rho_{0,k}^{\pi,\gamma}(s)
:=\frac1{\rho_0(s)}\sum_{t=0}^\infty\gamma^t
\frac{d\mathbb P_{\rho_{0,\mathrm{in}},k,\pi}(S_t\in\cdot)}{d\mu}(s),
\qquad
d_{0,k}^{\pi,\gamma}(a,s)
:=\rho_{0,k}^{\pi,\gamma}(s)\frac{\pi(a\mid s)}{\pi_0(a\mid s)}.
\]
For \(k=k_0\), these agree with the ratios defined in
Section~\ref{sec::applications}.

\begin{lemma}[Bounded discounted-occupancy ratios]
\label{lemma::boundedoccupancy}
Under Assumption~\ref{cond::boundedpositivity}, for every Markov policy
\(\pi\) and \(\delta\in[0,1)\),
\(d_{0,k_0}^{\pi,\delta}\in L^\infty(\lambda)\). For each fixed \(\delta\),
the bound is uniform over \(\pi\).
\end{lemma}

\begin{proof}
Let \(p_t^\pi\) denote the \(\mu\)-density of \(S_t\) under the target
policy \(\pi\), kernel \(k_0\), and initial density
\(\rho_{0,\mathrm{in}}\). Since
\(\rho_{0,\mathrm{in}}\le T\rho_0\), Assumption~\ref{cond::boundedpositivity}
gives \(p_0^\pi\le TC\). For every \(t\ge1\),
\[
p_t^\pi(s')
=\int\sum_{a\in\mathcal A}\pi(a\mid s)k_0(s'\mid a,s)
p_{t-1}^\pi(s)\,\mu(ds)
\le C.
\]
Hence
\[
\rho_{0,k_0}^{\pi,\delta}(s)
\le \frac{TC+C\delta/(1-\delta)}{c},
\qquad
d_{0,k_0}^{\pi,\delta}(a,s)
\le \frac{TC+C\delta/(1-\delta)}{c\eta}
\]
for \(\lambda\)-almost every \((a,s)\), proving the claim.
\end{proof}

\begin{lemma}[Discounted-occupancy identity]
\label{lemma::adjointident}
Under Assumption~\ref{cond::boundedpositivity}, for every Markov policy
\(\pi\), \(\delta\in[0,1)\), and \(f\in L^2(\lambda)\),
\[
\mathbb E_0\!\left[(\pi\mathcal T_{k_0,\pi,\delta}^{-1}f)(S_0)\right]
=E_0\!\left[d_{0,k_0}^{\pi,\delta}(A,S)f(A,S)\right].
\]
\end{lemma}

\begin{proof}
By Lemma~\ref{lemma::invertibility},
\[
\pi\mathcal T_{k_0,\pi,\delta}^{-1}f
=\sum_{t=0}^\infty\delta^t
\pi\mathcal M_{k_0,\pi}^tf
\quad\text{in }L^2(\mu).
\]
If \(p_t^\pi\) is the state density at time \(t\) under
\((\rho_{0,\mathrm{in}},k_0,\pi)\), then
\begin{align*}
\mathbb E_0[(\pi\mathcal T_{k_0,\pi,\delta}^{-1}f)(S_0)]
&=\sum_{t=0}^\infty\delta^t
\int p_t^\pi(s)\sum_a\pi(a\mid s)f(a,s)\,\mu(ds)\\
&=E_0[d_{0,k_0}^{\pi,\delta}(A,S)f(A,S)].
\end{align*}
The final equality is the definition of
\(d_{0,k_0}^{\pi,\delta}\); Lemma~\ref{lemma::boundedoccupancy} justifies
the interchange of the series and the integral.
\end{proof}

\begin{lemma}[Differentiability of the fixed-policy Bellman solution]
\label{lemma::bellmanfrechet}
Suppose $\mathcal{T}_{k,\pi,\gamma}:L^2(\lambda)\to L^2(\lambda)$ has a
bounded inverse. Then the map
$(r,k)\mapsto \mathcal{T}_{k,\pi,\gamma}^{-1} r$ is Fr\'echet differentiable
in $L^2$, with partial derivatives, for $g \in \addTK$ and
$h\in L^2(\lambda)$,
\[
\adk\bigl(\mathcal{T}_{k,\pi,\gamma}^{-1}\bigr)[g]
= \gamma \,\mathcal{T}_{k,\pi,\gamma}^{-1}\bigl(\mathcal{P}_{g}\Pi \bigr)\mathcal{T}_{k,\pi,\gamma}^{-1},
\qquad
\partial_r\bigl(\mathcal{T}_{k,\pi,\gamma}^{-1} r\bigr)[h] = \mathcal{T}_{k,\pi,\gamma}^{-1} h.
\]
Moreover, for all sufficiently small $g$ such that $k+g\in\mathcal K$,
\[
\|\mathcal T_{k+g,\pi,\gamma}^{-1}(r+h)-\mathcal T_{k,\pi,\gamma}^{-1}(r)-\adk\mathcal T_{k,\pi,\gamma}^{-1}[g](r)- \partial_r (\mathcal T_{k,\pi,\gamma}^{-1} r)[h]\|_{L^2(\lambda)}
\;\lesssim\; \|g\|_{\mathcal K}^2\,\|r\|_{L^2(\lambda)} \;+\; \|g\|_{\mathcal K}\,\|h\|_{L^2(\lambda)}.
\]
\end{lemma}
\begin{proof}
If \(\gamma=0\), then
\(\mathcal T_{k,\pi,0}=I\) for every \(k\). The kernel derivative and the
remainder in the lemma are therefore zero, while the reward derivative is the
identity. Hence the result holds when \(\gamma=0\). Suppose for the remainder
of the proof that \(\gamma>0\).

Set \(A=\mathcal T_{k,\pi,\gamma}\) and
\(\Delta_g=\mathcal T_{k+g,\pi,\gamma}-\mathcal T_{k,\pi,\gamma}\).
Lemma~\ref{lemma::inverseperturbation} gives
\begin{equation}
\adk\bigl(\mathcal{T}_{k,\pi,\gamma}^{-1}\bigr)[g]
= -\,\mathcal{T}_{k,\pi,\gamma}^{-1}\bigl(\adk \mathcal{T}_{k,\pi,\gamma}[g]\bigr)\mathcal{T}_{k,\pi,\gamma}^{-1},
\label{eqn::bellman-inverse-derivative}
\end{equation}
where $\adk \mathcal{T}_{k,\pi,\gamma}[g] = - \gamma \mathcal{P}_{g}\Pi $.
Adding and subtracting terms gives
\begin{align*}
&\mathcal{T}_{k+g,\pi,\gamma}^{-1} - \mathcal{T}_{k,\pi,\gamma}^{-1}
- \bigl(\adk \mathcal{T}_{k,\pi,\gamma}^{-1}\bigr)[g] \\
&\quad= \mathcal{T}_{k,\pi,\gamma}^{-1}\,\bigl(\mathcal{T}_{k+g,\pi,\gamma} - \mathcal{T}_{k,\pi,\gamma}\bigr)\,
\mathcal{T}_{k,\pi,\gamma}^{-1}\,\bigl(\mathcal{T}_{k+g,\pi,\gamma} - \mathcal{T}_{k,\pi,\gamma}\bigr)\,
\mathcal{T}_{k+g,\pi,\gamma}^{-1} \\
&\qquad\;+\;
\mathcal{T}_{k,\pi,\gamma}^{-1}\Bigl(\adk \mathcal{T}_{k,\pi,\gamma}[g] -
\bigl(\mathcal{T}_{k+g,\pi,\gamma} - \mathcal{T}_{k,\pi,\gamma}\bigr)\Bigr)\mathcal{T}_{k,\pi,\gamma}^{-1}.
\end{align*}
\[
\mathcal{T}_{k+g,\pi,\gamma} - \mathcal{T}_{k,\pi,\gamma} = - \gamma \mathcal{P}_{g}\Pi .
\]
\[
\adk \mathcal{T}_{k,\pi,\gamma}[g] -
\bigl(\mathcal{T}_{k+g,\pi,\gamma} - \mathcal{T}_{k,\pi,\gamma}\bigr) = 0,
\]
since \(k \mapsto \mathcal{T}_{k,\pi,\gamma} = I - \gamma \mathcal{P}_k \Pi\) is affine in \(k\).
Thus, the final term vanishes, and we obtain
\begin{align*}
&\mathcal{T}_{k+g,\pi,\gamma}^{-1} - \mathcal{T}_{k,\pi,\gamma}^{-1}
- \bigl(\adk \mathcal{T}_{k,\pi,\gamma}^{-1}\bigr)[g] \\
&\quad=  \gamma^2 \mathcal{T}_{k,\pi,\gamma}^{-1}\,\bigl( \mathcal{P}_{g}\Pi \bigr)\,
\mathcal{T}_{k,\pi,\gamma}^{-1}\,\bigl( \mathcal{P}_{g}\Pi \bigr)\,
\mathcal{T}_{k+g,\pi,\gamma}^{-1}.
\end{align*}
Taking operator norms and using submultiplicativity,
\[
\begin{aligned}
&\bigl\|\mathcal{T}_{k+g,\pi,\gamma}^{-1} - \mathcal{T}_{k,\pi,\gamma}^{-1}
- (\adk \mathcal{T}_{k,\pi,\gamma}^{-1})[g]\bigr\|_{L^2(\lambda)\to L^2(\lambda)}\\
&\quad\le
\|\mathcal{T}_{k,\pi,\gamma}^{-1}\|_{L^2(\lambda)\to L^2(\lambda)}^2
\|\mathcal{T}_{k+g,\pi,\gamma}^{-1}\|_{L^2(\lambda)\to L^2(\lambda)}\\
&\qquad\quad\times
\|\mathcal{P}_{g}\Pi \|_{L^2(\lambda)\to L^2(\lambda)}^{2}.
\end{aligned}
\]
Equation~\eqref{eqn::bellman-kernel-operator-bound}, proved below, gives
\(\|\mathcal P_g\Pi\|_{L^2(\lambda)\to L^2(\lambda)}
\le C\|g\|_{\mathcal K}\) for some \(C<\infty\). This bound gives
\[
\bigl\|\mathcal{T}_{k+g,\pi,\gamma}^{-1} - \mathcal{T}_{k,\pi,\gamma}^{-1}
- (\adk \mathcal{T}_{k,\pi,\gamma}^{-1})[g]\bigr\|_{L^2(\lambda)\to L^2(\lambda)}
\;\lesssim\; \,  \,\|\mathcal{T}_{k,\pi,\gamma}^{-1}\|_{L^2(\lambda)\to L^2(\lambda)}^{2}\,\|\mathcal{T}_{k+g,\pi,\gamma}^{-1}\|_{L^2(\lambda)\to L^2(\lambda)}\;
\|g\|_{\mathcal K}^{2}.
\]
Let \(M=\|\mathcal T_{k,\pi,\gamma}^{-1}\|_{L^2\to L^2}\). Since
\(\|\Delta_g\|_{L^2\to L^2}\le\gamma C\|g\|_{\mathcal K}\), the condition
\(\|g\|_{\mathcal K}<(2\gamma CM)^{-1}\) implies
\(\|A^{-1}\Delta_g\|<1/2\). Lemma~\ref{lemma::inverseperturbation} then
gives
\[
\|\mathcal T_{k+g,\pi,\gamma}^{-1}\|_{L^2\to L^2}\le2M.
\]
Consequently,
\begin{equation}
\bigl\|\mathcal{T}_{k+g,\pi,\gamma}^{-1} - \mathcal{T}_{k,\pi,\gamma}^{-1}
- (\adk \mathcal{T}_{k,\pi,\gamma}^{-1})[g]\bigr\|_{L^2(\lambda)\to L^2(\lambda)}
= O\!\left(\|g\|_{\mathcal K}^{2}\right),
\label{eqn::bellman-inverse-quadratic-remainder}
\end{equation}
as claimed.

Since $r \mapsto \mathcal T_{k,\pi,\gamma}^{-1} r$ is linear,
$\partial_r \big(\mathcal T_{k,\pi,\gamma}^{-1} r\big)[h]
=\mathcal T_{k,\pi,\gamma}^{-1}h$. Adding and subtracting gives
\[
\begin{aligned}
&\mathcal T_{k+g,\pi,\gamma}^{-1}(r+h)
-\mathcal T_{k,\pi,\gamma}^{-1}(r)
-\adk\mathcal T_{k,\pi,\gamma}^{-1}[g](r)
-\partial_r \big(\mathcal T_{k,\pi,\gamma}^{-1} r\big)[h]\\
&\quad =
\underbrace{
\big(\mathcal T_{k+g,\pi,\gamma}^{-1}
-\mathcal T_{k,\pi,\gamma}^{-1}
-\adk\mathcal T_{k,\pi,\gamma}^{-1}[g]\big)r
}_{\text{quadratic in }g}
+\underbrace{
\big(\mathcal T_{k+g,\pi,\gamma}^{-1}
-\mathcal T_{k,\pi,\gamma}^{-1}\big)h
}_{\text{linear in }g}.
\end{aligned}
\]
Hence, by the triangle inequality and operator-norm submultiplicativity,
\begin{align*}
\left\|
\begin{aligned}
&\mathcal T_{k+g,\pi,\gamma}^{-1}(r+h)
-\mathcal T_{k,\pi,\gamma}^{-1}(r)\\
&\quad -\adk\mathcal T_{k,\pi,\gamma}^{-1}[g](r)
-\mathcal T_{k,\pi,\gamma}^{-1}h
\end{aligned}
\right\|_{L^2(\lambda)}
&\le
\left\|
\begin{aligned}
&\mathcal T_{k+g,\pi,\gamma}^{-1}
-\mathcal T_{k,\pi,\gamma}^{-1}\\
&\quad -\adk\mathcal T_{k,\pi,\gamma}^{-1}[g]
\end{aligned}
\right\|_{L^2(\lambda)\to L^2(\lambda)}
\,\|r\|_{L^2(\lambda)} \\
&\quad + \|\mathcal T_{k+g,\pi,\gamma}^{-1}-\mathcal T_{k,\pi,\gamma}^{-1}\|_{L^2(\lambda)\to L^2(\lambda)}\,\|h\|_{L^2(\lambda)}.
\end{align*}
Equation~\eqref{eqn::bellman-inverse-quadratic-remainder} gives
\[
\|\mathcal T_{k+g,\pi,\gamma}^{-1}-\mathcal T_{k,\pi,\gamma}^{-1}-\adk\mathcal T_{k,\pi,\gamma}^{-1}[g]\|_{L^2(\lambda)\to L^2(\lambda)} \lesssim \|g\|_{\mathcal K}^2,
\]
and \eqref{eqn::bellman-inverse-derivative},
\eqref{eqn::bellman-kernel-operator-bound}, and the triangle inequality give
\[
\|\mathcal T_{k+g,\pi,\gamma}^{-1}-\mathcal T_{k,\pi,\gamma}^{-1}\|_{L^2(\lambda)\to L^2(\lambda)} \lesssim \|\adk\mathcal T_{k,\pi,\gamma}^{-1}[g]\|_{L^2(\lambda)\to L^2(\lambda)} + \|g\|_{\mathcal K}^2 \lesssim \|g\|_{\mathcal K} + \|g\|_{\mathcal K}^2.
\]
Hence,
\[
\|\mathcal T_{k+g,\pi,\gamma}^{-1}(r+h)-\mathcal T_{k,\pi,\gamma}^{-1}(r)-\adk\mathcal T_{k,\pi,\gamma}^{-1}[g](r)- \partial_r (\mathcal T_{k,\pi,\gamma}^{-1} r)[h]\|_{L^2(\lambda)}
\;\lesssim\; \|g\|_{\mathcal K}^2\,\|r\|_{L^2(\lambda)} \;+\; \|g\|_{\mathcal K}\,\|h\|_{L^2(\lambda)}.
\]

It remains to prove the claim that $\|\mathcal{P}_{g}\Pi \|_{L^2(\lambda) \rightarrow L^2(\lambda)} \lesssim \|g\|_{\mathcal{K}}$. For $(a,s)\in\mathcal{A}\times\mathcal{S}$, write
\[
(\mathcal{P}_{g}\Pi f)(a,s)
= \int_{\mathcal S} g(s' \mid a,s)\,h(s')\,\mu(ds'),
\qquad
h(s') := \sum_{a'\in\mathcal A} \pi(a' \mid s')\, f(a',s').
\]
By Cauchy--Schwarz in $L^2(\mu)$,
\[
|(\mathcal{P}_{g}\Pi f)(a,s)|
\;\le\; \Big(\int g(s'\mid a,s)^2\,\mu(ds')\Big)^{1/2}
       \Big(\int h(s')^2\,\mu(ds')\Big)^{1/2}.
\]
Next, by Cauchy--Schwarz over $a'\in\mathcal A$,
\[
h(s')^2 \;\le\; \Big(\sum_{a'} \pi(a'\mid s')^2\Big)\Big(\sum_{a'} f(a',s')^2\Big).
\]
Integrating in $s'$ and letting $c_\pi := \sup_{s'}\sum_{a'} \pi(a'\mid s')^2\le 1$, we get
\[
\int h(s')^2\,\mu(ds') \;\le\; c_\pi \int \sum_{a'} f(a',s')^2\,\mu(ds')
= c_\pi\,\|f\|_{L^2(\#_{\mathcal A}\otimes \mu)}^2.
\]
Therefore,
\[
|(\mathcal{P}_{g}\Pi f)(a,s)|^2
\;\le\; c_\pi\Big(\int g(s'\mid a,s)^2\,\mu(ds')\Big)\,\|f\|_{L^2(\#_{\mathcal A}\otimes \mu)}^2.
\]
Integrating over $(a,s)$ with respect to $\lambda(da,ds)$ yields
\[
\|\mathcal{P}_{g}\Pi f\|_{L^2(\lambda)}^2
\;\le\; c_\pi \Big(\int\!\!\int g(s'\mid a,s)^2\,\mu(ds')\,\lambda(da,ds)\Big)
          \|f\|_{L^2(\#_{\mathcal A}\otimes \mu)}^2
= c_\pi\,\|g\|_{L^2(\mu\otimes \lambda)}^2\,\|f\|_{L^2(\#_{\mathcal A}\otimes \mu)}^2.
\]
Taking square roots and using
$L^2(\#_{\mathcal A}\otimes \mu)=L^2(\lambda)$ gives
\[
\|\mathcal{P}_{g}\Pi f\|_{L^2(\lambda)}
\lesssim \|g\|_{L^2(\mu\otimes \lambda)}\,\|f\|_{L^2(\lambda)},
\]
and hence
\begin{equation}
\|\mathcal P_g\Pi\|_{L^2(\lambda)\to L^2(\lambda)}
\lesssim\|g\|_{\mathcal K}.
\label{eqn::bellman-kernel-operator-bound}
\end{equation}
This completes the proof of the derivative and remainder claims.
\end{proof}

\subsubsection{Lemmas for Example~\ref{example::norm}}
\label{lemmas::norm}

For $(r,k) \in L^\infty(\lambda) \times \mathcal K$, define the
normalized policy-value operator
\[
\mathcal{V}_k^g(r) := \Pi\,\mathcal{T}_{k,\pi,\gamma'}^{-1}\!\left\{g + (I - \nu)\,\mathcal{T}_{k,\nu,\gamma}^{-1}(r-g)\right\}.
\]
The quantity $\mathcal{V}_k^g(r)$ is the value of policy $\pi$ under the
$\nu$-normalized reward
\(r_{r,k,\nu,\gamma,g}=g+(I-\nu)\mathcal{T}_{k,\nu,\gamma}^{-1}(r-g)\).

\begin{lemma}[Differentiability of the normalized policy-value operator]
 \label{lemma::normalizedvaluedifferentiable}
Let $\pi,\nu$ be policies. Suppose $\mathcal{T}_{k,\pi,\gamma'}:L^2(\lambda)\to L^2(\lambda)$ and
$\mathcal{T}_{k,\nu,\gamma}:L^2(\lambda)\to L^2(\lambda)$ are invertible,
and that both inverse operators exist and are uniformly bounded over a
\(\|\cdot\|_{\mathcal K}\)-neighborhood of \(k\).
Then the map $(r,k) \mapsto \mathcal{V}_k^g(r)$ is Fr\'echet differentiable
from $L^2(\lambda)\times\mathcal K$ to $L^2(\mu)$, with derivatives
\begin{align*}
\partial_r \mathcal{V}_k^g(r)[h]
&= \Pi\,\mathcal{T}_{k,\pi,\gamma'}^{-1}(I-\nu)\,\mathcal{T}_{k,\nu,\gamma}^{-1}(h), \\[6pt]
\adk \mathcal{V}_k^g(r)[\eta]
&= \gamma'\,\Pi\,\mathcal{T}_{k,\pi,\gamma'}^{-1}(\mathcal{P}_\eta \Pi)\,
      \mathcal{T}_{k,\pi,\gamma'}^{-1}\!\left\{g + (I-\nu)\,\mathcal{T}_{k,\nu,\gamma}^{-1}(r-g)\right\} \\
&\quad+\; \gamma\,\Pi\,\mathcal{T}_{k,\pi,\gamma'}^{-1}(I-\nu)\,
      \mathcal{T}_{k,\nu,\gamma}^{-1}(\mathcal{P}_\eta \nu)\,
      \mathcal{T}_{k,\nu,\gamma}^{-1}(r-g).
\end{align*}
Moreover, the second-order remainder satisfies
\begin{align*}
&\|\mathcal V_{k+\eta}^g(r+h)-\mathcal V_{k}^g(r)
-\adk\mathcal V_{k}^g(r)[\eta]
-\partial_r\mathcal V_{k}^g(r)[h]\|_{L^2(\mu)}\\
&\qquad\lesssim \|\eta\|_{\mathcal K}^2
+\|\eta\|_{\mathcal K}\,\|h\|_{L^2(\lambda)},
\end{align*}
where the implicit constant is uniform on bounded \(L^2\)-neighborhoods of
\(r\) and depends on the local inverse bounds and the operator norms of
\(\Pi\) and \((I-\nu)\).
\end{lemma}
\begin{proof}
Linearity in $r$ gives
\[
\partial_r \mathcal{V}_k^g(r)[h]
=\Pi\mathcal{T}_{k,\pi,\gamma'}^{-1}(I-\nu)
\mathcal{T}_{k,\nu,\gamma}^{-1}h.
\]

We now turn to the additive kernel derivative. By
Lemma~\ref{lemma::bellmanfrechet}, for $\eta \in \addTK$,
\begin{equation}
\label{eqn::normalized-resolvent-derivatives}
\adk\bigl(\mathcal{T}_{k,\nu,\gamma}^{-1}\bigr)[\eta]
= \gamma \,\mathcal{T}_{k,\nu,\gamma}^{-1}
\bigl(\mathcal{P}_{\eta}\nu \bigr)
\mathcal{T}_{k,\nu,\gamma}^{-1},
\end{equation}
and, with \(\pi\) and \(\gamma'\) in place of \(\nu\) and \(\gamma\),
\begin{equation}
\label{eqn::evaluation-resolvent-derivatives}
\adk\bigl(\mathcal{T}_{k,\pi,\gamma'}^{-1}\bigr)[\eta]
\;=\; \gamma' \,\mathcal{T}_{k,\pi,\gamma'}^{-1}\bigl(\mathcal{P}_{\eta}\Pi \bigr)\mathcal{T}_{k,\pi,\gamma'}^{-1},
\end{equation}
The product rule and
\eqref{eqn::normalized-resolvent-derivatives}--
\eqref{eqn::evaluation-resolvent-derivatives} give
\begin{align*}
\adk \mathcal{V}_k^g(r)[\eta]
&= \Pi\Bigl(\adk \mathcal{T}_{k,\pi,\gamma'}^{-1}[\eta]\Bigr)
      \left\{g + (I-\nu)\,\mathcal{T}_{k,\nu,\gamma}^{-1}(r-g)\right\}
  \;+\; \Pi\,\mathcal{T}_{k,\pi,\gamma'}^{-1}
      (I-\nu)\,\Bigl(\adk \mathcal{T}_{k,\nu,\gamma}^{-1}[\eta]\Bigr)(r-g).
\end{align*}
Equivalently, using \(r_{r,k,\nu,\gamma,g}= g + (I-\nu)\,\mathcal{T}_{k,\nu,\gamma}^{-1}(r-g)\) and \(q_{r-g,k}^{\nu,\gamma} =  \mathcal{T}_{k,\nu,\gamma}^{-1}(r-g)\),
\begin{align*}
\adk \mathcal{V}_k^g(r)[\eta]
&= \gamma'\,\Pi\,\mathcal{T}_{k,\pi,\gamma'}^{-1}(\mathcal{P}_\eta \Pi)\,
      \mathcal{T}_{k,\pi,\gamma'}^{-1}\bigl(r_{r,k,\nu,\gamma,g}\bigr) \\
&\quad+\; \gamma\,\Pi\,\mathcal{T}_{k,\pi,\gamma'}^{-1}(I-\nu)\,
      \mathcal{T}_{k,\nu,\gamma}^{-1}(\mathcal{P}_\eta \nu)\,
      q_{r-g,k}^{\nu,\gamma}.
\end{align*}

We now turn to the second-order remainder. Adding and subtracting, we have
\begin{align*}
\mathcal{V}_{k+\eta}^g(r)-\mathcal{V}_k^g(r)
&=\Pi\big(\mathcal T_{k+\eta,\pi,\gamma'}^{-1}
-\mathcal T_{k,\pi,\gamma'}^{-1}\big)
\left\{g+(I-\nu)\mathcal T_{k,\nu,\gamma}^{-1}(r-g)\right\}\\
&\quad+\Pi\mathcal T_{k+\eta,\pi,\gamma'}^{-1}(I-\nu)
\big(\mathcal T_{k+\eta,\nu,\gamma}^{-1}
-\mathcal T_{k,\nu,\gamma}^{-1}\big)(r-g).
\end{align*}
Subtracting the linear term
\[
\adk \mathcal{V}_k^g(r)[\eta]
=\Pi\big(\adk\mathcal T_{k,\pi,\gamma'}^{-1}[\eta]\big)
\left\{g+(I-\nu)\mathcal T_{k,\nu,\gamma}^{-1}(r-g)\right\}
+\Pi\mathcal T_{k,\pi,\gamma'}^{-1}(I-\nu)
\big(\adk\mathcal T_{k,\nu,\gamma}^{-1}[\eta]\big)(r-g),
\]
and adding and subtracting $\Pi\,\mathcal{T}_{k,\pi,\gamma'}^{-1}(I-\nu)\big(\mathcal{T}_{k+\eta,\nu,\gamma}^{-1}-\mathcal{T}_{k,\nu,\gamma}^{-1}\big)(r-g)$, we obtain
\begin{align}
&\mathcal{V}_{k+\eta}^g(r) - \mathcal{V}_k^g(r) - \adk \mathcal{V}_k^g(r)[\eta] \\
&= \Pi\Big\{\big(\mathcal{T}_{k+\eta,\pi,\gamma'}^{-1}-\mathcal{T}_{k,\pi,\gamma'}^{-1}
      - \adk \mathcal{T}_{k,\pi,\gamma'}^{-1}[\eta]\big)
      \left\{g+(I-\nu)\mathcal T_{k,\nu,\gamma}^{-1}(r-g)\right\}\Big\} \\
&\quad+\; \Pi\Big\{\mathcal{T}_{k,\pi,\gamma'}^{-1}(I-\nu)\big(\mathcal{T}_{k+\eta,\nu,\gamma}^{-1}-\mathcal{T}_{k,\nu,\gamma}^{-1}
      - \adk \mathcal{T}_{k,\nu,\gamma}^{-1}[\eta]\big)\Big\}(r-g) \\
&\quad+\; \Pi\Big\{\big(\mathcal{T}_{k+\eta,\pi,\gamma'}^{-1}-\mathcal{T}_{k,\pi,\gamma'}^{-1}\big)
      (I-\nu)\,\big(\mathcal{T}_{k+\eta,\nu,\gamma}^{-1}-\mathcal{T}_{k,\nu,\gamma}^{-1}\big)\Big\}(r-g) .
\label{eqn::normalized-kernel-remainder-decomposition}
\end{align}

For the first summand on the right-hand side of
\eqref{eqn::normalized-kernel-remainder-decomposition},
Lemma~\ref{lemma::bellmanfrechet} gives
\[
\|\mathcal T_{k+\eta,\pi, \gamma'}^{-1}-\mathcal T_{k,\pi, \gamma'}^{-1}-(\adk\mathcal T_{k,\pi,\gamma'}^{-1})[\eta]\|_{L^2(\lambda)\to L^2(\lambda)}
= O\!\big(\|\eta\|_{\mathcal K}^2\big).
\]
Hence,
\begin{align*}
    &\Big\|\Pi\Big\{\big(\mathcal{T}_{k+\eta,\pi,\gamma'}^{-1}-\mathcal{T}_{k,\pi,\gamma'}^{-1}
      - (\adk \mathcal{T}_{k,\pi,\gamma'}^{-1})[\eta]\big)
      \left[g+(I-\nu)\mathcal T_{k,\nu,\gamma}^{-1}(r-g)\right]\Big\} \Big\|_{L^2(\mu)} \\
    &\qquad \le \|\mathcal{T}_{k+\eta,\pi,\gamma'}^{-1}-\mathcal{T}_{k,\pi,\gamma'}^{-1}
      - (\adk \mathcal{T}_{k,\pi,\gamma'}^{-1})[\eta]\|_{L^2(\lambda)\to L^2(\lambda)} \,
      \|g+(I-\nu)\mathcal T_{k,\nu,\gamma}^{-1}(r-g)\|_{L^2(\lambda)} \\
    &\qquad = O\!\big(\|\eta\|_{\mathcal K}^2\big),
\end{align*}
uniformly over bounded $L^2$-neighborhoods of $r$. Here the big-oh notation
absorbs the fixed function $g$, the local inverse bound, and the operator norm
of $(I-\nu)$; the latter is finite because
\(\|(I-\nu)\|_{L^2(\lambda)\to L^2(\lambda)} \leq 1 + \|\nu\|_{L^2(\lambda)\to L^2(\lambda)} < \infty\),
since \(\mathcal A\) is finite and \(\nu\) is a Markov kernel.

For the second summand on the right-hand side of
\eqref{eqn::normalized-kernel-remainder-decomposition},
Lemma~\ref{lemma::bellmanfrechet} gives
\[
\|\mathcal T_{k+\eta,\nu, \gamma}^{-1}-\mathcal T_{k,\nu, \gamma}^{-1}-(\adk\mathcal T_{k,\nu,\gamma}^{-1})[\eta]\|_{L^2(\lambda)\to L^2(\lambda)}
= O\!\big(\|\eta\|_{\mathcal K}^2\big).
\]
Hence,
\begin{align*}
    &\Big\|\Pi\Big\{\mathcal{T}_{k,\pi,\gamma'}^{-1}(I-\nu)\big(\mathcal{T}_{k+\eta,\nu,\gamma}^{-1}-\mathcal{T}_{k,\nu,\gamma}^{-1}
      - (\adk \mathcal{T}_{k,\nu,\gamma}^{-1})[\eta]\big)\Big\}(r-g) \Big\|_{L^2(\mu)} \\
    &\qquad \le \|\Pi\|_{L^2(\lambda)\to L^2(\mu)}
      \|\mathcal{T}_{k,\pi,\gamma'}^{-1}\|_{L^2(\lambda)\to L^2(\lambda)}
      \|(I-\nu)\|_{L^2(\lambda)\to L^2(\lambda)} \\
    &\hspace{35mm}\times
      \|\mathcal{T}_{k+\eta,\nu,\gamma}^{-1}-\mathcal{T}_{k,\nu,\gamma}^{-1}
      - (\adk \mathcal{T}_{k,\nu,\gamma}^{-1})[\eta]\|_{L^2(\lambda)\to L^2(\lambda)}
      \|r-g\|_{L^2(\lambda)} \\
    &\qquad = O\!\big(\|\eta\|_{\mathcal K}^2 \, \|r-g\|_{L^2(\lambda)}\big),
\end{align*}
where the implicit constant absorbs the displayed operator norms and the
fixed anchor $g$. The bound is therefore uniform over bounded
$L^2$-neighborhoods of $r$.

For the third summand on the right-hand side of
\eqref{eqn::normalized-kernel-remainder-decomposition}, the triangle
inequality and Lemma~\ref{lemma::bellmanfrechet} give
\[
\|\mathcal T_{k+\eta,\pi,\gamma'}^{-1}-\mathcal T_{k,\pi,\gamma'}^{-1}\|_{L^2(\lambda)\to L^2(\lambda)}
\lesssim \|\eta\|_{\mathcal K},
\]
since \(\|(\adk\mathcal T_{k,\pi,\gamma'}^{-1})[\eta]\|_{L^2(\lambda)\to L^2(\lambda)} \le C\,\|\eta\|_{\mathcal K}\) for some constant \(C\).
Applying Lemma~\ref{lemma::bellmanfrechet} with \((\nu,\gamma)\) in place
of \((\pi,\gamma')\) also gives
\[
\|\mathcal T_{k+\eta,\nu,\gamma}^{-1}-\mathcal T_{k,\nu,\gamma}^{-1}\|_{L^2(\lambda)\to L^2(\lambda)}
\lesssim \|\eta\|_{\mathcal K}.
\]
Hence, taking operator norms,
\begin{align*}
&\Big\|\Pi\big(\mathcal T_{k+\eta,\pi,\gamma'}^{-1}
-\mathcal T_{k,\pi,\gamma'}^{-1}\big)(I-\nu)
\big(\mathcal T_{k+\eta,\nu,\gamma}^{-1}
-\mathcal T_{k,\nu,\gamma}^{-1}\big)(r-g)\Big\|_{L^2(\mu)} \\
&\qquad \lesssim \|\mathcal T_{k+\eta,\pi,\gamma'}^{-1}-\mathcal T_{k,\pi,\gamma'}^{-1}\|_{L^2(\lambda)\to L^2(\lambda)} \,
\|\mathcal T_{k+\eta,\nu,\gamma}^{-1}-\mathcal T_{k,\nu,\gamma}^{-1}\|_{L^2(\lambda)\to L^2(\lambda)} \,\|r-g\|_{L^2(\lambda)} \\
&\qquad \lesssim \|\eta\|_{\mathcal K}^2,
\end{align*}
uniformly over bounded $L^2$-neighborhoods of $r$.
The joint perturbation decomposes as
\begin{align*}
   & \mathcal{V}_{k+\eta}^g(r+ h) - \mathcal{V}_k^g(r) - \adk \mathcal{V}_k^g(r)[\eta] - \partial_r \mathcal{V}_k^g(r)[h] \\
   &= \Big\{\mathcal{V}_{k+\eta}^g(r) - \mathcal{V}_k^g(r) - \adk \mathcal{V}_k^g(r)[\eta]\Big\}
      + \Big\{\mathcal{V}_{k+\eta}^g(r+h) - \mathcal{V}_{k+\eta}^g(r) - \partial_r \mathcal{V}_k^g(r)[h]\Big\}.
\end{align*}
It remains to prove that
\begin{equation}
\|\mathcal{V}_{k+\eta}^g(r+h) - \mathcal{V}_{k+\eta}^g(r) - \partial_r \mathcal{V}_k^g(r)[h]\|_{L^2(\mu)} \;\lesssim\; \|\eta\|_{\mathcal K}\,\|h\|_{L^2(\lambda)},
\label{eqn::normalization-mixed-remainder}
\end{equation}
in which case
\begin{align*}
   & \|\mathcal{V}_{k+\eta}^g(r+ h) - \mathcal{V}_k^g(r) - \adk \mathcal{V}_k^g(r)[\eta] - \partial_r \mathcal{V}_k^g(r)[h]\|_{L^2(\mu)} \\
   &\leq \|\mathcal{V}_{k+\eta}^g(r) - \mathcal{V}_k^g(r) - \adk \mathcal{V}_k^g(r)[\eta]\|_{L^2(\mu)}
      + \|\mathcal{V}_{k+\eta}^g(r+h) - \mathcal{V}_{k+\eta}^g(r) - \partial_r \mathcal{V}_k^g(r)[h]\|_{L^2(\mu)}\\
   &= O(\|\eta\|_{\mathcal K}^2)
   +O(\|\eta\|_{\mathcal K}\,\|h\|_{L^2(\lambda)}),
\end{align*}
which is the stated second-order remainder bound by
\eqref{eqn::normalization-mixed-remainder}.

\medskip
To prove \eqref{eqn::normalization-mixed-remainder}, recall that
\[
\partial_r \mathcal{V}_k^g(r)[h] = \Pi\,\mathcal{T}_{k,\pi,\gamma'}^{-1}(I-\nu)\,\mathcal{T}_{k,\nu,\gamma}^{-1}(h).
\]
Hence,
\begin{align*}
\mathcal{V}_{k+\eta}^g(r+h) - \mathcal{V}_{k+\eta}^g(r) - \partial_r \mathcal{V}_k^g(r)[h]
&= \Pi\Big(\mathcal{T}_{k+\eta,\pi,\gamma'}^{-1}-\mathcal{T}_{k,\pi,\gamma'}^{-1}\Big)(I-\nu)\,\mathcal{T}_{k+\eta,\nu,\gamma}^{-1}(h) \\
&\quad+\; \Pi\,\mathcal{T}_{k,\pi,\gamma'}^{-1}(I-\nu)\Big(\mathcal{T}_{k+\eta,\nu,\gamma}^{-1}-\mathcal{T}_{k,\nu,\gamma}^{-1}\Big)(h).
\end{align*}
Taking $L^2(\mu)$ norms gives
\begin{align*}
\big\|\mathcal{V}_{k+\eta}^g(r+h) - \mathcal{V}_{k+\eta}^g(r)
&\qquad - \partial_r \mathcal{V}_k^g(r)[h]\big\|_{L^2(\mu)} \\
&\le \|\mathcal{T}_{k+\eta,\pi,\gamma'}^{-1}-\mathcal{T}_{k,\pi,\gamma'}^{-1}\|_{L^2(\lambda)\to L^2(\lambda)}
   \,\|(I-\nu)\,\mathcal{T}_{k+\eta,\nu,\gamma}^{-1}(h)\|_{L^2(\lambda)} \\
&\quad+\; \|\mathcal{T}_{k,\pi,\gamma'}^{-1}\|_{L^2(\lambda)\to L^2(\lambda)} \,
   \|(I-\nu)\|_{L^2(\lambda)\to L^2(\lambda)} \\
&\qquad\times
   \|\mathcal{T}_{k+\eta,\nu,\gamma}^{-1}-\mathcal{T}_{k,\nu,\gamma}^{-1}\|_{L^2(\lambda)\to L^2(\lambda)} \,
   \|h\|_{L^2(\lambda)}.
\end{align*}

Lemma~\ref{lemma::bellmanfrechet}, applied with $(\pi,\gamma')$ and with
$(\nu,\gamma)$, respectively, gives
\[
\begin{aligned}
\|\mathcal{T}_{k+\eta,\pi,\gamma'}^{-1}-\mathcal{T}_{k,\pi,\gamma'}^{-1}\|_{L^2(\lambda)\to L^2(\lambda)}
&\lesssim \|\eta\|_{\mathcal K}, \\
\|\mathcal{T}_{k+\eta,\nu,\gamma}^{-1}-\mathcal{T}_{k,\nu,\gamma}^{-1}\|_{L^2(\lambda)\to L^2(\lambda)}
&\lesssim \|\eta\|_{\mathcal K}.
\end{aligned}
\]
Substituting these bounds and absorbing the operator norms of $(I-\nu)$,
$\Pi$, and $\mathcal{T}_{k,\pi,\gamma'}^{-1}$ into the implied constant gives
\[
\begin{aligned}
\|\mathcal{V}_{k+\eta}^g(r+h) - \mathcal{V}_{k+\eta}^g(r)
- \partial_r \mathcal{V}_k^g(r)[h]\|_{L^2(\mu)}
\lesssim \|\eta\|_{\mathcal K}\,\|h\|_{L^2(\lambda)}.
\end{aligned}
\]
This proves \eqref{eqn::normalization-mixed-remainder} and completes the
proof of the lemma.

\end{proof}

 \subsubsection{Lemmas for Example~\ref{example::1b}}

In the following lemma, we recall the softmax policy $\pi_{r,k}^\star$, its
log-partition function $\Phi_{r,k}^\star$, and the solution $v_{r,k}^\star$ of
the soft Bellman equation from Example~\ref{example::1b}. For each $(r,k)$,
define the operator $(\Pi_{r,k}^\star f)(s) := \sum_{\tilde a \in
\mathcal{A}^\star(s)} \pi_{r,k}^\star(\tilde a \mid
s)\,f(\tilde a,s)$. Define the conditional covariance under
$\pi_{r,k}^\star$ by
\[
\operatorname{Cov}_{\pi_{r,k}^\star}\!\big(f,g\big)(s)
:= \big(\Pi_{r,k}^\star[fg]\big)(s)
- \big(\Pi_{r,k}^\star f\big)(s)\,\big(\Pi_{r,k}^\star g\big)(s),
\qquad f,g \in L^\infty(\lambda).
\]

\begin{lemma}[Derivatives for soft Bellman operators]
\label{lemma::derivSoftBellmanStar}
The map $(r,k) \mapsto v_{r,k}^\star$ is sup-norm Fr\'echet differentiable as a map from $L^\infty(\lambda) \times L^\infty(\mu \otimes \lambda)$ to $L^\infty(\lambda)$, with
\[
\partial_r v_{r,k}^\star[h] = (\mathcal{T}_{r,k}^\star)^{-1}\,\mathcal{P}_k\,\Pi_{r,k}^\star h,
\qquad
\adk v_{r,k}^\star[w] = (\mathcal{T}_{r,k}^\star)^{-1}\,\mathcal{P}_w\,\Phi_{r,k}^\star.
\]
Moreover, for any test function $f$,
\[
\partial_r \Pi_{r,k}^\star[h] f
= \tfrac{1}{\tau^\star}\,
\operatorname{Cov}_{\pi_{r,k}^\star}\!\big(f,\,(\mathcal{T}_{r,k}^\star)^{-1}h\big),
\qquad
\adk \Pi_{r,k}^\star[w] f
= \tfrac{1}{\tau^\star}\,
\operatorname{Cov}_{\pi_{r,k}^\star}\!\big(f,\,\gamma\,\adk v_{r,k}^\star[w]\big).
\]
Consequently,
\[
\begin{aligned}
\partial_r \mathcal{T}_{r,k}^\star[h] f
&= -\,\tfrac{\gamma}{\tau^\star}\,
\mathcal{P}_k\,
\operatorname{Cov}_{\pi_{r,k}^\star}\!\big(f,\,(\mathcal{T}_{r,k}^\star)^{-1}h\big), \\[6pt]
\adk \mathcal{T}_{r,k}^\star[w] f
&= -\,\gamma\,\mathcal{P}_w\,\Pi_{r,k}^\star f
- \tfrac{\gamma}{\tau^\star}\,
\mathcal{P}_k\,
\operatorname{Cov}_{\pi_{r,k}^\star}\!\big(f,\,\gamma\,\adk v_{r,k}^\star[w]\big),
\end{aligned}
\]
and
\[
\partial_r \!\left[ \Pi_{r,k}^\star (\mathcal{T}_{r,k}^\star)^{-1}\right][h]
= \Big(I + \gamma\,\Pi_{r,k}^\star (\mathcal{T}_{r,k}^\star)^{-1}\mathcal{P}_k\Big)\,
\big(\partial_r \Pi_{r,k}^\star[h]\big)\,(\mathcal{T}_{r,k}^\star)^{-1}.
\]
\[
\adk \!\left[ \Pi_{r,k}^\star (\mathcal{T}_{r,k}^\star)^{-1}\right][w]
= \big(I-\gamma\,\Pi_{r,k}^\star \mathcal{P}_k\big)^{-1}
\Big[\,\big(\adk \Pi_{r,k}^\star[w]\big) + \gamma\,\Pi_{r,k}^\star \mathcal{P}_w \Pi_{r,k}^\star\,\Big]
(\mathcal{T}_{r,k}^\star)^{-1}.
\]
\end{lemma}

\begin{proof}
Because the available action sets do not vary with $(r,k)$, the proofs of
Lemmas~\ref{lemma::IFT}, \ref{lemma::qderiv},
and~\ref{lemma::operatorderivatives} apply with each action sum at state $s$
restricted to $\mathcal A^\star(s)$ and with $\tau$ replaced by
$\tau^\star$. This gives the displayed derivatives of $v_{r,k}^\star$ and
$\Pi_{r,k}^\star$.

Since $\mathcal{T}_{r,k}^\star = I - \gamma \mathcal{P}_k \Pi_{r,k}^\star$, applying the chain rule yields
\[
\begin{aligned}
\partial_r \mathcal{T}_{r,k}^\star[h] f
&= -\,\gamma\,\mathcal{P}_k\,\big(\partial_r \Pi_{r,k}^\star[h]\big)f
= -\,\tfrac{\gamma}{\tau^\star}\,\mathcal{P}_k\,\operatorname{Cov}_{\pi_{r,k}^\star}\!\big(f,\,(\mathcal{T}_{r,k}^\star)^{-1}h\big), \\[6pt]
\adk \mathcal{T}_{r,k}^\star[w] f
&= -\,\gamma\,\mathcal{P}_w\,\Pi_{r,k}^\star f
- \gamma\,\mathcal{P}_k\,\big(\adk \Pi_{r,k}^\star[w]\big)f \\[3pt]
&= -\,\gamma\,\mathcal{P}_w\,\Pi_{r,k}^\star f
- \tfrac{\gamma}{\tau^\star}\,\mathcal{P}_k\,\operatorname{Cov}_{\pi_{r,k}^\star}\!\big(f,\,\gamma\,\adk v_{r,k}^\star[w]\big).
\end{aligned}
\]
This establishes the derivative identities for $\mathcal{T}_{r,k}^\star$.

Let $G(r,k):=\Pi_{r,k}^\star(\mathcal{T}_{r,k}^\star)^{-1}$. The product
rule and the derivative formula in Lemma~\ref{lemma::inverseperturbation}
give
\[
\begin{aligned}
\partial_rG(r,k)[h]
&=\big(\partial_r\Pi_{r,k}^\star[h]\big)
(\mathcal T_{r,k}^\star)^{-1}
-\Pi_{r,k}^\star(\mathcal T_{r,k}^\star)^{-1}
\big(\partial_r\mathcal T_{r,k}^\star[h]\big)
(\mathcal T_{r,k}^\star)^{-1}\\
&=\Big(I+\gamma\Pi_{r,k}^\star(\mathcal T_{r,k}^\star)^{-1}
\mathcal P_k\Big)
\big(\partial_r\Pi_{r,k}^\star[h]\big)
(\mathcal T_{r,k}^\star)^{-1},
\end{aligned}
\]
where the second equality uses
\(\partial_r\mathcal T_{r,k}^\star[h]
=-\gamma\mathcal P_k\partial_r\Pi_{r,k}^\star[h]\).
For a kernel direction \(w\), Lemma~\ref{lemma::inverseperturbation} gives
\[
\adk(\mathcal{T}_{r,k}^\star)^{-1}[w]
= -\,(\mathcal{T}_{r,k}^\star)^{-1}\,\big(\adk \mathcal{T}_{r,k}^\star[w]\big)\,(\mathcal{T}_{r,k}^\star)^{-1},
\]
and therefore
\[
\adk G[w]
= \big(\adk \Pi_{r,k}^\star[w]\big)(\mathcal{T}_{r,k}^\star)^{-1}
+ \Pi_{r,k}^\star\,\adk(\mathcal{T}_{r,k}^\star)^{-1}[w].
\]
Since $\mathcal{T}_{r,k}^\star = I-\gamma\,\mathcal{P}_k \Pi_{r,k}^\star$,
\[
\adk \mathcal{T}_{r,k}^\star[w]
= -\gamma\big(\mathcal{P}_w \Pi_{r,k}^\star + \mathcal{P}_k\,\adk \Pi_{r,k}^\star[w]\big).
\]
Substituting gives
\[
\adk G[w]
= \big(\adk \Pi_{r,k}^\star[w]\big)(\mathcal{T}_{r,k}^\star)^{-1}
+ \gamma\,\Pi_{r,k}^\star(\mathcal{T}_{r,k}^\star)^{-1}
\big(\mathcal{P}_w \Pi_{r,k}^\star + \mathcal{P}_k\,\adk \Pi_{r,k}^\star[w]\big)
(\mathcal{T}_{r,k}^\star)^{-1}.
\]
Regrouping,
\[
\adk G[w]
= \Big(I+\gamma\,\Pi_{r,k}^\star(\mathcal{T}_{r,k}^\star)^{-1}\mathcal{P}_k\Big)\,
\big(\adk \Pi_{r,k}^\star[w]\big)\,(\mathcal{T}_{r,k}^\star)^{-1}
\;+\;
\gamma\,\Pi_{r,k}^\star(\mathcal{T}_{r,k}^\star)^{-1}\mathcal{P}_w \Pi_{r,k}^\star(\mathcal{T}_{r,k}^\star)^{-1}.
\]
Finally, by the resolvent identity $(I-AB)^{-1}=I+A(I-BA)^{-1}B$ applied with $A=\gamma\,\Pi_{r,k}^\star$, $B=\mathcal{P}_k$,
\[
I+\gamma\,\Pi_{r,k}^\star(\mathcal{T}_{r,k}^\star)^{-1}\mathcal{P}_k
= (I-\gamma\,\Pi_{r,k}^\star \mathcal{P}_k)^{-1},
\quad
\Pi_{r,k}^\star(\mathcal{T}_{r,k}^\star)^{-1}
= (I-\gamma\,\Pi_{r,k}^\star \mathcal{P}_k)^{-1}\Pi_{r,k}^\star,
\]
which yields the stated formula.

\end{proof}

\begin{lemma}[Linearization remainder for Example 1b]
\label{lemma::linearizationExamplestar}
Assume Assumption~\ref{cond::boundedpositivity}, fix
$M>\|r_0\|_{L^\infty(\lambda)}$ and
$M_k>\|k_0\|_{L^\infty(\mu\otimes\lambda)}$, and restrict attention to
rewards and kernels satisfying
\[
\|r\|_{L^\infty(\lambda)}\le M,
\qquad
\|k\|_{L^\infty(\mu\otimes\lambda)}\le M_k.
\]
Then there exists an $L^2$-ball around
$(r_0,k_0) \in \mathcal{H} \times \mathcal{K}$ such that
$(\mathcal{T}_{r,k}^\star)^{-1}: L^2(\lambda) \to L^2(\lambda)$ exists. Let
$m(s,r,k) = \Pi_{r,k}^\star\,(\mathcal{T}_{r,k}^\star)^{-1}(r)(s)$.
Then
\[
\operatorname{Rem}_0(r_0,k_0; r,k) \;\lesssim\; \|k - k_0\|_{\mathcal{K}}^{2}
+ \|r - r_0\|_{\mathcal{H}}^{2},
\]
with the implicit constant depending on $M$, $M_k$, $\gamma$, $\tau^\star$, the
finite action set, the constants in Assumption~\ref{cond::boundedpositivity},
and
$\|(\mathcal{T}_{r_0,k_0}^\star)^{-1}\|_{L^2(\lambda)\to L^2(\lambda)}$.
\end{lemma}
\begin{proof}
For this proof, suppress the star and write $v_{r,k}$, $\Pi_{r,k}$, and $\mathcal{T}_{r,k}$ for the soft Bellman objects associated with the state-dependent action set $\mathcal{A}^\star(s)$ and fixed temperature $\tau^\star$ from Example~\ref{example::1b}. Then $m(s,r,k) = \Pi_{r,k}\,\mathcal{T}_{r,k}^{-1}(r)(s)$.

Lemma~\ref{lemma::invertibility}, applied with
$\pi=\pi_{r_0,k_0}^\star$, gives the bounded inverse
$\mathcal T_{r_0,k_0}^{-1}$ on $L^2(\lambda)$. Lemmas~\ref{lemma::lipschitzvalue}
and~\ref{lemma::valuestarderivativereward}--\ref{lemma::jointderivative}
apply with each action sum at state $s$ restricted to
$\mathcal A^\star(s)$ and with temperature $\tau^\star$. They also give a
neighborhood of $(r_0,k_0)$ on which the required inverses exist.
Lemma~\ref{lemma::initialpooledbounds} and
Assumption~\ref{cond::boundedpositivity} imply
\begin{align*}
&\left|\mathbb E_0\!\left[
\left\{
\begin{aligned}
&\Pi_{r,k}\,\mathcal{T}_{r,k}^{-1}(r)
- \Pi_{r_0,k_0}\,\mathcal{T}_{r_0,k_0}^{-1}(r_0)\\
&\quad- \partial_r\!\big[\Pi_{r_0,k_0}\,\mathcal{T}_{r_0,k_0}^{-1}(r_0)\big][\,r-r_0\,]\\
&\quad- \adk\!\big[\Pi_{r_0,k_0}\,\mathcal{T}_{r_0,k_0}^{-1}(r_0)\big][\,k-k_0\,]
\end{aligned}
\right\}(S_0)\right]\right| \\
&\;\;\lesssim\;
\left\|
\begin{aligned}
&\Pi_{r,k}\,\mathcal{T}_{r,k}^{-1}(r)
- \Pi_{r_0,k_0}\,\mathcal{T}_{r_0,k_0}^{-1}(r_0)\\
&\quad- \partial_r\!\big[\Pi_{r_0,k_0}\,\mathcal{T}_{r_0,k_0}^{-1}(r_0)\big][\,r-r_0\,]\\
&\quad- \adk\!\big[\Pi_{r_0,k_0}\,\mathcal{T}_{r_0,k_0}^{-1}(r_0)\big][\,k-k_0\,]
\end{aligned}
\right\|_{L^1(\mu)}.
\end{align*}
By Lemma~\ref{lemma::jointderivative}, we have
\begin{align*}
&\left\|
\begin{aligned}
&\Pi_{r,k}\,\mathcal{T}_{r,k}^{-1}(r)
- \Pi_{r_0,k_0}\,\mathcal{T}_{r_0,k_0}^{-1}(r_0)\\
&\quad- \partial_r\!\big[\Pi_{r_0,k_0}\,\mathcal{T}_{r_0,k_0}^{-1}(r_0)\big][\,r-r_0\,]\\
&\quad- \adk\!\big[\Pi_{r_0,k_0}\,\mathcal{T}_{r_0,k_0}^{-1}(r_0)\big][\,k-k_0\,]
\end{aligned}
\right\|_{L^1(\mu)} \\
&\;\;=\; O\!\left(\|r-r_0\|_{L^2(\lambda)}^{2}
  + \|v_{r_0,k}-v_{r_0,k_0}\|_{L^2(\lambda)}^{2}
+ \|k-k_0\|_{\mathcal K}^{2}\right).
\end{align*}
Furthermore, by Lemma~\ref{lemma::lipschitzvalue},
\[
\|v_{r_0,k}-v_{r_0,k_0}\|_{L^2(\lambda)}^{2}
\;\lesssim\; \|k-k_0\|_{\mathcal K}^{2}.
\]
Hence,
\begin{align*}
&\big\|\Pi_{r,k}\,\mathcal{T}_{r,k}^{-1}(r)
- \Pi_{r_0,k_0}\,\mathcal{T}_{r_0,k_0}^{-1}(r_0)
- \partial_r\!\big[\Pi_{r_0,k_0}\,\mathcal{T}_{r_0,k_0}^{-1}(r_0)\big][\,r-r_0\,]
- \adk\!\big[\Pi_{r_0,k_0}\,\mathcal{T}_{r_0,k_0}^{-1}(r_0)\big][\,k-k_0\,]\big\|_{L^1(\mu)} \\
&\;\;=\; O\!\left(\|r-r_0\|_{L^2(\lambda)}^{2}
+ \|k-k_0\|_{\mathcal K}^{2}\right).
\end{align*}
We conclude that
\begin{align*}
&\Big|\mathbb E_0\!\Big[\Big\{\Pi_{r,k}\,\mathcal{T}_{r,k}^{-1}(r)
- \Pi_{r_0,k_0}\,\mathcal{T}_{r_0,k_0}^{-1}(r_0)
- \partial_r\!\big[\Pi_{r_0,k_0}\,\mathcal{T}_{r_0,k_0}^{-1}(r_0)\big][\,r-r_0\,] \\
&\qquad\qquad
- \adk\!\big[\Pi_{r_0,k_0}\,\mathcal{T}_{r_0,k_0}^{-1}(r_0)\big][\,k-k_0\,]\Big\}(S_0)\Big]\Big| \\
&\quad=\; O\!\left(\|r-r_0\|_{L^2(\lambda)}^{2}
+ \|k-k_0\|_{\mathcal K}^{2}\right).
\end{align*}
Hence,
\[
\operatorname{Rem}_0(r_0,k_0; r,k)\;\lesssim\;\|k - k_0\|_{\mathcal{K}}^{2}
+  \|r - r_0\|_{\mathcal{H}}^{2}.
\]
This proves Lemma~\ref{lemma::linearizationExamplestar}.
\end{proof}

\subsection{Proofs of EIFs in examples}

\label{appendix::eifexproofs}

\begin{proof}[Proof of Theorem~\ref{theorem::EIFvalue}]
Let
\[
m(s,r,k):=\pi\mathcal T_{k,\pi,\gamma}^{-1}r(s).
\]
Lemmas~\ref{lemma::invertibility} and~\ref{lemma::bellmanfrechet} establish
Conditions~\ref{cond::boundedfunc}--\ref{cond::lipschitz}. For
\(h\in T_{\mathcal H}\), Lemma~\ref{lemma::adjointident} gives
\[
\mathbb E_0\{\partial_r m(S_0,r_0,k_0)[h]\}
=\mathbb E_0\!\left[
\pi\mathcal T_{k_0,\pi,\gamma}^{-1}h(S_0)\right]
=E_0\{d_0^{\pi,\gamma}(A,S)h(A,S)\}.
\]
Thus \(\alpha_0=d_0^{\pi,\gamma}\). For
\(\beta\in T_{\mathcal K}\), Lemma~\ref{lemma::bellmanfrechet} and
conditional centering give
\begin{align*}
\mathbb E_0\{\partial_k m(S_0,r_0,k_0)[\beta]\}
&=E_0\!\left[
d_0^{\pi,\gamma}(A,S)\,
\gamma\mathcal P_{k_0\beta}
V_{r_0,k_0}^{\pi,\gamma}(A,S)\right]\\
&=E_0\!\left[
d_0^{\pi,\gamma}(A,S)
\{r_0(A,S)+\gamma V_{r_0,k_0}^{\pi,\gamma}(S')
-q_{r_0,k_0}^{\pi,\gamma}(A,S)\}
\beta(S',A,S)\right],
\end{align*}
where the second equality uses the conditional centering of \(\beta\) and
Bellman's equation. Thus
\[
\beta_0(s',a,s)
=d_0^{\pi,\gamma}(a,s)
\{r_0(a,s)+\gamma V_{r_0,k_0}^{\pi,\gamma}(s')
-q_{r_0,k_0}^{\pi,\gamma}(a,s)\}.
\]
Bellman's equation also gives
\[
E_0\{\beta_0(S',A,S)\mid A,S\}=0.
\]
Lemma~\ref{lemma::boundedoccupancy} and boundedness of the Bellman solution
give \(\beta_0\in L^2(P_0)\); hence \(\beta_0\in T_{\mathcal K}\).
Substitution into
Theorem~\ref{theorem::EIF} yields the stated trajectory-level EIF.

Finally, Lemmas~\ref{lemma::bellmanfrechet}
and~\ref{lemma::initialpooledbounds} give
\[
\operatorname{Rem}_0(r_0,k_0;r,k)
=O\!\left(\|k-k_0\|_{\mathcal K}^2
+\|k-k_0\|_{\mathcal K}\|r-r_0\|_{\mathcal H}\right).
\]
This proves the theorem.
\end{proof}

\begin{proof}[Proof of Theorem~\ref{theorem::EIFvaluenorm}]
Apply Theorem~\ref{theorem::EIFnorm} with
\[
\tilde m(s,\bar r,k)
:=\pi\mathcal T_{k,\pi,\gamma'}^{-1}\bar r(s).
\]
The two Bellman inverses are bounded by
Lemma~\ref{lemma::invertibility}, and the corresponding occupancy ratios are
bounded by Lemma~\ref{lemma::boundedoccupancy}.
Conditions~\ref{cond::boundedfunc::norm}--\ref{cond::lipschitz::norm} hold
by hypothesis. Lemmas~\ref{lemma::invertibility},
\ref{lemma::bellmanfrechet}, and~\ref{lemma::adjointident} identify the
derivative representers of this fixed-policy map. Applying
Lemma~\ref{lemma::adjointident} at discount \(\gamma'\) identifies the
corresponding partial-derivative representers as
\[
\tilde\alpha_0=d_0^{\pi,\gamma'}
\]
and
\[
\tilde\beta_0(s',a,s)
=d_0^{\pi,\gamma'}(a,s)
\{r_{0,\nu,\gamma,g}(a,s)
+\gamma'V_{r_0,k_0,\nu}^{\pi,\gamma'}(s')
-q_{r_0,k_0,\nu}^{\pi,\gamma'}(a,s)\}.
\]
Theorem~\ref{theorem::EIFnorm} therefore identifies the reward representer as
the unique solution \(\alpha_0^{\pi,\nu}\) of
\eqref{eqn::policyvaluenormbalance} and adds the kernel term
\[
\alpha_0^{\pi,\nu}(a,s)
\{r_0(a,s)-g(s)+\gamma V_{r_0-g,k_0}^{\nu,\gamma}(s')
-q_{r_0-g,k_0}^{\nu,\gamma}(a,s)\}.
\]
Substituting these representers into Theorem~\ref{theorem::EIF} yields the
stated trajectory-level EIF.
\end{proof}
\begin{proof}[Proof of Theorem~\ref{theorem::EIFsoftmax}]
For each $(r,k)$, let
\(q_{r,k}^\star:=q_{r,k}^{\pi_{r,k}^\star,\gamma}\) denote the ordinary
policy-evaluation $Q$-function of $\pi_{r,k}^\star$, and let
\(V_{r,k}^\star:=\Pi_{r,k}^\star q_{r,k}^\star\). These objects are distinct
from the soft action index \(r+\gamma v_{r,k}^\star\) that defines
\(\pi_{r,k}^\star\).
Let
\[
m(s,r,k):=V_{r,k}^{\star}(s).
\]
Abbreviate
\[
\Pi_0:=\Pi_{r_0,k_0}^\star,
\quad
\mathcal T_0:=\mathcal T_{r_0,k_0}^\star
=I-\gamma\mathcal P_{k_0}\Pi_0,
\quad
q_0^\star:=q_{r_0,k_0}^\star,
\quad
V_0^\star:=\Pi_0q_0^\star.
\]
Lemma~\ref{lemma::boundedoccupancy} gives
\(d_{r_0,k_0}^\star\in L^\infty(\lambda)\). The balance equation
\eqref{eqn::tildedbalance} is the adjoint equation
\[
\mathcal T_0^*\widetilde d_{r_0,k_0}^\star
=d_{r_0,k_0}^\star(q_0^\star-V_0^\star)
\]
in the design \(L^2\) space. Its right-hand side is square-integrable, and
Lemma~\ref{lemma::invertibility} gives
\((\mathcal T_0^*)^{-1}=(\mathcal T_0^{-1})^*\). Hence the equation has a
unique solution \(\widetilde d_{r_0,k_0}^\star\in T_{\mathcal H}\).

We first calculate the reward derivative. For bounded
\(h\in T_{\mathcal H}\), put \(\dot q_h=\mathcal T_0^{-1}h\).
Lemma~\ref{lemma::derivSoftBellmanStar}, the product rule, and
\eqref{eqn::first-resolvent-identity} give
\begin{align}
\partial_r m(\cdot,r_0,k_0)[h]
={}&\Pi_0\mathcal T_0^{-1}h\notag\\
&+\frac1{\tau^\star}
(I-\gamma\Pi_0\mathcal P_{k_0})^{-1}
\operatorname{Cov}_{\pi_{r_0,k_0}^\star}
(q_0^\star,\dot q_h).
\label{eqn::softmax-value-reward-derivative}
\end{align}
For every state function \(G\), the resolvent identity and
Lemma~\ref{lemma::adjointident} give
\begin{equation}
\mathbb E_0\!\left[
\{(I-\gamma\Pi_0\mathcal P_{k_0})^{-1}G\}(S_0)\right]
=E_0\{d_{r_0,k_0}^\star(A,S)G(S)\}.
\label{eqn::softmax-state-occupancy-identity}
\end{equation}
Moreover, the balance equation with \(h=\dot q_h\) gives
\begin{align}
&E_0\!\left[d_{r_0,k_0}^\star(A,S)
\{q_0^\star(A,S)-V_0^\star(S)\}\dot q_h(A,S)\right]\notag\\
&\qquad=E_0\{\widetilde d_{r_0,k_0}^\star(A,S)
(\mathcal T_0\dot q_h)(A,S)\}
=E_0\{\widetilde d_{r_0,k_0}^\star(A,S)h(A,S)\}.
\label{eqn::softmax-reward-balance-chain}
\end{align}
Applying Lemma~\ref{lemma::adjointident},
\eqref{eqn::softmax-state-occupancy-identity}, and
\eqref{eqn::softmax-reward-balance-chain} to
\eqref{eqn::softmax-value-reward-derivative} yields
\begin{align}
\mathbb E_0\{\partial_r m(S_0,r_0,k_0)[h]\}
={}&E_0\{d_{r_0,k_0}^\star(A,S)h(A,S)\}\notag\\
&+\frac1{\tau^\star}E_0\!\left[
d_{r_0,k_0}^\star(A,S)
\{q_0^\star(A,S)-V_0^\star(S)\}\dot q_h(A,S)
\right]\notag\\
={}&E_0\!\left[
\left\{d_{r_0,k_0}^\star(A,S)
+\frac1{\tau^\star}\widetilde d_{r_0,k_0}^\star(A,S)\right\}
h(A,S)\right],
\label{eqn::softmax-value-reward-riesz-identity}
\end{align}
Thus
\[
\alpha_0
=d_{r_0,k_0}^\star+\tau^{\star-1}\widetilde d_{r_0,k_0}^\star.
\]

We next calculate the kernel derivative. Let
\(\beta\in T_{\mathcal K}\) be bounded, set
\(g_\beta=k_0\beta\in\addTK\), and define
\[
\dot v_\beta
:=\mathcal T_0^{-1}\mathcal P_{g_\beta}\Phi_{r_0,k_0}^\star.
\]
Differentiating \(m=\Pi_{r,k}^\star q_{r,k}^\star\) with respect to the
kernel and applying Lemma~\ref{lemma::derivSoftBellmanStar} gives
\begin{align}
\partial_k m(\cdot,r_0,k_0)[\beta]
={}&(I-\gamma\Pi_0\mathcal P_{k_0})^{-1}
\Bigg\{
\frac{\gamma}{\tau^\star}
\operatorname{Cov}_{\pi_{r_0,k_0}^\star}(q_0^\star,\dot v_\beta)
+\gamma\Pi_0\mathcal P_{g_\beta}V_0^\star
\Bigg\}.
\label{eqn::softmax-value-kernel-derivative}
\end{align}
The occupancy identity and \eqref{eqn::tildedbalance} imply
\begin{align}
&E_0\!\left[d_{r_0,k_0}^\star(A,S)
\{q_0^\star(A,S)-V_0^\star(S)\}\dot v_\beta(A,S)\right]\notag\\
&\quad=E_0\{\widetilde d_{r_0,k_0}^\star(A,S)
(\mathcal T_0\dot v_\beta)(A,S)\}\notag\\
&\quad=E_0\{\widetilde d_{r_0,k_0}^\star(A,S)
(\mathcal P_{g_\beta}\Phi_{r_0,k_0}^\star)(A,S)\}\notag\\
&\quad=E_0\{\widetilde d_{r_0,k_0}^\star(A,S)
\Phi_{r_0,k_0}^\star(S')\beta(S',A,S)\}.
\label{eqn::softmax-kernel-balance-chain}
\end{align}
Conditional centering of \(\beta\) also gives
\begin{align}
E_0\{\widetilde d_{r_0,k_0}^\star(A,S)
v_{r_0,k_0}^\star(A,S)\beta(S',A,S)\}&=0,\notag\\
E_0\{d_{r_0,k_0}^\star(A,S)
\{q_0^\star(A,S)-r_0(A,S)\}\beta(S',A,S)\}&=0.
\label{eqn::softmax-kernel-centering-identities}
\end{align}
Equations~\eqref{eqn::softmax-state-occupancy-identity},
\eqref{eqn::softmax-kernel-balance-chain}, and
\eqref{eqn::softmax-kernel-centering-identities} therefore imply
\begin{align}
\mathbb E_0\{\partial_k m(S_0,r_0,k_0)[\beta]\}
={}&\frac{\gamma}{\tau^\star}E_0\!\left[
d_{r_0,k_0}^\star(A,S)
\{q_0^\star(A,S)-V_0^\star(S)\}\dot v_\beta(A,S)
\right]\notag\\
&+\gamma E_0\{d_{r_0,k_0}^\star(A,S)V_0^\star(S')
\beta(S',A,S)\}\notag\\
={}&\frac{\gamma}{\tau^\star}E_0\!\left[
\widetilde d_{r_0,k_0}^\star(A,S)
\{\Phi_{r_0,k_0}^\star(S')-v_{r_0,k_0}^\star(A,S)\}
\beta(S',A,S)\right]\notag\\
&+E_0\!\left[d_{r_0,k_0}^\star(A,S)
\{r_0(A,S)+\gamma V_0^\star(S')-q_0^\star(A,S)\}
\beta(S',A,S)\right].
\label{eqn::softmax-value-kernel-riesz-identity}
\end{align}
The first identity in
\eqref{eqn::softmax-kernel-centering-identities} permits the subtraction of
\(v_{r_0,k_0}^\star(A,S)=\mathcal P_{k_0}
\Phi_{r_0,k_0}^\star(A,S)\); the second permits the subtraction of
\(q_0^\star(A,S)-r_0(A,S)\). Therefore the kernel representer is
\begin{align*}
\beta_0(s',a,s)
={}&\frac{\gamma}{\tau^\star}
\widetilde d_{r_0,k_0}^{\star}(a,s)
\{\Phi_{r_0,k_0}^{\star}(s')-v_{r_0,k_0}^{\star}(a,s)\}\\
&+d_{r_0,k_0}^{\star}(a,s)
\{r_0(a,s)+\gamma V_{r_0,k_0}^{\star}(s')
-q_{r_0,k_0}^{\star}(a,s)\}.
\end{align*}
The Bellman equations for \(v_{r_0,k_0}^\star\) and
\(q_{r_0,k_0}^\star\) give
\begin{align}
E_0\{\Phi_{r_0,k_0}^\star(S')
-v_{r_0,k_0}^\star(A,S)\mid A,S\}&=0,\notag\\
E_0\{r_0(A,S)+\gamma V_0^\star(S')
-q_0^\star(A,S)\mid A,S\}&=0.
\label{eqn::softmax-kernel-representer-centering}
\end{align}
The first residual in
\eqref{eqn::softmax-kernel-representer-centering} is bounded, whereas
\(\widetilde d_{r_0,k_0}^\star\in L^2(P_0)\). The second residual and
\(d_{r_0,k_0}^\star\) are bounded. Hence
\(\beta_0\in L^2(P_0)\), and
\eqref{eqn::softmax-kernel-representer-centering} gives
\(\beta_0\in T_{\mathcal K}\).

The inner products in
\eqref{eqn::softmax-value-reward-riesz-identity} and
\eqref{eqn::softmax-value-kernel-riesz-identity} are continuous in
\(\|h\|_{\mathcal H,0}\) and \(\|\beta\|_{\mathcal K,0}\). The operator
bounds in Lemmas~\ref{lemma::derivSoftBellmanStar}
and~\ref{lemma::invertibility}, together with
Lemma~\ref{lemma::tangentapproximation}, therefore extend both identities
from bounded directions to \(T_{\mathcal H}\) and \(T_{\mathcal K}\).
Thus \(\alpha_0\) and \(\beta_0\) are the required Riesz representers, and
their Riesz identities prove
Condition~\ref{cond::boundedfunc2}.

It remains to verify the state-valued Lipschitz condition. For nearby feasible
perturbations, write \(k_j=(1+b_j)k_0\),
\(\Pi_j=\Pi_{r_j,k_j}^\star\),
\(\mathcal T_j=I-\gamma\mathcal P_{k_j}\Pi_j\),
\(q_j=\mathcal T_j^{-1}r_j\), and \(m_j=\Pi_jq_j\).
Fix common reward and kernel envelopes as in
Condition~\ref{cond::lipschitz}. Lemmas~\ref{lemma::lipschitzvalue},
\ref{lemma::softmaxLipschitz}, and~\ref{lemma::forwardopLipschitz} give
\begin{equation}
\|\mathcal T_1-\mathcal T_2\|_{L^2\to L^2}
+\|(\Pi_1-\Pi_2)q_2\|_{L^2(\mu)}
\lesssim
\|r_1-r_2\|_{\mathcal H,0}
+\|b_1-b_2\|_{\mathcal K,0}.
\label{eqn::softmax-value-local-operator-lipschitz}
\end{equation}
Lemma~\ref{lemma::inverseperturbation} makes the inverse norms locally
uniform. The first resolvent identity and
\eqref{eqn::softmax-value-local-operator-lipschitz} therefore give
\[
\|q_1-q_2\|_{L^2(\lambda)}
\lesssim
\|r_1-r_2\|_{\mathcal H,0}
+\|b_1-b_2\|_{\mathcal K,0}.
\]
Since
\[
m_1-m_2=(\Pi_1-\Pi_2)q_2+\Pi_1(q_1-q_2),
\]
Jensen's inequality and
\eqref{eqn::softmax-value-local-operator-lipschitz} yield
\begin{equation}
\|m_1-m_2\|_{L^2(\mu)}
\lesssim
\|r_1-r_2\|_{\mathcal H,0}
+\|b_1-b_2\|_{\mathcal K,0}.
\label{eqn::softmax-value-state-lipschitz}
\end{equation}
This proves Condition~\ref{cond::lipschitz}. Finally,
\eqref{eqn::softmax-value-reward-derivative},
\eqref{eqn::softmax-value-kernel-derivative}, and
Lemma~\ref{lemma::linearizationExamplestar} prove
Condition~\ref{cond::boundedfunc}.
Substitution into Theorem~\ref{theorem::EIF} gives the displayed
trajectory-level EIF. The remainder bound in
Lemma~\ref{lemma::linearizationExamplestar} gives
\[
\operatorname{Rem}_0(r_0,k_0;r,k)
=O\!\left(\|k-k_0\|_{\mathcal K}^2+\|r-r_0\|_{\mathcal H}^2\right).
\]
This proves the theorem.
\end{proof}
\begin{proof}[Proof of Theorem~\ref{theorem::EIFsoftmaxnorm}]
Write \(\bar r_0 := r_{0,\nu,\gamma,g}\), and let
\[
\tilde m(s,r,k) := V_{r,k}^{\star,\gamma'}(s),
\qquad
m(s,r,k) := \tilde m(s,r_{r,k,\nu,\gamma,g},k).
\]
Under the stated assumptions, Theorem~\ref{theorem::EIFsoftmax} applies with
\(r_0\) replaced by \(\bar r_0\) and \(\gamma\) by \(\gamma'\). Hence the
Riesz representers of the partial derivatives of
\((r,k) \mapsto \mathbb E_0[\tilde m(S_0,r,k)]\) at
\((\bar r_0,k_0)\) are
\[
\tilde\alpha_0(a,s)
=d_{\bar r_0,k_0}^{\star,\gamma'}(a,s)
+\frac{1}{\tau^\star}\widetilde d_{\bar r_0,k_0}^{\star,\gamma'}(a,s)
\]
and
\[
\begin{aligned}
\tilde\beta_0(s',a,s)
&= \frac{\gamma'}{\tau^\star}\,
\widetilde d_{\bar r_0,k_0}^{\star,\gamma'}(a,s)\,
\bigl\{\Phi_{\bar r_0,k_0}^{\star,\gamma'}(s')
- v_{\bar r_0,k_0}^{\star,\gamma'}(a,s)\bigr\} \\
&\quad + d_{\bar r_0,k_0}^{\star,\gamma'}(a,s)\,
\bigl\{\bar r_0(a,s) + \gamma' V_{\bar r_0,k_0}^{\star,\gamma'}(s')
- q_{\bar r_0,k_0}^{\star,\gamma'}(a,s)\bigr\}.
\end{aligned}
\]

Applying Theorem~\ref{theorem::EIFnorm} to the map \(\tilde m\) gives
the balance characterization of \(\alpha_0\) stated in
Theorem~\ref{theorem::EIFsoftmaxnorm}, together with
\[
\beta_0(s',a,s)
= \tilde\beta_0(s',a,s)
+ \alpha_0(a,s)\,
\bigl\{r_0(a,s) - g(s) + \gamma V_{r_0-g,k_0}^{\nu,\gamma}(s')
- q_{r_0-g,k_0}^{\nu,\gamma}(a,s)\bigr\}.
\]
Finally, substituting these representers into Theorem~\ref{theorem::EIF}
gives the stated trajectory-level EIF.
\end{proof}

\subsection{Functional von Mises expansions for examples}

In the following lemmas, we define
\[
V_q^\pi(s) := \sum_{a \in \mathcal{A}} \pi(a \mid s)\, q(a,s), \qquad d_{\rho,r}^{\pi,\gamma}(a,s) := \rho(s)\,\frac{\pi(a \mid s)}{\pi_r(a \mid s) },
\qquad
\pi_r(a \mid s) := \exp\{r(a,s)\}.
\]

\begin{lemma}[von Mises expansion for policy value]
\label{lemma::vonmisespolicyvalue}
For any $r, q \in L^\infty(\lambda)$ and $\rho \in L^\infty(\mu)$,
\begin{align*}
&\; \mathbb E_0\!\left[V_q^\pi(S_0)-V_{r_0,k_0}^\pi(S_0)\right] \\
&\quad + E_0\!\left[d_{\rho,r}^{\pi,\gamma}(A,S)
   \{r(A,S)+\gamma V_q^\pi(S')-q(A,S)\}\right] \\
&\quad + E_0\!\left[\rho(S)\Bigl\{\tfrac{\pi}{\pi_r}(A\mid S)-1\Bigr\}\right] \\[6pt]
&= E_0\!\Big[\{d_0^{\pi,\gamma}(A,S) - d_{\rho,r}^{\pi,\gamma}(A,S)\}\\
&\hspace{27mm}\times
\Bigg\{
\begin{aligned}
&q(A,S) - q_{r_0,k_0}^{\pi,\gamma}(A,S)\\[-1mm]
&\quad - \gamma\,(V_q^\pi(S') - V_{r_0,k_0}^{\pi,\gamma}(S'))
\end{aligned}
\Bigg\}\Big] \\
&\quad + O\!\left(\|r - r_0\|^2_{\mathcal{H}, 0}\right),
\end{align*}
where the big-oh notation depends on $\|r_0\|_{L^\infty(\lambda)}$, $\|r\|_{L^\infty(\lambda)}$, and $\|\rho\|_{L^\infty(\mu)}$.
\end{lemma}
\begin{proof}
    In what follows, we drop explicit dependence on $\pi$ and $\gamma$ in our notation, writing
    $q_{0} := q_{r_0,k_0}^{\pi,\gamma}$ and $d_0 := d_0^{\pi,\gamma}$.

Consider the quantity
\begin{equation}
\label{eqn::prooffullterm}
\begin{aligned}
\mathbb E_0[V_q^\pi(S_0)]-\mathbb E_0[V_{q_0}^\pi(S_0)]
&+ E_0\!\left[d_{\rho,r}(A,S)
\{r(A,S) + \gamma V_q^\pi(S') - q(A,S)\}\right] \\
&+ E_0\!\left[\rho(S)
\Bigl\{\tfrac{\pi}{\pi_r}(A \mid S) - 1\Bigr\}\right].
\end{aligned}
\end{equation}

Adding and subtracting $r_0(A,S)$ inside the expectation, we obtain
\begin{align*}
E_0\!\left[d_{\rho,r}(A,S)\{r(A,S) + \gamma V_q^\pi(S') - q(A,S)\}\right]
&= E_0\!\left[d_{\rho,r}(A,S)\{r_0(A,S) + \gamma V_q^\pi(S') - q(A,S)\}\right] \\
&\quad + E_0\!\left[d_{\rho,r}(A,S)\{r(A,S) - r_0(A,S)\}\right].
\end{align*}
Hence, the quantity in \eqref{eqn::prooffullterm} can be written
\begin{equation}
\label{eqn::prooffullterm2}
\begin{aligned}
\mathbb E_0[V_q^\pi(S_0)]-\mathbb E_0[V_{q_0}^\pi(S_0)]
+ E_0\!\left[d_{\rho,r}(A,S)\{r_0(A,S) + \gamma V_q^\pi(S') - q(A,S)\}\right]
\\ + E_0\!\left[d_{\rho,r}(A,S)\{r(A,S) - r_0(A,S)\}\right]
+ E_0\!\left[\rho(S)\Bigl\{\tfrac{\pi}{\pi_r}(A \mid S) - 1\Bigr\}\right].
\end{aligned}
\end{equation}
By Bellman's equation,
\[
r_0(A,S) \;=\; q_0(A,S) \;-\; \gamma\,E_0\!\left[V_{q_0}^\pi(S') \mid A,S\right].
\]
Thus, by the law of iterated expectations,
\begin{align*}
E_0\!\left[d_{\rho,r}^{\pi,\gamma}(A,S)\{r_0(A,S) + \gamma V_q^\pi(S') - q(A,S)\}\right]
&= E_0\!\Bigg[d_{\rho,r}^{\pi,\gamma}(A,S)
\Bigg\{
\begin{aligned}
&q_0(A,S) - \gamma V_{q_0}^\pi(S')\\[-1mm]
&\quad + \gamma V_q^\pi(S') - q(A,S)
\end{aligned}
\Bigg\}\Bigg] \\
&= E_0\!\Bigg[d_{\rho,r}^{\pi,\gamma}(A,S)
\Bigg\{
\begin{aligned}
&q_0(A,S) - q(A,S)\\[-1mm]
&\quad + \gamma (V_q^\pi(S') - V_{q_0}^\pi(S'))
\end{aligned}
\Bigg\}\Bigg].
\end{align*}
Moreover, for any $q$,
\[
\mathbb E_0[V_q^\pi(S_0)]
\;=\;E_0\!\left[d_0(A,S)\{q(A,S)-\gamma V_q^\pi(S')\}\right].
\]
Therefore,
\begin{align*}
\mathbb E_0[V_q^\pi(S_0)]-\mathbb E_0[V_{q_0}^\pi(S_0)]
&= E_0\!\left[d_0(A,S)\{q(A,S) - q_0(A,S) - \gamma (V_q^\pi(S') - V_{q_0}^\pi(S'))\}\right].
\end{align*}
Combining these results, expression~\eqref{eqn::prooffullterm2} can be written as
\begin{equation}
\label{eqn::prooffullterm3}
\begin{aligned}
& E_0\!\left[d_0(A,S)\{q(A,S) - q_0(A,S) - \gamma (V_q^\pi(S') - V_{q_0}^\pi(S'))\}\right] \\
&\quad + E_0\!\left[d_{\rho,r}(A,S)\{q_0(A,S) - q(A,S) + \gamma (V_q^\pi(S') - V_{q_0}^\pi(S'))\}\right] \\
&\quad + E_0\!\left[d_{\rho,r}(A,S)\{r(A,S) - r_0(A,S)\}\right]
   + E_0\!\left[\rho(S)\Bigl\{\tfrac{\pi}{\pi_r}(A \mid S) - 1\Bigr\}\right] \\
&= E_0\!\left[(d_0(A,S) - d_{\rho,r}(A,S))\{q(A,S) - q_0(A,S) - \gamma (V_q^\pi(S') - V_{q_0}^\pi(S'))\}\right] \\
&\quad + E_0\!\left[d_{\rho,r}(A,S)\{r(A,S) - r_0(A,S)\}\right]
   + E_0\!\left[\rho(S)\Bigl\{\tfrac{\pi}{\pi_r}(A \mid S) - 1\Bigr\}\right].
\end{aligned}
\end{equation}

We turn to the term $E_0\!\left[d_{\rho,r}^{\pi,\gamma}(A,S)\{r(A,S) - r_0(A,S)\}\right]$. Since
\(
\max\{\|r\|_{L^\infty(\lambda)}, \|r_0\|_{L^\infty(\lambda)}\} < M,
\)
Taylor's inequality gives
\[
\left|\exp\{r(A,S) - r_0(A,S)\} - 1 - \{r(A,S) - r_0(A,S)\}\right|
   \;\lesssim\; \{r(A,S) - r_0(A,S)\}^2,
\]
with the constant depending only on $M$. Hence, recalling $\pi_r(a \mid s) = \exp\{r(a,s)\}$, we have
\begin{align*}
E_0\!\left[d_{\rho,r}^{\pi,\gamma}(A,S)\{r(A,S) - r_0(A,S)\}\right]
&= E_0\!\left[d_{\rho,r}^{\pi,\gamma}(A,S)\{\exp\{r(A,S) - r_0(A,S)\} - 1\}\right] \\
&\quad + O\!\left(E_0\!\left[d_{\rho,r}^{\pi,\gamma}(A,S)\{r(A,S) - r_0(A,S)\}^2\right]\right),
\end{align*}
The exponential-ratio expectation in this display satisfies
\begin{align*}
E_0\!\left[d_{\rho,r}^{\pi,\gamma}(A,S)\{\exp\{r(A,S) - r_0(A,S)\} - 1\}\right]
&= E_0\!\left[\rho(S)\Bigl\{\tfrac{\pi}{\pi_0}(A \mid S) - \tfrac{\pi}{\pi_r}(A \mid S)\Bigr\}\right] \\
&= E_0\!\left[\rho(S)\Bigl\{1 - \tfrac{\pi}{\pi_r}(A \mid S)\Bigr\}\right],
\end{align*}
where we used the law of iterated expectations and the fact that $E_0\!\left[\tfrac{\pi}{\pi_0}(A \mid S)\mid S\right] = 1$.
Hence,
\begin{align*}
    E_0\!\left[d_{\rho,r}^{\pi,\gamma}(A,S)\{r(A,S) - r_0(A,S)\}\right]
&= E_0\!\left[\rho(S)\Bigl\{1 - \tfrac{\pi}{\pi_r}(A \mid S)\Bigr\}\right]\\
&\quad + O\!\Big(E_0\!\left[d_{\rho,r}^{\pi,\gamma}(A,S)\right.\\[-1mm]
&\hspace{34mm}\left.\times\{r(A,S) - r_0(A,S)\}^2\right]\Big).
\end{align*}
Equivalently,
\begin{equation}
\label{eqn::proofsecondterm}
\begin{aligned}
   E_0\!\left[d_{\rho,r}(A,S)\{r(A,S) - r_0(A,S)\}\right]
   + E_0\!\left[\rho(S)\Bigl\{\tfrac{\pi}{\pi_r}(A \mid S) - 1\Bigr\}\right]\\
=O\!\left(E_0\!\left[d_{\rho,r}^{\pi,\gamma}(A,S)\{r(A,S) - r_0(A,S)\}^2\right]\right)
\end{aligned}
\end{equation}

Plugging \eqref{eqn::proofsecondterm} into \eqref{eqn::prooffullterm3}, we obtain
\begin{align*}
&\; \mathbb E_0[V_q^\pi(S_0)]-\mathbb E_0[V_{q_0}^\pi(S_0)] \\
&\quad + E_0\!\left[d_{\rho,r}^{\pi,\gamma}(A,S)
   \{r(A,S) + \gamma V_q^\pi(S') - q(A,S)\}\right] \\
&\quad + E_0\!\left[\rho(S)\Bigl\{\tfrac{\pi}{\pi_r}(A \mid S) - 1\Bigr\}\right] \\[6pt]
&= E_0\!\Bigg[d_0(A,S)
\Bigg\{
\begin{aligned}
&q(A,S) - q_0(A,S)\\[-1mm]
&\quad - \gamma (V_q^\pi(S') - V_{q_0}^\pi(S'))
\end{aligned}
\Bigg\}\Bigg] \\
&\quad + E_0\!\Bigg[d_{\rho,r}^{\pi,\gamma}(A,S)
\Bigg\{
\begin{aligned}
&q_0(A,S) - q(A,S)\\[-1mm]
&\quad + \gamma (V_q^\pi(S') - V_{q_0}^\pi(S'))
\end{aligned}
\Bigg\}\Bigg] \\
&\quad + O\!\Big(E_0\!\left[d_{\rho,r}^{\pi,\gamma}(A,S)\right.\\[-1mm]
&\hspace{34mm}\left.\times\{r(A,S) - r_0(A,S)\}^2\right]\Big).
\end{align*}
We conclude that
\begin{align*}
&\; \mathbb E_0[V_q^\pi(S_0)]-\mathbb E_0[V_{q_0}^\pi(S_0)] \\
&\quad + E_0\!\left[d_{\rho,r}^{\pi,\gamma}(A,S)
   \{r(A,S) + \gamma V_q^\pi(S') - q(A,S)\}\right] \\
&\quad + E_0\!\left[\rho(S)\Bigl\{\tfrac{\pi}{\pi_r}(A \mid S) - 1\Bigr\}\right] \\[6pt]
&= E_0\!\Big[\{d_0(A,S) - d_{\rho,r}^{\pi,\gamma}(A,S)\}\\
&\hspace{27mm}\times
\{q(A,S) - q_0(A,S) - \gamma (V_q^\pi(S') - V_{q_0}^\pi(S'))\}\Big] \\
&\quad + O\!\left(E_0\!\left[d_{\rho,r}^{\pi,\gamma}(A,S)\{r(A,S) - r_0(A,S)\}^2\right]\right).
\end{align*}
Finally, since
\[
d_{\rho,r}^{\pi,\gamma}(a,s)
= \rho(s)\,\frac{\pi(a \mid s)}{\pi_r(a \mid s)}
= \rho(s)\,\frac{\pi(a \mid s)}{\pi_0(a \mid s)}\,\exp\{r_0(a,s)-r(a,s)\},
\]
the weight $d_{\rho,r}^{\pi,\gamma}$ is uniformly bounded by a constant depending only on
$\|\rho\|_{L^\infty(\mu)}$, $\|r\|_{L^\infty(\lambda)}$, and $\|r_0\|_{L^\infty(\lambda)}$.
Therefore,
\[
E_0\!\left[d_{\rho,r}^{\pi,\gamma}(A,S)\{r(A,S) - r_0(A,S)\}^2\right]
\lesssim \|r-r_0\|_{\mathcal{H},0}^2,
\]
which yields the stated remainder bound.

\end{proof}

\begin{lemma}[von Mises expansion for policy value with normalization]
\label{lemma::vonmisespolicyvaluenormalization}
Let \(d,\alpha,r,q^{\nu,\gamma},q_\nu^{\pi,\gamma'}\) be bounded, set
\(r_{\nu,g}:=g+(I-\nu)q^{\nu,\gamma}\), and define
\(\pi_r(a\mid s):=\exp\{r(a,s)\}\). Then
\begin{align*}
&\mathbb E_0\!\left[
V_{q_\nu^{\pi,\gamma'}}^\pi(S_0)
-V_{q_{0,\nu}^{\pi,\gamma'}}^\pi(S_0)\right]\\
&\quad+E_0\!\left[
d(A,S)\{r_{\nu,g}(A,S)
+\gamma'V_{q_\nu^{\pi,\gamma'}}^\pi(S')
-q_\nu^{\pi,\gamma'}(A,S)\}\right.\\
&\qquad\left.
+\alpha(A,S)\{r(A,S)-g(S)
+\gamma V_{q^{\nu,\gamma}}^\nu(S')
-q^{\nu,\gamma}(A,S)\}
+\alpha(A,S)-\sum_a\pi_r(a\mid S)\alpha(a,S)
\right]\\
&=O\!\left[
\{\|d-d_0^{\pi,\gamma'}\|_{\mathcal H,0}
+\|\alpha-\alpha_0^{\pi,\nu}\|_{\mathcal H,0}\}
\|\mathcal T_{k_0,\nu,\gamma}
(q^{\nu,\gamma}-q_0^{\nu,\gamma})\|_{\mathcal H,0}\right.\\
&\hspace{20mm}\left.
+\|r-r_0\|_{\mathcal H,0}^2
+\|d-d_0^{\pi,\gamma'}\|_{\mathcal H,0}
\|\mathcal T_{k_0,\pi,\gamma'}
(q_\nu^{\pi,\gamma'}-q_{0,\nu}^{\pi,\gamma'})
\|_{\mathcal H,0}
\right].
\end{align*}
The implicit constant is uniform over bounded subsets on which
\(\mathcal T_{k_0,\nu,\gamma}^{-1}\) is bounded.
\end{lemma}

\begin{proof}
Write
\(\delta_\nu=q^{\nu,\gamma}-q_0^{\nu,\gamma}\) and
\(\delta_\pi=q_\nu^{\pi,\gamma'}-q_{0,\nu}^{\pi,\gamma'}\).
The candidate and true normalized rewards differ by
\begin{equation}
r_{\nu,g}-r_{0,\nu,\gamma,g}=(I-\nu)\delta_\nu.
\label{eqn::normalized-vm-reward-difference}
\end{equation}
Consequently, the conditional mean of the candidate evaluation residual
satisfies
\begin{equation}
E_0\!\left[
r_{\nu,g}(A,S)+\gamma'V_{q_\nu^{\pi,\gamma'}}^\pi(S')
-q_\nu^{\pi,\gamma'}(A,S)\mid A,S\right]
=(I-\nu)\delta_\nu
-\mathcal T_{k_0,\pi,\gamma'}\delta_\pi.
\label{eqn::normalized-vm-evaluation-residual}
\end{equation}
The discounted-occupancy identity gives
\begin{equation}
\mathbb E_0\!\left[
V_{q_\nu^{\pi,\gamma'}}^\pi(S_0)
-V_{q_{0,\nu}^{\pi,\gamma'}}^\pi(S_0)\right]
=E_0\!\left[d_0^{\pi,\gamma'}(A,S)
\{\mathcal T_{k_0,\pi,\gamma'}\delta_\pi\}(A,S)\right].
\label{eqn::normalized-vm-occupancy-identity}
\end{equation}
Substitution of \eqref{eqn::normalized-vm-evaluation-residual} into the
evaluation correction and addition of
\eqref{eqn::normalized-vm-occupancy-identity} give
\begin{align}
&\mathbb E_0\!\left[
V_{q_\nu^{\pi,\gamma'}}^\pi(S_0)
-V_{q_{0,\nu}^{\pi,\gamma'}}^\pi(S_0)\right]\notag\\
&\quad+E_0\!\left[
d(A,S)\{r_{\nu,g}(A,S)
+\gamma'V_{q_\nu^{\pi,\gamma'}}^\pi(S')
-q_\nu^{\pi,\gamma'}(A,S)\}\right]\notag\\
&\qquad
=-E_0[(d-d_0^{\pi,\gamma'})(A,S)
\{\mathcal T_{k_0,\pi,\gamma'}\delta_\pi\}(A,S)]
+E_0[d(A,S)(I-\nu)\delta_\nu(A,S)].
\label{eqn::nuproofstart}
\end{align}

Because
\(\mathcal T_{k_0,\nu,\gamma}q_0^{\nu,\gamma}=r_0-g\),
the normalization correction and behavior-policy centering term equal
\begin{align}
&-E_0[\alpha(A,S)
\{\mathcal T_{k_0,\nu,\gamma}\delta_\nu\}(A,S)]
+E_0[\alpha(A,S)\{r-r_0\}(A,S)]\notag\\
&\qquad
+E_0\!\left[\alpha(A,S)
-\sum_a\pi_r(a\mid S)\alpha(a,S)\right].
\label{eq:step1proof}
\end{align}
Taking \(h=\delta_\nu\) in \eqref{eqn::policyvaluenormbalance}
and adding and subtracting \(d_0^{\pi,\gamma'}\) and
\(\alpha_0^{\pi,\nu}\) gives
\begin{align}
&E_0[d(A,S)(I-\nu)\delta_\nu(A,S)]
-E_0[\alpha(A,S)
\{\mathcal T_{k_0,\nu,\gamma}\delta_\nu\}(A,S)]\notag\\
&\quad
=E_0[(d-d_0^{\pi,\gamma'})(A,S)(I-\nu)\delta_\nu(A,S)]\notag\\
&\qquad
-E_0[(\alpha-\alpha_0^{\pi,\nu})(A,S)
\{\mathcal T_{k_0,\nu,\gamma}\delta_\nu\}(A,S)].
\label{eq:step3proof}
\end{align}
A second-order expansion of the centering term can be written exactly.  Put
$\delta r=r-r_0$.  Since $\pi_r=\pi_0e^{\delta r}$, iterated expectation
gives
\begin{align}
&E_0\!\left[
\alpha(A,S)\delta r(A,S)
+\alpha(A,S)-\sum_a\pi_r(a\mid S)\alpha(a,S)
\right]\notag\\
&\qquad=E_{0,S}\!\left[\sum_a\pi_0(a\mid S)\alpha(a,S)
\{1+\delta r(a,S)-e^{\delta r(a,S)}\}\right].
\label{eqn::policy-normalization-centering-cancellation}
\end{align}
The inequality $|e^u-1-u|\le e^{|u|}u^2/2$, equivalently
Lemma~\ref{lemma::quadraticloss}, applied to
\eqref{eqn::policy-normalization-centering-cancellation}, gives
\[
\left|E_0\!\left[
\alpha(A,S)\{r-r_0\}(A,S)
+\alpha(A,S)-\sum_a\pi_r(a\mid S)\alpha(a,S)
\right]\right|
\lesssim \|r-r_0\|_{\mathcal H,0}^2.
\]
Moreover, bounded invertibility of \(\mathcal T_{k_0,\nu,\gamma}\) and
boundedness of the action-centering operator imply
\begin{equation}
\|(I-\nu)\delta_\nu\|_{\mathcal H,0}
\lesssim
\|\mathcal T_{k_0,\nu,\gamma}\delta_\nu\|_{\mathcal H,0}.
\label{eqn::normalized-centering-operator-bound}
\end{equation}
Cauchy--Schwarz and
\eqref{eqn::normalized-centering-operator-bound} give
\begin{align*}
&\left|E_0[(d-d_0^{\pi,\gamma'})(A,S)
\{\mathcal T_{k_0,\pi,\gamma'}\delta_\pi\}(A,S)]\right|\\
&\qquad\le
\|d-d_0^{\pi,\gamma'}\|_{\mathcal H,0}
\|\mathcal T_{k_0,\pi,\gamma'}\delta_\pi\|_{\mathcal H,0},\\
&\left|E_0[(d-d_0^{\pi,\gamma'})(A,S)
(I-\nu)\delta_\nu(A,S)]\right|\\
&\qquad\lesssim
\|d-d_0^{\pi,\gamma'}\|_{\mathcal H,0}
\|\mathcal T_{k_0,\nu,\gamma}\delta_\nu\|_{\mathcal H,0},\\
&\left|E_0[(\alpha-\alpha_0^{\pi,\nu})(A,S)
\{\mathcal T_{k_0,\nu,\gamma}\delta_\nu\}(A,S)]\right|\\
&\qquad\le
\|\alpha-\alpha_0^{\pi,\nu}\|_{\mathcal H,0}
\|\mathcal T_{k_0,\nu,\gamma}\delta_\nu\|_{\mathcal H,0}.
\end{align*}
Combining these inequalities with \eqref{eqn::nuproofstart},
\eqref{eq:step1proof}, \eqref{eq:step3proof}, and
\eqref{eqn::policy-normalization-centering-cancellation} proves the stated
bound.

\end{proof}

For the following lemma, define
\[
\pi_{r,v^\star}^\star(a \mid s)
\;\propto\;
\mathbf{1}\{a \in \mathcal{A}^\star(s)\}
\exp\!\left\{
    \tfrac{1}{\tau^\star}\bigl(r(a,s) + \gamma v^\star(a,s)\bigr)
\right\}.
\]
Also define
\[
\Phi_{r,v^\star}^\star(s)
\;:=\;
\tau^\star \log\!\Biggl(
\sum_{a \in \mathcal{A}^\star(s)}
\exp\!\left\{
    \tfrac{1}{\tau^\star}\bigl(r(a,s) + \gamma v^\star(a,s)\bigr)
\right\}
\Biggr).
\]

\begin{lemma}[von Mises expansion for soft optimal value]
\label{lemma::vonmises_softstar}
Under Assumption~\ref{cond::boundedpositivity}, let
$q^\star, r, v^\star,\widetilde d^\star \in L^\infty(\lambda)$ and let
$\rho^\star \in L^\infty(\mu)$ be arbitrary functions. Then
    \begin{align*}
&\;\mathbb E_0\!\Big[
    \sum_{a \in \mathcal{A}^\star(S_0)}
        \pi_{r,v^\star}^\star(a \mid S_0)\,q^\star(a,S_0)
        -
    \sum_{a \in \mathcal{A}^\star(S_0)}
        \pi_{r_0,k_0}^\star(a \mid S_0)\,q_{r_0,k_0}^\star(a,S_0)
\Big]
\\
&\quad+\;
E_0\!\Big[
    \rho^\star(S)\,
    \tfrac{\pi_{r,v^\star}^\star}{\pi_r}(A \mid S)\,
    \Bigl\{ r(A,S)
    +\gamma\!\sum_{\tilde a\in\mathcal A^\star(S')}
       \pi_{r,v^\star}^\star(\tilde a\mid S')q^\star(\tilde a,S')
    -q^\star(A,S)\Bigr\}
\Big]
\\
&\quad+\;
E_0\!\Big[
    \rho^\star(S)\,
    \Bigl\{\tfrac{\pi_{r,v^\star}^\star}{\pi_r}(A \mid S)-1\Bigr\}
\Big]
\\
&\quad+\;
\frac{\gamma}{\tau^\star}\,
E_0\!\Big[
       \widetilde d^\star(A,S)\,
       \Bigl(\Phi_{r,v^\star}^\star(S') - v^\star(A,S)\Bigr)
\Big]
\\
&\quad+\;
\frac{1}{\tau^\star}\,
E_0\!\Big[
        \widetilde d^\star(A,S)
        -\sum_{a\in\mathcal A}\pi_r(a\mid S)\widetilde d^\star(a,S)
\Big]
\\[6pt]
&=\;
O\!\Big(
     \|\rho^\star - \rho_{r_0,k_0}^\star\|_{L^2(\rho_0)}\,
     \bigl\|\mathcal{T}_{r_0,k_0}^\star(q^\star - q_{r_0,k_0}^\star)\bigr\|_{\mathcal{H},0}
\\
&\hspace{2.2cm}
     +\;\|\rho^\star - \rho_{r_0,k_0}^\star\|_{L^2(\rho_0)}
     \bigl(\|r-r_0\|_{\mathcal H,0}
     +\|v^\star-v_{r_0,k_0}^\star\|_{\mathcal H,0}\bigr)
\\
&\hspace{2.2cm}
     +\;
     \bigl(\|r-r_0\|_{\mathcal{H},0} + \|v^\star - v_{r_0,k_0}^\star\|_{\mathcal{H},0}\bigr)\,
     \bigl\|\mathcal{T}_{r_0,k_0}^\star(q^\star - q_{r_0,k_0}^\star)\bigr\|_{\mathcal{H},0}
\\
&\hspace{2.2cm}
     +\;
     \|r-r_0\|_{\mathcal{H},0}^{2}
     +\;
     \|v^\star - v_{r_0,k_0}^\star\|_{\mathcal{H},0}^{2}
     +\;
     \|\widetilde d^\star - \widetilde d_{r_0,k_0}^\star\|_{\mathcal{H},0}
     \\
&\hspace{3.0cm}\times
     \Bigl(
     \|r-r_0\|_{\mathcal{H},0}
     +\|v^\star - v_{r_0,k_0}^\star\|_{\mathcal{H},0}
     \Bigr)
\Big),
\end{align*}
where the implicit constant in the big-$O$ term depends only on
\(\|q^\star\|_\infty\),
\(\|r\|_\infty\),
\(\|r_0\|_\infty\),
\(\|v^\star\|_\infty\),
\(\|v_{r_0,k_0}^\star\|_\infty\),
\(\|\rho^\star\|_\infty\),
\(\|\widetilde d^\star\|_\infty\),
\(\gamma\), \(\tau^\star\),
\(\|(\mathcal T_{r_0,k_0}^\star)^{-1}\|_{L^2\to L^2}\), and the constants
in Assumption~\ref{cond::boundedpositivity}.

\end{lemma}

\begin{proof}[Proof of Lemma~\ref{lemma::vonmises_softstar}]
Write \(v_0=v_{r_0,k_0}^\star\),
\(q_0^\star=q_{r_0,k_0}^\star\), and
\(\pi_0^\star=\pi_{r_0,v_0}^\star\). Also set
\[
\pi^\star:=\pi_{r,v^\star}^\star,
\quad e:=q^\star-q_0^\star,
\quad \delta r:=r-r_0,
\quad \delta v:=v^\star-v_0,
\quad \delta Q:=\delta r+\gamma\delta v,
\quad B:=\mathcal T_{r_0,k_0}^\star e.
\]
Here \(q_0^\star=q_{r_0,k_0}^{\pi_0^\star,\gamma}\) is the ordinary
policy-evaluation \(Q\)-function, whereas \(r_0+\gamma v_0\) is the soft
action index that generates \(\pi_0^\star\).
The direct value difference has the exact decomposition
\begin{align}
&\mathbb E_0\!\left[
\sum_a\pi^\star(a\mid S_0)q^\star(a,S_0)
-\sum_a\pi_0^\star(a\mid S_0)q_0^\star(a,S_0)\right]\notag\\
&\quad=
\mathbb E_0\{(\Pi_0e)(S_0)\}
+\mathbb E_0\{((\Pi^\star-\Pi_0)q_0^\star)(S_0)\}
+\mathbb E_0\{((\Pi^\star-\Pi_0)e)(S_0)\}.
\label{eqn::softstar-direct-decomposition}
\end{align}
Here \(\Pi^\star\) and \(\Pi_0\) denote averaging under \(\pi^\star\)
and \(\pi_0^\star\), respectively. The third term in
\eqref{eqn::softstar-direct-decomposition} is second order: the
softmax Lipschitz bound, Lemma~\ref{lemma::invertibility}, and
Lemma~\ref{lemma::initialpooledbounds} give
\begin{equation}
\left|\mathbb E_0\{((\Pi^\star-\Pi_0)e)(S_0)\}\right|
\lesssim
(\|\delta r\|_{\mathcal H,0}+\|\delta v\|_{\mathcal H,0})
\|B\|_{\mathcal H,0}.
\label{eqn::softstar-cross-bound}
\end{equation}

Define
\begin{align}
\mathcal E_{\mathrm{eval}}:={}&
\mathbb E_0\{(\Pi^\star q^\star)(S_0)-(\Pi_0q_0^\star)(S_0)\}
\notag\\
&+E_0\!\left[
\rho^\star(S)\frac{\pi^\star(A\mid S)}{\pi_r(A\mid S)}
\{r(A,S)+\gamma(\Pi^\star q^\star)(S')-q^\star(A,S)\}\right]
\notag\\
&+E_0\!\left[
\rho^\star(S)\left\{\frac{\pi^\star(A\mid S)}{\pi_r(A\mid S)}-1\right\}\right].
\label{eqn::softstar-evaluation-definition}
\end{align}
Define the fixed-policy and policy-change components by
\begin{align}
\mathcal E_{\mathrm{fp}}:={}&
\mathbb E_0\{(\Pi_0e)(S_0)\}
+E_0\!\left[
\rho^\star(S)\frac{\pi_0^\star(A\mid S)}{\pi_r(A\mid S)}
\{r(A,S)+\gamma(\Pi_0q^\star)(S')-q^\star(A,S)\}\right]
\notag\\
&+E_0\!\left[
\rho^\star(S)\left\{\frac{\pi_0^\star(A\mid S)}{\pi_r(A\mid S)}-1\right\}\right],
\label{eqn::softstar-fixed-policy-component}\\
\mathcal D_\pi:={}&
\mathbb E_0\{((\Pi^\star-\Pi_0)q_0^\star)(S_0)\}
\notag\\
&+E_0\!\left[
\rho^\star(S)\frac{(\pi^\star-\pi_0^\star)(A\mid S)}{\pi_r(A\mid S)}
\{r(A,S)+\gamma(\Pi_0q^\star)(S')-q^\star(A,S)+1\}\right]
\notag\\
&+\gamma E_0\!\left[
\rho^\star(S)\frac{\pi^\star(A\mid S)}{\pi_r(A\mid S)}
\{(\Pi^\star-\Pi_0)q^\star\}(S')\right].
\label{eqn::softstar-policy-change-component}
\end{align}
Adding and subtracting the corrections that use \(\pi_0^\star\) in
\eqref{eqn::softstar-evaluation-definition} gives the exact identity
\begin{equation}
\mathcal E_{\mathrm{eval}}
=\mathcal E_{\mathrm{fp}}+\mathcal D_\pi
+\mathbb E_0\{((\Pi^\star-\Pi_0)e)(S_0)\}.
\label{eqn::softstar-evaluation-decomposition}
\end{equation}
Set
\[
\delta\pi=\pi^\star-\pi_0^\star,
\quad \delta\Pi=\Pi^\star-\Pi_0,
\quad W=\rho^\star\pi^\star/\pi_r,
\quad W_0=d_{r_0,k_0}^\star,
\quad \rho_0^\star=\rho_{r_0,k_0}^\star,
\]
and define
\[
G_{\delta Q}(s)
:=\operatorname{Cov}_{\pi_0^\star}
\{q_0^\star,\delta Q\}(s)
=\Pi_0\{(q_0^\star-V_0^\star)\delta Q\}(s).
\]
Taylor's theorem for the finite-action softmax map gives
\begin{align}
(\delta\Pi q_0^\star)(s)
={}&\frac1{\tau^\star}G_{\delta Q}(s)+R_{\Pi}(s),
\notag\\
\|R_{\Pi}\|_{L^1(\mu)}
\lesssim{}&\|\delta Q\|_{\mathcal H,0}^2.
\label{eqn::softstar-policy-expansion}
\end{align}
Bellman's equation and \(B=\mathcal T_{r_0,k_0}^\star e\) give
\begin{equation}
r+\gamma\mathcal P_{k_0}\Pi_0q^\star-q^\star
=\delta r-B.
\label{eqn::softstar-candidate-bellman-residual}
\end{equation}
Substitution of \eqref{eqn::softstar-policy-expansion} and
\eqref{eqn::softstar-candidate-bellman-residual} into
\eqref{eqn::softstar-policy-change-component} gives the exact expansion
\begin{align}
\mathcal D_\pi
={}&\frac1{\tau^\star}\left[
\mathbb E_0\{G_{\delta Q}(S_0)\}
+\gamma E_0\{W_0(A,S)G_{\delta Q}(S')\}\right]
+R_{\mathrm{pol}},
\label{eqn::softstar-policy-change-linearization}\\
R_{\mathrm{pol}}
={}&\mathbb E_0\{R_{\Pi}(S_0)\}
+E_0\!\left[
\rho^\star(S)\frac{\delta\pi(A\mid S)}{\pi_r(A\mid S)}
\{\delta r(A,S)-B(A,S)+1\}\right]
\notag\\
&+\gamma E_0\{W_0(A,S)R_{\Pi}(S')\}
+\gamma E_0[(W-W_0)(A,S)\{\delta\Pi q_0^\star\}(S')]
\notag\\
&+\gamma E_0[W(A,S)\{\delta\Pi e\}(S')].
\label{eqn::softstar-policy-change-remainder-decomposition}
\end{align}
The apparently linear part of the second term in
\eqref{eqn::softstar-policy-change-remainder-decomposition} satisfies the
exact centering identity
\begin{align}
E_0\!\left[
\rho^\star(S)\frac{\delta\pi(A\mid S)}{\pi_r(A\mid S)}\right]
&=E_{0,S}\!\left[
\rho^\star(S)\sum_a
\{e^{-\delta r(a,S)}-1\}\delta\pi(a\mid S)\right],
\label{eqn::softstar-policy-centering-cancellation}
\end{align}
because \(\sum_a\delta\pi(a\mid S)=0\).  The softmax Lipschitz bound,
\eqref{eqn::softstar-policy-centering-cancellation}, and the density
comparisons in Assumption~\ref{cond::boundedpositivity} give the following
termwise bounds. First,
\begin{equation}
\left|\mathbb E_0\{R_\Pi(S_0)\}\right|
+\gamma\left|E_0\{W_0(A,S)R_\Pi(S')\}\right|
\lesssim \|\delta Q\|_{\mathcal H,0}^2.
\label{eqn::softstar-policy-remainder-taylor-bounds}
\end{equation}
Second, Cauchy--Schwarz and the softmax Lipschitz bound give
\begin{align}
&\left|E_0\!\left[
\rho^\star(S)\frac{\delta\pi(A\mid S)}{\pi_r(A\mid S)}
\{\delta r(A,S)-B(A,S)\}\right]\right|\notag\\
&\qquad\lesssim
\|\delta Q\|_{\mathcal H,0}
\{\|\delta r\|_{\mathcal H,0}+\|B\|_{\mathcal H,0}\},
\label{eqn::softstar-policy-remainder-residual-bound}\\
&\left|E_0\!\left[
\rho^\star(S)\frac{\delta\pi(A\mid S)}{\pi_r(A\mid S)}\right]\right|
\lesssim
\|\delta r\|_{\mathcal H,0}\|\delta Q\|_{\mathcal H,0}.
\label{eqn::softstar-policy-remainder-centering-bound}
\end{align}
For the term involving \(W-W_0\), the exact
decomposition
\begin{align}
W-W_0
={}&(\rho^\star-\rho_0^\star)\frac{\pi^\star}{\pi_r}
+\rho_0^\star\frac{\pi^\star-\pi_0^\star}{\pi_r}
+\rho_0^\star\pi_0^\star
\frac{\pi_0-\pi_r}{\pi_r\pi_0}
\label{eqn::softstar-weight-difference-decomposition}
\end{align}
implies
\begin{align}
\|W-W_0\|_{L^2(P_0)}
&\lesssim
\|\rho^\star-\rho_0^\star\|_{L^2(\rho_0)}
+\|\delta r\|_{\mathcal H,0}
+\|\delta Q\|_{\mathcal H,0},\notag\\
\|\delta\Pi q_0^\star\|_{L^2(\mu)}
&\lesssim \|\delta Q\|_{\mathcal H,0}.
\label{eqn::softstar-weight-and-policy-bounds}
\end{align}
It follows that
\begin{align}
\left|E_0[(W-W_0)(A,S)\{\delta\Pi q_0^\star\}(S')]\right|
\lesssim{}&
\{\|\rho^\star-\rho_0^\star\|_{L^2(\rho_0)}
+\|\delta r\|_{\mathcal H,0}
+\|\delta Q\|_{\mathcal H,0}\}
\|\delta Q\|_{\mathcal H,0},
\label{eqn::softstar-policy-remainder-weight-bound}\\
\left|E_0[W(A,S)\{\delta\Pi e\}(S')]\right|
\lesssim{}&
\|\delta Q\|_{\mathcal H,0}\|e\|_{\mathcal H,0}
\lesssim
\|\delta Q\|_{\mathcal H,0}\|B\|_{\mathcal H,0}.
\label{eqn::softstar-policy-remainder-cross-bound}
\end{align}
The last inequality uses bounded invertibility of
\(\mathcal T_{r_0,k_0}^\star\). Combining
\eqref{eqn::softstar-policy-remainder-taylor-bounds}--
\eqref{eqn::softstar-policy-remainder-cross-bound}, using
\(\|\delta Q\|_{\mathcal H,0}
\lesssim\|\delta r\|_{\mathcal H,0}
+\|\delta v\|_{\mathcal H,0}\), and applying \(2xy\le x^2+y^2\)
to products involving only \(\delta r\) and \(\delta v\), yields
\begin{align}
|R_{\mathrm{pol}}|
\lesssim{}&
\|\delta Q\|_{\mathcal H,0}\|B\|_{\mathcal H,0}
+\|\rho^\star-\rho_{r_0,k_0}^\star\|_{L^2(\rho_0)}
\|\delta Q\|_{\mathcal H,0}
\notag\\
&+\|\delta r\|_{\mathcal H,0}^2
+\|\delta v\|_{\mathcal H,0}^2.
\label{eqn::softstar-policy-change-remainder}
\end{align}
Finally, Lemma~\ref{lemma::adjointident} and
\eqref{eqn::tildedbalance} give
\begin{align}
&\mathbb E_0\{G_{\delta Q}(S_0)\}
+\gamma E_0\{W_0(A,S)G_{\delta Q}(S')\}\notag\\
&\quad=E_0\{W_0(A,S)(q_0^\star-V_0^\star)(A,S)
\delta Q(A,S)\}\notag\\
&\quad=E_0\{\widetilde d_{r_0,k_0}^\star(A,S)
(\mathcal T_{r_0,k_0}^\star\delta Q)(A,S)\}.
\label{eqn::softstar-policy-leading-identity}
\end{align}
Lemma~\ref{lemma::vonmisespolicyvalue}, applied to
\eqref{eqn::softstar-fixed-policy-component} with target policy
\(\pi_0^\star\), gives
\begin{equation}
|\mathcal E_{\mathrm{fp}}|
\lesssim
\{\|\rho^\star-\rho_{r_0,k_0}^\star\|_{L^2(\rho_0)}
+\|\delta r\|_{\mathcal H,0}\}\|B\|_{\mathcal H,0}
+\|\delta r\|_{\mathcal H,0}^2.
\label{eqn::softstar-fixed-policy-bound}
\end{equation}
Combining \eqref{eqn::softstar-evaluation-decomposition},
\eqref{eqn::softstar-cross-bound},
\eqref{eqn::softstar-fixed-policy-bound}, and
\eqref{eqn::softstar-policy-change-linearization}--
\eqref{eqn::softstar-policy-leading-identity} gives
\begin{equation}
\mathcal E_{\mathrm{eval}}
=
\frac1{\tau^\star}E_0\!\left[
\widetilde d_{r_0,k_0}^\star(A,S)
\{\mathcal T_{r_0,k_0}^\star\delta Q\}(A,S)\right]
+R_{\mathrm{eval}},
\label{eqn::softstar-master-expansion}
\end{equation}
where
\begin{align}
|R_{\mathrm{eval}}|
\lesssim{}&
\{\|\rho^\star-\rho_{r_0,k_0}^\star\|_{L^2(\rho_0)}
+\|\delta r\|_{\mathcal H,0}
+\|\delta v\|_{\mathcal H,0}\}\|B\|_{\mathcal H,0}
\notag\\
&+\|\rho^\star-\rho_{r_0,k_0}^\star\|_{L^2(\rho_0)}
\{\|\delta r\|_{\mathcal H,0}+\|\delta v\|_{\mathcal H,0}\}
\notag\\
&+\|\delta r\|_{\mathcal H,0}^2
+\|\delta v\|_{\mathcal H,0}^2.
\label{eqn::softstar-evaluation-remainder}
\end{align}
The terms involving \(B\) in
\eqref{eqn::softstar-evaluation-remainder} are precisely those in
\eqref{eqn::softstar-fixed-policy-bound},
\eqref{eqn::softstar-policy-change-remainder}, and
\eqref{eqn::softstar-cross-bound}.  The product of the occupancy-ratio and
policy errors, as well as the two squared errors, is given explicitly in
\eqref{eqn::softstar-policy-change-remainder}.

We now display the cancellations in the leading term of
\eqref{eqn::softstar-master-expansion}.
For \(d\in L^2(\lambda)\), define
\begin{align}
C_v(d)&:=\frac{\gamma}{\tau^\star}E_0\!\left[
d(A,S)\{\Phi_{r,v^\star}^\star(S')-v^\star(A,S)\}\right],
\label{eqn::softstar-Cv-definition}\\
C_r(d)&:=\frac1{\tau^\star}E_0\!\left[
d(A,S)-\sum_a\pi_r(a\mid S)d(a,S)\right].
\label{eqn::softstar-Cr-definition}
\end{align}
Put
\(\Delta_d=\widetilde d^\star-\widetilde d_{r_0,k_0}^\star\).
Define the log-sum-exp remainder
\begin{equation}
R_\Phi(s)
:=\Phi_{r,v^\star}^\star(s)-\Phi_{r_0,v_0}^\star(s)
-(\Pi_0\delta Q)(s).
\label{eqn::softstar-logsumexp-remainder-definition}
\end{equation}
The Hessian of the finite-action log-sum-exp map is
\(\tau^{\star-1}\) times a covariance matrix.  Taylor's theorem therefore
gives the pointwise bound
\begin{equation}
0\le R_\Phi(s)
\le \frac1{2\tau^\star}
\sum_{a\in\mathcal A^\star(s)}\delta Q(a,s)^2.
\label{eqn::softstar-logsumexp-pointwise-bound}
\end{equation}
Because \(\pi_r=\pi_0\exp(\delta r)\), iterated expectation gives
\begin{equation}
C_r(\widetilde d^\star)
=-\frac1{\tau^\star}E_0\!\left[
\widetilde d^\star\delta r\right]+R_r(\widetilde d^\star),
\qquad
|R_r(\widetilde d^\star)|
\lesssim\|\widetilde d^\star\|_\infty
\|\delta r\|_{\mathcal H,0}^2,
\label{eqn::softstar-reward-cancellation}
\end{equation}
where the remainder bound follows from
\(|e^x-1-x|\le e^{|x|}x^2/2\) and the boundedness assumed in the lemma.
Conditional expectation,
\(v_0=\mathcal P_{k_0}\Phi_{r_0,v_0}^\star\),
and \eqref{eqn::softstar-logsumexp-remainder-definition} give
\begin{equation}
C_v(\widetilde d^\star)
=\frac{\gamma}{\tau^\star}E_0\!\left[
\widetilde d^\star
\{\mathcal P_{k_0}(\Pi_0\delta Q+R_\Phi)-\delta v\}\right].
\label{eqn::softstar-bellman-correction-expansion}
\end{equation}
Since
\begin{equation}
\mathcal T_{r_0,k_0}^\star\delta Q
=\delta r+\gamma\delta v
-\gamma\mathcal P_{k_0}\Pi_0\delta Q,
\label{eqn::softstar-linear-bellman-expansion}
\end{equation}
equations~\eqref{eqn::softstar-reward-cancellation},
\eqref{eqn::softstar-bellman-correction-expansion}, and
\eqref{eqn::softstar-linear-bellman-expansion} yield
\begin{equation}
\begin{aligned}
&\frac1{\tau^\star}E_0\!\left[
\widetilde d_{r_0,k_0}^\star
\{\mathcal T_{r_0,k_0}^\star\delta Q\}\right]
+C_v(\widetilde d^\star)+C_r(\widetilde d^\star)\\
&\quad=-\frac1{\tau^\star}E_0\!\left[
\Delta_d\{\mathcal T_{r_0,k_0}^\star\delta Q\}\right]
+\frac{\gamma}{\tau^\star}E_0\!\left[
\widetilde d^\star\{\mathcal P_{k_0}R_\Phi\}\right]
+R_r(\widetilde d^\star).
\end{aligned}
\label{eqn::softstar-candidate-cancellation}
\end{equation}
The boundedness of \(\mathcal T_{r_0,k_0}^\star\),
\eqref{eqn::softstar-logsumexp-pointwise-bound}, and the density comparisons
in Assumption~\ref{cond::boundedpositivity} give
\begin{align}
\|\mathcal T_{r_0,k_0}^\star\delta Q\|_{\mathcal H,0}
&\lesssim\|\delta r\|_{\mathcal H,0}
+\|\delta v\|_{\mathcal H,0},
\label{eqn::softstar-linear-bellman-bound}\\
E_0|\{\mathcal P_{k_0}R_\Phi\}(A,S)|
&\lesssim \|\delta r\|_{\mathcal H,0}^2
+\|\delta v\|_{\mathcal H,0}^2.
\label{eqn::softstar-logsumexp-integrated-bound}
\end{align}
Cauchy--Schwarz applied to the expectation involving \(\Delta_d\) in
\eqref{eqn::softstar-candidate-cancellation}, followed by
\eqref{eqn::softstar-linear-bellman-bound} and
\eqref{eqn::softstar-logsumexp-integrated-bound}, gives
\begin{equation}
\begin{aligned}
&\left|\frac1{\tau^\star}E_0\!\left[
\widetilde d_{r_0,k_0}^\star
\{\mathcal T_{r_0,k_0}^\star\delta Q\}\right]
+C_v(\widetilde d^\star)+C_r(\widetilde d^\star)\right|\\
&\qquad\lesssim
\|\widetilde d^\star-\widetilde d_{r_0,k_0}^\star\|_{\mathcal H,0}
\{\|\delta r\|_{\mathcal H,0}+\|\delta v\|_{\mathcal H,0}\}
+\|\delta r\|_{\mathcal H,0}^2
+\|\delta v\|_{\mathcal H,0}^2.
\end{aligned}
\label{eqn::softstar-signed-ratio-remainder}
\end{equation}
Combining \eqref{eqn::softstar-master-expansion},
\eqref{eqn::softstar-evaluation-remainder}, and
\eqref{eqn::softstar-signed-ratio-remainder} gives the bound stated in the
lemma.
\end{proof}

\subsection{Proofs for Section~\ref{sec::examplesestimators}}

\begin{proof}[Proof of Theorem~\ref{theorem::ALex1}]
Let \(\phi_n(O)\) be the trajectory-level uncentered influence function formed
by \(V_n^{\pi,\gamma}(S_0)\) plus the average over \(t=0,\ldots,T-1\) of the
transition correction in the estimator, and define \(\phi_0\) analogously at
the true nuisances. Then
\(\psi_n=\mathbb P_n\phi_n\), \(\mathbb P_0\phi_0=\psi_0\), and
\(\chi_0=\phi_0-\psi_0\). The decomposition in the proof of
Theorem~\ref{thm:asymptotic_linearity} gives
\[
\psi_n-\psi_0
=\mathbb P_n\chi_0
+(\mathbb P_n-\mathbb P_0)(\phi_n-\phi_0)
+\{\mathbb P_0\phi_n-\psi_0\}.
\]
Assumptions~\ref{assump:boundedness::policy},
\ref{assump:splitting::policy}, and~\ref{assump:consistency::policy}, together
with the Lipschitz continuity of \(x\mapsto e^{-x}\) on the uniformly bounded
range of \(r_n\), give
\begin{align}
\|\phi_n-\phi_0\|_{L^2(\mathbb P_0)}
\lesssim{}&
\|\rho_n^{\pi,\gamma}-\rho_0^{\pi,\gamma}\|_{L^2(\rho_0)}
+\|r_n-r_0\|_{\mathcal H,0}\notag\\
&+\|q_n^{\pi,\gamma}-q_{r_0,k_0}^{\pi,\gamma}\|_{\mathcal H,0}
=o_p(1).
\label{eqn::policy-correction-L2-continuity}
\end{align}
Here and below, Assumption~\ref{cond::boundedpositivity} converts the pooled
one-transition norms into the corresponding trajectory-component norms, and
the fixed trajectory length \(T\) converts componentwise bounds into
\(L^2(\mathbb P_0)\) bounds.
Lemma~\ref{lemma::crossfitprocess}, applied foldwise, then makes the
empirical-process term \(o_p(n^{-1/2})\). By
Lemma~\ref{lemma::vonmisespolicyvalue},
\begin{align*}
|\mathbb P_0\phi_n-\psi_0|
\lesssim{}&
\{\|\rho_n^{\pi,\gamma}-\rho_0^{\pi,\gamma}\|_{L^2(\rho_0)}
+\|r_n-r_0\|_{\mathcal H,0}\}\\
&\times
\|\mathcal T_{k_0,\pi,\gamma}
(q_n^{\pi,\gamma}-q_{r_0,k_0}^{\pi,\gamma})\|_{\mathcal H,0}
+\|r_n-r_0\|_{\mathcal H,0}^2.
\end{align*}
This is \(o_p(n^{-1/2})\) by
Condition~\ref{assump:rates::policy} and
Condition~\ref*{cond::reward}. Substitution into the decomposition proves
the expansion in Theorem~\ref{theorem::ALex1}; its normality and efficiency
claims follow from Theorems~\ref{thm:asymptotic_linearity}
and~\ref{theorem::EIFvalue}.
\end{proof}

\begin{proof}[Proof of Theorem~\ref{theorem::ALex1b}]
Let \(\phi_n(O)\) be \(V_n^\star(S_0)\) plus the average over the trajectory
of the four transition corrections in the estimator, and define \(\phi_0\)
at the true nuisances. Then
\(\psi_n=\mathbb P_n\phi_n\), \(\mathbb P_0\phi_0=\psi_0\), and
\(\chi_0=\phi_0-\psi_0\). As in the proof of
Theorem~\ref{thm:asymptotic_linearity},
\[
\psi_n-\psi_0
=\mathbb P_n\chi_0
+(\mathbb P_n-\mathbb P_0)(\phi_n-\phi_0)
+\{\mathbb P_0\phi_n-\psi_0\}.
\]
Assumptions~\ref{assump:boundedness::softstar},
\ref{assump:splitting::softstar}, and~\ref{assump:consistency::softstar},
together with the finite-action softmax and log-sum-exp Lipschitz bounds, give
\begin{align}
\|\phi_n-\phi_0\|_{L^2(\mathbb P_0)}
\lesssim{}&
\|\rho_n^\star-\rho_{r_0,k_0}^\star\|_{L^2(\rho_0)}
+\|\widetilde d_n^\star-\widetilde d_{r_0,k_0}^\star\|_{\mathcal H,0}\notag\\
&+\|q_n^\star-q_{r_0,k_0}^\star\|_{\mathcal H,0}
+\|v_n^\star-v_{r_0,k_0}^\star\|_{\mathcal H,0}
+\|r_n-r_0\|_{\mathcal H,0}
=o_p(1).
\label{eqn::softstar-correction-L2-continuity}
\end{align}
Lemma~\ref{lemma::crossfitprocess}, applied foldwise, therefore makes the
empirical-process term \(o_p(n^{-1/2})\).
Lemma~\ref{lemma::vonmises_softstar}, evaluated at the estimated nuisances,
bounds the population term by
\begin{align*}
O_p\!\Big(&
\|\rho_n^\star-\rho_{r_0,k_0}^\star\|_{L^2(\rho_0)}
\|\mathcal T_{r_0,k_0}^\star(q_n^\star-q_{r_0,k_0}^\star)\|_{\mathcal H,0}\\
&+\|\rho_n^\star-\rho_{r_0,k_0}^\star\|_{L^2(\rho_0)}
\{\|r_n-r_0\|_{\mathcal H,0}
+\|v_n^\star-v_{r_0,k_0}^\star\|_{\mathcal H,0}\}\\
&+\{\|r_n-r_0\|_{\mathcal H,0}
+\|v_n^\star-v_{r_0,k_0}^\star\|_{\mathcal H,0}\}
\|\mathcal T_{r_0,k_0}^\star(q_n^\star-q_{r_0,k_0}^\star)\|_{\mathcal H,0}\\
&+\|r_n-r_0\|_{\mathcal H,0}^2
+\|v_n^\star-v_{r_0,k_0}^\star\|_{\mathcal H,0}^2\\
&+\|\widetilde d_n^\star-\widetilde d_{r_0,k_0}^\star\|_{\mathcal H,0}
\{\|r_n-r_0\|_{\mathcal H,0}
+\|v_n^\star-v_{r_0,k_0}^\star\|_{\mathcal H,0}\}\Big).
\end{align*}
Condition~\ref*{cond::reward} and Condition~\ref{assump:rates::softstar}, in
particular part~\ref{assump:rho-policy-product-softstar}, make every term in
this display \(o_p(n^{-1/2})\). Hence
\(\mathbb P_0\phi_n-\psi_0=o_p(n^{-1/2})\), and substitution into the
preceding decomposition proves the expansion in
Theorem~\ref{theorem::ALex1b}. Its normality and efficiency conclusions
follow from Theorems~\ref{thm:asymptotic_linearity}
and~\ref{theorem::EIFsoftmax}.
\end{proof}

\begin{proof}[Proof of Theorem~\ref{theorem::ALex2}]
Let \(\phi_n(O)\) be \(V_{n,\nu}^{\pi,\gamma'}(S_0)\) plus the average over
the trajectory of the three transition corrections in the estimator, and
define \(\phi_0\) at the true nuisances. Then
\(\psi_n=\mathbb P_n\phi_n\), \(\mathbb P_0\phi_0=\psi_0\), and
\(\chi_0=\phi_0-\psi_0\). The trajectory-level decomposition gives
\[
\psi_n-\psi_0
=\mathbb P_n\chi_0
+(\mathbb P_n-\mathbb P_0)(\phi_n-\phi_0)
+\{\mathbb P_0\phi_n-\psi_0\}.
\]
Assumptions~\ref{assump:boundedness::policynorm},
\ref{assump:splitting::policynorm}, and
\ref{assump:consistency::policynorm}, together with boundedness of the
action-averaging operators, give
\begin{align}
\|\phi_n-\phi_0\|_{L^2(\mathbb P_0)}
\lesssim{}&
\|d_n^{\pi,\gamma'}-d_0^{\pi,\gamma'}\|_{\mathcal H,0}
+\|\alpha_n^{\pi,\nu}-\alpha_0^{\pi,\nu}\|_{\mathcal H,0}
+\|r_n-r_0\|_{\mathcal H,0}\notag\\
&+\|q_n^{\nu,\gamma}-q_0^{\nu,\gamma}\|_{\mathcal H,0}
+\|q_{n,\nu}^{\pi,\gamma'}-q_{0,\nu}^{\pi,\gamma'}\|_{\mathcal H,0}
=o_p(1).
\label{eqn::normalized-policy-correction-L2-continuity}
\end{align}
Lemma~\ref{lemma::crossfitprocess}, applied foldwise, therefore makes the
empirical-process term \(o_p(n^{-1/2})\). By
Lemma~\ref{lemma::vonmisespolicyvaluenormalization},
\begin{align*}
|\mathbb P_0\phi_n-\psi_0|
\lesssim{}&
\|d_n^{\pi,\gamma'}-d_0^{\pi,\gamma'}\|_{\mathcal H,0}
\|\mathcal T_{k_0,\pi,\gamma'}
(q_{n,\nu}^{\pi,\gamma'}-q_{0,\nu}^{\pi,\gamma'})\|_{\mathcal H,0}\\
&+\{\|d_n^{\pi,\gamma'}-d_0^{\pi,\gamma'}\|_{\mathcal H,0}
+\|\alpha_n^{\pi,\nu}-\alpha_0^{\pi,\nu}\|_{\mathcal H,0}\}\\
&\hspace{13mm}\times
\|\mathcal T_{k_0,\nu,\gamma}
(q_n^{\nu,\gamma}-q_0^{\nu,\gamma})\|_{\mathcal H,0}
+\|r_n-r_0\|_{\mathcal H,0}^2.
\end{align*}
This is \(o_p(n^{-1/2})\) by
Condition~\ref{assump:rates::policynorm} and
Condition~\ref*{cond::reward}. Substitution into the decomposition proves
the expansion in Theorem~\ref{theorem::ALex2}; its normality and efficiency
claims follow from Theorems~\ref{thm:asymptotic_linearity}
and~\ref{theorem::EIFvaluenorm}.
\end{proof}

\subsection{Structural coefficients under reward normalization}

\begin{proof}[Proof of Theorem~\ref{theorem::EIFcoefnorm}]
For \(j=1,\ldots,p\), define
\[
\tilde m_j(s,r,k)
:=\sum_{a\in\mathcal A}\omega(a\mid s)x_j(a,s)r(a,s).
\]
Its kernel derivative is zero, while its reward derivative satisfies
\[
\mathbb E_0\{\partial_r\tilde m_j(S_0,\bar r_0,k_0)[h]\}
=\mathbb E_0\!\left[
\sum_{a\in\mathcal A}\omega(a\mid S_0)x_j(a,S_0)h(a,S_0)\right].
\]
Because \(x\) is bounded and \(\omega\) is fixed, each \(\tilde m_j\) is
bounded linear in the reward and independent of the kernel. Hence
Conditions~\ref{cond::boundedfunc::norm}--\ref{cond::lipschitz::norm} hold
with zero second-order remainder.
We first verify that the balance equation defining
\(\alpha_{0,j}^x\) has a unique solution. Define the bounded linear
functional
\[
L_j(h):=\mathbb E_0\!\left[
\sum_{a\in\mathcal A}\omega(a\mid S_0)x_j(a,S_0)
\{h(a,S_0)-\nu h(S_0)\}\right].
\]
Lemma~\ref{lemma::initialpooledbounds}, boundedness of \(x\), and boundedness
of \(I-\nu\) show that \(L_j\) is continuous on \(T_{\mathcal H}\).
Lemma~\ref{lemma::invertibility} shows that
\(\mathcal T_{k_0,\nu,\gamma}^{-1}\) is bounded. Hence
\(f\mapsto L_j(\mathcal T_{k_0,\nu,\gamma}^{-1}f)\) is a continuous linear
functional on \(T_{\mathcal H}\). By the Riesz representation theorem, there
is a unique \(\alpha_{0,j}^x\in T_{\mathcal H}\) such that
\begin{equation}
E_0\{\alpha_{0,j}^x(A,S)f(A,S)\}
=L_j(\mathcal T_{k_0,\nu,\gamma}^{-1}f),
\qquad f\in T_{\mathcal H}.
\label{eqn::coef-representer-riesz-construction}
\end{equation}
Taking \(f=\mathcal T_{k_0,\nu,\gamma}h\) in
\eqref{eqn::coef-representer-riesz-construction} gives
\eqref{eqn::coef-representer-balance}; surjectivity of
\(\mathcal T_{k_0,\nu,\gamma}\) gives uniqueness in that balance equation.

For a reward direction $h$, the derivative of the normalized reward is
\begin{equation}
D_r\bar r_0[h]
=(I-\nu)\mathcal T_{k_0,\nu,\gamma}^{-1}h.
\label{eqn::coef-normalized-reward-derivative}
\end{equation}
Substituting
$f=\mathcal T_{k_0,\nu,\gamma}^{-1}h$ into
\eqref{eqn::coef-representer-balance} gives
\begin{align}
&\mathbb E_0\!\left[\sum_a\omega(a\mid S_0)x_j(a,S_0)
\{D_r\bar r_0[h]\}(a,S_0)\right]\notag\\
&\qquad=E_0\!\left[\alpha_{0,j}^x(A,S)
\{f(A,S)-\gamma\nu f(S')\}\right]
=E_0\{\alpha_{0,j}^x(A,S)h(A,S)\}.
\label{eqn::coef-reward-riesz-identity}
\end{align}
Thus, $\alpha_{0,j}^x$ is the reward representer.  For a bounded kernel
score $\beta$, differentiation of the fixed-policy Bellman equation gives
\begin{equation}
\mathcal T_{k_0,\nu,\gamma}\dot q_\beta
=\gamma\mathcal P_{k_0\beta}V_{r_0-g,k_0}^{\nu,\gamma}.
\label{eqn::coef-kernel-bellman-derivative}
\end{equation}
Using \eqref{eqn::coef-representer-balance} with $h=\dot q_\beta$ and then
\eqref{eqn::coef-kernel-bellman-derivative}, we obtain
\begin{align}
&\mathbb E_0\!\left[\sum_a\omega(a\mid S_0)x_j(a,S_0)
\{(I-\nu)\dot q_\beta\}(a,S_0)\right]\notag\\
&\quad=\gamma E_0\!\left[\alpha_{0,j}^x(A,S)\beta(S',A,S)
V_{r_0-g,k_0}^{\nu,\gamma}(S')\right]\notag\\
&\quad=E_0\!\left[\beta(S',A,S)\alpha_{0,j}^x(A,S)
\{r_0(A,S)-g(S)+\gamma V_{r_0-g,k_0}^{\nu,\gamma}(S')
-q_{r_0-g,k_0}^{\nu,\gamma}(A,S)\}\right].
\label{eqn::coef-kernel-riesz-identity}
\end{align}
The last equality uses $E_0\{\beta(S',A,S)\mid A,S\}=0$ and Bellman's
equation.  Hence the kernel representer is
\[
\alpha_{0,j}^x(a,s)
\{r_0(a,s)-g(s)+\gamma V_{r_0-g,k_0}^{\nu,\gamma}(s')
-q_{r_0-g,k_0}^{\nu,\gamma}(a,s)\}.
\]
Bellman's equation shows that the residual inside braces has conditional
mean zero given \((A,S)\). The residual is bounded and
\(\alpha_{0,j}^x\in L^2(P_0)\), so this representer belongs to
\(T_{\mathcal K}\). Theorem~\ref{theorem::EIFnorm}, applied coordinatewise,
now gives the displayed trajectory-level EIF for \(b_{0,j}\); stacking the
coordinates gives \(\chi_{b,0}\).

Finally, the influence function of \(G_0\) is
\[
\sum_{a\in\mathcal A}\omega(a\mid S_0)x(a,S_0)x(a,S_0)^\top-G_0.
\]
The matrix delta method for \(\theta_0=G_0^{-1}b_0\) therefore gives
\[
\chi_{\theta,0}
=G_0^{-1}\{\chi_{b,0}-\chi_{G,0}\theta_0\},
\]
which is the stated result.
\end{proof}

\begin{lemma}[von Mises expansion for structural coefficient moments]
\label{lemma::vonmisescoefnorm}
Let \(r,q\in L^\infty(\lambda)\) and
\(\alpha^x\in L^\infty(\lambda)^p\). Write
\[
q_0^{\nu,\gamma}:=q_{r_0-g,k_0}^{\nu,\gamma},
\qquad
\bar r_q:=g+q-V_q^\nu,
\qquad
\pi_r(a\mid s):=\exp\{r(a,s)\}.
\]
Then
\begin{align*}
\Big\|&
\mathbb E_0\!\left[
\sum_{a\in\mathcal A}\omega(a\mid S_0)x(a,S_0)\bar r_q(a,S_0)
\right]-b_0\\
&+E_0\!\left[\alpha^x(A,S)
-\sum_a\pi_r(a\mid S)\alpha^x(a,S)\right]\\
&+E_0\!\left[\alpha^x(A,S)
\{r(A,S)-g(S)+\gamma V_q^\nu(S')-q(A,S)\}\right]
\Big\|_2\\
&\lesssim
\|\alpha^x-\alpha_0^x\|_{\mathcal H,0}
\|\mathcal T_{k_0,\nu,\gamma}(q-q_0^{\nu,\gamma})\|_{\mathcal H,0}
+\|r-r_0\|_{\mathcal H,0}^2.
\end{align*}
The implicit constant is uniform over bounded \(r\), \(r_0\), and
\(\alpha^x\).
\end{lemma}

\begin{proof}
Set
\[
\delta_q
:=\mathcal T_{k_0,\nu,\gamma}(q-q_0^{\nu,\gamma})
=q-\gamma\mathcal P_{k_0}V_q^\nu-r_0+g.
\]
Since
\(\bar r_q-\bar r_0=(I-\nu)(q-q_0^{\nu,\gamma})\), the balance
characterization of \(\alpha_0^x\) gives
\begin{align*}
&\mathbb E_0\!\left[
\sum_a\omega(a\mid S_0)x(a,S_0)
\{\bar r_q-\bar r_0\}(a,S_0)\right]\\
&\qquad=E_0\{\alpha_0^x(A,S)\delta_q(A,S)\}.
\end{align*}
Adding and subtracting \(\alpha^x\), and using
\[
\delta_q(A,S)
=E_0\!\left[
q(A,S)-\gamma V_q^\nu(S')-r(A,S)+g(S)\mid A,S
\right]+(r-r_0)(A,S),
\]
reduces the left-hand side of the lemma to
\[
E_0\{(\alpha_0^x-\alpha^x)(A,S)\delta_q(A,S)\}
+E_0\{\alpha^x(A,S)(r-r_0)(A,S)\}
+E_0\!\left[\alpha^x(A,S)-\sum_a\pi_r(a\mid S)\alpha^x(a,S)\right].
\]
Cauchy--Schwarz bounds
\(E_0\{(\alpha_0^x-\alpha^x)(A,S)\delta_q(A,S)\}\) by the product in
the statement.
Put $\delta r=r-r_0$.  Since $\pi_r=\pi_0e^{\delta r}$, iterated
expectation gives the exact identity
\begin{align}
&E_0\{\alpha^x(A,S)\delta r(A,S)\}
+E_0\!\left[\alpha^x(A,S)-\sum_a\pi_r(a\mid S)\alpha^x(a,S)\right]\notag\\
&\qquad=E_{0,S}\!\left[\sum_a\pi_0(a\mid S)\alpha^x(a,S)
\{1+\delta r(a,S)-e^{\delta r(a,S)}\}\right].
\label{eqn::coef-normalization-cancellation}
\end{align}
The inequality $|e^u-1-u|\le e^{|u|}u^2/2$, equivalently the remainder
bound in Lemma~\ref{lemma::quadraticloss}, applied to
\eqref{eqn::coef-normalization-cancellation}, bounds its right-hand side by
$O(\|r-r_0\|_{\mathcal H,0}^2)$.  This proves the stated expansion.
\end{proof}

\begin{proof}[Proof of Theorem~\ref{theorem::ALex3}]
Let
\[
\begin{aligned}
\phi_n^b(O)
:={}&\sum_{a\in\mathcal A}\omega(a\mid S_0)x(a,S_0)\bar r_n(a,S_0)\\
&+\frac1T\sum_{t=0}^{T-1}\left[
\alpha_n^x(A_t,S_t)
-\sum_{\tilde a}\pi_n(\tilde a\mid S_t)\alpha_n^x(\tilde a,S_t)\right.\\
&\hspace{24mm}\left.
+\alpha_n^x(A_t,S_t)
\{r_n(A_t,S_t)-g(S_t)+\gamma V_n^{\nu,\gamma}(S_{t+1})
-q_n^{\nu,\gamma}(A_t,S_t)\}\right],
\end{aligned}
\]
and define \(\phi_0^b\) at the true nuisances. Then
\(b_n=\mathbb P_n\phi_n^b\),
\(\mathbb P_0\phi_0^b=b_0\), and
\(\chi_{b,0}=\phi_0^b-b_0\). Hence
\[
b_n-b_0
=\mathbb P_n\chi_{b,0}
+(\mathbb P_n-\mathbb P_0)(\phi_n^b-\phi_0^b)
+\{\mathbb P_0\phi_n^b-b_0\}.
\]
Assumptions~\ref{assump:boundedness::coefnorm},
\ref{assump:splitting::coefnorm}, and~\ref{assump:consistency::coefnorm},
together with boundedness of \(x\), \(I-\nu\), and the candidate nuisances,
give, coordinatewise for fixed \(p\),
\begin{align}
\|\phi_n^b-\phi_0^b\|_{L^2(\mathbb P_0)}
\lesssim{}&
\|\alpha_n^x-\alpha_0^x\|_{\mathcal H,0}
+\|q_n^{\nu,\gamma}-q_0^{\nu,\gamma}\|_{\mathcal H,0}
+\|r_n-r_0\|_{\mathcal H,0}
=o_p(1).
\label{eqn::coefficient-correction-L2-continuity}
\end{align}
Lemma~\ref{lemma::crossfitprocess}, applied foldwise, therefore makes the
empirical-process term \(o_p(n^{-1/2})\).
Lemma~\ref{lemma::vonmisescoefnorm},
Condition~\ref{assump:rates::coefnorm}, and
Condition~\ref*{cond::reward} give
\[
\|\mathbb P_0\phi_n^b-b_0\|_2
=O_p\!\left(
\|\alpha_n^x-\alpha_0^x\|_{\mathcal H,0}
\|\mathcal T_{k_0,\nu,\gamma}
(q_n^{\nu,\gamma}-q_0^{\nu,\gamma})\|_{\mathcal H,0}
+\|r_n-r_0\|_{\mathcal H,0}^2\right)
=o_p(n^{-1/2}).
\]
Thus
\begin{equation}
b_n-b_0=\frac1n\sum_{i=1}^n\chi_{b,0}(O_i)+o_p(n^{-1/2}).
\label{eqn::coef-b-linear-expansion}
\end{equation}

Moreover,
\begin{equation}
G_n-G_0
=\frac1n\sum_{i=1}^n\left[
\sum_{a\in\mathcal A}\omega(a\mid S_{i,0})
x(a,S_{i,0})x(a,S_{i,0})^\top-G_0\right].
\label{eqn::coef-G-empirical-expansion}
\end{equation}
Let $\Delta_G=G_n-G_0$, $\Delta_b=b_n-b_0$, and
$d_n=\Delta_b-\Delta_G\theta_0$.  The identity
$G_n^{-1}-G_0^{-1}=-G_n^{-1}\Delta_GG_0^{-1}$ gives
\begin{align}
\theta_n-\theta_0
&=G_0^{-1}d_n+R_n,
&R_n&=-G_n^{-1}\Delta_GG_0^{-1}d_n,
\label{eqn::coef-matrix-expansion}\\
\|R_n\|_2
&\le \|G_n^{-1}\|_{\mathrm{op}}\|G_0^{-1}\|_{\mathrm{op}}
\|\Delta_G\|_{\mathrm{op}}
\{\|\Delta_b\|_2+\|\Delta_G\|_{\mathrm{op}}\|\theta_0\|_2\}.
\label{eqn::coef-matrix-remainder}
\end{align}
Equations~\eqref{eqn::coef-b-linear-expansion} and
\eqref{eqn::coef-G-empirical-expansion} imply
$\|\Delta_b\|_2+\|\Delta_G\|_{\mathrm{op}}=O_p(n^{-1/2})$ and
$\|G_n^{-1}\|_{\mathrm{op}}=O_p(1)$.  Thus
\eqref{eqn::coef-matrix-remainder} is $O_p(n^{-1})$, and substitution into
\eqref{eqn::coef-matrix-expansion} yields
\[
\theta_n-\theta_0
=\frac1n\sum_{i=1}^n\chi_{\theta,0}(O_i)+o_p(n^{-1/2}).
\]
The multivariate central limit theorem gives the stated normal limit with
\(\Sigma_0=\operatorname{Var}_{\mathbb P_0}\{\chi_{\theta,0}(O)\}\).
\end{proof}

\section{Smoothness of the Soft Bellman Equation and Its Solutions}

This section collects the smoothness results for the soft Bellman equation
used in the EIF and von Mises arguments for Example~\ref{example::1b}. We
first establish existence, first-order differentiability, and local stability
of the fixed point, and then derive the operator and second-order bounds used
in the application proofs.

\subsection{Notation}

We study the functional smoothness of the soft Bellman equation and its fixed-point solution for a fixed temperature parameter $\tau \in (0,\infty)$. Let $\lambda := \# \otimes \mu$ denote the product measure on $\mathcal{A}\times\mathcal{S}$, where $\#$ is counting measure on the finite set $\mathcal{A}$ and $\mu$ is a finite dominating measure on $\mathcal{S}$. For any measurable $B \subseteq \mathcal{A}\times\mathcal{S}$,
\[
\lambda(B) = \sum_{a \in \mathcal{A}} \int 1_B(a,s)\,\mu(ds).
\]
For a linear operator $B:L^p(\xi)\to L^q(\zeta)$, write
$\|B\|_{L^p(\xi)\to L^q(\zeta)}
:=\sup_{\|f\|_{L^p(\xi)}\leq 1}\|Bf\|_{L^q(\zeta)}$; when
$\xi=\zeta=\lambda$, we abbreviate this norm as
$\|B\|_{L^p\to L^q}$.

For $r \in \mathcal{H}$ and $k \in \mathcal{K}$, let $v_{r,k}$ denote the
fixed point, shown below to exist uniquely in $L^\infty(\lambda)$, of the soft
Bellman equation
\[
v_{r,k}(a,s) = \int \tau \log\!\left(\sum_{a'\in\mathcal{A}}\exp\!\left(\frac{r(a',s')+\gamma v_{r,k}(a',s')}{\tau}\right)\right)k(s'\mid a,s)\,\mu(ds').
\]
Define $F: L^\infty(\lambda)\times L^\infty(\lambda)\times\mathcal{K}\to L^\infty(\lambda)$ by
\[
F(r,v,k)(a,s) := v(a,s) - \int \tau \log\!\left(\sum_{a'\in\mathcal{A}}\exp\!\left(\frac{r(a',s')+\gamma v(a',s')}{\tau}\right)\right)k(s'\mid a,s)\,\mu(ds'),
\]
and for each $k \in \mathcal{K}$, let $F_k(r,v):=F(r,v,k)$.

We write
\[
\operatorname{logsumexp}_{\tau}(q)(s) := \tau \log\!\left( \sum_{a' \in \mathcal{A}} \exp\!\left\{ \frac{q(a', s)}{\tau} \right\} \right),
\]
\[
\pi_{r,k}(a'\mid s):=\frac{\exp\!\left(\frac{r(a',s)+\gamma v_{r,k}(a',s)}{\tau}\right)}{\sum_{b\in\mathcal{A}}\exp\!\left(\frac{r(b,s)+\gamma v_{r,k}(b,s)}{\tau}\right)},
\qquad
q_{r,k}(a,s):=r(a,s)+\gamma v_{r,k}(a,s),
\]
and for general $q$ let
\[
\widetilde{\pi}_{q,\tau}(a'\mid s):=\frac{\exp\!\left(\frac{q(a',s)}{\tau}\right)}{\sum_{b\in\mathcal{A}}\exp\!\left(\frac{q(b,s)}{\tau}\right)}.
\]
For fixed $s'$, the conditional covariance under $\pi$ is
\[
\begin{aligned}
\operatorname{Cov}_{\pi(\cdot\mid s')}(g(\cdot,s'),h(\cdot,s'))
&:= \sum_{a\in\mathcal{A}}\pi(a\mid s')g(a,s')h(a,s') \\
&\quad - \Big(\sum_{a\in\mathcal{A}}\pi(a\mid s')g(a,s')\Big)
\Big(\sum_{a\in\mathcal{A}}\pi(a\mid s')h(a,s')\Big).
\end{aligned}
\]
Throughout this subsection, the temperature $\tau$ is fixed and all implicit
constants may depend on $\tau$.

\subsection{Smoothness results}

\label{appendix::technicalsoftmax}

Throughout this section, additive kernel perturbations range over \(\addTK\), equipped with
the norm \(\|\cdot\|_{\mathcal K}\). Their additive Fr\'echet derivative is
denoted by \(\adk\); \(\partial_k\) remains reserved for the semiparametric
score derivative used in the EIF calculations.

We begin with existence and first-order smoothness of the soft Bellman fixed
point.

\begin{lemma}[Existence and uniqueness of solution to soft Bellman equation]
    For each \(r \in L^\infty(\lambda)\) and $k \in \mathcal{K}$, there exists a unique solution \(v_{r,k} \in L^\infty(\lambda)\) to the soft Bellman equation $F_{k}(r, v) = 0$ over \(v \in L^\infty(\lambda)\).
\end{lemma}
\begin{proof}
For fixed \(r\in L^\infty(\lambda)\), define
\(B_{r,k}(v):=v-F_k(r,v)\). Since
\[
\|B_{r,k}(0)\|_\infty
\le \|r\|_\infty+\tau\log|\mathcal A|,
\]
\(B_{r,k}\) maps \(L^\infty(\lambda)\) into itself. The log-sum-exp map is
one-Lipschitz in supremum norm, so \(B_{r,k}\) is a contraction with modulus
\(\gamma<1\). Banach's fixed-point theorem therefore gives a
unique \(v_{r,k}\) satisfying \(v_{r,k}=B_{r,k}(v_{r,k})\), or equivalently
\(F_k(r,v_{r,k})=0\).
\end{proof}

The following lemma establishes Fr\'echet differentiability of $F_{k}$ for a fixed $k$.

\begin{lemma}[Fr\'echet differentiability of $F_k$]
\label{lemma::softbellmanissmooth}
Let $k \in \mathcal{K}$. The map $F_{k}$ is Fr\'echet differentiable in supremum norm, with partial derivatives at \((r,v) \in L^\infty(\lambda) \times L^\infty(\lambda)\), for directions $u,h \in L^\infty(\lambda)$, given by
\begin{align*}
\partial_v F_{k}(r,v)[h](a,s)
&= h(a,s) - \gamma \int \sum_{a' \in \mathcal{A}} \widetilde{\pi}_{\,r+\gamma v,\tau}(a' \mid s')\, h(a',s')\, k(s' \mid a,s) \mu(ds'),\\
\partial_r F_{k}(r,v)[u](a,s)
&= - \int \sum_{a' \in \mathcal{A}} \widetilde{\pi}_{\,r+\gamma v,\tau}(a' \mid s')\, u(a',s')\, k(s' \mid a,s) \mu(ds').
\end{align*}
In particular, there exists a fixed constant \(C_{\tau,k}<\infty\), depending
only on the fixed temperature $\tau$ and $\|k\|_{L^\infty}$, such that,
uniformly over $r,v,h,u$,
\[
\|\,F_{k}(r+u,v+h) - F_{k}(r,v) - \partial_r F_{k}(r,v)[u] - \partial_v F_{k}(r,v)[h]\,\|_{L^\infty(\lambda)}
\;\le\; C_{\tau,k} \,\|u + \gamma h\|_{L^2(\lambda)}^{2}.
\]
\end{lemma}
\begin{proof}
Direct differentiation of log-sum-exp gives the displayed G\^ateaux
derivatives. To verify Fr\'echet differentiability, Taylor's theorem gives, for
each direction \(f \in L^\infty(\lambda)\),
\begin{align*}
\Big|
\operatorname{logsumexp}_{\tau}(q+f)(s')
-\operatorname{logsumexp}_{\tau}(q)(s')\notag\\
\qquad
-\sum_{a'\in\mathcal A}\widetilde{\pi}_{q,\tau}(a'\mid s')\,f(a',s')
\Big|
&\le \frac{1}{\tau} \int_0^1(1-t)
\sum_{a'\in\mathcal A}\widetilde{\pi}_{\,q+t f,\tau}(a'\mid s')\,
|f(a',s')|^2\,dt\\
&\le \frac{1}{2\tau} \sum_{a'\in\mathcal A}|f(a',s')|^2.
\end{align*}
Substituting \(q:=r+\gamma v\) and \(f:=u+\gamma h\) yields
\begin{align*}
&\big|F_{k}(r+u,v+h)(a,s)-F_{k}(r,v)(a,s)
-\partial_v F_{k}(r,v)[h](a,s)-\partial_r F_{k}(r,v)[u](a,s)\big|\\
&\qquad\le \frac{1}{2\tau}\int \sum_{a'\in\mathcal A} |u(a',s')+\gamma h(a',s')|^2 \, k(s'\mid a,s) \mu(ds').
\end{align*}
Because $k\in\mathcal K$ is essentially bounded,
$k(s'\mid a,s)\mu(ds')\lesssim\mu(ds')$. Hence,
\begin{align*}
&\big|F_{k}(r+u,v+h)(a,s)-F_{k}(r,v)(a,s)
-\partial_v F_{k}(r,v)[h](a,s)-\partial_r F_{k}(r,v)[u](a,s)\big|\\
&\qquad\lesssim  \int \sum_{a'\in\mathcal A} |u(a',s')+\gamma h(a',s')|^2 \, \mu(ds')\\
&\qquad\lesssim \| u + \gamma h\|_{L^2(\lambda)}^2.
\end{align*}
Taking supremum norms proves the stated remainder bound and hence Fr\'echet
differentiability.

\end{proof}

\begin{lemma}[Fr\'echet differentiability of $F$]
\label{lemma::bellmansmoothnessfull}
The map $F$, extended to
$L^\infty(\lambda)\times L^\infty(\lambda)\times
L^\infty(\mu\otimes\lambda)$, is Fr\'echet differentiable in supremum norm,
with partial derivatives at $(r,v,k)$ given by
\begin{align*}
\adk F(r,v,k)[k' - k](a,s)
&= - \int \operatorname{logsumexp}_{\tau}(r+\gamma v)(s')\, (k'-k)(s'\mid a,s)\, \mu(ds'),\\
\partial_v F(r,v,k)[h](a,s) &= \partial_v F_{k}(r,v)[h](a,s),\\
\partial_r F(r,v,k)[u](a,s) &= \partial_r F_{k}(r,v)[u](a,s),
\end{align*}
for $k' \in L^\infty(\mu\otimes\lambda)$ and
$h,u \in L^\infty(\lambda)$. Define
\[
R:=F(r+u,v+h,k')-F(r,v,k)-\partial_r F(r,v,k)[u]
-\partial_v F(r,v,k)[h]-\adk F(r,v,k)[k'-k].
\]
Locally uniformly in $(r,v,k)$,
\begin{align*}
\|R\|_{L^2(\lambda)}
&\lesssim \|u+\gamma h\|_{L^2(\lambda)}^2
+\{\|u\|_{L^2(\lambda)}+\|h\|_{L^2(\lambda)}\}
\|k'-k\|_{L^2(\mu\otimes\lambda)},\\
\|R\|_{L^\infty(\lambda)}
&\lesssim \|u+\gamma h\|_{L^2(\lambda)}^2
+\{\|u\|_{L^2(\lambda)}+\|h\|_{L^2(\lambda)}\}
\|k'-k\|_{L^\infty(\mu\otimes\lambda)}.
\end{align*}

\end{lemma}
\begin{proof}
Add and subtract $F(r,v,k')$. The Taylor argument in
Lemma~\ref{lemma::softbellmanissmooth}, with $|k'|$ in place of $k$, gives
\[
F(r+u,v+h,k')-F(r,v,k')
= \partial_v F_{k'}(r,v)[h] + \partial_r F_{k'}(r,v)[u]
+R_0,
\]
where $\|R_0\|_{L^\infty(\lambda)}
\lesssim\|u+\gamma h\|_{L^2(\lambda)}^2$. The bound is locally uniform
because $\|k'\|_{L^\infty}$ is bounded in a supremum-norm neighborhood of
$k$. Linearity in the kernel gives
\[
F(r,v,k')-F(r,v,k)=\adk F(r,v,k)[k'-k].
\]
Hence,
\begin{align*}
R
&=\partial_v F_{k'}(r,v)[h]-\partial_v F_k(r,v)[h]\\
&\quad+\partial_r F_{k'}(r,v)[u]-\partial_r F_k(r,v)[u]+R_0.
\end{align*}
It remains to control the changes in the two partial derivatives. Set
$g:=k'-k$. The derivative formulas give
\begin{align*}
\left|\partial_r F_{k'}(r,v)[u]-\partial_r F_k(r,v)[u]\right|(a,s)
&= \left|\int\sum_{a'\in\mathcal A}
\widetilde{\pi}_{r+\gamma v,\tau}(a'\mid s')
u(a',s')g(s'\mid a,s)\,\mu(ds')\right|\\
&\le \int\sum_{a'\in\mathcal A}
|u(a',s')|\,|g(s'\mid a,s)|\,\mu(ds').
\end{align*}
Cauchy--Schwarz, the finiteness of $\mathcal A$, and the finiteness of
$\mu$ and $\lambda$ give the two operator-norm bounds
\begin{align}
&\left\|\partial_r F_{k'}(r,v)[u]-\partial_r F_k(r,v)[u]\right\|_{L^2(\lambda)}
+\left\|\partial_v F_{k'}(r,v)[h]-\partial_v F_k(r,v)[h]\right\|_{L^2(\lambda)}\notag\\
&\qquad\lesssim
\{\|u\|_{L^2(\lambda)}+\|h\|_{L^2(\lambda)}\}
\|g\|_{L^2(\mu\otimes\lambda)},
\label{eqn::bellman-partial-kernel-continuity-L2}\\
&\left\|\partial_r F_{k'}(r,v)[u]-\partial_r F_k(r,v)[u]\right\|_{L^\infty(\lambda)}
+\left\|\partial_v F_{k'}(r,v)[h]-\partial_v F_k(r,v)[h]\right\|_{L^\infty(\lambda)}\notag\\
&\qquad\lesssim
\{\|u\|_{L^2(\lambda)}+\|h\|_{L^2(\lambda)}\}
\|g\|_{L^\infty(\mu\otimes\lambda)}.
\label{eqn::bellman-partial-kernel-continuity-Linfty}
\end{align}
Combining these bounds with the bound for $R_0$ proves both remainder
inequalities.

Finally, the softmax map is locally Lipschitz in supremum norm.
For the kernel derivative, log-sum-exp Lipschitz continuity gives
\begin{align}
&\|\adk F(r_1,v_1,k)[w]-\adk F(r_2,v_2,k)[w]\|_{L^\infty(\lambda)}
\notag\\
&\qquad\lesssim
\{\|r_1-r_2\|_{L^\infty(\lambda)}
+\gamma\|v_1-v_2\|_{L^\infty(\lambda)}\}
\|w\|_{L^\infty(\mu\otimes\lambda)}.
\label{eqn::bellman-partial-kernel-map-continuity}
\end{align}
Equations~\eqref{eqn::bellman-partial-kernel-continuity-L2} and
\eqref{eqn::bellman-partial-kernel-continuity-Linfty}, together with the
displayed formulas for the reward and value derivatives and
\eqref{eqn::bellman-partial-kernel-map-continuity}, show that each partial
derivative of \(F\) varies continuously in operator norm with \((r,v,k)\).
Thus \(F\) is continuously Fr\'echet differentiable on the stated product
space.

\end{proof}

\begin{lemma}[Sup-norm Fr\'echet differentiability of the soft value function]
\label{lemma::IFT}
Let $(r,k) \in \mathcal{H} \times \mathcal{K}$. Suppose $\gamma < 1$, so that $\mathcal{T}_{r,k}: L^\infty(\lambda) \to L^\infty(\lambda)$ is invertible. Viewed as a map from $L^\infty(\lambda)\times L^\infty(\mu\otimes\lambda)$ to $L^\infty(\lambda)$, \((r,k) \mapsto v_{r,k}\) is Fr\'echet differentiable, with partial derivatives
\[
\partial_r v_{r,k} \;=\; \mathcal T_{r,k}^{-1}\, \mathcal P_{r,k},
\qquad
\adk v_{r,k}[w] \;=\; \mathcal T_{r,k}^{-1}\, \mathcal{P}_{w}\,
\Phi(r + \gamma v_{r,k}),
\]
where $\Phi(q):=\operatorname{logsumexp}_{\tau}(q)$ and
\[
\mathcal{P}_{w}\,\Phi(r + \gamma v_{r,k})(a,s)
:= \int \Phi(r + \gamma v_{r,k})(s')\,w(s'\mid a,s)\,\mu(ds').
\]
\end{lemma}
\begin{proof}
By Lemma~\ref{lemma::bellmansmoothnessfull}, $F$ is continuously Fr\'echet
differentiable between the stated supremum-norm spaces. Moreover,
\[
\partial_v F_{k}(r,v_{r,k}): L^\infty(\lambda) \to L^\infty(\lambda)
\]
equals $\mathcal T_{r,k}$ and is a bounded linear isomorphism because
$\gamma<1$. The implicit function theorem for Banach spaces therefore gives
the derivatives
\begin{align*}
    \partial_r v_{r,k}[u]
&= -\big(\partial_v F_{k}(r,v_{r,k})\big)^{-1}\!\big(\partial_r F_{k}(r,v_{r,k})[u]\big),\\
\adk v_{r,k}[w]
&= -\big(\partial_v F_{k}(r,v_{r,k})\big)^{-1}\!\big(\adk F(r,v_{r,k},k)[w]\big).
\end{align*}
Since $\partial_v F_{k}(r,v)=\mathcal T_{r,k}$,
$\partial_r F_{k}(r,v)=-\mathcal P_{r,k}$, and
$\adk F(r,v,k)[w]=-\mathcal P_w\Phi(r+\gamma v)$, these identities reduce to
\[
\partial_r v_{r,k} \;=\; \mathcal T_{r,k}^{-1}\, \mathcal P_{r,k},
\qquad
\adk v_{r,k}[w] \;=\; \mathcal T_{r,k}^{-1}\mathcal P_w
\Phi(r+\gamma v_{r,k}),
\]
as claimed.
\end{proof}

\begin{lemma}[Reward derivative of the soft action value]
\label{lemma::qderiv}
Under the conditions of Lemma~\ref{lemma::IFT},
\[
\partial_r q_{r,k}[u] = \mathcal{T}_{r,k}^{-1}u
\]
for every $u \in L^\infty(\lambda)$.
\end{lemma}

\begin{proof}
By Lemma~\ref{lemma::IFT},
\begin{align*}
    \partial_r q_{r,k}[u]
    &= u + \gamma \,\partial_r v_{r,k}(u) \\
    &= \left(I + \gamma \,\mathcal{T}_{r,k}^{-1}\mathcal{P}_{r,k}\right)(u) \\
    &= \mathcal{T}_{r,k}^{-1}\left(\mathcal{T}_{r,k} + \gamma \mathcal{P}_{r,k}\right)(u) \\
    &= \mathcal{T}_{r,k}^{-1}\left(I - \gamma \mathcal{P}_{r,k} + \gamma \mathcal{P}_{r,k}\right)(u) \\
    &= \mathcal{T}_{r,k}^{-1} u.
\end{align*}
This is the asserted derivative formula.
\end{proof}

In the local operator statements below, kernel neighborhoods are intersected
with
\[
\mathcal K(M_k):=\{k\in\mathcal K:
\|k\|_{L^\infty(\mu\otimes\lambda)}\le M_k\}
\]
for a fixed \(M_k<\infty\). Constants may depend on \(M_k\).

\begin{lemma}[Lipschitz continuity of the soft Bellman fixed point]
\label{lemma::lipschitzvalue}
Suppose that $\gamma<1$. For fixed $k\in\mathcal K$ and bounded rewards
$r,r'$, we have
\[
\|v_{r',k}-v_{r,k}\|_{L^\infty(\lambda)}
\le C_k(1-\gamma)^{-1}\|r'-r\|_{L^2(\lambda)}.
\]
Under Assumption~\ref{cond::boundedpositivity}, there are a neighborhood
$\mathcal U$ of $(r_0,k_0)$ in
$L^2(\lambda)\times(\mathcal K,\|\cdot\|_{\mathcal K})$ and a constant
$L<\infty$ such that, for every $(r,k),(r',k')\in\mathcal U$,
\begin{equation}
\|v_{r',k'}-v_{r,k}\|_{L^2(\lambda)}
\le L\{\|r'-r\|_{L^2(\lambda)}+\|k'-k\|_{\mathcal K}\}.
\label{eqn::local-value-lipschitz}
\end{equation}
\end{lemma}
\begin{proof}
The log-sum-exp map is one-Lipschitz in the supremum norm. The fixed-point
equations and Cauchy--Schwarz therefore give
\[
\|v_{r',k}-v_{r,k}\|_\infty
\le C_k\|r'-r\|_{L^2(\lambda)}
+\gamma\|v_{r',k}-v_{r,k}\|_\infty,
\]
which proves the first assertion.

For the second assertion, let $(r,k),(r',k')\in\mathcal U$, write
$v=v_{r,k}$ and $v'=v_{r',k'}$, and let $\Phi$ and $\Phi'$ denote their
log-sum-exp state values. Define the Markov policy
\[
\bar\pi(a\mid s)
:=\int_0^1\pi_{q+t(q'-q)}(a\mid s)\,dt,
\qquad q=r+\gamma v,\quad q'=r'+\gamma v'.
\]
The fundamental theorem of calculus gives
\begin{equation}
\Phi'-\Phi=\bar\pi\{r'-r+\gamma(v'-v)\}.
\label{eqn::logsumexp-mean-value}
\end{equation}
Using $v'=\mathcal P_{k'}\Phi'$ and $v=\mathcal P_k\Phi$, we obtain
\begin{equation}
(I-\gamma\mathcal P_k\bar\pi)(v'-v)
=\mathcal P_{k'-k}\Phi'
+\mathcal P_k\bar\pi(r'-r).
\label{eqn::local-value-difference}
\end{equation}
Since the action set and the dominating measures are finite,
log-sum-exp Lipschitz continuity gives
\[
\|\Phi'\|_{L^2(\mu)}
\le C\{1+\|r'\|_{L^2(\lambda)}+\|v'\|_{L^2(\lambda)}\}.
\]
To bound \(v'\) without using this inequality recursively, define
\[
\bar\pi'(a\mid s):=\int_0^1\widetilde\pi_{tq',\tau}(a\mid s)\,dt.
\]
The fundamental theorem of calculus, now applied between \(0\) and \(q'\),
gives
\(\Phi'=\tau\log|\mathcal A|+\bar\pi' q'\). The fixed-point equation
therefore implies the exact identity
\begin{equation}
(I-\gamma\mathcal P_{k'}\bar\pi')v'
=\mathcal P_{k'}(\tau\log|\mathcal A|)
+\mathcal P_{k'}\bar\pi' r'.
\label{eqn::local-value-level-equation}
\end{equation}
For an arbitrary Markov policy \(\pi\), write
\[
I-\gamma\mathcal P_k\pi
=I-\gamma\mathcal P_{k_0}\pi
-\gamma\mathcal P_{k-k_0}\pi.
\]
Lemma~\ref{lemma::invertibility} bounds the inverse of the first operator
uniformly in \(\pi\), and the Hilbert--Schmidt bound gives
\(\|\mathcal P_{k-k_0}\pi\|_{L^2\to L^2}
\lesssim\|k-k_0\|_{\mathcal K}\).  After shrinking \(\mathcal U\),
Lemma~\ref{lemma::inverseperturbation} thus gives
\begin{equation}
\sup_{(\widetilde r,\widetilde k)\in\mathcal U}
\sup_{\pi}
\|(I-\gamma\mathcal P_{\widetilde k}\pi)^{-1}\|_{L^2\to L^2}
<\infty.
\label{eqn::local-policy-resolvent-bound}
\end{equation}
Applying \eqref{eqn::local-policy-resolvent-bound} to
\eqref{eqn::local-value-level-equation}, and using
\[
\sup_{(\widetilde r,\widetilde k)\in\mathcal U}
\|\widetilde k\|_{\mathcal K}<\infty
\]
with the Hilbert--Schmidt bound, yields
\begin{equation}
\|v'\|_{L^2(\lambda)}
\lesssim 1+\|r'\|_{L^2(\lambda)}.
\label{eqn::local-value-level-bound}
\end{equation}
Thus \(\|v'\|_{L^2(\lambda)}\) is uniformly bounded on \(\mathcal U\).
The Hilbert--Schmidt bound for \(\mathcal P_{k'-k}\) and
\eqref{eqn::local-policy-resolvent-bound}, applied to
\eqref{eqn::local-value-difference}, now yield
\[
\|v'-v\|_{L^2(\lambda)}
\le C\left[\|r'-r\|_{L^2(\lambda)}
+\|k'-k\|_{\mathcal K}
\{1+\|r'\|_{L^2(\lambda)}+\|v'\|_{L^2(\lambda)}\}\right].
\]
Applying \eqref{eqn::local-value-level-bound} to this inequality proves
\eqref{eqn::local-value-lipschitz}.

\end{proof}

\begin{lemma}[Second-order reward expansion of the soft value function]
 \label{lemma::valuefirstderivativeexpansion}
 Suppose $\gamma < 1$. For $r,r' \in \mathcal{H}$ and $k \in \mathcal{K}$, we have
\[
\big\|v_{r',k} - v_{r,k} - \partial_r v_{r,k}[\,r' - r\,]\big\|_{L^\infty(\lambda)}
\;\lesssim\; \|r' - r\|_{L^2(\lambda)}^2.
\]
\end{lemma}
\begin{proof}
Let $q_{r,k}:=r+\gamma v_{r,k}$. Since
$F(r',v_{r',k},k)=F(r,v_{r,k},k)=0$,
Lemma~\ref{lemma::bellmansmoothnessfull} gives
\begin{align*}
0
&= F(r', v_{r',k}, k) - F(r, v_{r,k}, k) \\
&= \partial_{r,v} F(r, v_{r,k}, k)\big[r' - r,\, v_{r',k} - v_{r,k}\big] + R(r' - r) \\
&= \partial_r F(r, v_{r,k}, k)[\,r' - r\,]
  + \partial_v F(r, v_{r,k}, k)[\,v_{r',k} - v_{r,k}\,] + R(r' - r),
\end{align*}
where $\|R(r'-r)\|_{L^\infty(\lambda)}
=O\!\big(\|q_{r',k}-q_{r,k}\|_{L^2(\lambda)}^2\big)$ by
Lemma~\ref{lemma::softbellmanissmooth}. The derivative identity in
Lemma~\ref{lemma::IFT} gives
\[
\partial_{r,v}F(r,v_{r,k},k)\big[r' - r,\,\partial_r v_{r,k}[\,r' - r\,]\big]
= \partial_r F(r,v_{r,k},k)[\,r' - r\,]
+ \partial_v F(r,v_{r,k},k)[\,\partial_r v_{r,k}[\,r' - r\,]\,]
= 0.
\]
Hence,
\[
0
= \partial_v F(r, v_{r,k}, k)\big[v_{r',k} - v_{r,k} - \partial_r v_{r,k}[\,r' - r\,]\big]
   + R(r' - r).
\]
Equivalently,
\[
\mathcal{T}_{r,k}\,\big(v_{r',k} - v_{r,k} - \partial_r v_{r,k}[\,r' - r\,]\big) = -\,R(r' - r).
\]
Since $\gamma < 1$, the map $\mathcal{T}_{r,k}: L^\infty(\lambda) \rightarrow L^\infty(\lambda)$ is invertible by Banach's fixed point theorem. Hence,
\[
v_{r',k} - v_{r,k} - \partial_r v_{r,k}[\,r' - r\,]
= -\,\mathcal{T}_{r,k}^{-1} R(r' - r),
\]
so that
\[
\big\|v_{r',k} - v_{r,k} - \partial_r v_{r,k}[\,r' - r\,]\big\|_{L^\infty(\lambda)}
\;\le\; \|\mathcal{T}_{r,k}^{-1}\|_{L^\infty\to L^\infty}\,
\|R(r' - r)\|_{L^\infty(\lambda)}
\;\lesssim\; \|q_{r',k} - q_{r,k}\|_{L^2(\lambda)}^2.
\]
Finally, by Lemma~\ref{lemma::lipschitzvalue}, $ \|q_{r',k} - q_{r,k}\|_{L^2(\lambda)}^2 \lesssim  \|r' - r\|_{L^2(\lambda)}^2$. Hence,
\[
 \big\|v_{r',k} - v_{r,k} - \partial_r v_{r,k}[\,r' - r\,]\big\|_{L^\infty(\lambda)}
 \;\lesssim\; \|r' - r\|_{L^2(\lambda)}^2.
\]
This proves the stated quadratic remainder bound.
\end{proof}

\begin{lemma}[Kernel derivative of the soft value function]
 \label{lemma::valuefirstderivativeexpansion::kernel}
Suppose Assumption~\ref{cond::boundedpositivity} holds and
$\mathcal T_{r,k}:L^2(\lambda)\to L^2(\lambda)$ has a bounded inverse.
Let $(r,k)$ and $(r,k')$ lie in the neighborhood $\mathcal U$ from
Lemma~\ref{lemma::lipschitzvalue}. Then
\[
\begin{aligned}
&\|v_{r,k'}-v_{r,k}-\adk v_{r,k}[k'-k]\|_{L^2(\lambda)}\\
&\qquad\lesssim
\|\mathcal T_{r,k}^{-1}\|_{L^2\to L^2}
\left\{\|v_{r,k'}-v_{r,k}\|_{L^2(\lambda)}^2
+\|v_{r,k'}-v_{r,k}\|_{L^2(\lambda)}
\|k'-k\|_{L^2(\mu\otimes\lambda)}\right\}
\lesssim \|k'-k\|_{\mathcal K}^2.
\end{aligned}
\]
Consequently, $k\mapsto v_{r,k}$ is Fr\'echet differentiable at $k$ as
a map from $(\mathcal K,\|\cdot\|_{\mathcal K})$ into $L^2(\lambda)$.
\end{lemma}
\begin{proof}
Since $F(r,v_{r,k'},k')=F(r,v_{r,k},k)=0$,
Lemma~\ref{lemma::bellmansmoothnessfull} gives
\begin{align*}
0
&= F(r, v_{r,k'}, k') - F(r, v_{r,k}, k) \\
&= \partial_v F(r, v_{r,k}, k)[\,v_{r,k'} - v_{r,k}\,]
  + \adk F(r, v_{r,k}, k)[\,k' - k\,] + R_2,
\end{align*}
where Lemma~\ref{lemma::bellmansmoothnessfull} gives
\[
\|R_2\|_{L^2(\lambda)}
=O\!\left(\|v_{r,k'}-v_{r,k}\|_{L^2(\lambda)}^2
+\|v_{r,k'}-v_{r,k}\|_{L^2(\lambda)}
\|k'-k\|_{L^2(\mu\otimes\lambda)}\right).
\]
The derivative identity in Lemma~\ref{lemma::IFT} gives
\[
\partial_v F(r,v_{r,k},k)[\,\adk v_{r,k}[\,k' - k\,]\,] + \adk F(r,v_{r,k},k)[\,k' - k\,]
= 0.
\]
Hence,
\[
0
= \partial_v F(r, v_{r,k}, k)\big[v_{r,k'} - v_{r,k} - \adk v_{r,k}[\,k' - k\,]\big]
   + R_2.
\]
Equivalently,
\[
\mathcal{T}_{r,k}\,\big(v_{r,k'} - v_{r,k} - \adk v_{r,k}[\,k' - k\,]\big) = -\,R_2.
\]
Applying the bounded inverse of $\mathcal T_{r,k}$ gives the first
inequality in the statement.  Equation~\eqref{eqn::local-value-lipschitz},
with $r'=r$, gives
\(\|v_{r,k'}-v_{r,k}\|_{L^2(\lambda)}\lesssim
\|k'-k\|_{\mathcal K}\), and hence gives the second inequality. Moreover,
the derivative formula in Lemma~\ref{lemma::IFT} and the Hilbert--Schmidt
inequality give
\[
\|\adk v_{r,k}[g]\|_{L^2(\lambda)}
\le
\|\mathcal T_{r,k}^{-1}\|_{L^2\to L^2}
\|g\|_{\mathcal K}\|\Phi_{r,k}\|_{L^2(\mu)}.
\]
Thus the derivative is bounded in \(\|\cdot\|_{\mathcal K}\). Together with
the quadratic remainder, this proves the Fr\'echet conclusion.

\end{proof}

\begin{lemma}[Lipschitz properties of softmax operator]
\label{lemma::softmaxLipschitz}
For all $r,r'\in L^\infty(\lambda)$ and $k,k'\in\mathcal K$,
\[
\bigl\|\Pi_{r',k'}-\Pi_{r,k}\bigr\|_{L^2(\lambda)\to L^1(\mu)}
\lesssim \|r'-r\|_{L^2(\lambda)}
+\|v_{r,k'}-v_{r,k}\|_{L^2(\lambda)}.
\]
\end{lemma}
\begin{proof}
The $\tau^{-1}$-Lipschitz property of the temperature-$\tau$ softmax and
Cauchy--Schwarz give
\begin{align*}
   \bigl|\Pi_{r',k'}(f)-\Pi_{r,k}(f)\bigr|(s)
   &= \Bigl|\sum_{a \in \mathcal{A}} f(a,s)\,\bigl(\pi_{r',k'}(a\mid s)-\pi_{r,k}(a\mid s)\bigr)\Bigr| \\
   &= \bigl|\langle f(\cdot,s),\,\pi_{r',k'}(\cdot\mid s)-\pi_{r,k}(\cdot\mid s)\rangle_{\ell_2(\mathcal A)}\bigr| \\
   &\le \|f(\cdot,s)\|_{\ell_2(\mathcal A)} \,\|\pi_{r',k'}(\cdot\mid s)-\pi_{r,k}(\cdot\mid s)\|_{\ell_2(\mathcal A)} \\
   &\lesssim \tau^{-1}\,\|f(\cdot,s)\|_{\ell_2(\mathcal A)} \\
   &\qquad\times
   \Biggl(\sum_{a\in\mathcal A}
      \bigl(|r'(a,s)-r(a,s)| + |v_{r',k'}(a,s)-v_{r,k}(a,s)|\bigr)^2
   \Biggr)^{1/2},
\end{align*}
where $ \langle g,h\rangle_{\ell_2(\mathcal A)} := \sum_{a\in\mathcal A} g(a)h(a)$.

Let $L^2(\mu;\ell_2(\mathcal A))$ denote the space of measurable functions
$g:\mathcal{A}\times\mathcal{S}\to\mathbb{R}$ with finite mixed norm
\(\|g\|_{L^2(\mu;\ell_2(\mathcal A))}^2
   := \int_{\mathcal S} \sum_{a\in\mathcal A} |g(a,s)|^2 \, \mu(ds)\).
Then, by Cauchy--Schwarz,
\begin{align*}
\bigl\|\Pi_{r',k'}(f)-\Pi_{r,k}(f)\bigr\|_{L^1(\mu)}
&= \int \bigl|\Pi_{r',k'}(f)-\Pi_{r,k}(f)\bigr|(s)\, \mu(ds) \\
&\le \int \|f(\cdot,s)\|_{\ell_2(\mathcal A)} \,
        \|\pi_{r',k'}(\cdot\mid s)-\pi_{r,k}(\cdot\mid s)\|_{\ell_2(\mathcal A)}\,\mu(ds) \\
&\le \|f\|_{L^2(\mu;\ell_2(\mathcal A))}\;
    \|\pi_{r',k'}-\pi_{r,k}\|_{L^2(\mu;\ell_2(\mathcal A))} \\
&\lesssim \|f\|_{L^2(\mu;\ell_2(\mathcal A))}\;
    \|(r'-r)+(v_{r',k'}-v_{r,k})\|_{L^2(\mu;\ell_2(\mathcal A))} \\
&\lesssim \|f\|_{L^2(\mu;\ell_2(\mathcal A))}\;
   \Big( \|r'-r\|_{L^2(\mu;\ell_2(\mathcal A))}
       + \|v_{r',k'}-v_{r,k}\|_{L^2(\mu;\ell_2(\mathcal A))} \Big).
\end{align*}
Finally, since $\mathcal{A}$ is finite, the $L^2(\mu;\ell_2(\mathcal A))$
and $L^2(\lambda)$ norms agree. Hence,
\begin{align*}
\bigl\|\Pi_{r',k'}(f)-\Pi_{r,k}(f)\bigr\|_{L^1(\mu)} &\lesssim \|f\|_{L^2(\lambda)}\;
   \Big( \|r'-r\|_{L^2(\lambda)}
       + \|v_{r',k'}-v_{r,k}\|_{L^2(\lambda)} \Big).
\end{align*}
Lemma~\ref{lemma::lipschitzvalue} also gives
\begin{align*}
\|v_{r',k'}-v_{r,k}\|_{L^2(\lambda)}
&\leq \|v_{r',k'}-v_{r,k'}\|_{L^2(\lambda)}
   + \|v_{r,k'}-v_{r,k}\|_{L^2(\lambda)} \\
&\lesssim \|r'-r\|_{L^2(\lambda)}
   + \|v_{r,k'}-v_{r,k}\|_{L^2(\lambda)}.
\end{align*}
Combining these bounds gives
\begin{align*}
\bigl\|\Pi_{r',k'}(f)-\Pi_{r,k}(f)\bigr\|_{L^1(\mu)}
   \;\lesssim\; \|f\|_{L^2(\lambda)}
   \Big( \|r'-r\|_{L^2(\lambda)}
       + \|v_{r,k'}-v_{r,k}\|_{L^2(\lambda)} \Big).
\end{align*}
Taking the supremum over the unit ball of $L^2(\lambda)$ proves the stated
operator bound.

\end{proof}

\begin{lemma}[Lipschitz continuity of the soft transition operator]
\label{lemma::forwardopLipschitz}
On every parameter neighborhood whose kernels belong to
\(\mathcal K(M_k)\), there exists a constant \(C<\infty\), uniform on that
neighborhood,
such that
\[
\|\mathcal{P}_{r',k'} - \mathcal{P}_{r,k}\|_{L^2(\lambda) \to L^2(\lambda)}
\;\le\; C\Big(\|k' - k\|_{\mathcal K} + \|r' - r\|_{L^2(\lambda)}
+ \|v_{r,k'}-v_{r,k}\|_{L^2(\lambda)}\Big).
\]
\end{lemma}

\begin{proof}
Since $\mathcal P_{r,k}=\mathcal P_k\Pi_{r,k}$,
\[
\mathcal P_{r',k'}-\mathcal P_{r,k}
=\mathcal P_{k'-k}\Pi_{r',k'}
+\mathcal P_k(\Pi_{r',k'}-\Pi_{r,k}).
\]
The Hilbert--Schmidt bound used in Lemma~\ref{lemma::bellmanfrechet} and
Jensen's inequality imply
\[
\|\mathcal P_{k'-k}\Pi_{r',k'}\|_{L^2(\lambda)\to L^2(\lambda)}
\lesssim\|k'-k\|_{\mathcal K}.
\]
Moreover, boundedness of $k$ and finiteness of the dominating measures give
$\|\mathcal P_k f\|_{L^2(\lambda)}\lesssim\|f\|_{L^1(\mu)}$ for every
state function $f$. Lemma~\ref{lemma::softmaxLipschitz} therefore yields
\[
\|\mathcal P_k(\Pi_{r',k'}-\Pi_{r,k})\|_{L^2(\lambda)\to L^2(\lambda)}
\lesssim\|r'-r\|_{L^2(\lambda)}
+\|v_{r,k'}-v_{r,k}\|_{L^2(\lambda)}.
\]
Combining these two bounds proves
\[
\|\mathcal{P}_{r',k'} - \mathcal{P}_{r,k}\|_{L^2(\lambda) \to L^2(\lambda)}
\;\lesssim\; \|k' - k\|_{\mathcal K} + \|r' - r\|_{L^2(\lambda)}
+ \|v_{r,k'}-v_{r,k}\|_{L^2(\lambda)}.
\]
\end{proof}

\begin{lemma}[Derivatives and remainder bounds for the softmax and soft transition operators]
\label{lemma::operatorderivatives}
Let $r, r' \in L^\infty(\lambda)$, $k, k' \in \mathcal{K}$, and $f \in L^\infty(\lambda)$.
Suppose $\|\mathcal{T}_{r,k}^{-1}\|_{L^2 \rightarrow L^2} < \infty$.
Then the partial G\^ateaux derivatives of the maps $(r,k) \mapsto \Pi_{r, k} f$
and $(r,k) \mapsto \mathcal{P}_{r, k} f$ in the directions $r'-r$ and
$k'-k$ are
\begin{align*}
\partial_r \Pi_{r, k}[\,r'-r\,](f)(s')
&= \frac{1}{\tau}\,\operatorname{Cov}_{\pi_{r, k}(\cdot \mid s')}
   \big(f(\cdot, s'),\, \partial_r q_{r,k}[\,r'-r\,](\cdot, s')\big), \\[0.8ex]
\partial_r \mathcal{P}_{r,k}[\,r'-r\,](f)(a,s)
&= \frac{1}{\tau}\int \operatorname{Cov}_{\pi_{r, k}(\cdot\mid s')}\!\Big(
       f(\cdot,s'),\, \partial_r q_{r,k}[\,r'-r\,](\cdot,s')
     \Big)\,k(s'\mid a,s)\,\mu(ds'), \\[1.2ex]
\adk \Pi_{r, k}[\,k'-k\,](f)(s')
&= \frac{1}{\tau}\,\operatorname{Cov}_{\pi_{r, k}(\cdot \mid s')}
   \big(f(\cdot, s'),\, \adk q_{r,k}[\,k'-k\,](\cdot, s')\big), \\[0.8ex]
\adk \mathcal{P}_{r,k}[\,k'-k\,](f)(a,s)
&= \frac{1}{\tau}\int \operatorname{Cov}_{\pi_{r, k}(\cdot\mid s')}\!\Big(
       f(\cdot,s'),\, \adk q_{r,k}[\,k'-k\,](\cdot,s')
     \Big)\,k(s'\mid a,s)\,\mu(ds') \\
&\quad + \int \Pi_{r,k}(f)(s')\,(k'-k)(s' \mid a, s)\,\mu(ds').
\end{align*}
Moreover, the following bounds hold:
\begin{align*}
\left\|
\begin{aligned}
&\Pi_{r', k} f - \Pi_{r, k} f\\[-1mm]
&\quad - \partial_r \Pi_{r, k}[\,r'-r\,](f)
\end{aligned}
\right\|_{L^1(\mu)}
&\;\lesssim\; \|f\|_{L^\infty(\lambda)} \, \|r'-r\|_{L^2(\lambda)}^2, \\[1.0ex]
\left\|
\begin{aligned}
&\mathcal{P}_{r', k} f - \mathcal{P}_{r, k} f\\[-1mm]
&\quad - \partial_r \mathcal{P}_{r, k}[\,r'-r\,](f)
\end{aligned}
\right\|_{L^\infty(\lambda)}
&\;\lesssim\; \|f\|_{L^\infty(\lambda)} \, \|r'-r\|_{L^2(\lambda)}^2, \\[1.2ex]
\left\|
\begin{aligned}
&\Pi_{r, k'} f - \Pi_{r, k} f\\[-1mm]
&\quad - \adk \Pi_{r, k}[\,k'-k\,](f)
\end{aligned}
\right\|_{L^1(\mu)}
&\;\lesssim\; \|f\|_{L^\infty(\lambda)} \Big\{
      \|v_{r, k'} - v_{r,k}\|_{L^2(\lambda)}^2 \notag\\
&\qquad\qquad
      + \|\mathcal{T}_{r,k}^{-1}\|_{L^2(\lambda) \to L^2(\lambda)} \,
        \|v_{r,k'} - v_{r,k}\|_{L^2(\lambda)}\,
        \|k' - k\|_{L^2(\mu \otimes \lambda)}
      \Big\}, \\[1.2ex]
\left\|
\begin{aligned}
&\mathcal{P}_{r,k'}(f) - \mathcal{P}_{r,k}(f)\\
&\quad - \adk \mathcal{P}_{r,k}[\,k'-k\,](f)
\end{aligned}
\right\|_{L^2(\lambda)}
&\;\lesssim\; \|f\|_{L^\infty(\lambda)} \Big\{
      \|v_{r, k'} - v_{r,k}\|_{L^2(\lambda)}^2 \notag\\
&\qquad
      + \big(1 + \|\mathcal{T}_{r,k}^{-1}\|_{L^2(\lambda) \to L^2(\lambda)}\big)\notag\\
&\qquad\quad \times
        \|v_{r,k'} - v_{r,k}\|_{L^2(\lambda)}\,
        \|k' - k\|_{L^2(\mu \otimes \lambda)}
      \Big\}.
\end{align*}
\end{lemma}

\begin{proof}
\noindent \textbf{Differentiability of softmax operator.} Define the operator $\widetilde{\Pi}_{q,\tau}: L^2(\lambda) \rightarrow L^2(\mu)$ by
\[
(\widetilde{\Pi}_{q,\tau} f)(s) := \sum_{a' \in \mathcal{A}} \widetilde{\pi}_{q,\tau}(a' \mid s)\, f(a', s),
\qquad
\widetilde{\pi}_{q,\tau}(a' \mid s) := \frac{\exp\!\left(\frac{q(a',s)}{\tau}\right)}{\sum_{b \in \mathcal{A}} \exp\!\left(\frac{q(b,s)}{\tau}\right)}.
\]
For fixed $s' \in \mathcal{S}$, let
\[
\Phi(t) := (\widetilde{\Pi}_{q + t f_1,\tau} f)(s')
= \sum_{a' \in \mathcal{A}} \widetilde{\pi}_{\,q + t f_1,\tau}(a' \mid s')\, f(a', s').
\]
The derivative of $\widetilde{\pi}_{\,q + t f_1,\tau}(a' \mid s')$ with respect to $t$ is
\[
\frac{d}{dt} \widetilde{\pi}_{\,q + t f_1,\tau}(a' \mid s')
= \frac{1}{\tau}\,\widetilde{\pi}_{\,q + t f_1,\tau}(a' \mid s') \big[ f_1(a', s') - E_{\widetilde{\pi}_{\,q + t f_1,\tau}}[f_1(\cdot, s')] \big].
\]
Hence, $\Phi'(0) = \tfrac{1}{\tau}\operatorname{Cov}_{\widetilde{\pi}_{q,\tau}(\cdot \mid s')}(f(\cdot, s'), f_1(\cdot, s'))$.
The second derivative $\Phi''(t)$ is uniformly bounded in absolute value by
\(2\,\tau^{-2}\,\|f(\cdot, s')\|_\infty
\|f_1(\cdot, s')\|_{L^2(\widetilde{\pi}_{\,q + t f_1,\tau}(\cdot \mid s'))}^2\).
By Taylor's theorem with integral remainder,
\begin{align}
\left| \Phi(t) - \Phi(0) - t\,\Phi'(0) \right|
&\le t^2\,\tau^{-2}\,\|f(\cdot, s')\|_\infty
   \sup_{u \in [0,1]} \|f_1(\cdot, s')\|_{L^2(\widetilde{\pi}_{\,q + u t f_1,\tau}(\cdot \mid s'))}^2 \\
&\le t^2\,\tau^{-2}\,\|f(\cdot, s')\|_\infty \sum_{a' \in \mathcal{A}} \{f_1(a', s')\}^2 .
\label{eqn::softmax-operator-taylor-bound}
\end{align}
Setting \(t=1\) in \eqref{eqn::softmax-operator-taylor-bound} gives
\begin{align*}
    \left|
\sum_{a' \in \mathcal{A}} \widetilde{\pi}_{\,q + f_1,\tau}(a' \mid s') f(a', s')
- \sum_{a' \in \mathcal{A}} \widetilde{\pi}_{q,\tau}(a' \mid s') f(a', s')
- \frac{1}{\tau}\operatorname{Cov}_{\widetilde{\pi}_{q,\tau}(\cdot \mid s')}\big(f(\cdot, s'), f_1(\cdot, s')\big)
\right|\\
\le \tau^{-2}\max_{a' \in \mathcal{A}} |f(a', s')| \sum_{a' \in \mathcal{A}} \big\{ f_1(a', s') \big\}^2.
\end{align*}
Thus, letting
\[
\partial_q \widetilde{\Pi}_{q,\tau}[f_1](f) := \big(s' \mapsto \tfrac{1}{\tau}\operatorname{Cov}_{\widetilde{\pi}_{q,\tau}(\cdot \mid s')}\big(f(\cdot, s'), f_1(\cdot, s')\big)\big),
\]
we obtain the pointwise bound
\begin{align}
    \left|(\widetilde{\Pi}_{q + f_1,\tau} f)(s') - (\widetilde{\Pi}_{q,\tau} f)(s') - \partial_q \widetilde{\Pi}_{q,\tau}[f_1](f)(s') \right|
&\lesssim  \max_{a' \in \mathcal{A}} |f(a', s')| \sum_{a' \in \mathcal{A}} \big\{ f_1(a', s') \big\}^2.
\label{eqn::softmax-pointwise-taylor-bound}
\end{align}

\noindent \textbf{Reward derivative of $\Pi_{r,k}$.}
The chain rule gives
$\partial_r\Pi_{r,k}[u]
=\partial_q\widetilde\Pi_{q_{r,k},\tau}[\partial_r q_{r,k}[u]]$ and hence
\begin{align*}
    \partial_r \Pi_{r, k}[u](f)(s') &=  \frac{1}{\tau}\operatorname{Cov}_{\pi_{r, k}(\cdot \mid s')}\big(f(\cdot, s'), \partial_r q_{r,k}[u](\cdot, s')\big) .
\end{align*}
Adding and subtracting the linear term gives
\begin{align*}
&\left| \Pi_{r + u, k} f - \Pi_{r, k} f - \partial_r \Pi_{r, k}[u](f) \right| \\
&= \left| \Pi_{r + u, k} f - \Pi_{r, k} f - \partial_q \widetilde{\Pi}_{q_{r,k},\tau}[\partial_r q_{r,k}[u]](f) \right| \\
&\leq \left| \widetilde{\Pi}_{q_{r + u, k},\tau} f - \widetilde{\Pi}_{q_{r,k},\tau} f - \partial_q \widetilde{\Pi}_{q_{r,k},\tau}[q_{r + u, k} - q_{r,k}] (f)\right| \\
&\quad + \left| \partial_q \widetilde{\Pi}_{q_{r,k},\tau}[q_{r + u, k} - q_{r,k} - \partial_r q_{r,k}[u]](f) \right|,
\end{align*}
where
\(q_{r + u, k} - q_{r,k} - \partial_r q_{r,k}[u]
= \gamma \big(v_{r + u, k} - v_{r,k} - \partial_r v_{r,k}[u]\big)\).
Equation~\eqref{eqn::softmax-pointwise-taylor-bound} therefore gives
\begin{align*}
&\left| \Pi_{r + u, k} f - \Pi_{r, k} f - \partial_r \Pi_{r, k}[u](f) \right| \\
&\lesssim \|f\|_{L^\infty(\lambda)}  \sum_{a' \in \mathcal{A}} (q_{r + u, k} - q_{r,k})^2(a' , \cdot)
   + \left| \partial_q \widetilde{\Pi}_{q_{r,k},\tau}[q_{r + u, k} - q_{r,k} - \partial_r q_{r,k}[u]](f) \right| \\
&\lesssim \|f\|_{L^\infty(\lambda)} \,\sum_{a' \in \mathcal{A}} (q_{r + u, k} - q_{r,k})^2(a' , \cdot)
   + \gamma\,\|f\|_{L^\infty(\lambda)} \cdot \Pi_{r,k}\left|v_{r + u, k} - v_{r,k} - \partial_r v_{r,k}[u] \right| ,
\end{align*}
where the last inequality follows from the covariance formula defining
$\partial_q\widetilde\Pi_{q_{r,k},\tau}$. Lemma~\ref{lemma::lipschitzvalue}
also gives
\[
\|q_{r',k} - q_{r,k}\|_{L^2(\lambda)}
\lesssim \|r' - r\|_{L^2(\lambda)}
+ \gamma \|v_{r',k} - v_{r,k}\|_{L^\infty(\lambda)}
\lesssim \|r' - r\|_{L^2(\lambda)}.
\]
Lemma~\ref{lemma::valuefirstderivativeexpansion} gives
$\|v_{r+u,k}-v_{r,k}-\partial_rv_{r,k}[u]\|_{L^\infty(\lambda)}
\lesssim\|u\|_{L^2(\lambda)}^2$. Consequently,
\begin{align}
    \label{eqn::boundPi_reward}
    \| \Pi_{r + u, k} f - \Pi_{r, k} f - \partial_r \Pi_{r, k}[u](f) \|_{L^1(\mu)}
\;\lesssim\; \|f\|_{L^\infty(\lambda)} \,\|u\|_{L^2(\lambda)}^2.
\end{align}

\noindent \textbf{Kernel derivative of $\Pi_{r,k}$.}
The chain rule gives
$\adk\Pi_{r,k}[k'-k]
=\partial_q\widetilde\Pi_{q_{r,k},\tau}[\adk q_{r,k}[k'-k]]$ and hence
\begin{align*}
    \adk \Pi_{r, k}[k'-k](f)(s') &=  \frac{1}{\tau}\operatorname{Cov}_{\pi_{r, k}(\cdot \mid s')}\big(f(\cdot, s'), \adk q_{r,k}[k'-k](\cdot, s')\big) .
\end{align*}
Adding and subtracting the linear term gives
\begin{align*}
&\left| \Pi_{r, k'} f - \Pi_{r, k} f - \adk \Pi_{r, k}[k'-k](f) \right| \\
&= \left| \Pi_{r , k'} f - \Pi_{r, k} f - \partial_q \widetilde{\Pi}_{q_{r,k},\tau}\big[\adk q_{r,k}[k'-k]\big](f) \right| \\
&\leq \left| \widetilde{\Pi}_{q_{r , k'},\tau} f - \widetilde{\Pi}_{q_{r,k},\tau} f
        - \partial_q \widetilde{\Pi}_{q_{r,k},\tau}\big[q_{r, k'} - q_{r,k}\big](f)\right| \\
&\quad + \left| \partial_q \widetilde{\Pi}_{q_{r,k},\tau}\big[q_{r, k'} - q_{r,k} - \adk q_{r,k}[k'-k]\big](f) \right|.
\end{align*}
Since $q_{r,k}=r+\gamma v_{r,k}$ and $r$ is fixed here, we have
\[
q_{r, k'} - q_{r,k} - \adk q_{r,k}[k'-k]
= \gamma \big(v_{r, k'} - v_{r,k} - \adk v_{r,k}[k'-k]\big).
\]
Equation~\eqref{eqn::softmax-pointwise-taylor-bound} gives
\begin{align*}
&\left| \Pi_{r, k'} f - \Pi_{r, k} f - \adk \Pi_{r, k}[k'-k](f) \right| \\
&\lesssim \|f\|_{L^\infty(\lambda)}  \sum_{a' \in \mathcal{A}} \big(q_{r, k'} - q_{r,k}\big)^2(a' , \cdot)
   + \left| \partial_q \widetilde{\Pi}_{q_{r,k},\tau}\big[q_{r, k'} - q_{r,k} - \adk q_{r,k}[k'-k]\big](f) \right| \\
&\lesssim \|f\|_{L^\infty(\lambda)} \,\sum_{a' \in \mathcal{A}} \big(v_{r, k'} - v_{r,k}\big)^2(a' , \cdot)
   + \gamma\,\|f\|_{L^\infty(\lambda)}  \Pi_{r,k}\left|v_{r, k'} - v_{r,k} - \adk v_{r,k}[k'-k] \right| .
\end{align*}
Lemma~\ref{lemma::valuefirstderivativeexpansion::kernel} bounds the last
term by
\[
\begin{aligned}
&\|v_{r,k'}-v_{r,k}-\adk v_{r,k}[k'-k]\|_{L^2(\lambda)}\\
&\quad\lesssim\|\mathcal T_{r,k}^{-1}\|_{L^2\to L^2}
\left\{\|v_{r,k'}-v_{r,k}\|_{L^2(\lambda)}^2
+\|v_{r,k'}-v_{r,k}\|_{L^2(\lambda)}
\|k'-k\|_{L^2(\mu\otimes\lambda)}\right\}.
\end{aligned}
\]
Thus, Cauchy--Schwarz gives
\begin{align}
\label{eqn::boundPi_kernel}
\| \Pi_{r, k'} f - \Pi_{r, k} f - \adk \Pi_{r, k}[k'-k](f) \|_{L^1(\mu)}
&\lesssim \|f\|_{L^\infty(\lambda)} \Big\{
    \|v_{r, k'} - v_{r,k}\|_{L^2(\lambda)}^2 \\
&\qquad\qquad
    + \|\mathcal{T}_{r,k}^{-1}\|_{L^2 \rightarrow L^2} \\
&\qquad\qquad\quad\times
      \|v_{r,k'} - v_{r,k}\|_{L^2(\lambda)}
      \|k' - k\|_{L^2(\mu \otimes \lambda)}
  \Big\}.
\end{align}

\noindent \textbf{Reward derivative of $\mathcal P_{r,k}$.}
By definition,
$(\mathcal P_{r,k}f)(a,s)=\int\Pi_{r,k}(f)(s')
k(s'\mid a,s)\,\mu(ds')$.
By linearity and the chain rule,
\begin{align*}
\partial_r \mathcal{P}_{r,k}[\,u\,](f)(a,s)
&= \int \partial_r \Pi_{r,k}[\,u\,](f)(s')\,k(s'\mid a,s)\,\mu(ds') \\
&= \frac{1}{\tau}\int \operatorname{Cov}_{\pi_{r,k}(\cdot\mid s')}
   \!\Big(f(\cdot,s'),\,\partial_r q_{r,k}[\,u\,](\cdot,s')\Big)\,
   k(s'\mid a,s)\,\mu(ds').
\end{align*}
Therefore,
\begin{align*}
&\left| \mathcal{P}_{r+u,k}f - \mathcal{P}_{r,k}f - \partial_r \mathcal{P}_{r,k}[\,u\,](f) \right|(a,s) \\
&\qquad= \left|\int \Big(\Pi_{r+u,k}f - \Pi_{r,k}f - \partial_r \Pi_{r,k}[\,u\,](f)\Big)(s')\,k(s'\mid a,s)\,\mu(ds')\right| \\
&\qquad\le \|k(\cdot\mid a,s)\|_{L^\infty(\mu)}\,
          \left\|\Pi_{r+u,k}f - \Pi_{r,k}f - \partial_r \Pi_{r,k}[\,u\,](f)\right\|_{L^1(\mu)} \\
&\qquad\lesssim \left\|\Pi_{r+u,k}f - \Pi_{r,k}f - \partial_r \Pi_{r,k}[\,u\,](f)\right\|_{L^1(\mu)}
\;\lesssim\; \|f\|_{L^\infty(\lambda)}\,\|u\|_{L^2(\lambda)}^2,
\end{align*}
where the last step uses \eqref{eqn::boundPi_reward} and the essential
boundedness of the fixed kernel $k$; the implied constant may depend on
$\|k\|_{L^\infty}$.

\noindent In particular, taking the supremum over $(a,s)$ yields
\[
\big\|\mathcal{P}_{r+u,k}f - \mathcal{P}_{r,k}f - \partial_r \mathcal{P}_{r,k}[\,u\,](f)\big\|_{L^\infty(\lambda)}
\;\lesssim\; \|f\|_{L^\infty(\lambda)}\,\|u\|_{L^2(\lambda)}^2.
\]

\noindent \textbf{Kernel derivative of $\mathcal P_{r,k}$.}
By the chain rule,
\begin{align*}
\adk \mathcal{P}_{r,k}[\,k'-k\,](f)(a,s)
&= \int \adk \Pi_{r,k}[\,k'-k\,](f)(s')\,k(s'\mid a,s)\,\mu(ds') \\
&\quad + \int \Pi_{r,k}(f)(s')\,(k'-k)(s'\mid a,s)\,\mu(ds') \\
&= \frac{1}{\tau}\int \operatorname{Cov}_{\pi_{r,k}(\cdot\mid s')}\!\Big(f(\cdot,s'),\,\adk q_{r,k}[\,k'-k\,](\cdot,s')\Big)\,k(s'\mid a,s)\,\mu(ds') \\
&\quad + \int \Pi_{r,k}(f)(s')\,(k'-k)(s'\mid a,s)\,\mu(ds').
\end{align*}
Hence,
\begin{align*}
&\Big|\mathcal{P}_{r,k'}(f)(a,s)-\mathcal{P}_{r,k}(f)(a,s)-\adk \mathcal{P}_{r,k}[\,k'-k\,](f)(a,s)\Big| \\
&\qquad= \Bigg|\int \Big(\Pi_{r,k'}(f)-\Pi_{r,k}(f)-\adk \Pi_{r,k}[\,k'-k\,](f)\Big)(s')\,k(s'\mid a,s)\,\mu(ds') \\
&\qquad\qquad\quad + \int \big(\Pi_{r,k'}(f)-\Pi_{r,k}(f)\big)(s')\,(k'-k)(s'\mid a,s)\,\mu(ds')\Bigg|.
\end{align*}
By \eqref{eqn::boundPi_kernel}, we have
\begin{align*}
&\Bigg|\int \Big(\Pi_{r,k'}(f)-\Pi_{r,k}(f)-\adk \Pi_{r,k}[\,k'-k\,](f)\Big)(s')\,k(s'\mid a,s)\,\mu(ds')\Bigg| \\
&\qquad\lesssim \|f\|_{L^\infty(\lambda)}\Big\{ \|v_{r,k'}-v_{r,k}\|_{L^2(\lambda)}^2
+ \|\mathcal{T}_{r,k}^{-1}\|_{L^2(\lambda)\to L^2(\lambda)}\,\|v_{r,k'}-v_{r,k}\|_{L^2(\lambda)}\,\|k'-k\|_{L^2(\mu\otimes\lambda)}\Big\}.
\end{align*}
For the second integral, using that the temperature-$\tau$ softmax map is $\tau^{-1}$-Lipschitz from $(\ell_\infty,\ell_1)$,
\[
\begin{aligned}
\big|\Pi_{r,k'}(f)(s')-\Pi_{r,k}(f)(s')\big|
&\le \|f\|_{L^\infty(\lambda)}\,
\big\|\pi_{r,k'}(\cdot\mid s')-\pi_{r,k}(\cdot\mid s')\big\|_1\\
&\le \frac{\gamma}{\tau}\,\|f\|_{L^\infty(\lambda)}\,
\sum_{a' \in \mathcal{A}}\big|v_{r,k'}(a',s')-v_{r,k}(a',s')\big|.
\end{aligned}
\]
For \(g:=k'-k\), the Hilbert--Schmidt inequality gives
\begin{align*}
\left\|
\begin{aligned}
(a,s)\mapsto \int
&\big(\Pi_{r,k'}(f)-\Pi_{r,k}(f)\big)(s')\\
&\times (k'-k)(s'\mid a,s)\,\mu(ds')
\end{aligned}
\right\|_{L^2(\lambda)} \\
&\leq \|g\|_{L^2(\mu\otimes\lambda)}
\|\Pi_{r,k'}(f)-\Pi_{r,k}(f)\|_{L^2(\mu)}\\
&\lesssim \|f\|_{L^\infty(\lambda)}\,
\|v_{r,k'}-v_{r,k}\|_{L^2(\lambda)}\,
\|k'-k\|_{L^2(\mu\otimes\lambda)}.
\end{align*}
Combining the two bounds,
\begin{align*}
\left\|
\begin{aligned}
&\mathcal{P}_{r,k'}(f)-\mathcal{P}_{r,k}(f)\\
&\quad - \adk \mathcal{P}_{r,k}[\,k'-k\,](f)
\end{aligned}
\right\|_{L^2(\lambda)}
\;\lesssim\; &\|f\|_{L^\infty(\lambda)}\Big\{
\|v_{r,k'}-v_{r,k}\|_{L^2(\lambda)}^2 \\
&+ \big(1+\|\mathcal{T}_{r,k}^{-1}\|_{L^2(\lambda)\to L^2(\lambda)}\big) \\
&\quad \times
\|v_{r,k'}-v_{r,k}\|_{L^2(\lambda)}\,
\|k'-k\|_{L^2(\mu\otimes\lambda)}\Big\}.
\end{align*}
These bounds prove the stated derivative formulas and remainder estimates.
\end{proof}

\begin{lemma}[Reward remainder for the soft policy value]
\label{lemma::valuestarderivativereward}
Let $r,h\in\mathcal H$, and fix $k$.
Suppose $\mathcal{T}_{r,k}:L^2(\lambda) \to L^2(\lambda)$ has a bounded
inverse. For all sufficiently small $h$ such that $r+h\in\mathcal H$,
$\mathcal T_{r+h,k}$ is also invertible and
\[
\big\|\Pi_{r+h,k}\,\mathcal{T}_{r+h,k}^{-1}(r+h)
 - \Pi_{r,k}\,\mathcal{T}_{r,k}^{-1}(r)
 - \partial_r\!\big[\Pi_{r,k}\,\mathcal{T}_{r,k}^{-1}(r)\big][h]\big\|_{L^1(\mu)}
= O(\|h\|_{L^2(\lambda)}^2),
\]
where the big-$O$ constant is locally uniform in $h$.
\end{lemma}
\begin{proof}
Lemmas~\ref{lemma::lipschitzvalue}
and~\ref{lemma::forwardopLipschitz} imply
\(\|\mathcal T_{r+h,k}-\mathcal T_{r,k}\|_{L^2\to L^2}
=O(\|h\|_{L^2(\lambda)})\). Thus, for all sufficiently small \(h\),
\begin{equation}
\|\mathcal T_{r,k}^{-1}
(\mathcal T_{r+h,k}-\mathcal T_{r,k})\|_{L^2\to L^2}\le\tfrac12.
\label{eqn::reward-local-inverse-radius}
\end{equation}
Lemma~\ref{lemma::inverseperturbation} and
\eqref{eqn::inverse-perturbation-bound} then show that
\(\mathcal T_{r+h,k}\) is invertible and
\(\|\mathcal T_{r+h,k}^{-1}\|_{L^2\to L^2}
\leq2\|\mathcal T_{r,k}^{-1}\|_{L^2\to L^2}\).

By the chain rule,
\begin{align*}
\partial_r\!\big[\Pi_{r,k}\,\mathcal{T}_{r,k}^{-1}(r)\big][h]
&= \big(\partial_r \Pi_{r,k}[h]\big)\,\mathcal{T}_{r,k}^{-1}(r)
\;+\;\Pi_{r,k}\,\big(\partial_r \mathcal{T}_{r,k}^{-1}[h]\big)(r)
\;+\;\Pi_{r,k}\,\mathcal{T}_{r,k}^{-1}(h),
\end{align*}
where $\partial_r \mathcal{T}_{r,k}^{-1}[h] = -\,\mathcal{T}_{r,k}^{-1}\big(\partial_r \mathcal{T}_{r,k}[h]\big)\mathcal{T}_{r,k}^{-1}$.
By adding and subtracting, we obtain
\begin{equation}
\begin{aligned}
&\Pi_{r+h,k}\,\mathcal{T}_{r+h,k}^{-1}(r+h) - \Pi_{r,k}\,\mathcal{T}_{r,k}^{-1}(r) - \partial_r\!\big[\Pi_{r,k}\,\mathcal{T}_{r,k}^{-1}(r)\big][h] \\
&= \big(\Pi_{r+h,k}-\Pi_{r,k}-\partial_r \Pi_{r,k}[h]\big)\,\mathcal{T}_{r,k}^{-1}(r) \\
&\quad+\;\Pi_{r,k}\,\big(\mathcal{T}_{r+h,k}^{-1}-\mathcal{T}_{r,k}^{-1}-\partial_r \mathcal{T}_{r,k}^{-1}[h]\big)(r) \\
&\quad+\;\big(\Pi_{r+h,k}-\Pi_{r,k}\big)\big(\mathcal{T}_{r+h,k}^{-1}-\mathcal{T}_{r,k}^{-1}\big)(r) \\
&\quad+\;\big(\Pi_{r+h,k}\,\mathcal{T}_{r+h,k}^{-1}-\Pi_{r,k}\,\mathcal{T}_{r,k}^{-1}\big)(h).
\end{aligned}
\label{eqn::soft-value-reward-remainder-decomposition}
\end{equation}
Taking norms and applying Lemma~\ref{lemma::operatorderivatives}, we have
\begin{align*}
& \big\|\Pi_{r+h,k}\,\mathcal{T}_{r+h,k}^{-1}(r+h) - \Pi_{r,k}\,\mathcal{T}_{r,k}^{-1}(r) - \partial_r\!\big[\Pi_{r,k}\,\mathcal{T}_{r,k}^{-1}(r)\big][h]\big\|_{L^1(\mu)} \\
&\lesssim \big\|\Pi_{r+h,k}-\Pi_{r,k}-\partial_r \Pi_{r,k}[h]\big\|_{L^\infty(\lambda) \to L^1(\mu)} \,\big\|\mathcal{T}_{r,k}^{-1}(r)\big\|_{L^\infty(\lambda)} \\
&\quad+ \big\|\mathcal{T}_{r+h,k}^{-1}-\mathcal{T}_{r,k}^{-1}-\partial_r \mathcal{T}_{r,k}^{-1}[h]\big\|_{L^\infty(\lambda) \to L^2(\lambda)} \,\|r\|_{L^\infty(\lambda)} \\
&\quad+\;\big\|\Pi_{r+h,k}-\Pi_{r,k}\big\|_{L^2(\lambda) \to L^1(\mu)}\,\big\|\big(\mathcal{T}_{r+h,k}^{-1}-\mathcal{T}_{r,k}^{-1}\big)(r)\big\|_{L^2(\lambda)} \\
&\quad+\;\big\|\big(\Pi_{r+h,k}\,\mathcal{T}_{r+h,k}^{-1}-\Pi_{r,k}\,\mathcal{T}_{r,k}^{-1}\big)(h)\big\|_{L^1(\mu)}.
\end{align*}

We bound each summand in
\eqref{eqn::soft-value-reward-remainder-decomposition} by a locally uniform
constant times $\|h\|_{L^2(\lambda)}^2$.

\medskip
\noindent \textbf{Direct softmax remainder.}
By Lemma~\ref{lemma::operatorderivatives},
\[
\big\|\Pi_{r+h,k}-\Pi_{r,k}-\partial_r \Pi_{r,k}[h]\big\|_{L^\infty(\lambda) \to L^1(\mu)} \lesssim \|h\|_{L^2(\lambda)}^2.
\]
Hence,
\[
\big\|\Pi_{r+h,k}-\Pi_{r,k}-\partial_r \Pi_{r,k}[h]\big\|_{L^\infty(\lambda) \to L^1(\mu)} \,\big\|\mathcal{T}_{r,k}^{-1}(r)\big\|_{L^\infty(\lambda)}
\lesssim \big\|\mathcal{T}_{r,k}^{-1}(r)\big\|_{L^\infty(\lambda)} \,\|h\|_{L^2(\lambda)}^2.
\]

\medskip
\noindent \textbf{Inverse remainder.}
Lemma~\ref{lemma::inverseperturbation}, specifically
\eqref{eqn::generalized-resolvent-identity} with
\(L=\partial_r\mathcal T_{r,k}[h]\), gives
\[
\begin{aligned}
&\mathcal{T}_{r+h,k}^{-1} - \mathcal{T}_{r,k}^{-1}
+ \mathcal{T}_{r,k}^{-1}\big(\partial_r \mathcal{T}_{r,k}[h]\big)\mathcal{T}_{r,k}^{-1} \\
&\qquad=- \mathcal{T}_{r,k}^{-1}\Big( \mathcal{T}_{r+h,k} - \mathcal{T}_{r,k} - \partial_r \mathcal{T}_{r,k}[h] \Big)\mathcal{T}_{r,k}^{-1} \\
&\qquad\quad +\; \mathcal{T}_{r,k}^{-1}\big(\mathcal{T}_{r+h,k} - \mathcal{T}_{r,k}\big)\mathcal{T}_{r,k}^{-1}\big(\mathcal{T}_{r+h,k} - \mathcal{T}_{r,k}\big)\mathcal{T}_{r+h,k}^{-1}.
\end{aligned}
\]
Hence,
\[
\begin{aligned}
& \big\|\mathcal{T}_{r+h,k}^{-1} - \mathcal{T}_{r,k}^{-1}
+ \mathcal{T}_{r,k}^{-1}(\partial_r \mathcal{T}_{r,k}[h])\mathcal{T}_{r,k}^{-1}\big\|_{L^\infty \to L^\infty} \\
&\;\;\lesssim \|\mathcal{T}_{r,k}^{-1}\|_{L^\infty \to L^\infty}^{2}\,
   \big\|\,\mathcal{T}_{r+h,k} - \mathcal{T}_{r,k} - \partial_r \mathcal{T}_{r,k}[h]\,\big\|_{L^\infty \to L^\infty}
   \;+\; \|\mathcal{T}_{r,k}^{-1}\|_{L^\infty \to L^\infty}^2\,
   \|\mathcal{T}_{r+h,k}^{-1}\|_{L^\infty \to L^\infty}\,\|h\|_{L^2(\lambda)}^2 .
\end{aligned}
\]
The remainder bound in Lemma~\ref{lemma::operatorderivatives} for $\mathcal{P}_{r,k}$ implies
\[
\big\|\,\mathcal{T}_{r+h,k} - \mathcal{T}_{r,k} - \partial_r \mathcal{T}_{r,k}[h]\,\big\|_{L^\infty \to L^\infty} \lesssim \|h\|_{L^2(\lambda)}^2.
\]
Consequently,
\[
\big\|\mathcal{T}_{r+h,k}^{-1} - \mathcal{T}_{r,k}^{-1}
+ \mathcal{T}_{r,k}^{-1}(\partial_r \mathcal{T}_{r,k}[h])\mathcal{T}_{r,k}^{-1}\big\|_{L^\infty \to L^\infty} \lesssim \|h\|_{L^2(\lambda)}^2,
\]
and thus
\[
\big\|\mathcal{T}_{r+h,k}^{-1}-\mathcal{T}_{r,k}^{-1}-\partial_r \mathcal{T}_{r,k}^{-1}[h]\big\|_{L^\infty(\lambda) \to L^2(\lambda)} \,\|r\|_{L^\infty(\lambda)} \lesssim \|h\|_{L^2(\lambda)}^2 \,\|r\|_{L^\infty(\lambda)}.
\]

\medskip
\noindent \textbf{Product of the policy and inverse changes.}
We have
\[
\|\mathcal{T}_{r+h,k}^{-1} - \mathcal{T}_{r,k}^{-1}\|_{L^\infty \to L^\infty}
\;\le\; \|\mathcal{T}_{r,k}^{-1}\|_{L^\infty \to L^\infty}\,
        \|\mathcal{T}_{r+h,k} - \mathcal{T}_{r,k}\|_{L^\infty \to L^\infty}\,
        \|\mathcal{T}_{r+h,k}^{-1}\|_{L^\infty \to L^\infty}.
\]
Furthermore, by Lemma~\ref{lemma::operatorderivatives},
\[
\big\| \mathcal{P}_{r + h, k}  - \mathcal{P}_{r, k} \big\|_{L^\infty \to L^\infty}
     \;\lesssim\; \|\partial_r \mathcal{P}_{r, k}[h]\|_{L^\infty \to L^\infty}
     + \|h\|_{L^2(\lambda)}^2,
\]
and $\|\partial_r \mathcal{P}_{r, k}[h]\|_{L^\infty \to L^\infty} \lesssim \|h\|_{L^2(\lambda)}$. Hence,
\[
\|\mathcal{P}_{r+h,k} - \mathcal{P}_{r,k}\|_{L^\infty \to L^\infty}
\;\lesssim\; \|h\|_{L^2(\lambda)},
\qquad
\|\mathcal{T}_{r+h,k} - \mathcal{T}_{r,k}\|_{L^\infty \to L^\infty}
\;\lesssim\; \|h\|_{L^2(\lambda)}.
\]
Therefore,
\[
\|\mathcal{T}_{r+h,k}^{-1} - \mathcal{T}_{r,k}^{-1}\|_{L^\infty \to L^\infty}
\;\lesssim \|h\|_{L^2(\lambda)},
\]
and
\[
\big\|\big(\mathcal{T}_{r+h,k}^{-1}-\mathcal{T}_{r,k}^{-1}\big)(r)\big\|_{L^2(\lambda)}
\lesssim \|h\|_{L^2(\lambda)} \,\|r\|_{L^\infty(\lambda)}.
\]

Finally, by Lemma~\ref{lemma::softmaxLipschitz},
\[
\big\|\Pi_{r+h,k}-\Pi_{r,k}\big\|_{L^2(\lambda) \to L^1(\mu)} \lesssim \|h\|_{L^2(\lambda)}.
\]
Combining these bounds,
\[
\big\|\Pi_{r+h,k}-\Pi_{r,k}\big\|_{L^2(\lambda) \to L^1(\mu)}
\,\big\|\big(\mathcal{T}_{r+h,k}^{-1}-\mathcal{T}_{r,k}^{-1}\big)(r)\big\|_{L^2(\lambda)}
\;\lesssim \|h\|_{L^2(\lambda)}^2 \,\|r\|_{L^\infty(\lambda)}.
\]

\medskip
\noindent \textbf{Perturbed direct term.}
\begin{align*}
&\big\|\Pi_{r+h,k}\,\mathcal{T}_{r+h,k}^{-1}(h)
-\Pi_{r,k}\,\mathcal{T}_{r,k}^{-1}(h)\big\|_{L^1(\mu)} \\
&= \big\|\big(\Pi_{r+h,k}-\Pi_{r,k}\big)\,\mathcal{T}_{r+h,k}^{-1}(h)
\;+\;\Pi_{r,k}\,\big(\mathcal{T}_{r+h,k}^{-1}-\mathcal{T}_{r,k}^{-1}\big)(h)\big\|_{L^1(\mu)} \\
&\le \big\|\Pi_{r+h,k}-\Pi_{r,k}\big\|_{L^2(\lambda)\to L^1(\mu)}\;
        \big\|\mathcal{T}_{r+h,k}^{-1}\big\|_{L^2(\lambda)\to L^2(\lambda)}\;\|h\|_{L^2(\lambda)} \\
&\quad\;+\;\big\|\Pi_{r,k}\big\|_{L^2(\lambda)\to L^1(\mu)}\;
        \big\|\mathcal{T}_{r+h,k}^{-1}-\mathcal{T}_{r,k}^{-1}\big\|_{L^2(\lambda)\to L^2(\lambda)}\;\|h\|_{L^2(\lambda)}.
\end{align*}
By Lemma~\ref{lemma::softmaxLipschitz},
$\big\|\Pi_{r+h,k}-\Pi_{r,k}\big\|_{L^2(\lambda)\to L^1(\mu)}
\lesssim \|h\|_{L^2(\lambda)}$. By the first resolvent identity,
\begin{align*}
&\|\mathcal T_{r+h,k}^{-1}-\mathcal T_{r,k}^{-1}\|_{L^2\to L^2}\\
&\quad\le
\|\mathcal T_{r,k}^{-1}\|_{L^2\to L^2}
\|\mathcal T_{r+h,k}-\mathcal T_{r,k}\|_{L^2\to L^2}
\|\mathcal T_{r+h,k}^{-1}\|_{L^2\to L^2}
\lesssim\|h\|_{L^2(\lambda)},
\end{align*}
where the final inequality follows from
Lemma~\ref{lemma::forwardopLipschitz} and the local inverse bound established
at the start of the proof. Jensen's and Cauchy--Schwarz inequalities give
\(\|\Pi_{r,k}\|_{L^2(\lambda)\to L^1(\mu)}
\le\mu(\mathcal S)^{1/2}\).
Therefore,
\[
\big\|\big(\Pi_{r+h,k}\,\mathcal{T}_{r+h,k}^{-1}
- \Pi_{r,k}\,\mathcal{T}_{r,k}^{-1}\big)(h)\big\|_{L^1(\mu)}
\;\lesssim\; \|h\|_{L^2(\lambda)}^2.
\]
Combining the four bounds and using the local inverse bound established at
the start of the proof yields
\[
\big\|\Pi_{r+h,k}\,\mathcal{T}_{r+h,k}^{-1}(r+h)
- \Pi_{r,k}\,\mathcal{T}_{r,k}^{-1}(r)
- \partial_r\!\big[\Pi_{r,k}\,\mathcal{T}_{r,k}^{-1}(r)\big][h]\big\|_{L^1(\mu)}
= O\!\left(\|h\|_{L^2(\lambda)}^{2}\right),
\]
with a constant depending only on the base inverse bound and the other
quantities stated in the lemma.
This proves the lemma.
\end{proof}

\begin{lemma}[Kernel remainder for the soft policy value]
\label{lemma::valuestarderivativekernel}
Let $r\in\mathcal H$ and $k \in \mathcal{K}$, and suppose
$\mathcal{T}_{r,k}:L^2(\lambda) \to L^2(\lambda)$ is invertible. Along any
sequence $g\to0$ in $\addTK$ such that $k+g\in\mathcal K$ and
$v_{r,k+g}\to v_{r,k}$ in $L^2(\lambda)$,
\[
\big\|\Pi_{r,k+g}\,\mathcal{T}_{r,k+g}^{-1}(r)
- \Pi_{r,k}\,\mathcal{T}_{r,k}^{-1}(r)
- \adk\!\big[\Pi_{r,k}\,\mathcal{T}_{r,k}^{-1}(r)\big][g]\big\|_{L^1(\mu)}
= O\!\left(\|v_{r,k+g} - v_{r,k}\|_{L^2(\lambda)}^{2} + \|g\|_{\mathcal K}^{2}\right),
\]
where $\mathcal T_{r,k+g}$ is invertible for all sufficiently small $g$ and
the big-$O$ constant depends on $\gamma$, $\|r\|_{L^\infty(\lambda)}$, and
$\|\mathcal{T}_{r,k}^{-1}\|_{L^2(\lambda)\to L^2(\lambda)}$.
\end{lemma}

\begin{proof}
Lemma~\ref{lemma::forwardopLipschitz} gives
\[
\|\mathcal T_{r,k+g}-\mathcal T_{r,k}\|_{L^2(\lambda)\to L^2(\lambda)}
\lesssim \|v_{r,k+g}-v_{r,k}\|_{L^2(\lambda)}+\|g\|_{\mathcal K}=o(1).
\]
For all sufficiently small perturbations, the left-hand side above is at
most \(1/\{2\|\mathcal T_{r,k}^{-1}\|_{L^2\to L^2}\}\).  Hence
Lemma~\ref{lemma::inverseperturbation} and
\eqref{eqn::inverse-perturbation-bound} imply that
$\mathcal T_{r,k+g}$ is invertible and
$\|\mathcal T_{r,k+g}^{-1}\|_{L^2\to L^2}\leq
2\|\mathcal T_{r,k}^{-1}\|_{L^2\to L^2}$.

By the chain rule,
\begin{align*}
\adk\!\big[\Pi_{r,k}\,\mathcal{T}_{r,k}^{-1}(r)\big][g]
&= \big(\adk \Pi_{r,k}[g]\big)\,\mathcal{T}_{r,k}^{-1}(r)
\;+\;\Pi_{r,k}\,\big(\adk \mathcal{T}_{r,k}^{-1}[g]\big)(r),
\end{align*}
where $\adk \mathcal{T}_{r,k}^{-1}[g] = -\,\mathcal{T}_{r,k}^{-1}\big(\adk \mathcal{T}_{r,k}[g]\big)\mathcal{T}_{r,k}^{-1}$.
By adding and subtracting, we obtain
\begin{equation}
\begin{aligned}
&\Pi_{r,k+g}\,\mathcal{T}_{r,k+g}^{-1}(r) - \Pi_{r,k}\,\mathcal{T}_{r,k}^{-1}(r) - \adk\!\big[\Pi_{r,k}\,\mathcal{T}_{r,k}^{-1}(r)\big][g] \\
&= \big(\Pi_{r,k+g}-\Pi_{r,k}-\adk \Pi_{r,k}[g]\big)\,\mathcal{T}_{r,k}^{-1}(r) \\
&\quad+\;\Pi_{r,k}\,\big(\mathcal{T}_{r,k+g}^{-1}-\mathcal{T}_{r,k}^{-1}-\adk \mathcal{T}_{r,k}^{-1}[g]\big)(r) \\
&\quad+\;\big(\Pi_{r,k+g}-\Pi_{r,k}\big)\big(\mathcal{T}_{r,k+g}^{-1}-\mathcal{T}_{r,k}^{-1}\big)(r).
\end{aligned}
\label{eqn::soft-value-kernel-remainder-decomposition}
\end{equation}
Taking norms and applying the kernel-direction bounds in
Lemma~\ref{lemma::operatorderivatives}, we have
\begin{align*}
& \big\|\Pi_{r,k+g}\,\mathcal{T}_{r,k+g}^{-1}(r) - \Pi_{r,k}\,\mathcal{T}_{r,k}^{-1}(r) - \adk\!\big[\Pi_{r,k}\,\mathcal{T}_{r,k}^{-1}(r)\big][g]\big\|_{L^1(\mu)} \\
&\lesssim \big\|\Pi_{r,k+g}-\Pi_{r,k}-\adk \Pi_{r,k}[g]\big\|_{L^\infty(\lambda) \to L^1(\mu)} \,\big\|\mathcal{T}_{r,k}^{-1}(r)\big\|_{L^\infty(\lambda)} \\
&\quad+ \big\|\mathcal{T}_{r,k+g}^{-1}-\mathcal{T}_{r,k}^{-1}-\adk \mathcal{T}_{r,k}^{-1}[g]\big\|_{L^\infty(\lambda) \to L^2(\lambda)} \,\|r\|_{L^\infty(\lambda)} \\
&\quad+\;\big\|\Pi_{r,k+g}-\Pi_{r,k}\big\|_{L^2(\lambda) \to L^1(\mu)}\,\big\|\big(\mathcal{T}_{r,k+g}^{-1}-\mathcal{T}_{r,k}^{-1}\big)(r)\big\|_{L^2(\lambda)}.
\end{align*}

We bound the three summands in
\eqref{eqn::soft-value-kernel-remainder-decomposition} separately.

\medskip
\noindent \textbf{Direct softmax remainder.}
By Lemma~\ref{lemma::operatorderivatives}, we have
\begin{align*}
\left\|
\begin{aligned}
&\Pi_{r, k+g} f - \Pi_{r, k} f\\[-1mm]
&\quad - \adk \Pi_{r, k}[\,g\,](f)
\end{aligned}
\right\|_{L^1(\mu)}
&\;\lesssim\; \|f\|_{L^\infty(\lambda)} \Big\{
      \|v_{r, k+g} - v_{r,k}\|_{L^2(\lambda)}^2 \\
&\hspace{6em}
      + \|\mathcal{T}_{r,k}^{-1}\|_{L^2(\lambda) \to L^2(\lambda)} \,
        \|v_{r,k+g} - v_{r,k}\|_{L^2(\lambda)}\,
        \|g\|_{L^2(\mu \otimes \lambda)} \Big\}.
\end{align*}
Hence,
\begin{align*}
 \big\|\Pi_{r,k+g}-\Pi_{r,k}-\adk \Pi_{r,k}[g]\big\|_{L^\infty(\lambda) \to L^1(\mu)} \,
 \big\|\mathcal{T}_{r,k}^{-1}(r)\big\|_{L^\infty(\lambda)}
&\lesssim  \|v_{r, k+g} - v_{r,k}\|_{L^2(\lambda)}^2 \\
&\quad+  \|v_{r,k+g} - v_{r,k}\|_{L^2(\lambda)}\,\|g\|_{\mathcal K},
\end{align*}
where the implicit constant depends on $\|\mathcal{T}_{r,k}^{-1}\|_{L^2 \to L^2}$ and $\|r\|_\infty$.

\noindent \textbf{Inverse remainder.}
Lemma~\ref{lemma::inverseperturbation}, specifically
\eqref{eqn::generalized-resolvent-identity} with
\(L=\adk\mathcal T_{r,k}[g]\), gives
\[
\begin{aligned}
&\mathcal{T}_{r,k+g}^{-1} - \mathcal{T}_{r,k}^{-1}
+ \mathcal{T}_{r,k}^{-1}\big(\adk \mathcal{T}_{r,k}[g]\big)\mathcal{T}_{r,k}^{-1} \\
&\qquad=- \mathcal{T}_{r,k}^{-1}\Big( \mathcal{T}_{r,k+g} - \mathcal{T}_{r,k} - \adk \mathcal{T}_{r,k}[g] \Big)\mathcal{T}_{r,k}^{-1} \\
&\qquad\quad +\; \mathcal{T}_{r,k}^{-1}\big(\mathcal{T}_{r,k+g} - \mathcal{T}_{r,k}\big)\mathcal{T}_{r,k}^{-1}\big(\mathcal{T}_{r,k+g} - \mathcal{T}_{r,k}\big)\mathcal{T}_{r,k+g}^{-1}.
\end{aligned}
\]
Hence,
\[
\begin{aligned}
& \big\|\mathcal{T}_{r,k+g}^{-1} - \mathcal{T}_{r,k}^{-1}
+ \mathcal{T}_{r,k}^{-1}(\adk \mathcal{T}_{r,k}[g])\mathcal{T}_{r,k}^{-1}\big\|_{L^\infty(\lambda) \to L^2(\lambda)} \\
&\;\;\lesssim \|\mathcal{T}_{r,k}^{-1}\|_{L^2(\lambda) \to L^2(\lambda)}\,
   \big\|\,\mathcal{T}_{r,k+g} - \mathcal{T}_{r,k} - \adk \mathcal{T}_{r,k}[g]\,\big\|_{L^\infty(\lambda) \to L^2(\lambda)} \,\|\mathcal{T}_{r,k}^{-1}\|_{L^\infty(\lambda) \to L^\infty(\lambda)} \\
&\hspace{6em}
   +\; \|\mathcal{T}_{r,k}^{-1}\|_{L^2(\lambda) \to L^2(\lambda)}^2\,
   \|\mathcal{T}_{r,k+g}^{-1}\|_{L^2(\lambda) \to L^2(\lambda)}\,
   \big\|\mathcal{T}_{r,k+g} - \mathcal{T}_{r,k}\big\|_{L^2(\lambda) \to L^2(\lambda)}^{\,2}.
\end{aligned}
\]
The remainder bound in Lemma~\ref{lemma::operatorderivatives} for $\mathcal{P}_{r,k}$ implies
\[
\big\|\,\mathcal{T}_{r,k+g} - \mathcal{T}_{r,k} - \adk \mathcal{T}_{r,k}[g]\,\big\|_{L^\infty(\lambda) \to L^2(\lambda)}
\;\lesssim\;  \|v_{r, k+g} - v_{r,k}\|_{L^2(\lambda)}^2
+  \|v_{r,k+g} - v_{r,k}\|_{L^2(\lambda)}\,\|g\|_{\mathcal{K}},
\]
whereas Lemma~\ref{lemma::forwardopLipschitz} gives
\[
\big\|\,\mathcal{T}_{r,k+g} - \mathcal{T}_{r,k}\,\big\|_{L^2(\lambda) \to L^2(\lambda)}
\;\lesssim\;  \|v_{r, k+g} - v_{r,k}\|_{L^2(\lambda)}
+  \|g\|_{\mathcal{K}}.
\]
For the quadratic product, we also use the continuous embedding
$L^\infty(\lambda)\subset L^2(\lambda)$, since $\lambda$ is finite.
Consequently,
\[
\begin{aligned}
&\big\|\mathcal{T}_{r,k+g}^{-1} - \mathcal{T}_{r,k}^{-1}
+ \mathcal{T}_{r,k}^{-1}(\adk \mathcal{T}_{r,k}[g])\mathcal{T}_{r,k}^{-1}\big\|_{L^\infty(\lambda) \to L^2(\lambda)} \\
&\lesssim\; \|\mathcal{T}_{r,k}^{-1}\|_{L^2(\lambda) \to L^2(\lambda)}^2 \notag\\
&\qquad \times \Big( \|v_{r, k+g} - v_{r,k}\|_{L^2(\lambda)}^2
+  \|v_{r,k+g} - v_{r,k}\|_{L^2(\lambda)}\,\|g\|_{\mathcal{K}} \Big) \\
&\quad +\; \|\mathcal{T}_{r,k}^{-1}\|_{L^2(\lambda) \to L^2(\lambda)}^2
   \,\|\mathcal{T}_{r,k+g}^{-1}\|_{L^2(\lambda) \to L^2(\lambda)} \notag\\
&\qquad \times
   \Big(
      \|v_{r, k+g} - v_{r,k}\|_{L^2(\lambda)}
      + \|g\|_{\mathcal{K}}
   \Big)^{\!2}.
\end{aligned}
\]
Therefore,
\[
\begin{aligned}
\big\|\mathcal{T}_{r,k+g}^{-1}-\mathcal{T}_{r,k}^{-1}
&\qquad - \adk \mathcal{T}_{r,k}^{-1}[g]\big\|_{L^\infty(\lambda) \to L^2(\lambda)} \\
&\qquad \times \|r\|_{L^\infty(\lambda)}
\lesssim\;& \|v_{r, k+g} - v_{r,k}\|_{L^2(\lambda)}^2 \\
&+  \|v_{r,k+g} - v_{r,k}\|_{L^2(\lambda)}\,\|g\|_{\mathcal{K}}
+\|g\|_{\mathcal K}^2.
\end{aligned}
\]

\noindent \textbf{Product of the policy and inverse changes.}
We have
\[
\|\mathcal{T}_{r,k+g}^{-1} - \mathcal{T}_{r,k}^{-1}\|_{L^2(\lambda) \to L^2(\lambda)}
\;\le\; \|\mathcal{T}_{r,k}^{-1}\|_{L^2(\lambda) \to L^2(\lambda)}\,
        \|\mathcal{T}_{r,k+g} - \mathcal{T}_{r,k}\|_{L^2(\lambda) \to L^2(\lambda)}\,
        \|\mathcal{T}_{r,k+g}^{-1}\|_{L^2(\lambda) \to L^2(\lambda)}.
\]
Furthermore, by Lemma~\ref{lemma::forwardopLipschitz},
\[
\|\mathcal{P}_{r,k+g} - \mathcal{P}_{r,k}\|_{L^2(\lambda) \to L^2(\lambda)}
\;\lesssim\; \|v_{r,k+g} - v_{r,k}\|_{L^2(\lambda)}
+ \|g\|_{\mathcal K},
\]
which implies
\[
\|\mathcal{T}_{r,k+g} - \mathcal{T}_{r,k}\|_{L^2(\lambda) \to L^2(\lambda)}
\;\lesssim\; \|v_{r,k+g} - v_{r,k}\|_{L^2(\lambda)} + \|g\|_{\mathcal K}.
\]
Therefore,
\[
\|\mathcal{T}_{r,k+g}^{-1} - \mathcal{T}_{r,k}^{-1}\|_{L^2(\lambda) \to L^2(\lambda)}
\;\lesssim\; \big(\|v_{r,k+g} - v_{r,k}\|_{L^2(\lambda)} + \|g\|_{\mathcal K}\big).
\]
Consequently,
\[
\big\|\big(\mathcal{T}_{r,k+g}^{-1}-\mathcal{T}_{r,k}^{-1}\big)(r)\big\|_{L^2(\lambda)}
\;\lesssim\; \big(\|v_{r,k+g} - v_{r,k}\|_{L^2(\lambda)} + \|g\|_{\mathcal K}\big)\,
\|r\|_{L^\infty(\lambda)}.
\]

Finally, by Lemma~\ref{lemma::softmaxLipschitz},
\[
\big\|\Pi_{r,k+g}-\Pi_{r,k}\big\|_{L^2(\lambda) \to L^1(\mu)}
\;\lesssim\; \|v_{r,k+g} - v_{r,k}\|_{L^2(\lambda)} + \|g\|_{\mathcal K}.
\]
Combining these bounds,
\[
\begin{aligned}
\big\|\Pi_{r,k+g}-\Pi_{r,k}\big\|_{L^2(\lambda) \to L^1(\mu)}
&\times \big\|\big(\mathcal{T}_{r,k+g}^{-1}-\mathcal{T}_{r,k}^{-1}\big)(r)\big\|_{L^2(\lambda)} \\
&\lesssim\; \big(\|v_{r,k+g} - v_{r,k}\|_{L^2(\lambda)} + \|g\|_{\mathcal K}\big)^2\,
\|r\|_{L^\infty(\lambda)}.
\end{aligned}
\]

\noindent \textbf{Final bound.}
Adding the bounds established for Terms 1--3 gives
\[
\begin{aligned}
\left\|
\begin{aligned}
&\Pi_{r,k+g}\,\mathcal{T}_{r,k+g}^{-1}(r)
- \Pi_{r,k}\,\mathcal{T}_{r,k}^{-1}(r)\\[-1mm]
&\quad- \adk\!\big[\Pi_{r,k}\,\mathcal{T}_{r,k}^{-1}(r)\big][g]
\end{aligned}
\right\|_{L^1(\mu)}
&\;\lesssim\; \big(\|v_{r,k+g} - v_{r,k}\|_{L^2(\lambda)} + \|g\|_{\mathcal K}\big)^{2}\,
\|r\|_{L^\infty(\lambda)}.
\end{aligned}
\]
Equivalently (absorbing $\|r\|_{L^\infty(\lambda)}$ into the implicit constant),
\[
\begin{aligned}
\left\|
\begin{aligned}
&\Pi_{r,k+g}\,\mathcal{T}_{r,k+g}^{-1}(r)
- \Pi_{r,k}\,\mathcal{T}_{r,k}^{-1}(r)\\[-1mm]
&\quad- \adk\!\big[\Pi_{r,k}\,\mathcal{T}_{r,k}^{-1}(r)\big][g]
\end{aligned}
\right\|_{L^1(\mu)}
&\;\lesssim\; \|v_{r,k+g} - v_{r,k}\|_{L^2(\lambda)}^{2} + \|g\|_{\mathcal K}^{2}.
\end{aligned}
\]
The inverse bound obtained from
\eqref{eqn::inverse-perturbation-bound} makes the implicit constant uniform
over all sufficiently small perturbations and completes the argument.

\end{proof}

\begin{lemma}[Mixed reward--kernel operator bounds]
\label{lemma::mixedoperatorbounds}
Fix \(r,r+h\in\mathcal H\), \(k\in\mathcal K\), and
\(g\in\addTK\). Suppose that
\(\mathcal T_{r,k}^{-1}\) and \(\mathcal T_{r+h,k}^{-1}\) exist on
\(L^2(\lambda)\), with locally bounded inverse norms. For
\(\eta\in\{0,h\}\), define
\[
\begin{gathered}
A_\eta:=\mathcal T_{r+\eta,k}^{-1},
\qquad P_\eta:=\Pi_{r+\eta,k},
\qquad Q_\eta[g]:=\adk q_{r+\eta,k}[g],\\
B_\eta:=\adk\Pi_{r+\eta,k}[g],
\qquad H_\eta:=\adk\mathcal P_{r+\eta,k}[g],
\qquad C_\eta:=\adk\mathcal T_{r+\eta,k}^{-1}[g].
\end{gathered}
\]
As \(h\to0\) in \(L^2(\lambda)\), locally uniformly at \((r,k)\),
\begin{align}
&\|P_h-P_0\|_{L^2\to L^1}
+\|A_h-A_0\|_{L^2\to L^2}
=O(\|h\|_{L^2(\lambda)}),
\label{eqn::joint-first-order-operator-bounds}\\
&\|B_h-B_0\|_{L^\infty\to L^1}
+\|C_h-C_0\|_{L^\infty\to L^2}
=O\{\|h\|_{L^2(\lambda)}\|g\|_{\mathcal K}\},
\label{eqn::joint-mixed-operator-bounds}\\
&\|B_h\|_{L^2\to L^1}+\|C_h\|_{L^2\to L^2}
+\|B_0\|_{L^2\to L^1}+\|C_0\|_{L^2\to L^2}
=O(\|g\|_{\mathcal K}).
\label{eqn::joint-undifferenced-operator-bounds}
\end{align}
\end{lemma}

\begin{proof}
Write \(q_\eta=q_{r+\eta,k}\),
\(\Phi_\eta=\operatorname{logsumexp}_\tau(q_\eta)\), and
\(\pi_\eta=\pi_{r+\eta,k}\). Lemmas~\ref{lemma::lipschitzvalue}
and~\ref{lemma::softmaxLipschitz} give
\begin{equation}
\|q_h-q_0\|_{L^2(\lambda)}
+\|\Phi_h-\Phi_0\|_{L^2(\mu)}
+\|\pi_h-\pi_0\|_{L^2(\mu;\ell_2)}
=O(\|h\|_{L^2(\lambda)}).
\label{eqn::mixed-basic-reward-bounds}
\end{equation}
The first resolvent identity and
Lemma~\ref{lemma::forwardopLipschitz} give
\[
A_h-A_0
=A_0(\mathcal T_{r,k}-\mathcal T_{r+h,k})A_h,
\qquad
\|A_h-A_0\|_{L^2\to L^2}=O(\|h\|_{L^2(\lambda)}).
\]
Together with Lemma~\ref{lemma::softmaxLipschitz}, this proves
\eqref{eqn::joint-first-order-operator-bounds}.

The kernel derivative of \(q\) satisfies
\(Q_\eta[g]=\gamma A_\eta\mathcal P_g\Phi_\eta\). Hence
\begin{equation}
Q_h[g]-Q_0[g]
=\gamma(A_h-A_0)\mathcal P_g\Phi_0
+\gamma A_h\mathcal P_g(\Phi_h-\Phi_0).
\label{eqn::mixed-kernel-q-decomposition}
\end{equation}
The Hilbert--Schmidt inequality
\(\|\mathcal P_gz\|_{L^2(\lambda)}
\le\|g\|_{\mathcal K}\|z\|_{L^2(\mu)}\),
\eqref{eqn::mixed-basic-reward-bounds}, and the local inverse bounds imply
\begin{equation}
\|Q_h[g]-Q_0[g]\|_{L^2(\lambda)}
=O\{\|h\|_{L^2(\lambda)}\|g\|_{\mathcal K}\},
\qquad
\max_{\eta\in\{0,h\}}\|Q_\eta[g]\|_{L^2(\lambda)}
=O(\|g\|_{\mathcal K}).
\label{eqn::mixed-kernel-q-bounds}
\end{equation}

For a statewise signed action measure \(\delta\pi\), write
\((\delta\pi)f=\sum_a\delta\pi(a\mid\cdot)f(a,\cdot)\). The covariance
formula in Lemma~\ref{lemma::operatorderivatives} gives the exact identity
\begin{align}
\tau(B_h-B_0)f
={}&\operatorname{Cov}_{\pi_h}\{f,Q_h[g]-Q_0[g]\}
+(\pi_h-\pi_0)(fQ_0[g])\notag\\
&-\{(\pi_h-\pi_0)f\}\,\pi_hQ_0[g]
-(\pi_0f)\{(\pi_h-\pi_0)Q_0[g]\}.
\label{eqn::mixed-softmax-derivative-decomposition}
\end{align}
Cauchy--Schwarz, \eqref{eqn::mixed-basic-reward-bounds}, and
\eqref{eqn::mixed-kernel-q-bounds} therefore give
\begin{equation}
\|B_h-B_0\|_{L^\infty\to L^1}
=O\{\|h\|_{L^2(\lambda)}\|g\|_{\mathcal K}\},
\qquad
\max_{\eta\in\{0,h\}}\|B_\eta\|_{L^2\to L^1}
=O(\|g\|_{\mathcal K}).
\label{eqn::mixed-softmax-derivative-bounds}
\end{equation}

The product rule for
\(\mathcal P_{r+\eta,k}=\mathcal P_kP_\eta\) gives
\begin{equation}
H_h-H_0=\mathcal P_k(B_h-B_0)+\mathcal P_g(P_h-P_0).
\label{eqn::mixed-transition-derivative-decomposition}
\end{equation}
The bound \(\|\mathcal P_kz\|_{L^2(\lambda)}
\lesssim\|z\|_{L^1(\mu)}\), the Hilbert--Schmidt inequality, and
\eqref{eqn::mixed-softmax-derivative-bounds} imply
\begin{equation}
\|H_h-H_0\|_{L^\infty\to L^2}
=O\{\|h\|_{L^2(\lambda)}\|g\|_{\mathcal K}\},
\qquad
\max_{\eta\in\{0,h\}}\|H_\eta\|_{L^2\to L^2}
=O(\|g\|_{\mathcal K}).
\label{eqn::mixed-transition-derivative-bounds}
\end{equation}

Finally, since
\(\adk\mathcal T_{r+\eta,k}[g]=-\gamma H_\eta\), the inverse derivative
formula gives \(C_\eta=\gamma A_\eta H_\eta A_\eta\). Adding and
subtracting intermediate products yields
\begin{align}
C_h-C_0
={}&\gamma(A_h-A_0)H_hA_h
+\gamma A_0(H_h-H_0)A_h\notag\\
&+\gamma A_0H_0(A_h-A_0).
\label{eqn::mixed-inverse-derivative-decomposition}
\end{align}
Using \eqref{eqn::joint-first-order-operator-bounds} for \(A_h-A_0\),
\eqref{eqn::mixed-transition-derivative-bounds} for \(H_h-H_0\), and the
locally uniform bounds for \(A_0,A_h,H_0,H_h\) in
\eqref{eqn::mixed-transition-derivative-bounds}, the three products in
\eqref{eqn::mixed-inverse-derivative-decomposition} give
\[
\|C_h-C_0\|_{L^\infty\to L^2}
=O\{\|h\|_{L^2(\lambda)}\|g\|_{\mathcal K}\},
\qquad
\max_{\eta\in\{0,h\}}\|C_\eta\|_{L^2\to L^2}
=O(\|g\|_{\mathcal K}).
\]
Combining these bounds proves
\eqref{eqn::joint-mixed-operator-bounds} and
\eqref{eqn::joint-undifferenced-operator-bounds}. The calculation applies
first to bounded additive directions \(g\); continuity in
\(\|g\|_{\mathcal K}\) and the definition of \(\addTK\) extend it to
\(g\in\addTK\).
\end{proof}

\begin{lemma}[Joint remainder for the soft policy value]
\label{lemma::jointderivative}
Let $r\in\mathcal H$ and $k \in \mathcal{K}$, and suppose the linear
operator $\mathcal{T}_{r,k}:L^2(\lambda) \to L^2(\lambda)$ is invertible.
Along any sequence $(h,g)\to(0,0)$ in
$L^2(\lambda)\times(\addTK,\|\cdot\|_{\mathcal K})$ such that
$r+h\in\mathcal H$, $k+g\in\mathcal K$, and
$v_{r,k+g}\to v_{r,k}$ in $L^2(\lambda)$, and suppose
\[
M_r:=\sup_{(h,g)}\|r+h\|_{L^\infty(\lambda)}<\infty,
\qquad
M_k:=\sup_{(h,g)}\|k+g\|_{L^\infty(\mu\otimes\lambda)}<\infty
\]
along this sequence. Then
\begin{align*}
&\big\|\Pi_{r+h,k+g}\,\mathcal{T}_{r+h,k+g}^{-1}(r+h)
- \Pi_{r,k}\,\mathcal{T}_{r,k}^{-1}(r)
- \partial_r\!\big[\Pi_{r,k}\,\mathcal{T}_{r,k}^{-1}(r)\big][h]
- \adk\!\big[\Pi_{r,k}\,\mathcal{T}_{r,k}^{-1}(r)\big][g]\big\|_{L^1(\mu)} \\
&\;\;=\; O\!\left(\|h\|_{L^2(\lambda)}^{2}
+ \|v_{r,k+g}-v_{r,k}\|_{L^2(\lambda)}^{2}
+ \|g\|_{\mathcal K}^{2}\right),
\end{align*}
with the big-$O$ constant depending on $\gamma$, $M_r$, $M_k$,
$\|r\|_{L^\infty(\lambda)}$, and
$\|\mathcal{T}_{r,k}^{-1}\|_{L^2(\lambda)\to L^2(\lambda)}$; moreover,
each inverse in the left-hand side exists for all sufficiently small
perturbations.
\end{lemma}
\begin{proof}
\noindent \textbf{Invertibility of the perturbed operators.}
We first show that $\mathcal{T}_{r+h,k}:L^2(\lambda)\to L^2(\lambda)$ and
$\mathcal{T}_{r+h,k+g}:L^2(\lambda)\to L^2(\lambda)$ are invertible whenever
the perturbation norm displayed in the lemma is sufficiently small.

Since $\mathcal{T}_{r,k}$ is invertible, it suffices to control the perturbations
\[
\Delta_1:=\mathcal{T}_{r+h,k}-\mathcal{T}_{r,k},
\qquad
\Delta_2:=\mathcal{T}_{r+h,k+g}-\mathcal{T}_{r,k}.
\]
By Lemma~\ref{lemma::forwardopLipschitz},
\begin{align*}
\|\Delta_1\|_{L^2(\lambda)\to L^2(\lambda)} &= O\!\big(\|h\|_{L^2(\lambda)}\big),\\
\|\Delta_2\|_{L^2(\lambda)\to L^2(\lambda)} &= O\!\Big(\|h\|_{L^2(\lambda)}+\|v_{r,k+g}-v_{r,k}\|_{L^2(\lambda)}+\|g\|_{\mathcal K}\Big).
\end{align*}
If $\|\mathcal{T}_{r,k}^{-1}\|_{L^2(\lambda)\to L^2(\lambda)}\,\|\Delta_i\|_{L^2(\lambda)\to L^2(\lambda)}<1$ for $i=1,2$,
then Lemma~\ref{lemma::inverseperturbation} shows that both $\mathcal{T}_{r+h,k}$ and
$\mathcal{T}_{r+h,k+g}$ are invertible, with
\begin{align*}
    \|\mathcal{T}_{r+h,k}^{-1}\|_{L^2(\lambda)\to L^2(\lambda)}
\le \frac{\|\mathcal{T}_{r,k}^{-1}\|_{L^2(\lambda)\to L^2(\lambda)}}{1-\|\mathcal{T}_{r,k}^{-1}\|_{L^2(\lambda)\to L^2(\lambda)}\,\|\Delta_1\|_{L^2(\lambda)\to L^2(\lambda)}},\\
\|\mathcal{T}_{r+h,k+g}^{-1}\|_{L^2(\lambda)\to L^2(\lambda)}
\le \frac{\|\mathcal{T}_{r,k}^{-1}\|_{L^2(\lambda)\to L^2(\lambda)}}{1-\|\mathcal{T}_{r,k}^{-1}\|_{L^2(\lambda)\to L^2(\lambda)}\,\|\Delta_2\|_{L^2(\lambda)\to L^2(\lambda)}}.
\end{align*}
For sufficiently small perturbations,
$\|\mathcal T_{r,k}^{-1}\|_{L^2\to L^2}
\|\Delta_i\|_{L^2\to L^2}<1$ for $i=1,2$.
Thus $\mathcal T_{r+h,k}$ and $\mathcal T_{r+h,k+g}$ are invertible, and
their inverse norms are locally bounded.

\noindent \textbf{Remainder bound.}
Lemma~\ref{lemma::valuestarderivativereward} gives
\[
\big\|\Pi_{r+h,k}\,\mathcal{T}_{r+h,k}^{-1}(r+h)
 - \Pi_{r,k}\,\mathcal{T}_{r,k}^{-1}(r)
 - \partial_r\!\big[\Pi_{r,k}\,\mathcal{T}_{r,k}^{-1}(r)\big][h]\big\|_{L^1(\mu)}
= O(\|h\|_{L^2(\lambda)}^2).
\]
Lemma~\ref{lemma::valuestarderivativekernel} gives
\[
\big\|\Pi_{r,k+g}\,\mathcal{T}_{r,k+g}^{-1}(r)
- \Pi_{r,k}\,\mathcal{T}_{r,k}^{-1}(r)
- \adk\!\big[\Pi_{r,k}\,\mathcal{T}_{r,k}^{-1}(r)\big][g]\big\|_{L^1(\mu)}
= O\!\left(\|v_{r,k+g} - v_{r,k}\|_{L^2(\lambda)}^{2} + \|g\|_{\mathcal K}^{2}\right).
\]
The fixed-k reward bound in Lemma~\ref{lemma::lipschitzvalue}, applied at
$k$ and $k+g$, also gives
\[
\|v_{r+h,k+g}-v_{r+h,k}\|_{L^2(\lambda)}
\leq \|v_{r,k+g}-v_{r,k}\|_{L^2(\lambda)}
+O\!\left(\|h\|_{L^2(\lambda)}\right).
\]
Thus the kernel remainder lemma may be applied at $(r+h,k)$ with the same
order after squaring. The uniform reward-envelope assumption bounds
\(\|r+h\|_{L^\infty(\lambda)}\), so the constant in
Lemma~\ref{lemma::valuestarderivativekernel} is uniform over this
neighborhood.
Therefore, adding and subtracting,
\begin{align*}
&\big\|\Pi_{r+h,k+g}\,\mathcal{T}_{r+h,k+g}^{-1}(r+h)
- \Pi_{r,k}\,\mathcal{T}_{r,k}^{-1}(r)
- \partial_r\!\big[\Pi_{r,k}\,\mathcal{T}_{r,k}^{-1}(r)\big][h]
- \adk\!\big[\Pi_{r,k}\,\mathcal{T}_{r,k}^{-1}(r)\big][g]\big\|_{L^1(\mu)} \\
&\;\;\le
\big\|\Pi_{r+h,k+g}\,\mathcal{T}_{r+h,k+g}^{-1}(r+h)
      - \Pi_{r+h,k}\,\mathcal{T}_{r+h,k}^{-1}(r+h)
      - \adk\!\big[\Pi_{r+h,k}\,\mathcal{T}_{r+h,k}^{-1}(r+h)\big][g]\big\|_{L^1(\mu)} \\
&\qquad +
\big\|\Pi_{r+h,k}\,\mathcal{T}_{r+h,k}^{-1}(r+h)
      - \Pi_{r,k}\,\mathcal{T}_{r,k}^{-1}(r)
      - \partial_r\!\big[\Pi_{r,k}\,\mathcal{T}_{r,k}^{-1}(r)\big][h]\big\|_{L^1(\mu)} \\
&\qquad +
\big\|\big(\adk\!\big[\Pi_{r+h,k}\,\mathcal{T}_{r+h,k}^{-1}(r+h)\big]
      - \adk\!\big[\Pi_{r,k}\,\mathcal{T}_{r,k}^{-1}(r)\big]\big)[g]\big\|_{L^1(\mu)}.
\end{align*}
Applying the lemmas, we obtain
\begin{align*}
&\big\|\Pi_{r+h,k+g}\,\mathcal{T}_{r+h,k+g}^{-1}(r+h)
- \Pi_{r,k}\,\mathcal{T}_{r,k}^{-1}(r)
- \partial_r\!\big[\Pi_{r,k}\,\mathcal{T}_{r,k}^{-1}(r)\big][h]
- \adk\!\big[\Pi_{r,k}\,\mathcal{T}_{r,k}^{-1}(r)\big][g]\big\|_{L^1(\mu)} \\
&\;\;\le
O\!\left(\|v_{r,k+g} - v_{r,k}\|_{L^2(\lambda)}^{2} + \|g\|_{\mathcal K}^{2}\right)
+ O\!\left(\|h\|_{L^2(\lambda)}^2\right) \\
&\qquad +
\big\|\big(\adk\!\big[\Pi_{r+h,k}\,\mathcal{T}_{r+h,k}^{-1}(r+h)\big]
      - \adk\!\big[\Pi_{r,k}\,\mathcal{T}_{r,k}^{-1}(r)\big]\big)[g]\big\|_{L^1(\mu)}.
\end{align*}
It remains to prove that
\begin{equation}
\big\|\big(\adk\![\Pi_{r+h,k}\mathcal{T}_{r+h,k}^{-1}(r{+}h)]
      - \adk\![\Pi_{r,k}\mathcal{T}_{r,k}^{-1}(r)]\big)[g]\big\|_{L^1(\mu)}
= O\!\left(\|h\|_{L^2(\lambda)}^2 + \|v_{r,k+g}-v_{r,k}\|_{L^2(\lambda)}^2 + \|g\|_{\mathcal K}^2\right).
\label{eqn::joint-derivative-mixed-remainder}
\end{equation}
Equation~\eqref{eqn::joint-derivative-mixed-remainder} therefore implies
\begin{align*}
&\big\|\Pi_{r+h,k+g}\,\mathcal{T}_{r+h,k+g}^{-1}(r+h)
- \Pi_{r,k}\,\mathcal{T}_{r,k}^{-1}(r)
- \partial_r\!\big[\Pi_{r,k}\,\mathcal{T}_{r,k}^{-1}(r)\big][h]
- \adk\!\big[\Pi_{r,k}\,\mathcal{T}_{r,k}^{-1}(r)\big][g]\big\|_{L^1(\mu)} \\
&\;\;=\; O\!\left(\|h\|_{L^2(\lambda)}^2 + \|v_{r,k+g}-v_{r,k}\|_{L^2(\lambda)}^2 + \|g\|_{\mathcal K}^2\right),
\end{align*}
which is the asserted remainder bound.

To prove \eqref{eqn::joint-derivative-mixed-remainder}, write
\(M(r,k)=\Pi_{r,k}\mathcal T_{r,k}^{-1}r\), and set
\[
A_0=\mathcal T_{r,k}^{-1},\quad
A_h=\mathcal T_{r+h,k}^{-1},\quad
P_0=\Pi_{r,k},\quad P_h=\Pi_{r+h,k},
\]
\[
B_0=\adk\Pi_{r,k}[g],\quad B_h=\adk\Pi_{r+h,k}[g],\quad
C_0=\adk\mathcal T_{r,k}^{-1}[g],\quad
C_h=\adk\mathcal T_{r+h,k}^{-1}[g].
\]
The product rule gives
\begin{equation}
\adk M(r,k)[g]=B_0A_0r+P_0C_0r,
\label{eqn::joint-kernel-product-rule}
\end{equation}
and
\[
\adk M(r+h,k)[g]=B_hA_h(r+h)+P_hC_h(r+h).
\]
Adding and subtracting intermediate products yields the exact decomposition
\begin{align}
&\adk M(r+h,k)[g]-\adk M(r,k)[g]\notag\\
&\quad=(B_h-B_0)A_0r+B_h(A_h-A_0)r
+(P_h-P_0)C_0r+P_h(C_h-C_0)r\notag\\
&\qquad\quad +(B_hA_h+P_hC_h)h.
\label{eqn::joint-five-product-decomposition}
\end{align}
Lemma~\ref{lemma::mixedoperatorbounds}, with the operators defined above,
gives \eqref{eqn::joint-first-order-operator-bounds},
\eqref{eqn::joint-mixed-operator-bounds}, and
\eqref{eqn::joint-undifferenced-operator-bounds}.
Apply \eqref{eqn::joint-mixed-operator-bounds} to the first and fourth
products in \eqref{eqn::joint-five-product-decomposition},
\eqref{eqn::joint-first-order-operator-bounds} and
\eqref{eqn::joint-undifferenced-operator-bounds} to the second and third,
and \eqref{eqn::joint-undifferenced-operator-bounds} to the last.  The local
bounds for \(A_0,A_h,P_0,P_h\) then yield
\begin{equation}
\|\adk M(r+h,k)[g]-\adk M(r,k)[g]\|_{L^1(\mu)}
=O\{\|h\|_{L^2(\lambda)}\|g\|_{\mathcal K}\}.
\label{eqn::joint-mixed-derivative-bound}
\end{equation}
Young's inequality applied to
\eqref{eqn::joint-mixed-derivative-bound} proves
\eqref{eqn::joint-derivative-mixed-remainder} and completes the proof.

\end{proof}

\begin{lemma}[Second-order partial Fr\'echet differentiability of $r \mapsto v_{r,k}$]
\label{lemma::valuederivsecond}
Suppose $\gamma < 1$.
Then the map $r \mapsto \partial_r v_{r,k} = \mathcal{T}_{r,k}^{-1}\mathcal{P}_{r,k}$
is Fr\'echet differentiable from $L^\infty(\lambda)$ to
$\mathcal L\{L^\infty(\lambda),L^\infty(\lambda)\}$, with derivative
\[
\partial_r^2 v_{r,k}[h_1,h_2]
:= \left\{\partial_r\!\big[\mathcal{T}_{r,k}^{-1}\mathcal{P}_{r,k}\big][h_2]\right\}(h_1)
= \frac{1}{\tau}\,\mathcal{T}_{r,k}^{-1}\,\mathcal{P}_{r,k}\!\left[
\operatorname{Cov}_{\pi_{r,k}}\!\Big(\mathcal{T}_{r,k}^{-1}h_1,\, \mathcal{T}_{r,k}^{-1}h_2
\Big)\right].
\]
In particular, the following quadratic expansion holds:
\begin{align*}
 \|\partial_{r} v_{r+h,k} - \partial_{r} v_{r,k} - \partial_r^2 v_{r,k}[\cdot, h]\|_{L^\infty \to L^\infty} \;\lesssim\; \|h\|_{L^2(\lambda)}^2 .
\end{align*}
\end{lemma}
\begin{proof}
By the chain rule, \(r\mapsto\mathcal T_{r,k}\) is G\^ateaux differentiable,
with
\[
\partial_r \mathcal{T}_{r,k}[h] \;=\; -\,\gamma\,\partial_r \mathcal{P}_{r,k}[h],
\qquad\text{i.e.}\quad
\big(\partial_r \mathcal{T}_{r,k}[h]\big)(f)
= -\,\gamma\,\big(\partial_r \mathcal{P}_{r,k}[h]\big)(f).
\]
The derivative formula in Lemma~\ref{lemma::inverseperturbation} gives
\[
\partial_r \mathcal{T}_{r,k}^{-1}[h]
= -\,\mathcal{T}_{r,k}^{-1}\,\big(\partial_r \mathcal{T}_{r,k}[h]\big)\,\mathcal{T}_{r,k}^{-1}.
\]
The product rule then gives
\[
\partial_r\!\big[\mathcal{T}_{r,k}^{-1}\mathcal{P}_{r,k}\big][h]
= - \mathcal{T}_{r,k}^{-1}\,\big(\partial_r \mathcal{T}_{r,k}[h]\big)\,\mathcal{T}_{r,k}^{-1}\,\mathcal{P}_{r,k}
\;+\;
\mathcal{T}_{r,k}^{-1}\,\big(\partial_r \mathcal{P}_{r,k}[h]\big).
\]
Applying this identity in directions $h_1,h_2$ yields
\begin{align*}
\partial_r^2 v_{r,k}[h_1,h_2]
&= -\,\mathcal{T}_{r,k}^{-1}\,(\partial_r \mathcal{T}_{r,k}[h_2])\,\mathcal{T}_{r,k}^{-1}\,\mathcal{P}_{r,k}[h_1]
\;+\;
\mathcal{T}_{r,k}^{-1}\,(\partial_r \mathcal{P}_{r,k}[h_2])[h_1] \\
&= \gamma\,\mathcal{T}_{r,k}^{-1}\,(\partial_r \mathcal{P}_{r,k}[h_2])\,\mathcal{T}_{r,k}^{-1}\,\mathcal{P}_{r,k}[h_1]
\;+\;
\mathcal{T}_{r,k}^{-1}\,(\partial_r \mathcal{P}_{r,k}[h_2])[h_1] \\
&= \mathcal{T}_{r,k}^{-1}\,(\partial_r \mathcal{P}_{r,k}[h_2])\,
\big(I+\gamma\,\mathcal{T}_{r,k}^{-1}\mathcal{P}_{r,k}\big)[h_1]\\
&= \mathcal{T}_{r,k}^{-1}\,(\partial_r \mathcal{P}_{r,k}[h_2])\,
\big(I+\gamma\,\partial_r v_{r,k}\big)[h_1]\\
&= \frac{1}{\tau}\,\mathcal{T}_{r,k}^{-1}\!\Big[(a,s)\mapsto
\int \operatorname{Cov}_{\pi_{r,k}(\cdot\mid s')}\!\Big(
\partial_r q_{r,k}[h_1](\cdot,s'),\; \partial_r q_{r,k}[h_2](\cdot,s')
\Big)\,k(s'\mid a,s) \mu(ds')\Big],
\end{align*}
where
$\partial_r q_{r,k}[h]:=h+\gamma\,\partial_r v_{r,k}[h]$. Equivalently,
\[
\partial_r^2 v_{r,k}[h_1,h_2]
= \frac{1}{\tau}\,\mathcal{T}_{r,k}^{-1}\,\mathcal{P}_{r,k}\!\left[(a',s')\mapsto
\operatorname{Cov}_{\pi_{r,k}(\cdot\mid s')}\!\Big(
\partial_r q_{r,k}[h_1](\cdot,s'),\; \partial_r q_{r,k}[h_2](\cdot,s')
\Big)\right].
\]
Lemma~\ref{lemma::qderiv} gives
$\partial_r q_{r,k}[h]=\mathcal T_{r,k}^{-1}h$, and therefore
\[
\partial_r^2 v_{r,k}[h_1,h_2]
= \frac{1}{\tau}\,\mathcal{T}_{r,k}^{-1}\,\mathcal{P}_{r,k}\!\left[
\operatorname{Cov}_{\pi_{r,k}}\!\Big(\mathcal{T}_{r,k}^{-1}h_1,\, \mathcal{T}_{r,k}^{-1}h_2
\Big)\right].
\]

We next bound the remainder in the first-order expansion of
$r\mapsto\mathcal T_{r,k}^{-1}\mathcal P_{r,k}$, namely,
$\mathcal{T}_{r+h,k}^{-1}\mathcal{P}_{r+h,k}
 - \mathcal{T}_{r,k}^{-1}\mathcal{P}_{r,k}
 - \partial_r\!\big[\mathcal{T}_{r,k}^{-1}\mathcal{P}_{r,k}\big][h]$.
Adding and subtracting the derivative terms gives the exact decomposition
\begin{align*}
&\mathcal{T}_{r+ h,k}^{-1}\mathcal{P}_{r+ h,k}
 - \mathcal{T}_{r,k}^{-1}\mathcal{P}_{r,k}
 - \partial_r\!\big[\mathcal{T}_{r,k}^{-1}\mathcal{P}_{r,k}\big][h] \\
&= \underbrace{\Big[\mathcal{T}_{r+h,k}^{-1} - \mathcal{T}_{r,k}^{-1}
         + \mathcal{T}_{r,k}^{-1}\,\big(\partial_r \mathcal{T}_{r,k}[h]\big)\,
           \mathcal{T}_{r,k}^{-1}\Big]\mathcal{P}_{r,k}}_{\text{Inverse-operator remainder}} \\
&\quad + \underbrace{\mathcal{T}_{r,k}^{-1}\Big[\mathcal{P}_{r+h,k} - \mathcal{P}_{r,k}
           - \partial_r \mathcal{P}_{r,k}[h]\Big]}_{\text{Direct $\mathcal{P}$ remainder}} \\
&\quad + \underbrace{\Big(\mathcal{T}_{r+h,k}^{-1} - \mathcal{T}_{r,k}^{-1}\Big)
       \Big(\mathcal{P}_{r+h,k} - \mathcal{P}_{r,k}\Big)}_{\text{Mixed interaction term}}.
\end{align*}

We bound the direct, mixed, and inverse-operator remainders in turn. Since
$\gamma<1$, the Bellman operators $\mathcal T_{r+h,k}$ and
$\mathcal T_{r,k}$ have bounded inverses on $L^\infty(\lambda)$.

\noindent \textbf{Direct $\mathcal P$ remainder.}
Lemma~\ref{lemma::operatorderivatives} gives
\begin{align*}
\big\|\mathcal{T}_{r,k}^{-1}\big[\mathcal{P}_{r+h,k} - \mathcal{P}_{r,k}
           - \partial_r \mathcal{P}_{r,k}[h]\big]\big\|_{L^\infty \to L^\infty}
&\leq \|\mathcal{T}_{r,k}^{-1}\|_{L^\infty \to L^\infty}\;
   \|\mathcal{P}_{r+h,k} - \mathcal{P}_{r,k}
           - \partial_r \mathcal{P}_{r,k}[h]\|_{L^\infty \to L^\infty} \\
&\lesssim
   \|h\|_{L^2(\lambda)}^2,
\end{align*}
where we used that $ \|\mathcal{T}_{r,k}^{-1}\|_{L^\infty \to L^\infty} < \infty$ since $\gamma < 1$.

\noindent \textbf{Mixed interaction term.}
Submultiplicativity gives
\begin{align*}
   \big\|\big(\mathcal{T}_{r+h,k}^{-1} - \mathcal{T}_{r,k}^{-1}\big)
       \big(\mathcal{P}_{r+h,k} - \mathcal{P}_{r,k}\big)\big\|_{L^\infty \to L^\infty}
   &\leq  \|\mathcal{T}_{r+h,k}^{-1} - \mathcal{T}_{r,k}^{-1} \|_{L^\infty \to L^\infty}\;
       \|\mathcal{P}_{r+h,k} - \mathcal{P}_{r,k}\|_{L^\infty \to L^\infty}.
\end{align*}
Moreover,
\[
\|\mathcal{T}_{r+h,k}^{-1} - \mathcal{T}_{r,k}^{-1}\|_{L^\infty \to L^\infty}
\;\le\; \|\mathcal{T}_{r,k}^{-1}\|_{L^\infty \to L^\infty}\,
        \|\mathcal{T}_{r+h,k} - \mathcal{T}_{r,k}\|_{L^\infty \to L^\infty}\,
        \|\mathcal{T}_{r+h,k}^{-1}\|_{L^\infty \to L^\infty}.
\]
Furthermore, Lemma~\ref{lemma::operatorderivatives} gives
\begin{align*}
    \big\| \mathcal{P}_{r + h, k}  - \mathcal{P}_{r, k}
     \big\|_{L^\infty \to L^\infty}
     \;\lesssim\; \|\partial_r \mathcal{P}_{r, k}[h]\|_{L^\infty \to L^\infty}
     + \|h\|_{L^2(\lambda)}^2.
\end{align*}
Moreover, $\|\partial_r \mathcal{P}_{r, k}[h]\|_{L^\infty \to L^\infty} \lesssim \|h\|_{L^2(\lambda)}$. Applying this bound, we obtain
\[
\|\mathcal{P}_{r+h,k} - \mathcal{P}_{r,k}\|_{L^\infty \to L^\infty}
\;\lesssim\; \|h\|_{L^2(\lambda)},
\qquad
\|\mathcal{T}_{r+h,k} - \mathcal{T}_{r,k}\|_{L^\infty \to L^\infty}
\;\lesssim\; \|h\|_{L^2(\lambda)}.
\]
Hence,
\begin{align*}
   \big\|\big(\mathcal{T}_{r+h,k}^{-1} - \mathcal{T}_{r,k}^{-1}\big)
       \big(\mathcal{P}_{r+h,k} - \mathcal{P}_{r,k}\big)\big\|_{L^\infty \to L^\infty}
   &\;\lesssim\; \big(1 + \|\mathcal{T}_{r,k}^{-1}\|_{L^\infty \to L^\infty}\big)\,
      \|h\|_{L^2(\lambda)}^2\\
       &\;\lesssim \|h\|_{L^2(\lambda)}^2.
\end{align*}

\noindent \textbf{Inverse-operator remainder.}
Equation~\eqref{eqn::generalized-resolvent-identity}, with
\(L=\partial_r\mathcal T_{r,k}[h]\), gives
\[
\begin{aligned}
&\mathcal{T}_{r+h,k}^{-1} - \mathcal{T}_{r,k}^{-1}
  + \mathcal{T}_{r,k}^{-1}\big(\partial_r \mathcal{T}_{r,k}[h]\big)\mathcal{T}_{r,k}^{-1} \\
&\qquad=-  \mathcal{T}_{r,k}^{-1}\Big( \mathcal{T}_{r+h,k} - \mathcal{T}_{r,k}  - \partial_r \mathcal{T}_{r,k}[h] \Big)\mathcal{T}_{r,k}^{-1} \\
&\qquad\quad +\; \mathcal{T}_{r,k}^{-1}\big(\mathcal{T}_{r+h,k} - \mathcal{T}_{r,k}\big)\mathcal{T}_{r,k}^{-1}\big(\mathcal{T}_{r+h,k} - \mathcal{T}_{r,k}\big)\mathcal{T}_{r+h,k}^{-1}.
\end{aligned}
\]
Taking operator norms yields
\[
\begin{aligned}
& \big\|\mathcal{T}_{r+h,k}^{-1} - \mathcal{T}_{r,k}^{-1}
+ \mathcal{T}_{r,k}^{-1}(\partial_r \mathcal{T}_{r,k}[h])\mathcal{T}_{r,k}^{-1}\big\|_{L^\infty \to L^\infty} \\
&\;\;\le  \|\mathcal{T}_{r,k}^{-1}\|_{L^\infty \to L^\infty}^{2}\,
   \big\|\,\mathcal{T}_{r+h,k} - \mathcal{T}_{r,k} - \partial_r \mathcal{T}_{r,k}[h]\,\big\|_{L^\infty \to L^\infty}  \\
&\qquad + \|\mathcal{T}_{r,k}^{-1}\|_{L^\infty \to L^\infty}\,
   \|\mathcal{T}_{r+h,k} - \mathcal{T}_{r,k}\|_{L^\infty \to L^\infty}\,
   \|\mathcal{T}_{r,k}^{-1}\|_{L^\infty \to L^\infty}\,
   \|\mathcal{T}_{r+h,k} - \mathcal{T}_{r,k}\|_{L^\infty \to L^\infty}\,
   \|\mathcal{T}_{r+h,k}^{-1}\|_{L^\infty \to L^\infty} \\
&\;\;\lesssim \|\mathcal{T}_{r,k}^{-1}\|_{L^\infty \to L^\infty}^{2}\,
   \big\|\,\mathcal{T}_{r+h,k} - \mathcal{T}_{r,k} - \partial_r \mathcal{T}_{r,k}[h]\,\big\|_{L^\infty \to L^\infty}
   \;+\; \|\mathcal{T}_{r,k}^{-1}\|_{L^\infty \to L^\infty}^2\,
   \|\mathcal{T}_{r+h,k}^{-1}\|_{L^\infty \to L^\infty}\,\|h\|_{L^2(\lambda)}^2 .
\end{aligned}
\]
The remainder bound in Lemma~\ref{lemma::operatorderivatives} implies
\[
\big\|\mathcal T_{r+h,k}-\mathcal T_{r,k}
-\partial_r\mathcal T_{r,k}[h]\big\|_{L^\infty\to L^\infty}
\lesssim\|h\|_{L^2(\lambda)}^2.
\]
Consequently,
\[
\begin{aligned}
& \big\|\mathcal{T}_{r+h,k}^{-1} - \mathcal{T}_{r,k}^{-1}
+ \mathcal{T}_{r,k}^{-1}(\partial_r \mathcal{T}_{r,k}[h])\mathcal{T}_{r,k}^{-1}\big\|_{L^\infty \to L^\infty} \lesssim   \|h\|_{L^2(\lambda)}^2.
\end{aligned}
\]

Combining the three bounds and using the local inverse bound from
Lemma~\ref{lemma::inverseperturbation} and
\eqref{eqn::inverse-perturbation-bound} gives
 \begin{align*}
 \|\partial_{r} v_{r + h,k} - \partial_{r} v_{r,k} - \partial_r^2 v_{r, k}[\cdot, h]  \|_{L^\infty \rightarrow L^\infty}  &= \|\mathcal{T}_{r+ h,k}^{-1}\mathcal{P}_{r+ h,k}
 - \mathcal{T}_{r,k}^{-1}\mathcal{P}_{r,k}
 - \partial_r\!\big[\mathcal{T}_{r,k}^{-1}\mathcal{P}_{r,k}\big][h]\|_{L^\infty \rightarrow L^\infty}\\
 & \lesssim \|h\|_{L^2(\lambda)}^2 .
 \end{align*}
This proves the stated operator-norm remainder bound.
\end{proof}

\begin{lemma}[Mixed reward--kernel derivative of the soft value function]
\label{lemma::valuederivsecond::kernel}
Suppose Assumption~\ref{cond::boundedpositivity} holds, and let
$(r,k)$ and $(r,k')$ lie in the neighborhood $\mathcal U$ from
Lemma~\ref{lemma::lipschitzvalue}, on which
$\mathcal T_{r,k'}^{-1}:L^2(\lambda)\to L^2(\lambda)$ exists and is
uniformly bounded. Then $k\mapsto\partial_r v_{r,k}$ is Fr\'echet
differentiable into
$\mathcal L\{L^\infty(\lambda),L^2(\lambda)\}$, with derivative
\[
\begin{aligned}
\adk\partial_r v_{r,k}\big[\,h,\,k'-k\,\big]
&= \mathcal{T}_{r,k}^{-1}\Bigg[
\frac{1}{\tau}\,\mathcal{P}_{r,k}\!\left[(a',s')\mapsto
\operatorname{Cov}_{\pi_{r,k}(\cdot\mid s')}\!\Big(
\partial_r q_{r,k}[h](\cdot,s'),\; \adk q_{r,k}[\,k'-k\,](\cdot,s')
\Big)\right] \\
&\qquad\qquad
+ \mathcal{P}_{k'-k}\,\Pi_{r,k}\,\partial_r q_{r,k}[h]
\Bigg].
\end{aligned}
\]
Here $\partial_r q_{r,k}[h](\cdot,s') := \mathcal{T}_{r,k}^{-1}h(\cdot,s')$ and $\adk q_{r,k}[k'-k](\cdot,s') := \gamma\,\adk v_{r,k}[k'-k](\cdot,s')$.
In particular, the following expansion holds:
\begin{align*}
 \|\partial_r v_{r,k'} - \partial_r v_{r,k} -  \adk \partial_r v_{r,k}[\cdot, k' - k]  \|_{L^\infty \rightarrow L^2}
 &\lesssim  \|k'-k\|_{L^2(\mu\otimes\lambda)}^2
      + \|v_{r,k'}-v_{r,k}\|_{L^2(\lambda)}^2
 \lesssim \|k'-k\|_{\mathcal K}^2.
\end{align*}

\end{lemma}
\begin{proof}
By the chain rule, \(k \mapsto \mathcal{T}_{r,k}\) is G\^ateaux differentiable with
\[
\adk \mathcal{T}_{r,k}[k'-k] \;=\; -\,\gamma\,\adk \mathcal{P}_{r,k}[k'-k],
\qquad\text{i.e.}\quad
\big(\adk \mathcal{T}_{r,k}[k'-k]\big)(f)
= -\,\gamma\,\big(\adk \mathcal{P}_{r,k}[k'-k]\big)(f).
\]
Applying the derivative formula in
Lemma~\ref{lemma::inverseperturbation},
\[
\adk \mathcal{T}_{r,k}^{-1}[k'-k]
= -\,\mathcal{T}_{r,k}^{-1}\,\big(\adk \mathcal{T}_{r,k}[k'-k]\big)\,\mathcal{T}_{r,k}^{-1}.
\]
By the product rule and differentiability of \(k\mapsto \mathcal{T}_{r,k}\) and \(k\mapsto \mathcal{P}_{r,k}\),
\[
\adk\!\big[\mathcal{T}_{r,k}^{-1}\mathcal{P}_{r,k}\big][k'-k]
= -\,\mathcal{T}_{r,k}^{-1}\,\big(\adk \mathcal{T}_{r,k}[k'-k]\big)\,\mathcal{T}_{r,k}^{-1}\,\mathcal{P}_{r,k}
\;+\;
\mathcal{T}_{r,k}^{-1}\,\big(\adk \mathcal{P}_{r,k}[k'-k]\big).
\]
Hence, for directions $h\in L^\infty(\lambda)$ and
$k' - k$, the mixed partial satisfies
\begin{align*}
\adk\partial_r v_{r,k}[\,h,\,k' - k\,]
&= -\,\mathcal{T}_{r,k}^{-1}\,\big(\adk \mathcal{T}_{r,k}[k' - k]\big)\,
     \mathcal{T}_{r,k}^{-1}\,\mathcal{P}_{r,k}[h]
 \;+\; \mathcal{T}_{r,k}^{-1}\,\big(\adk \mathcal{P}_{r,k}[k' - k]\big)[h] \\
&= \gamma\,\mathcal{T}_{r,k}^{-1}\,\big(\adk \mathcal{P}_{r,k}[k' - k]\big)\,
      \mathcal{T}_{r,k}^{-1}\,\mathcal{P}_{r,k}[h]
 \;+\; \mathcal{T}_{r,k}^{-1}\,\big(\adk \mathcal{P}_{r,k}[k' - k]\big)[h] \\
&= \mathcal{T}_{r,k}^{-1}\,\big(\adk \mathcal{P}_{r,k}[k' - k]\big)\,
   \big(I+\gamma\,\mathcal{T}_{r,k}^{-1}\mathcal{P}_{r,k}\big)[h] \\
&= \mathcal{T}_{r,k}^{-1}\,\big(\adk \mathcal{P}_{r,k}[k' - k]\big)\,
   \big(I+\gamma\,\partial_r v_{r,k}\big)[h].
\end{align*}
Here we used $\adk \mathcal{T}_{r,k}[k'-k]
=-\gamma\adk \mathcal{P}_{r,k}[k'-k]$ and
$I+\gamma\mathcal{T}_{r,k}^{-1}\mathcal{P}_{r,k}
=I+\gamma\partial_r v_{r,k}$. Substituting the operator derivatives from
Lemma~\ref{lemma::operatorderivatives} gives
\[
\begin{aligned}
\adk\partial_r v_{r,k}\big[\,h,\,k'-k\,\big]
&= \mathcal{T}_{r,k}^{-1}\Bigg[
\frac{1}{\tau}\,\mathcal{P}_{r,k}\!\left[(a',s')\mapsto
\operatorname{Cov}_{\pi_{r,k}(\cdot\mid s')}\!\Big(
\partial_r q_{r,k}[h](\cdot,s'),\; \adk q_{r,k}[\,k'-k\,](\cdot,s')
\Big)\right] \\
&\qquad\qquad
+ \mathcal{P}_{k'-k}\,\Pi_{r,k}\,\partial_r q_{r,k}[h]
\Bigg].
\end{aligned}
\]
Here $\adk q_{r,k}[k' - k](\cdot,s') := \gamma\,\adk v_{r,k}[k' - k](\cdot,s')$ and
$\pi_{r,k}(\cdot\mid s')$ denotes the softmax distribution induced by $q_{r,k}(\cdot,s')$.

\noindent \textbf{Quadratic remainder.}
Adding and subtracting the three first-order products gives the exact
decomposition
\begin{align*}
&\mathcal{T}_{r,k'}^{-1}\mathcal{P}_{r,k'}
 - \mathcal{T}_{r,k}^{-1}\mathcal{P}_{r,k}
 - \adk\!\big[\mathcal{T}_{r,k}^{-1}\mathcal{P}_{r,k}\big][\,k' - k\,] \\
&= \underbrace{\Big[\mathcal{T}_{r,k'}^{-1} - \mathcal{T}_{r,k}^{-1}
         + \mathcal{T}_{r,k}^{-1}\,\big(\adk \mathcal{T}_{r,k}[\,k' - k\,]\big)\,
           \mathcal{T}_{r,k}^{-1}\Big]\mathcal{P}_{r,k}}_{\text{Inverse-operator remainder}} \\
&\quad + \underbrace{\mathcal{T}_{r,k}^{-1}\Big[\mathcal{P}_{r,k'} - \mathcal{P}_{r,k}
           - \adk \mathcal{P}_{r,k}[\,k' - k\,]\Big]}_{\text{Direct $\mathcal{P}$ remainder}} \\
&\quad + \underbrace{\Big(\mathcal{T}_{r,k'}^{-1} - \mathcal{T}_{r,k}^{-1}\Big)
       \Big(\mathcal{P}_{r,k'} - \mathcal{P}_{r,k}\Big)}_{\text{Mixed interaction term}}.
\end{align*}
For the inverse-operator remainder,
Lemma~\ref{lemma::inverseperturbation}, specifically
\eqref{eqn::generalized-resolvent-identity} with
\(A=\mathcal T_{r,k}\),
\(\Delta=\mathcal T_{r,k'}-\mathcal T_{r,k}\), and
\(L=\adk\mathcal T_{r,k}[k'-k]\), gives
\[
\begin{aligned}
&\mathcal{T}_{r,k'}^{-1} - \mathcal{T}_{r,k}^{-1}
  + \mathcal{T}_{r,k}^{-1}\big(\adk \mathcal{T}_{r,k}[\,k' - k\,]\big)\mathcal{T}_{r,k}^{-1} \\
&\qquad= - \mathcal{T}_{r,k}^{-1}\Big(\mathcal{T}_{r,k'} - \mathcal{T}_{r,k}  - \adk \mathcal{T}_{r,k}[\,k' - k\,]\Big) \mathcal{T}_{r,k}^{-1} \\
&\qquad\quad +\; \mathcal{T}_{r,k}^{-1}\big(\mathcal{T}_{r,k'} - \mathcal{T}_{r,k}\big)\mathcal{T}_{r,k}^{-1}\big(\mathcal{T}_{r,k'} - \mathcal{T}_{r,k}\big)\mathcal{T}_{r,k'}^{-1}.
\end{aligned}
\]
Substitution gives
\begin{align*}
&\mathcal{T}_{r,k'}^{-1}\mathcal{P}_{r,k'}
 - \mathcal{T}_{r,k}^{-1}\mathcal{P}_{r,k}
 - \adk\!\big[\mathcal{T}_{r,k}^{-1}\mathcal{P}_{r,k}\big][\,k' - k\,] \\
&=
\underbrace{- \mathcal{T}_{r,k}^{-1}\Big(\mathcal{T}_{r,k'} - \mathcal{T}_{r,k}  - \adk \mathcal{T}_{r,k}[\,k' - k\,]\Big) \mathcal{T}_{r,k}^{-1}\mathcal P_{r,k} }_{\text{Inverse linear remainder}} \\
&\qquad+\;
\underbrace{\mathcal{T}_{r,k}^{-1}\big(\mathcal{T}_{r,k'} - \mathcal{T}_{r,k}\big)\mathcal{T}_{r,k}^{-1}\big(\mathcal{T}_{r,k'} - \mathcal{T}_{r,k}\big)\mathcal{T}_{r,k'}^{-1}\,\mathcal{P}_{r,k}}_{\text{Inverse quadratic product}} \\
&\qquad+\;
\underbrace{\mathcal{T}_{r,k}^{-1}\Big[\mathcal{P}_{r,k'} - \mathcal{P}_{r,k}
           - \adk \mathcal{P}_{r,k}[\,k' - k\,]\Big]}_{\text{Direct $\mathcal{P}$ remainder}} \\
&\qquad+\;
\underbrace{\Big(\mathcal{T}_{r,k'}^{-1} - \mathcal{T}_{r,k}^{-1}\Big)
       \Big(\mathcal{P}_{r,k'} - \mathcal{P}_{r,k}\Big)}_{\text{Mixed interaction term}}.
\end{align*}

\noindent \textbf{Inverse linear remainder.}
We have
\begin{align*}
  &\big\| \mathcal{T}_{r,k}^{-1}\Big(\mathcal{T}_{r,k'} - \mathcal{T}_{r,k}
      - \adk \mathcal{T}_{r,k}[\,k' - k\,]\Big)\mathcal{T}_{r,k}^{-1}\mathcal P_{r,k}
   \big\|_{L^\infty \rightarrow L^2}  \\
  &\leq \|\mathcal{T}_{r,k}^{-1}\|_{L^2 \rightarrow L^2}\;
       \big\|\mathcal{T}_{r,k'} - \mathcal{T}_{r,k}
          - \adk \mathcal{T}_{r,k}[\,k' - k\,]\big\|_{L^\infty \rightarrow L^2}\;
       \|\mathcal{T}_{r,k}^{-1}\|_{L^\infty \rightarrow L^\infty}
       \|\mathcal P_{r,k}\|_{L^\infty\to L^\infty} \\
  &\lesssim \big\|\mathcal{T}_{r,k'} - \mathcal{T}_{r,k}
          - \adk \mathcal{T}_{r,k}[\,k' - k\,]\big\|_{L^\infty \rightarrow L^2}.
\end{align*}
Here we used that $\|\mathcal{T}_{r,k}^{-1}\|_{L^2 \rightarrow L^2} < \infty$ and absorbed its dependence into the notation $\lesssim$. By Lemma~\ref{lemma::operatorderivatives}, we thus obtain
\begin{align*}
  &\big\| \mathcal{T}_{r,k}^{-1}\Big(\mathcal{T}_{r,k'} - \mathcal{T}_{r,k}
      - \adk \mathcal{T}_{r,k}[\,k' - k\,]\Big)\mathcal{T}_{r,k}^{-1}\mathcal P_{r,k}
   \big\|_{L^\infty \rightarrow L^2} \\
  &\lesssim \|v_{r,k'} - v_{r,k}\|_{L^2(\lambda)}^2
          + \|v_{r,k'} - v_{r,k}\|_{L^2(\lambda)}\,
            \|k' - k\|_{L^2(\mu \otimes \lambda)}.
\end{align*}

\noindent \textbf{Inverse quadratic product.}
Lemma~\ref{lemma::forwardopLipschitz} and
$\mathcal T_{r,k}=I-\gamma\mathcal P_{r,k}$ give
\[
\|\mathcal T_{r,k'}-\mathcal T_{r,k}\|_{L^2(\lambda)\to L^2(\lambda)}
\lesssim \|k'-k\|_{\mathcal K}
+\|v_{r,k'}-v_{r,k}\|_{L^2(\lambda)}.
\]
Submultiplicativity of operator norms therefore gives
\begin{align*}
&\big\|\mathcal{T}_{r,k}^{-1}\big(\mathcal{T}_{r,k'}-\mathcal{T}_{r,k}\big)
  \mathcal{T}_{r,k}^{-1}\big(\mathcal{T}_{r,k'}-\mathcal{T}_{r,k}\big)
  \mathcal{T}_{r,k'}^{-1}\,\mathcal{P}_{r,k}\big\|_{L^2 \rightarrow L^2} \\
&\qquad\le
\|\mathcal{T}_{r,k}^{-1}\|_{L^2(\lambda)\to L^2(\lambda)}^2\,
\|\mathcal{T}_{r,k'}^{-1}\|_{L^2(\lambda)\to L^2(\lambda)}\,
\|\mathcal{P}_{r,k}\|_{L^2(\lambda)\to L^2(\lambda)}\,
\|\mathcal{T}_{r,k'}-\mathcal{T}_{r,k}\|_{L^2(\lambda)\to L^2(\lambda)}^2 \\
&\qquad\lesssim
\Big(\|k'-k\|_{L^2(\mu\otimes\lambda)}
      + \|v_{r,k'}-v_{r,k}\|_{L^2(\lambda)}\Big)^2\\
      &\qquad\lesssim
\|k'-k\|_{L^2(\mu\otimes\lambda)}^2
      + \|v_{r,k'}-v_{r,k}\|_{L^2(\lambda)}^2.
\end{align*}
Here the $\lesssim$ notation absorbs
$\|\mathcal{T}_{r,k}^{-1}\|_{L^2(\lambda)\to L^2(\lambda)}$,
$\|\mathcal{T}_{r,k'}^{-1}\|_{L^2(\lambda)\to L^2(\lambda)}$, and
$\|\mathcal{P}_{r,k}\|_{L^2(\lambda)\to L^2(\lambda)}$.

\noindent \textbf{Direct operator remainder.}
Submultiplicativity and the kernel remainder bound in
Lemma~\ref{lemma::operatorderivatives} give
\begin{align*}
  &\big\| \mathcal{T}_{r,k}^{-1}\Big(\mathcal{P}_{r,k'} - \mathcal{P}_{r,k}
      - \adk \mathcal{P}_{r,k}[\,k' - k\,]\Big)
   \big\|_{L^\infty \rightarrow L^2} \\
  &\le \|\mathcal T_{r,k}^{-1}\|_{L^2\to L^2}
  \big\|\mathcal{P}_{r,k'} - \mathcal{P}_{r,k}
      - \adk \mathcal{P}_{r,k}[\,k' - k\,]\big\|_{L^\infty\to L^2}\\
  &\lesssim \|v_{r,k'} - v_{r,k}\|_{L^2(\lambda)}^2
          + \|v_{r,k'} - v_{r,k}\|_{L^2(\lambda)}\,
            \|k' - k\|_{L^2(\mu \otimes \lambda)}.
\end{align*}

\noindent \textbf{Mixed interaction term.}
We have
\begin{align*}
   \big\|\big(\mathcal{T}_{r,k'}^{-1} - \mathcal{T}_{r,k}^{-1}\big)
       \big(\mathcal{P}_{r,k'} - \mathcal{P}_{r,k}\big)\big\|_{L^\infty \to L^2}
   &\lesssim  \|\mathcal{T}_{r,k'}^{-1} - \mathcal{T}_{r,k}^{-1} \|_{L^2(\lambda) \to L^2(\lambda)}\;
       \|\mathcal{P}_{r,k'} - \mathcal{P}_{r,k}\|_{L^\infty \to L^2}.
\end{align*}
Equation~\eqref{eqn::first-resolvent-identity} gives
\[
\begin{aligned}
\|\mathcal{T}_{r,k'}^{-1} - \mathcal{T}_{r,k}^{-1}\|_{L^2(\lambda) \to L^2(\lambda)}
&= \|\mathcal{T}_{r,k}^{-1}\big(\mathcal{T}_{r,k} - \mathcal{T}_{r,k'}\big)\mathcal{T}_{r,k'}^{-1}\|_{L^2(\lambda) \to L^2(\lambda)} \\
&\le \|\mathcal{T}_{r,k}^{-1}\|_{L^2(\lambda) \to L^2(\lambda)}\,
\|\mathcal{T}_{r,k'} - \mathcal{T}_{r,k}\|_{L^2(\lambda) \to L^2(\lambda)}\,
\|\mathcal{T}_{r,k'}^{-1}\|_{L^2(\lambda) \to L^2(\lambda)}.
\end{aligned}
\]
Hence,
\begin{align*}
   \big\|\big(\mathcal{T}_{r,k'}^{-1} - \mathcal{T}_{r,k}^{-1}\big)
       \big(\mathcal{P}_{r,k'} - \mathcal{P}_{r,k}\big)\big\|_{L^\infty \to L^2}
   &\lesssim  \|\mathcal{T}_{r,k}^{-1}\|_{L^2(\lambda) \to L^2(\lambda)}
       \|\mathcal{T}_{r,k'}^{-1}\|_{L^2(\lambda) \to L^2(\lambda)} \\
   &\qquad \times \|\mathcal{T}_{r,k'} - \mathcal{T}_{r,k} \|_{L^2(\lambda) \to L^2(\lambda)}\;
       \|\mathcal{P}_{r,k'} - \mathcal{P}_{r,k}\|_{L^\infty \to L^2}.
\end{align*}
Lemma~\ref{lemma::forwardopLipschitz} also gives
\[
\|\mathcal P_{r,k'}-\mathcal P_{r,k}\|_{L^\infty\to L^2}
\lesssim\|\mathcal P_{r,k'}-\mathcal P_{r,k}\|_{L^2\to L^2}
\lesssim\|k'-k\|_{\mathcal K}
+\|v_{r,k'}-v_{r,k}\|_{L^2(\lambda)}.
\]
Consequently,
\begin{align*}
   \big\|\big(\mathcal{T}_{r,k'}^{-1} - \mathcal{T}_{r,k}^{-1}\big)
&\qquad \times \big(\mathcal{P}_{r,k'} - \mathcal{P}_{r,k}\big)\big\|_{L^\infty \to L^2} \\
   &\lesssim  \|\mathcal{T}_{r,k}^{-1}\|_{L^2(\lambda) \to L^2(\lambda)}
       \|\mathcal{T}_{r,k'}^{-1}\|_{L^2(\lambda) \to L^2(\lambda)} \\
   &\qquad \times \left\{ \|k'-k\|_{L^2(\mu\otimes\lambda)}
    +  \|v_{r,k'}-v_{r,k}\|_{L^2(\lambda)} \right\}^2 \\
     &\lesssim    \|k'-k\|_{L^2(\mu\otimes\lambda)}^2
    +  \|v_{r,k'}-v_{r,k}\|_{L^2(\lambda)}^2.
\end{align*}

Combining the four bounds gives
\begin{align*}
 \|\partial_r v_{r,k'} - \partial_r v_{r,k}
&\qquad -  \adk \partial_r v_{r,k}[\cdot, k' - k]  \|_{L^\infty \rightarrow L^2} \\
&= \left \|\mathcal{T}_{r,k'}^{-1}\mathcal{P}_{r,k'}
 - \mathcal{T}_{r,k}^{-1}\mathcal{P}_{r,k}
 - \adk\!\big[\mathcal{T}_{r,k}^{-1}\mathcal{P}_{r,k}\big][\,k' - k\,] \right\|_{L^\infty \rightarrow L^2} \\
 & \lesssim  \|k'-k\|_{L^2(\mu\otimes\lambda)}^2
      + \|v_{r,k'}-v_{r,k}\|_{L^2(\lambda)}^2,
\end{align*}
which is the first inequality in the lemma.  The second follows from
\eqref{eqn::local-value-lipschitz}.  The resulting quadratic remainder in
\(\|k'-k\|_{\mathcal K}\) proves the stated Fr\'echet differentiability.
\end{proof}

\section{Reusable analytic lemmas}
\label{app::reusablelemmas}

\begin{lemma}[Perturbation bounds for bounded inverse operators]
\label{lemma::inverseperturbation}
Let \(A\) be a boundedly invertible operator on a Banach space, let
\(\Delta\) be bounded, and suppose \(\|A^{-1}\Delta\|<1\). Then
\(A+\Delta\) is invertible and
\begin{equation}
\|(A+\Delta)^{-1}\|
\le \frac{\|A^{-1}\|}{1-\|A^{-1}\Delta\|}.
\label{eqn::inverse-perturbation-bound}
\end{equation}
Moreover,
\begin{align}
(A+\Delta)^{-1}-A^{-1}
&=-A^{-1}\Delta(A+\Delta)^{-1},
\label{eqn::first-resolvent-identity}\\
(A+\Delta)^{-1}-A^{-1}+A^{-1}\Delta A^{-1}
&=A^{-1}\Delta A^{-1}\Delta(A+\Delta)^{-1}.
\label{eqn::quadratic-resolvent-identity}
\end{align}
More generally, for every bounded operator \(L\),
\begin{align}
&(A+\Delta)^{-1}-A^{-1}+A^{-1}LA^{-1}\notag\\
&\qquad=-A^{-1}(\Delta-L)A^{-1}
+A^{-1}\Delta A^{-1}\Delta(A+\Delta)^{-1}.
\label{eqn::generalized-resolvent-identity}
\end{align}
In particular, the inverse map is Fr\'echet differentiable at \(A\), with
derivative \(\Delta\mapsto-A^{-1}\Delta A^{-1}\), and its remainder is at
most
\[
\frac{\|A^{-1}\|^3\|\Delta\|^2}
{1-\|A^{-1}\Delta\|}.
\]
\end{lemma}

\begin{proof}
Write \(A+\Delta=A(I+A^{-1}\Delta)\). The Neumann series for
\((I+A^{-1}\Delta)^{-1}\) proves invertibility and the first bound. Writing
\(B=A+\Delta\), we have
\[
B^{-1}-A^{-1}=A^{-1}(A-B)B^{-1}=-A^{-1}\Delta B^{-1},
\]
which proves \eqref{eqn::first-resolvent-identity}. Consequently,
\[
\begin{aligned}
B^{-1}-A^{-1}+A^{-1}\Delta A^{-1}
&=A^{-1}\Delta(A^{-1}-B^{-1})\\
&=A^{-1}\Delta A^{-1}\Delta B^{-1},
\end{aligned}
\]
which proves \eqref{eqn::quadratic-resolvent-identity} and its norm bound.
Adding
\(A^{-1}(L-\Delta)A^{-1}\) to
\eqref{eqn::quadratic-resolvent-identity} gives
\eqref{eqn::generalized-resolvent-identity}.
\end{proof}

\begin{lemma}[Approximation of tangent directions]
\label{lemma::tangentapproximation}
Under Assumption~\ref{cond::boundedpositivity}, the map
\(\beta\mapsto k_0\beta\) is a bounded linear isomorphism from
\(T_{\mathcal K}\) to \(\addTK\). In particular,
\[
\|k_0\beta\|_{\mathcal K}\asymp
\|\beta\|_{\mathcal K,0},
\qquad \beta\in T_{\mathcal K}.
\]
The bounded elements of \(T_{\mathcal H}\) are dense in
\(T_{\mathcal H}\), and the bounded, conditionally centered elements of
\(T_{\mathcal K}\) are dense in \(T_{\mathcal K}\). Consequently, two
continuous linear functionals on either tangent space that agree on the
corresponding bounded directions agree on the entire tangent space.
\end{lemma}

\begin{proof}
Because \(\mathcal H\subset L^\infty(\lambda)\) is dense in its completion
\(T_{\mathcal H}\), bounded reward directions are dense in
\(T_{\mathcal H}\). For
\(\beta\in T_{\mathcal K}\), define
\[
c_M:=\beta\,\mathbbm 1\{|\beta|\le M\},
\qquad
\beta_M:=c_M-E_0(c_M\mid A,S).
\]
Then \(\beta_M\) is bounded and
\(E_0(\beta_M\mid A,S)=0\). Since
\(E_0(\beta\mid A,S)=0\), conditional Jensen's inequality gives
\[
\|\beta_M-\beta\|_{\mathcal K,0}
\le 2\|\beta\,\mathbbm 1\{|\beta|>M\}\|_{\mathcal K,0}
\longrightarrow0.
\]
For each \(M\), let \(g_M:=k_0\beta_M\). Then \(g_M\) is bounded and
\[
\int g_M(s'\mid a,s)\,\mu(ds')
=E_0\{\beta_M(S',A,S)\mid A=a,S=s\}=0.
\]
Thus \(g_M\) belongs to the bounded generating set of \(\addTK\). Moreover,
the uniform upper and positive lower bounds on \(k_0\), the pooled state
density, and \(\pi_0\) in Assumption~\ref{cond::boundedpositivity} give
\begin{equation}
\|k_0\beta\|_{\mathcal K}\asymp\|\beta\|_{\mathcal K,0}.
\label{eqn::kernel-score-additive-norm-equivalence}
\end{equation}
Hence \(g_M\to k_0\beta\) in \(\|\cdot\|_{\mathcal K}\), so
\(k_0\beta\in\addTK\).

Conversely, let \(g\in\addTK\), and choose bounded, conditionally centered
functions \(g_M\) from the defining generating set of \(\addTK\), with
\(g_M\to g\) in \(\|\cdot\|_{\mathcal K}\). Define
\(\widetilde\beta_M=g_M/k_0\). Then
\(E_0(\widetilde\beta_M\mid A,S)=0\), and
\eqref{eqn::kernel-score-additive-norm-equivalence} shows that
\(\widetilde\beta_M\) is Cauchy in \(T_{\mathcal K}\). Its limit is
\(g/k_0\), which proves that multiplication by \(k_0\) maps
\(T_{\mathcal K}\) onto \(\addTK\) with a bounded inverse. Taking limits of the bounded
approximations and using continuity of both linear functionals proves
their equality on the full tangent space.
\end{proof}

\begin{lemma}[Initial-state and pooled-transition bounds]
\label{lemma::initialpooledbounds}
Under Assumption~\ref{cond::boundedpositivity}, there is a constant
\(C<\infty\) such that, for every \(f\in L^2(\mu)\) and every
\(g\in L^2(P_0)\),
\[
\|f(S_0)\|_{L^2(\mathbb P_0)}\le C\|f\|_{L^2(\mu)},
\qquad
\left\|\frac1T\sum_{t=0}^{T-1}g(Z_t)\right\|_{L^2(\mathbb P_0)}
\le \|g\|_{L^2(P_0)}.
\]
\end{lemma}

\begin{proof}
Because the initial-state density is one summand in the pooled density,
\(\rho_{0,\mathrm{in}}\le T\rho_0\le TC_0\). Hence
\[
\|f(S_0)\|_{L^2(\mathbb P_0)}^2
\le TC_0\|f\|_{L^2(\mu)}^2.
\]
For the second inequality, Jensen's inequality gives
\[
\mathbb E_0\!\left[\left\{\frac1T\sum_{t=0}^{T-1}g(Z_t)\right\}^2\right]
\le \frac1T\sum_{t=0}^{T-1}\mathbb E_0\{g(Z_t)^2\}
=E_0\{g(Z)^2\},
\]
where the last equality is the definition of the pooled transition law.
\end{proof}

\begin{lemma}[Score identities for the time-homogeneous Markov model]
\label{lemma::markovscoreidentities}
Let
\[
\ell(O)=u(S_0)+\sum_{t=0}^{T-1}
\{h(A_t,S_t)+b(S_{t+1},A_t,S_t)\},
\]
where \(\mathbb E_0\{u(S_0)\}=0\), \(\pi_0h(S)=0\), and
\(E_0\{b(S',A,S)\mid A,S\}=0\). Every regular submodel of
\(\mathfrak M_T\) has a score of this form, every bounded triple with these
centering properties generates a regular submodel, and the tangent space is
the closure of the bounded centered triples. For any
square-integrable \(f\) of one transition and any regular submodel with
score \(\ell\),
\begin{equation}
\left.\frac{d}{d\varepsilon}P_\varepsilon f\right|_{\varepsilon=0}
=\mathbb E_0\!\left[\ell(O)\frac1T\sum_{t=0}^{T-1}f(Z_t)\right].
\label{eqn::pooledscorelifting}
\end{equation}
If \(x(S_0)\), \(a(A,S)\), and \(c(S',A,S)\) are square-integrable and
satisfy the same respective centering restrictions, then
\begin{align}
&\mathbb E_0\!\left[\ell(O)\left\{x(S_0)
+\frac1T\sum_{t=0}^{T-1}
\bigl(a(A_t,S_t)+c(S_{t+1},A_t,S_t)\bigr)\right\}\right]\\
&\qquad=\mathbb E_0\{u(S_0)x(S_0)\}
+E_0\{h(A,S)a(A,S)\}
+E_0\{b(S',A,S)c(S',A,S)\}.
\label{eqn::markov-score-inner-product}
\end{align}
\end{lemma}

\begin{proof}
Differentiating the initial-density, policy, and transition-density factors
of a regular submodel gives the displayed decomposition of \(\ell\), and
normalization of each factor gives the three centering restrictions. For a
bounded triple, the multiplicative factors
\(1+\varepsilon u\), \(1+\varepsilon h\), and
\(1+\varepsilon b\) define normalized nonnegative densities for all
sufficiently small \(|\varepsilon|\). Their density ratios relative to the
true factors are then uniformly bounded above and away from zero. Because
the horizon is finite, the same is true of the resulting pooled state-density
ratio. The strict interior bounds in
Assumption~\ref{cond::boundedpositivity} therefore imply that the path lies in
\(\mathfrak M_T\) for all sufficiently small \(|\varepsilon|\), and its
trajectory score is \(\ell\). Taking closures proves the tangent-space
statement.

The pooled law is the average of the laws of \(Z_t\), so differentiating
this finite average proves \eqref{eqn::pooledscorelifting}.  To prove
\eqref{eqn::markov-score-inner-product}, let
\(\mathcal G_t=\sigma(S_0,A_0,S_1,\ldots,A_{t-1},S_t)\), and set
\[
\xi_t=h(A_t,S_t)+b(S_{t+1},A_t,S_t),\qquad
\zeta_t=a(A_t,S_t)+c(S_{t+1},A_t,S_t).
\]
The centering restrictions give
\begin{equation}
\mathbb E_0(\xi_t\mid\mathcal G_t)=0,
\qquad
\mathbb E_0(\zeta_t\mid\mathcal G_t)=0.
\label{eqn::markov-increment-centering}
\end{equation}
For \(t<j\), \(\xi_t\) is \(\mathcal G_j\)-measurable, so
\begin{equation}
\mathbb E_0(\xi_t\zeta_j)
=\mathbb E_0\{\xi_t\mathbb E_0(\zeta_j\mid\mathcal G_j)\}=0.
\label{eqn::markov-offdiagonal-orthogonality}
\end{equation}
For \(j<t\), \(\zeta_j\) is \(\mathcal G_t\)-measurable, and hence
\begin{equation}
\mathbb E_0(\xi_t\zeta_j)
=\mathbb E_0\{\zeta_j\mathbb E_0(\xi_t\mid\mathcal G_t)\}=0.
\label{eqn::markov-reverse-offdiagonal-orthogonality}
\end{equation}
Likewise, because \(u(S_0)\) and \(x(S_0)\) are
\(\mathcal G_t\)-measurable,
\begin{equation}
\mathbb E_0\{u(S_0)\zeta_t\}=0,
\qquad
\mathbb E_0\{x(S_0)\xi_t\}=0.
\label{eqn::markov-initial-increment-orthogonality}
\end{equation}
At a common time index,
\begin{align}
\mathbb E_0\{h(A_t,S_t)c(S_{t+1},A_t,S_t)\}
&=\mathbb E_0[h(A_t,S_t)
\mathbb E_0\{c(S_{t+1},A_t,S_t)\mid A_t,S_t\}]=0,\notag\\
\mathbb E_0\{b(S_{t+1},A_t,S_t)a(A_t,S_t)\}
&=\mathbb E_0[a(A_t,S_t)
\mathbb E_0\{b(S_{t+1},A_t,S_t)\mid A_t,S_t\}]=0.
\label{eqn::markov-withintime-orthogonality}
\end{align}
Therefore,
\[
\frac1T\sum_{t=0}^{T-1}\mathbb E_0(\xi_t\zeta_t)
=E_0\{h(A,S)a(A,S)\}+E_0\{b(S',A,S)c(S',A,S)\},
\]
where \(E_0\) on the right is expectation under the pooled transition law.
Equations~\eqref{eqn::markov-offdiagonal-orthogonality}--
\eqref{eqn::markov-initial-increment-orthogonality} and
\eqref{eqn::markov-withintime-orthogonality} now give
\eqref{eqn::markov-score-inner-product}.
\end{proof}

\begin{lemma}[Empirical-process bound under cross-fitting]
\label{lemma::crossfitprocess}
Let \(I_1,\ldots,I_J\) be a fixed number of evaluation folds. For each
\(j\), let \(f_{n,j}\) be estimated without using observations in \(I_j\),
and suppose \(P_0f_{n,j}^2=o_P(1)\). Then
\[
\sum_{j=1}^J\frac{|I_j|}{n}
(\mathbb P_{n,j}-\mathbb P_0)f_{n,j}=o_P(n^{-1/2}),
\]
where \(\mathbb P_{n,j}\) is the empirical measure on \(I_j\). The same
conclusion holds for an independent training sample by taking \(J=1\).
\end{lemma}

\begin{proof}
Conditionally on the training data for fold \(j\), the observations in
\(I_j\) are independent of \(f_{n,j}\), and
the weighted fold contribution satisfies
\[
\begin{aligned}
&\operatorname{Var}\!\left\{
\frac{|I_j|}{n}(\mathbb P_{n,j}-\mathbb P_0)f_{n,j}
\,\middle|\, f_{n,j}\right\}\\
&\qquad\le \frac{|I_j|}{n^2}P_0f_{n,j}^2
\le \frac1nP_0f_{n,j}^2=o_P(n^{-1}).
\end{aligned}
\]
Conditional Chebyshev's inequality implies that each weighted contribution
is \(o_P(n^{-1/2})\). Since \(J\) is fixed, their sum is
\(o_P(n^{-1/2})\), without requiring balanced fold sizes.
\end{proof}

\clearpage
{\singlespacing
\bibliographystyle{abbrvnat}
\bibliography{ref}
}
\end{document}